\documentclass{article}

\usepackage{microtype}
\usepackage{graphicx}
\usepackage{subcaption}
\usepackage{booktabs} 

\usepackage{hyperref}

\usepackage[preprint]{icml2026}

\usepackage{amsmath}
\usepackage{amssymb}
\usepackage{mathtools}
\usepackage{amsthm}

\makeatletter
\renewcommand{\Notice@String}{}
\def\@copyrightspace{}
\makeatother

\usepackage[capitalize,noabbrev]{cleveref}

\theoremstyle{plain}

\theoremstyle{definition}

\theoremstyle{remark}

\usepackage[textsize=tiny]{todonotes}
\usepackage{xcolor}
\usepackage{subcaption}
\usepackage{enumitem}
\usepackage[table]{xcolor} 
\usepackage{multirow}
\usepackage{float}
\definecolor{commentgrey}{RGB}{0,102,204}
\icmltitlerunning{Linear-DPO: Linear Direct Preference Optimization for Diffusion and Flow-Matching Generative Models}

\begin{document}

\icmlcorrespondingauthor{}{}



\twocolumn[
\begin{center}
{\LARGE \bf
Linear-DPO: Linear Direct Preference Optimization for Diffusion and Flow-Matching Generative Models
\par}
\vskip 0.3in

{\large
Kesong Li$^{1,2}$ \quad
Yixuan Xu$^{2}$ \quad
Kuo-kun Tseng$^{1}$ \quad
Weiyi Lu$^{2}$ \quad
Kan Liu$^{2}$ \quad
Tao Lan$^{2}$
\par}
\vskip 0.15in

{\normalsize
$^{1}$School of Computer Science and Technology, Harbin Institute of Technology, Shenzhen, China \\
$^{2}$Alibaba Group, Hangzhou, Zhejiang, China
\par}
\vskip 0.1in

{\normalsize
\texttt{kktseng@hit.edu.cn, weiyi.lwy@alibaba-inc.com}
\par}
\vskip 0.3in
\end{center}
]
\thispagestyle{empty}



\begingroup
\renewcommand{\footnotetext}[1]{} 
\printAffiliationsAndNotice{}     
\endgroup                         
\begin{abstract}
Direct Preference Optimization (DPO) is successful for alignment in LLMs but still faces challenges in text-to-image generation. Existing studies are confined to denoising diffusion models while overlooking flow-matching, and suffer from an objective mismatch when applying discrete NLP-based DPO to regression-based generative tasks.\ In this paper, we derive a generalized DPO objective that covers both diffusion and flow-matching via a unified reverse-time SDE framework, and point out from a gradient perspective that the standard DPO objective is suboptimal for text-to-image generation. Consequently, we propose Linear-DPO, which replaces the aggressive sigmoid-based utility function with a sustained linear utility and incorporates an EMA-updated reference model. Qualitative and quantitative experiments on diffusion models (SD1.5, SDXL) and flow-matching model (SD3-Medium) demonstrate the superiority of our approach over existing baselines.
Code and model weights are available at \url{https://github.com/Whynot0101/Linear-DPO}.
\end{abstract}
\vskip -0.1in
\begin{figure*}[h]
\centering
\begin{minipage}{0.87\textwidth}
\centering
\vskip -0.1in
\begin{minipage}{0.2\textwidth}\includegraphics[width=\linewidth]{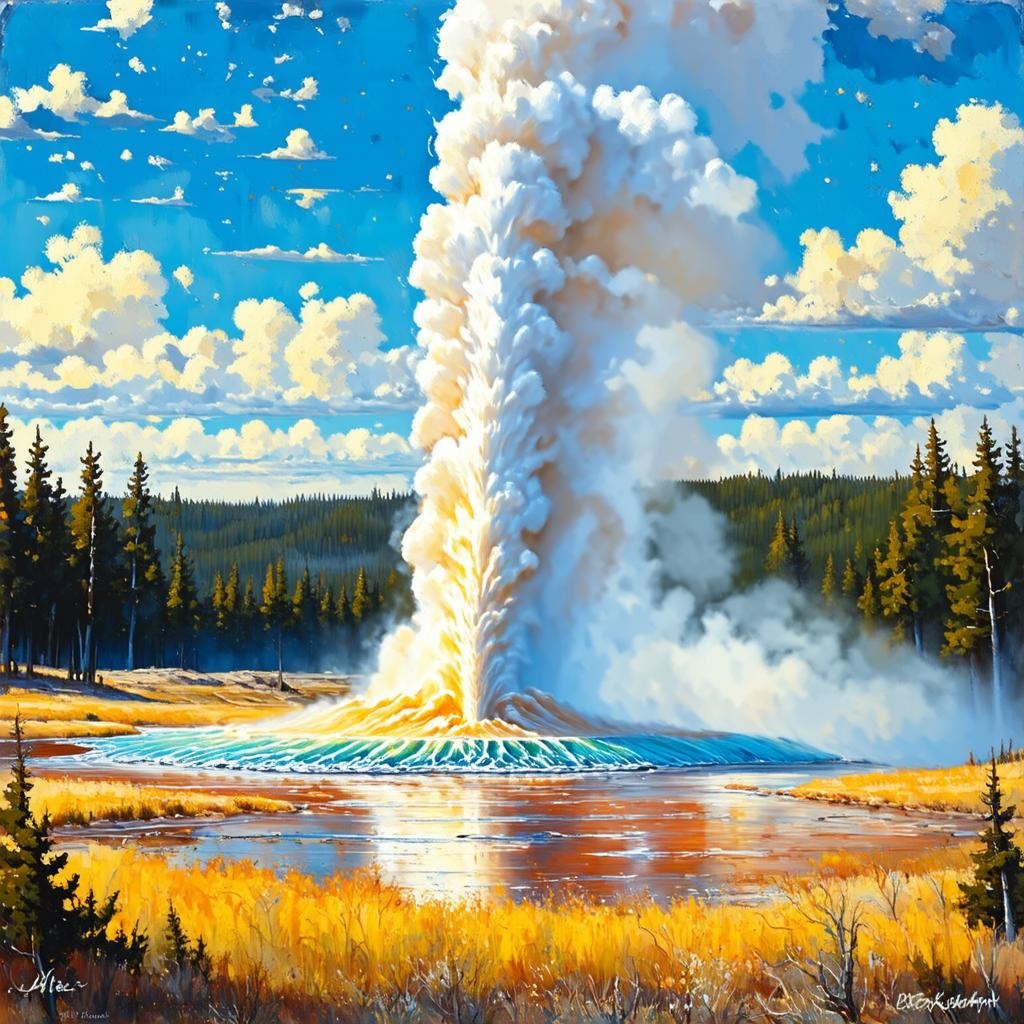}\end{minipage}%
\begin{minipage}{0.2\textwidth}\includegraphics[width=\linewidth]{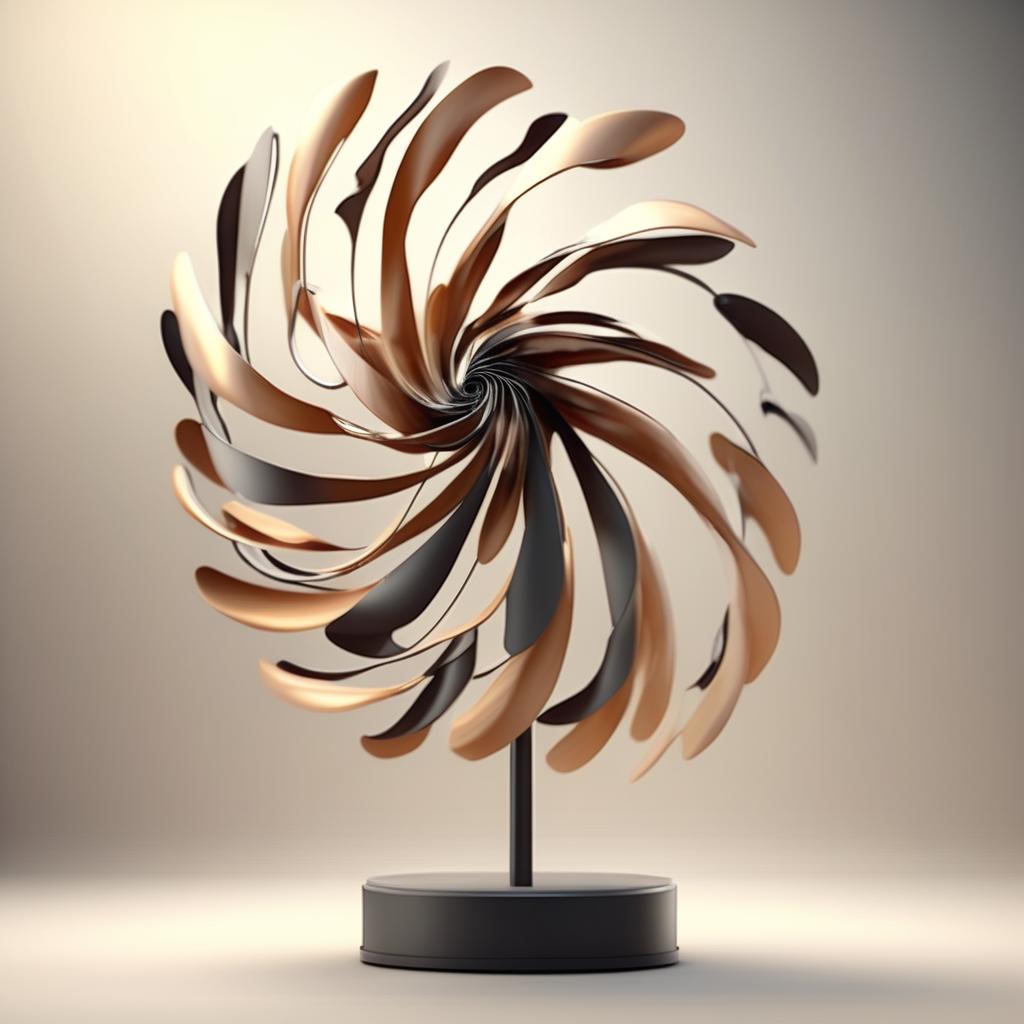}\end{minipage}%
\begin{minipage}{0.2\textwidth}\includegraphics[width=\linewidth]{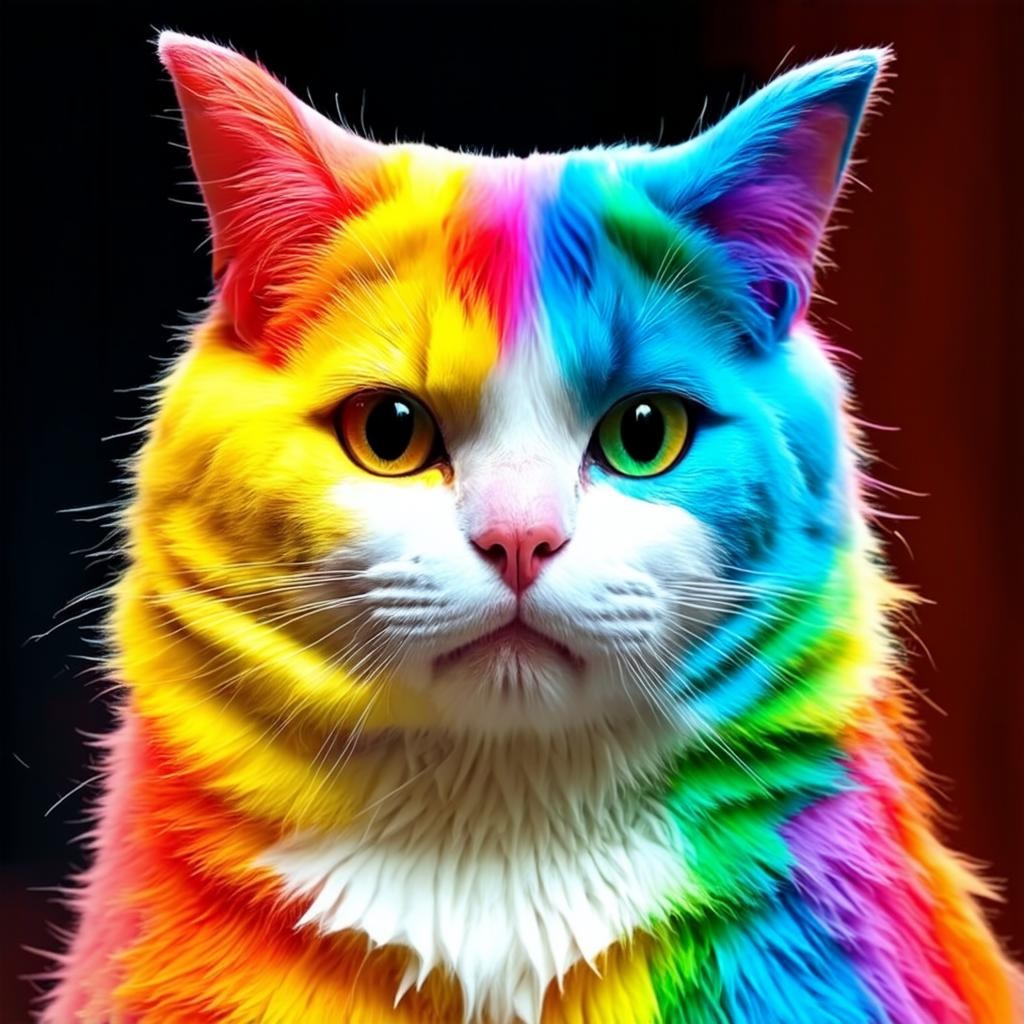}\end{minipage}%
\begin{minipage}{0.2\textwidth}\includegraphics[width=\linewidth]{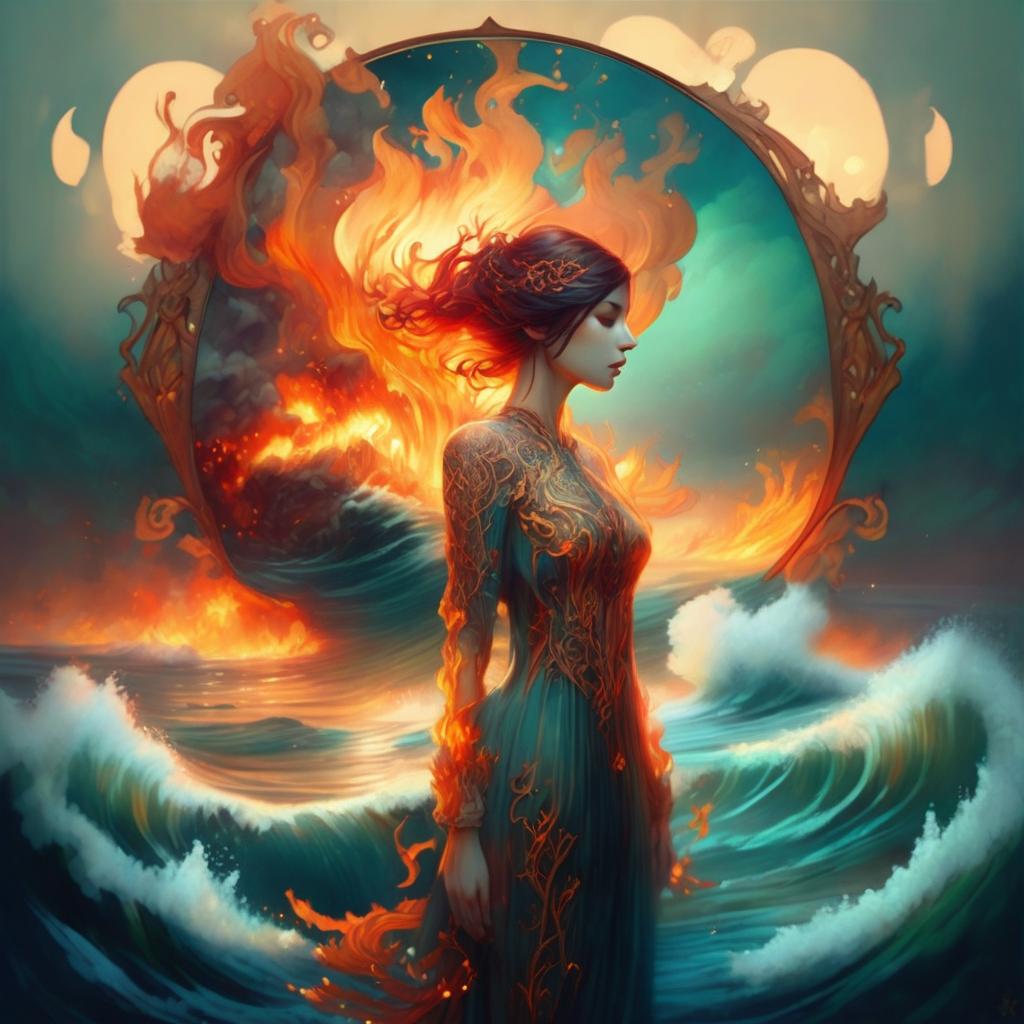}\end{minipage}%
\begin{minipage}{0.2\textwidth}\includegraphics[width=\linewidth]{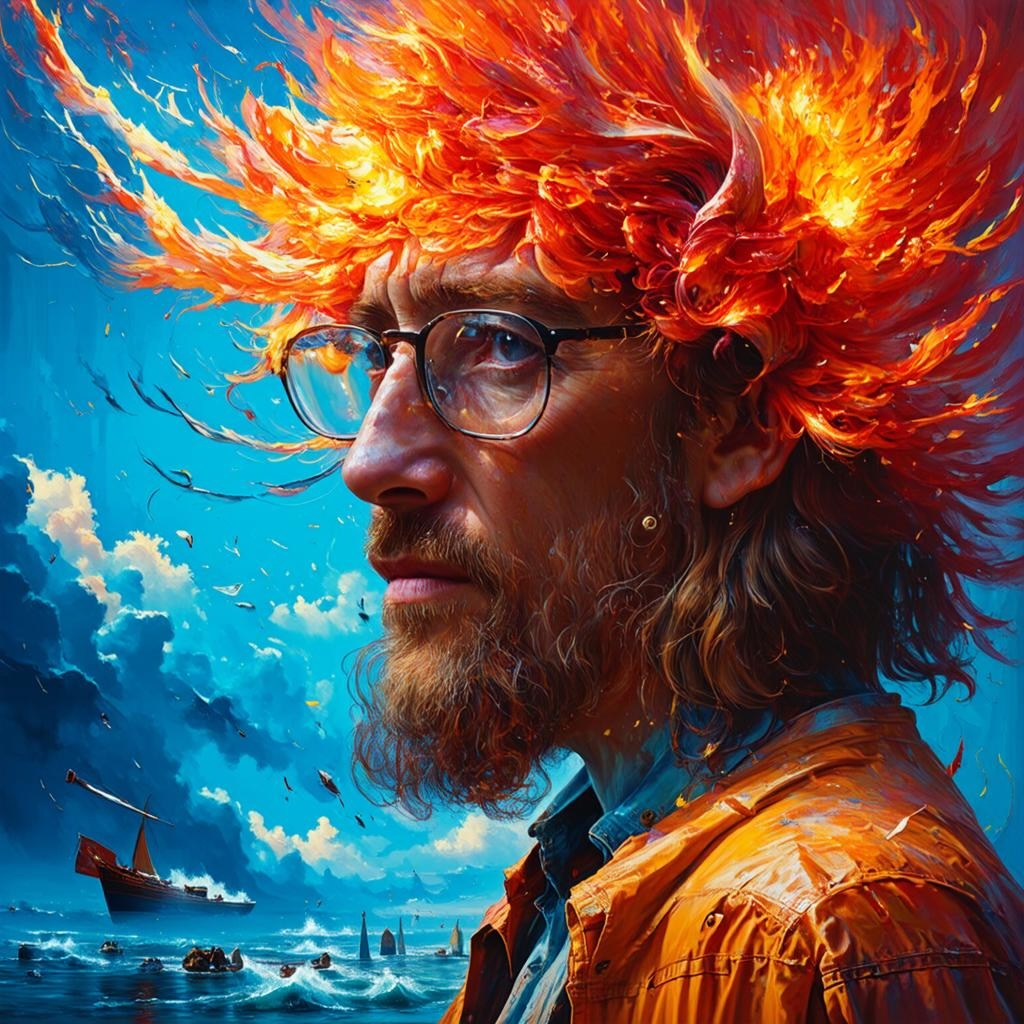}\end{minipage}%

\begin{minipage}{0.2\textwidth}\includegraphics[width=\linewidth]{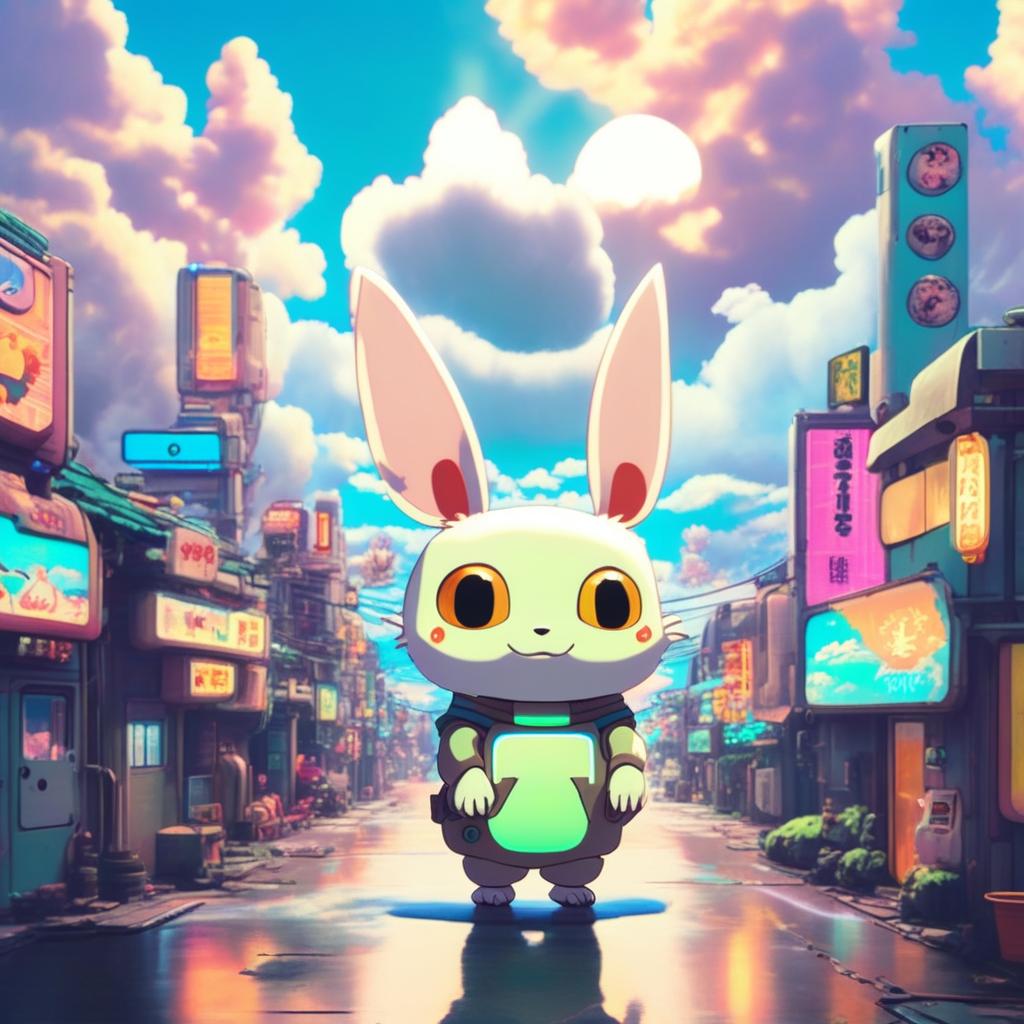}\end{minipage}%
\begin{minipage}{0.2\textwidth}\includegraphics[width=\linewidth]{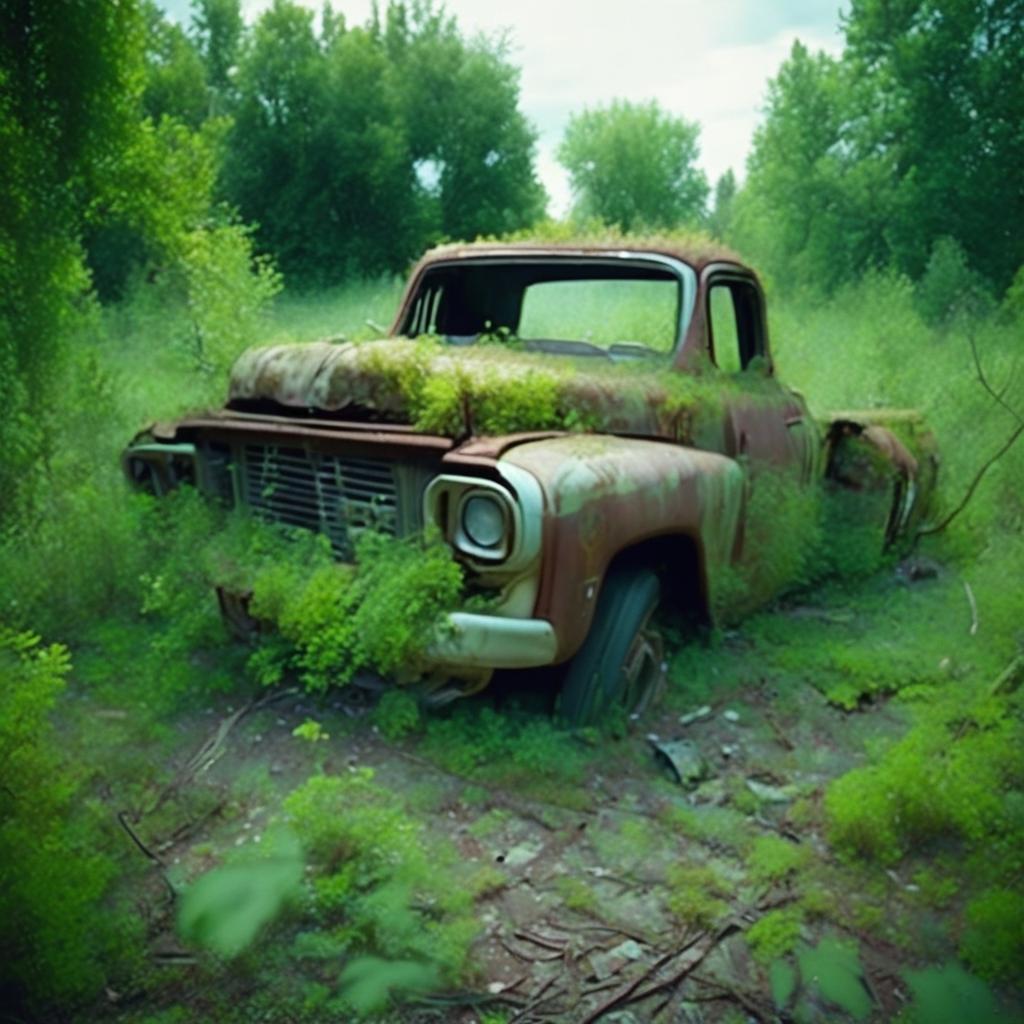}\end{minipage}%
\begin{minipage}{0.2\textwidth}\includegraphics[width=\linewidth]{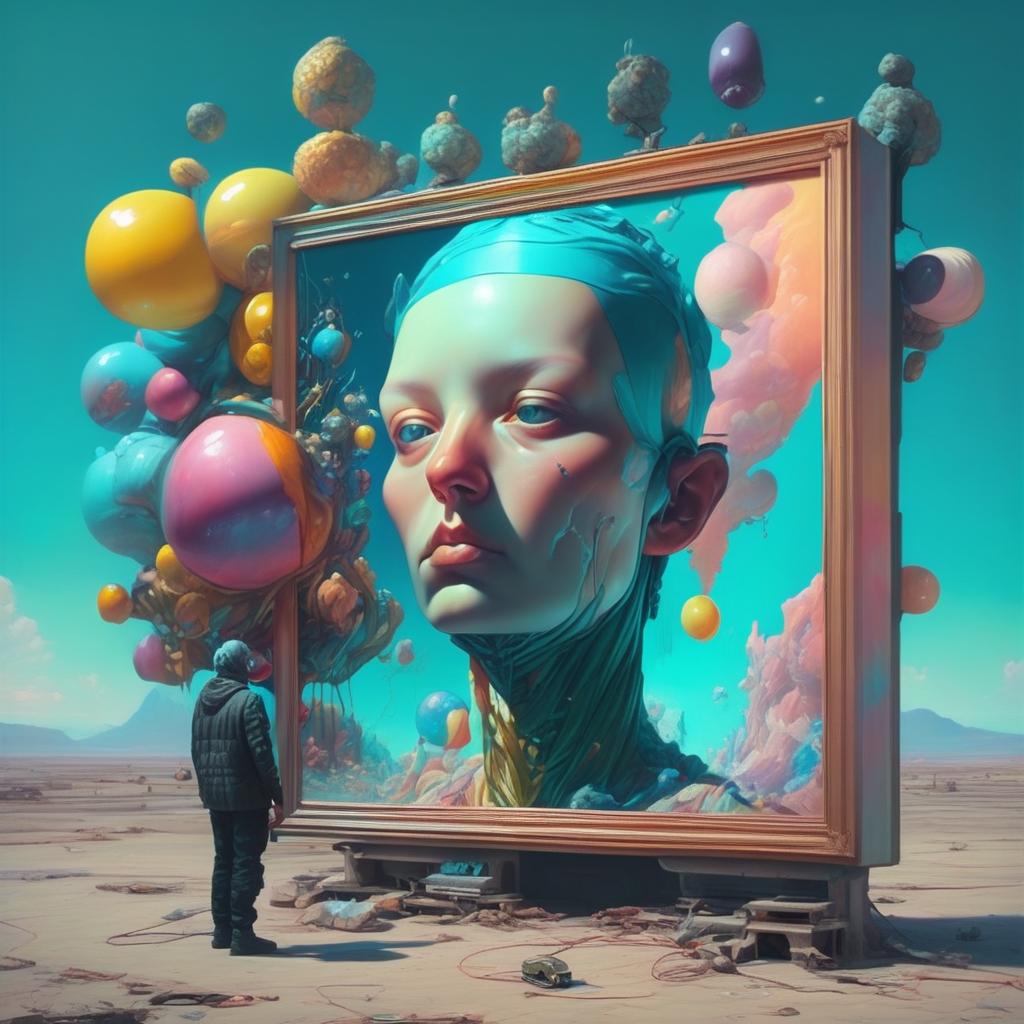}\end{minipage}%
\begin{minipage}{0.2\textwidth}\includegraphics[width=\linewidth]{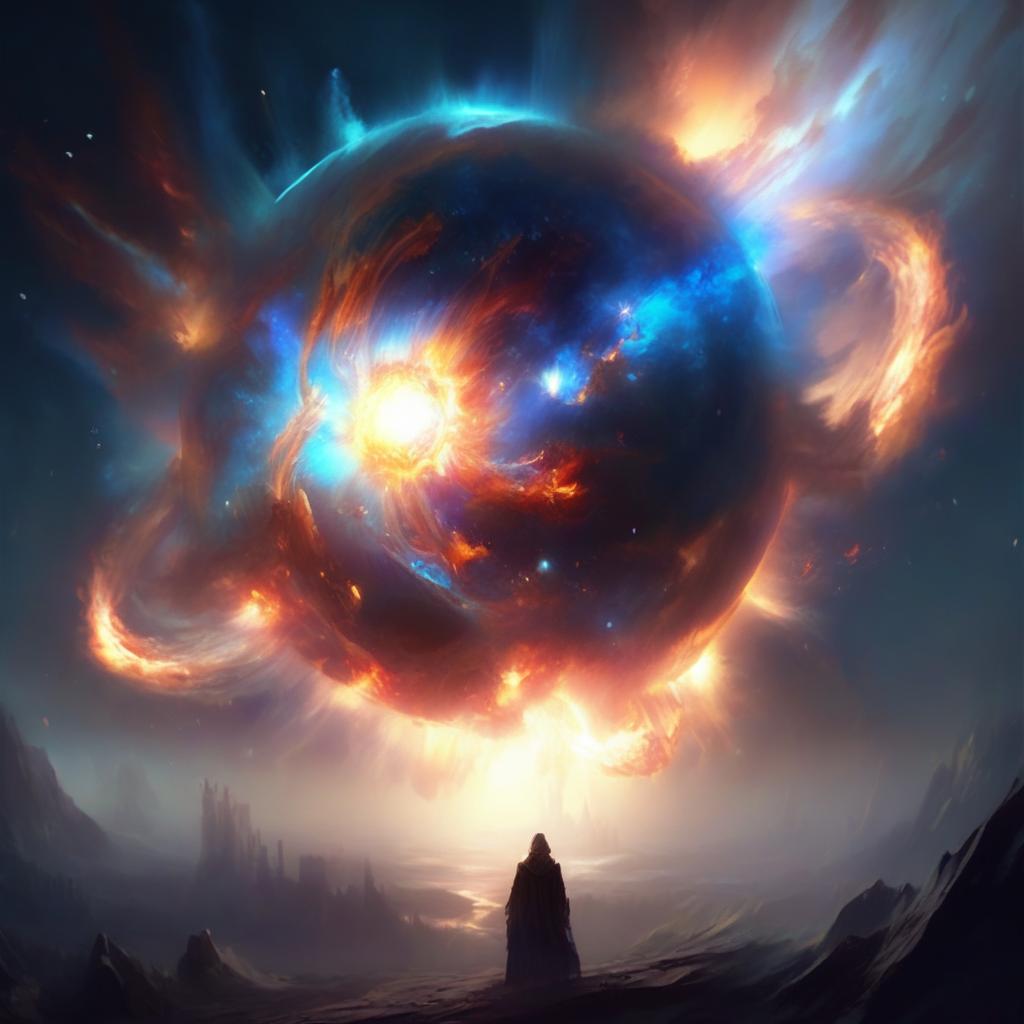}\end{minipage}%
\begin{minipage}{0.2\textwidth}\includegraphics[width=\linewidth]{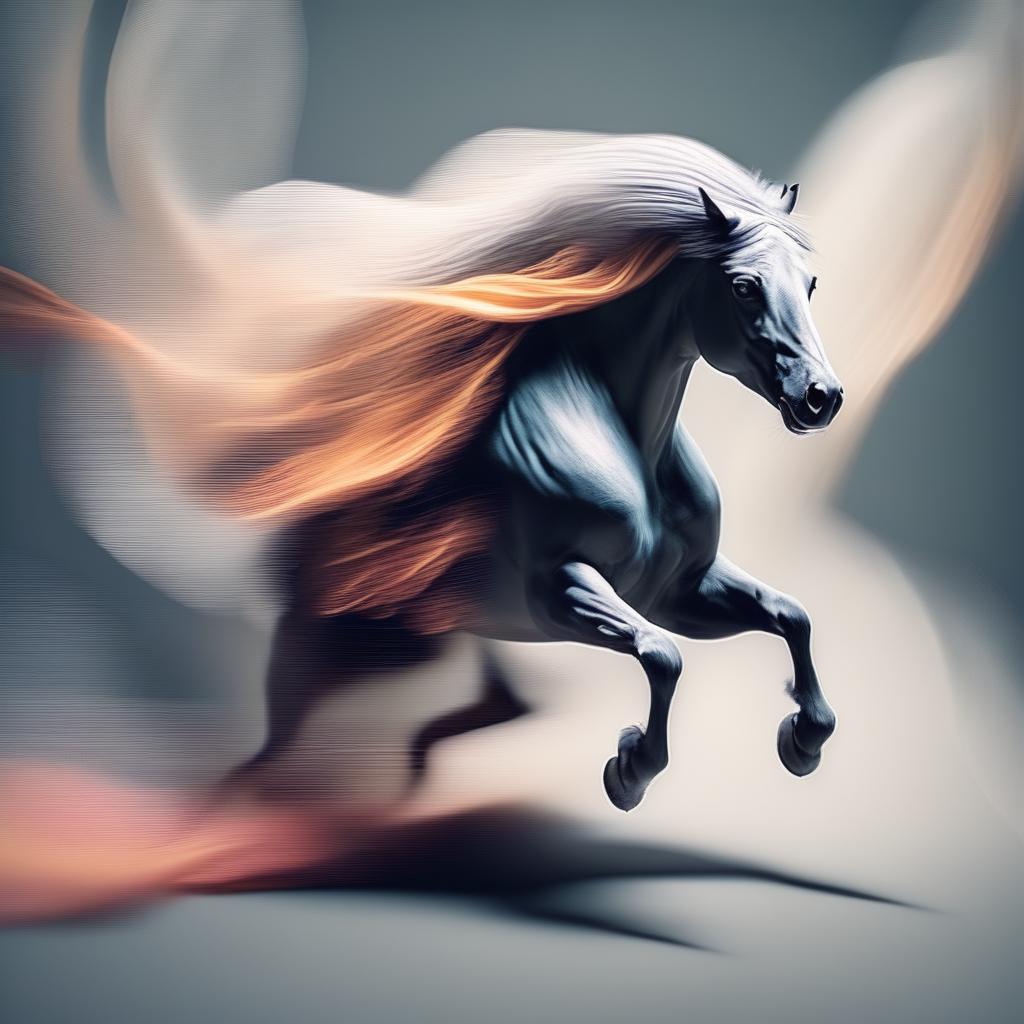}\end{minipage}%

\begin{minipage}{0.2\textwidth}\includegraphics[width=\linewidth]{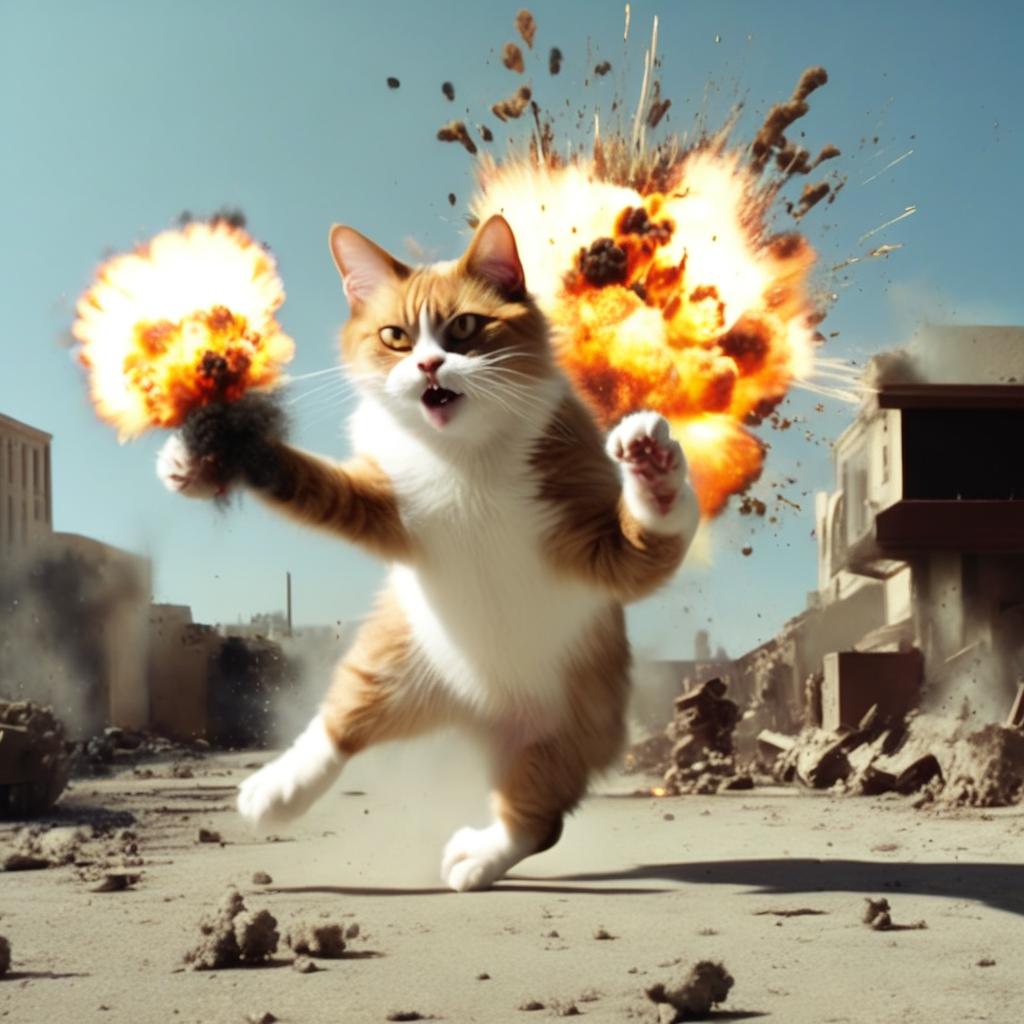}\end{minipage}%
\begin{minipage}{0.2\textwidth}\includegraphics[width=\linewidth]{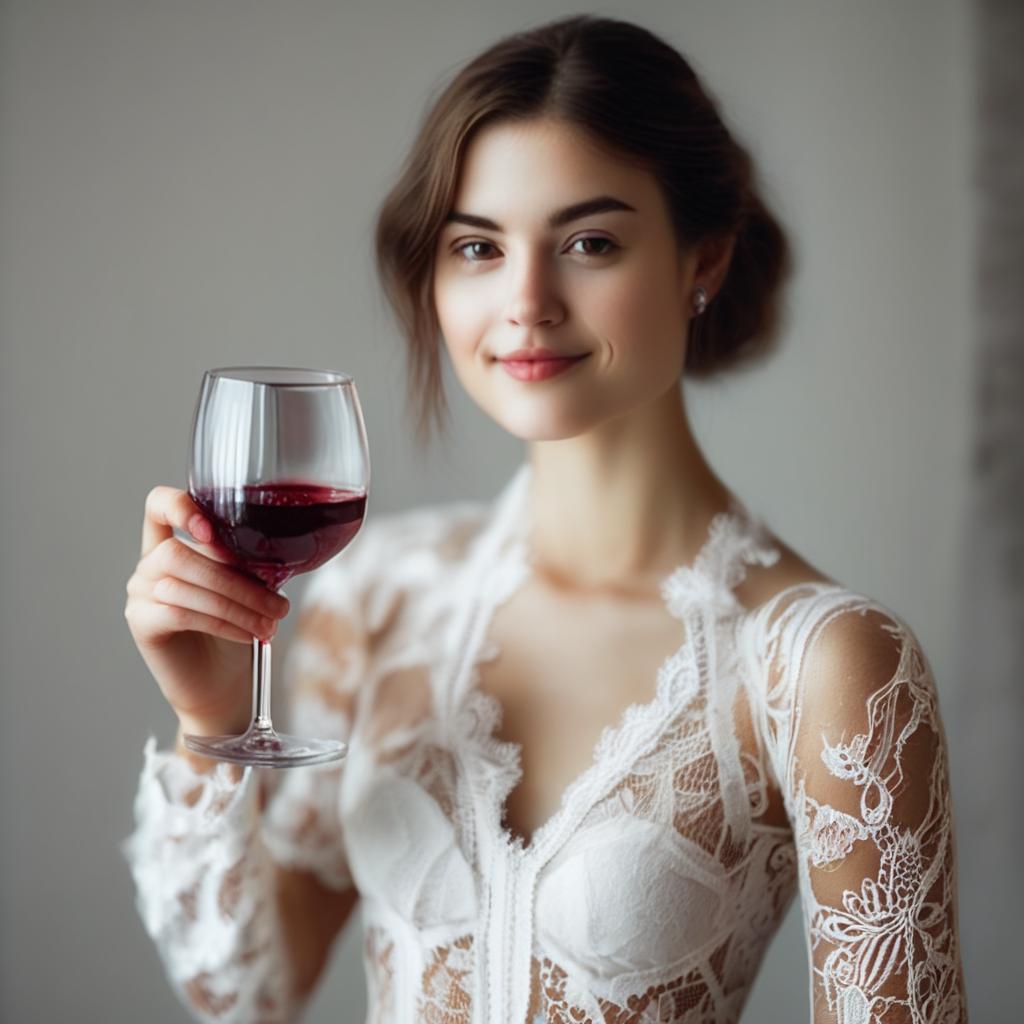}\end{minipage}%
\begin{minipage}{0.2\textwidth}\includegraphics[width=\linewidth]{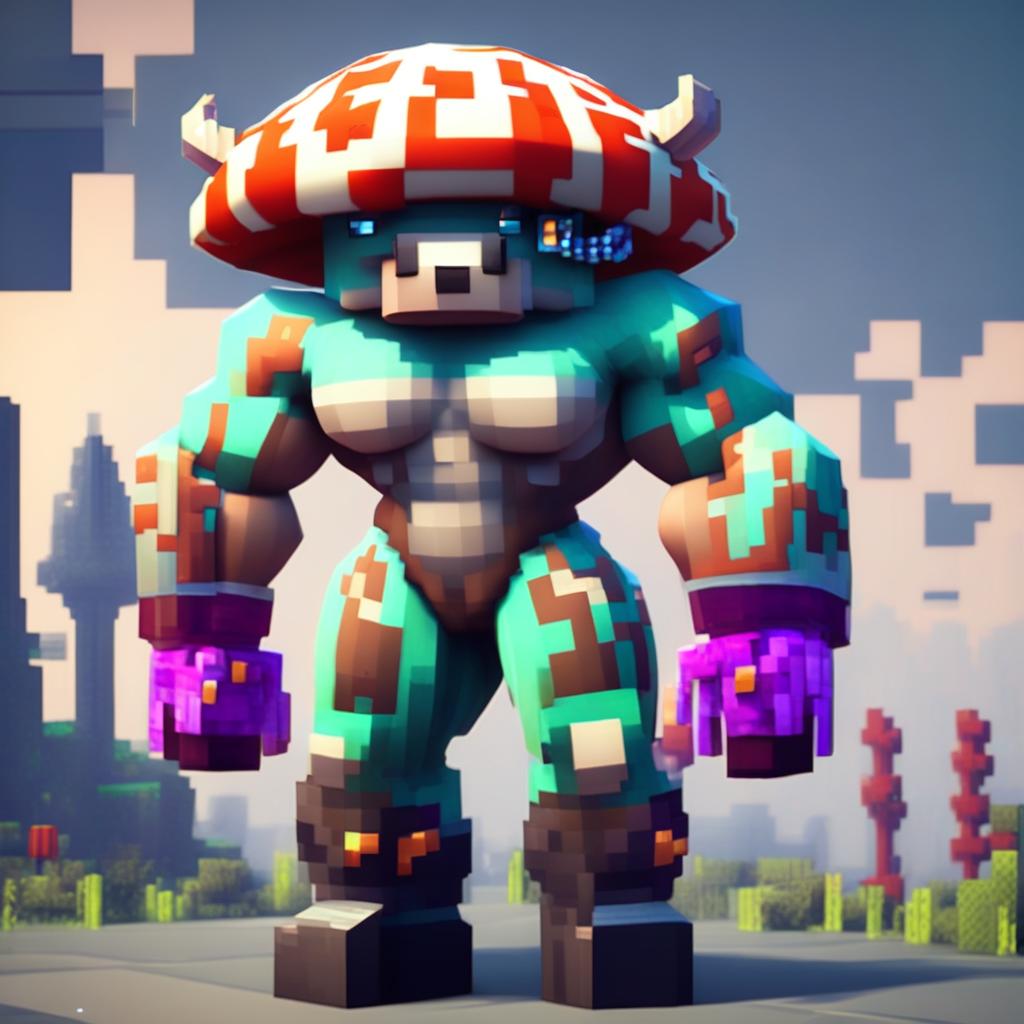}\end{minipage}%
\begin{minipage}{0.2\textwidth}\includegraphics[width=\linewidth]{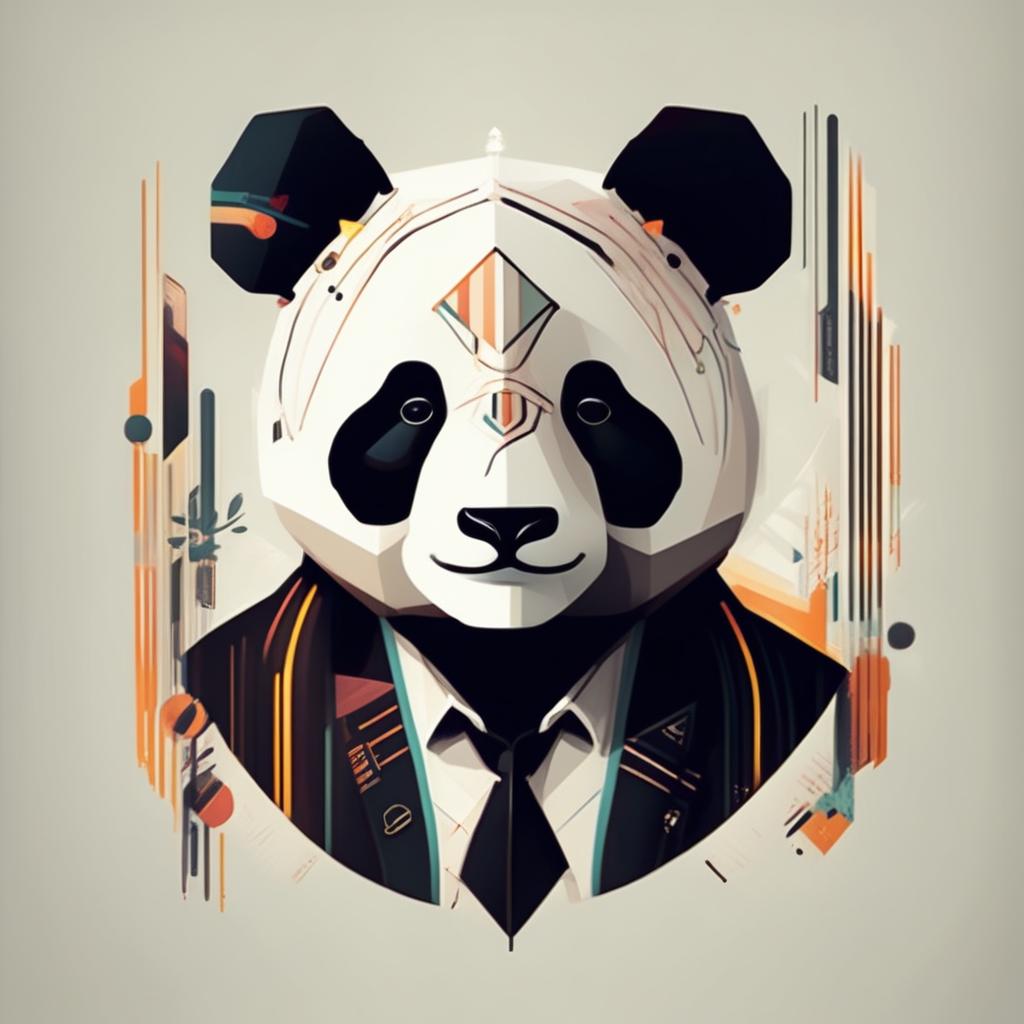}\end{minipage}%
\begin{minipage}{0.2\textwidth}\includegraphics[width=\linewidth]{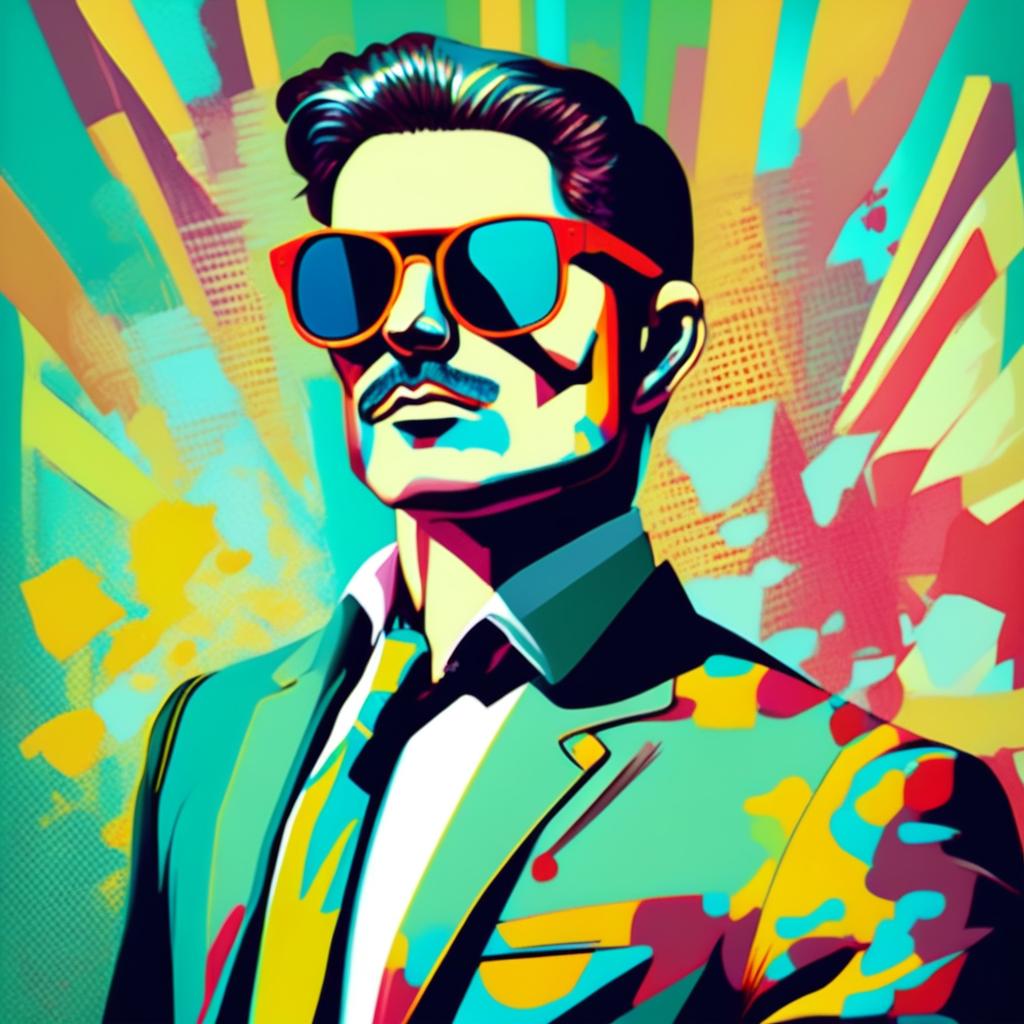}\end{minipage}%
\end{minipage}
\caption{Samples from SDXL and SD3-M fine-tuned with our proposed \textbf{Linear-DPO}. Linear-DPO is a more powerful direct preference optimization method designed for diffusion and flow-matching generative models; the results show significant improvements in visual appeal, detail richness, and alignment with human preferences.} 
\vskip -0.15in
\end{figure*}

\section{Introduction}
Diffusion~\cite{sohl2015deep, ho2020denoising} and flow-matching~\cite{lipman2022flow, liu2022flow} models have achieved remarkable progress in text-to-image (T2I) generation and underpin many large-scale pretrained foundation models~\cite{rombach2022high, podell2023sdxl, esser2024scaling, wu2025qwen}. Despite their impressive generative capabilities, these models are typically trained to match web-scale image–text distributions rather than human preferences. Motivated by the success of Reinforcement Learning from Human Feedback (RLHF)~\cite{christiano2017deep, schulman2017proximal, shao2024deepseekmath} in large language models, recent studies~\cite{black2023training, fan2023dpok, liu2025flow, xu2023imagereward} have begun to explore human preference alignment for T2I generation. However, existing approaches remain constrained by high computational overhead and a heavy dependency on high-quality reward models.

In particular, Direct Preference Optimization (DPO) ~\cite{rafailov2023direct} starts from the RLHF objective and analytically derives an implicit reward that depends only on the policy model and reference model. This allows one to optimize directly on offline human preference pairs, eliminating the need for a pre-trained reward model.
Diffusion-DPO~\cite{wallace2024diffusion} further extends this idea to diffusion models by formulating the DPO loss in terms of standard denoising losses. Despite improvements over the original Diffusion-DPO formulation, subsequent variants~\cite{li2024aligning,zhu2025dspo,hong2024margin,li2025divergence,fu2025diffusion} still exhibit critical limitations:

\textbf{(a) Limited to diffusion models.} Their theoretical analysis and validation are confined to U-Net-based denoising diffusion models such as SD1.5~\cite{rombach2022high} and SDXL~\cite{podell2023sdxl}. In contrast, state-of-the-art models (e.g., SD3-Medium~\cite{esser2024scaling}, Qwen-Image~\cite{wu2025qwen}) have shifted toward flow-matching training paradigms~\cite{lipman2022flow} and MMDiT architectures. This creates an urgent need for a unified DPO formulation covering both diffusion and flow-matching generative models which can be rigorously evaluated on large-scale, modern generative models.

\textbf{(b) Optimization objective mismatch.} Direct transfer of the DPO formulation to regression-based training objectives overlooks a critical domain gap. The DPO objective implies a discrete ranking logic tailored for language modeling: once the probability gap between chosen and rejected tokens is sufficiently large, the gradients vanish to stabilize training. However, this "margin-maximization" behavior is fundamentally ill-suited for the regression nature of diffusion and flow-matching models. Since text-to-image generation requires long-horizon, stable updates to continuously refine fine-grained visual details, the rapid gradient decay in standard DPO creates a "pseudo-convergence" trap, causing the model to become rigid too early in the training process.

To address the challenges above, we seek to answer: \emph{Can we design a single, principled direct preference optimization objective that applies universally to diffusion, score-based, and flow-matching models, while fundamentally suiting such regression-based training objectives?}

First, we cast flow-matching into a Stochastic Differential Equation (SDE) perspective via a reverse-time SDE formulated by vector fields, thereby introducing the necessary stochasticity to derive conditional probabilities. Based on this unified SDE framework, we derive a generalized DPO objective that covers diffusion and flow-matching generative models.
Subsequently, we analyze this unified objective from a gradient perspective and show that its optimization can be viewed as weighted Supervised Fine-Tuning (SFT), where the gradient is the difference between the SFT gradients of chosen and rejected samples, dynamically modulated by a sigmoid-based utility function. To better suit generative tasks, we introduce \textbf{Linear-DPO}, which replaces this sigmoid gating with a more sustained linear utility function and substitutes the fixed reference model with an EMA-updated copy of the policy model to encourage continuous optimization.
We evaluate Linear-DPO on diffusion (SD1.5, SDXL) and flow-matching (SD3-M) models, achieving consistent gains over pretrained models, SFT, and prior Diffusion-DPO variants across human preference benchmarks and automated metrics. Our contributions are summarized as:
\begin{itemize}[leftmargin=*, itemsep=-5pt, topsep=0pt]
\item We establish a generalized DPO objective bridging diffusion and flow-matching via a unified SDE perspective.
\item We analyze DPO via a gradient lens and propose Linear-DPO, a novel approach better suited for T2I generation.
\item We demonstrate the effectiveness and scalability of Linear-DPO across models ranging from diffusion (SD1.5, SDXL) to advanced flow-matching models (SD3-M).
\end{itemize}

\section{Related works}
\subsection{Diffusion and Flow-Matching Generative Models}
Diffusion models~\cite{sohl2015deep, ho2020denoising} generate data by learning to reverse a forward stochastic process that incrementally transforms data into Gaussian noise. A pivotal unification~\cite{song2020score} of score matching~\cite{song2019generative} and probabilistic modeling via Stochastic Differential Equations (SDEs) reveals the underlying continuous-time trajectory of these models, while~\citet{karras2022elucidating} further deconstruct the design space and enhances performance stability. Concurrently, Flow-Matching frameworks~\cite{lipman2022flow,albergo2022building} circumvent the structural constraints of traditional flow models by directly regressing vector fields onto probability paths, significantly improving inference efficiency through path linearization~\cite{liu2022flow}. Subsequent studies~\cite{ma2024sit,albergo2023stochastic} have further bridged the gap between these two paradigms through the stochastic interpolant framework, proving their equivalence in marginal distribution evolution and demonstrating that flow-matching models can also incorporate SDE sampling. 

\subsection{Alignment with Human Preferences of T2I}
Inspired by the success of RLHF in NLP, recent studies~\cite{black2023training, fan2023reinforcement, liu2025flow, xue2025dancegrpo} have begun exploring human preference alignment for generative models. Direct Preference Optimization (DPO)~\cite{rafailov2023direct} eliminates the need for explicit reward models, enabling direct learning from offline preference pairs. Diffusion-DPO~\cite{wallace2024diffusion} reformulates the likelihood function using the evidence lower bound (ELBO), enabling diffusion models to learn directly from the standard denoising loss. Diffusion-KTO~\cite{li2024aligning} draws from prospect theory to model alignment as the maximization of human utility, allowing for the use of simpler binary feedback. DSPO~\cite{zhu2025dspo} identifies that the direct adaptation of DPO to diffusion leads to specific estimation errors and proposes a score-matching objective to maintain consistency with the pretraining phase. Meanwhile, DMPO~\cite{li2025divergence} utilizes reverse KL divergence to avoid the "mean-seeking" trap and aligns more closely with the original RL objective. To enhance flexibility and stability, MaPO~\cite{hong2024margin} introduces a reference-free, margin-aware optimization scheme that addresses the distribution mismatch between reference models and preference data.
Without exception, existing studies focus exclusively on diffusion models, leaving the flow-matching paradigm undiscussed.
\section{Preliminaries}
\subsection{Direct Preference Optimization}

The objective of Reinforcement Learning from Human Feedback (RLHF) is to maximize the expected reward score while constraining the distribution of the policy model $p_\theta$ and the reference model $p_\text{ref}$ using KL divergence \cite{jaques2017sequence, jaques2020human}:
\begin{align}
    \mathcal{J}_\text{RLHF} &= \max_{p_\theta} \mathbb{E}_{c\sim\mathcal{D},x\sim p_\theta(x|c)}[r(x,c)] \notag\\
    &-\beta \mathbb{D}_\text{KL}[p_\theta(x|c)\|p_\text{ref}(x|c)],
\end{align} 
where \(\beta\) is a hyperparameter controlling the constraint strength, \(c \sim \mathcal{D}\) is the input context (prompt) sampled from the dataset $\mathcal{D}$, and \(x \sim p_\theta(\cdot \mid c)\) is the generated output.

Based on the above objective $\mathcal{J}_\text{RLHF}$, Direct Preference Optimization (DPO) \cite{rafailov2023direct} derives an implicit reward function $r^*(x,c) = \beta \log\frac{p_\theta(x|c)}{p_\text{ref}(x|c)}+\beta \log Z(c)$ that eliminates the need for an additional pre-trained reward model, where $Z(c)$ is the partition function.

Using the Bradley-Terry \cite{bradley1952rank} Model, the final loss function of DPO is the negative log-likelihood of the preferred sample $x^w$ over the dispreferred sample $x^l$ based on the implicit reward given the condition $c$:
\begin{align}
    \mathcal{L}_\text{DPO}(\theta) =& -\mathbb{E}_{c,x^w,x^l\sim \mathcal{D}}\log\sigma( \notag \\ 
    &\beta \log\frac{p_{\theta}(x^w|c)}{p_\text{ref}(x^w|c)}- \beta \log\frac{p_\theta(x^l|c)}{p_\text{ref}(x^l|c)}).
\end{align}
\subsection{Diffusion Models and Flow-Matching}
The forward process of diffusion models is defined as:
\begin{equation}
    x_t = \alpha_t x_0 + \sigma_t \epsilon,
\end{equation} 
where $\epsilon \sim \mathcal{N}(0,\mathbf{I})$. The forward process can be formulated as a SDE: $\mathrm{d}x_t=f(t)x_t\,\mathrm{d}t+g(t)\mathrm{d}w_t$ in the continuous-time setting. Following \citet{song2020score}, the corresponding reverse SDE that governs the evolution of the marginal probability $p(x_t)$ is:
\begin{equation}
\mathrm{d}x_t = \left[f(t)x_t - g^2(t)\nabla_{x_t}\log p(x_t)\right]\mathrm{d}t + g(t)\mathrm{d}\bar{w}_t,
\label{eq.revers_sde}
\end{equation}
where \(\mathrm{d}\bar{w}_t\) denotes the reverse Wiener process corresponding to \(\mathrm{d}w_t\), and coefficients $f(t), g(t)$ are derived from the noise schedule $\alpha_t, \sigma_t$.

Parallel to diffusion, flow-matching~\cite{lipman2022flow} models the transformation between image $x_0\sim p_\text{data}$ and noise $x_1\sim p_\text{noise}$ via a deterministic ODE. Specifically, Rectified Flow~\cite{liu2022flow} defines the forward path as a linear interpolation: $x_t = (1-t)x_0 + tx_1$. 
The model predicts a velocity field \(v_\theta(x_t,t)\), which is trained to match the target velocity \(v=x_1-x_0\):
\begin{equation}
    \mathcal{L}_\text{RF}(\theta) = \mathbb{E}_{t,\,x_0\sim p_\text{data},\, x_1\sim p_\text{noise}}[\|v-v_\theta(x_t,t)\|^2_2].
\end{equation}

\subsection{Diffusion-DPO}
The probability of generating the final clean image, \(p_\theta(x_0|c)\), is not tractable, 
Diffusion-DPO~\cite{wallace2024diffusion} uses the whole generation process $x_{0:T}$ to get the implicit reward $r^*(x_{0:T}, c)$, and formulates an approximate objective based on the standard denoising loss:
\begin{align}
L_1(\theta)
&= -\mathbb{E}_{\substack{
    (x^w_0, x^l_0, c) \sim \mathcal{D},\; t \sim \mathcal{U}(0,T),\\
    \scriptscriptstyle x^w_t \sim q(x^w_t \mid x^w_0),\; x^l_t \sim q(x^l_t \mid x^l_0)
}}\log\sigma\Bigl(
        -\beta T \psi(\frac{\alpha_t^2}{\sigma_t^2})\bigl(  \nonumber \\
&\quad \|\epsilon^w - \epsilon_\theta(x_t^w,t,c)\|^2_2
- \|\epsilon^w - \epsilon_\text{ref}(x_t^w,t,c)\|^2_2 \nonumber \\
&- (\|\epsilon^l - \epsilon_\theta(x_t^l,t,c)\|^2_2
- \|\epsilon^l - \epsilon_\text{ref}(x_t^l,t,c)\|^2_2)      
        \bigr)
    \Bigr),
\label{eq.diffdpo}
\end{align}
where $T$ represents the total number of timesteps during training, and $\psi(\frac{\alpha_t^2}{\sigma_t^2})$ is a predefined weighting function~\cite{ho2020denoising}.
\section{Method}
\subsection{The Unified DPO for Diffusion and Flow-Matching}
The key requirement in Diffusion-DPO is explicit access to one-step conditional distributions for sampling
$q(x_{t-1}\mid x_t)$, $p_\theta(x_{t-1}\mid x_t)$, and $p_\text{ref}(x_{t-1}\mid x_t)$,
which are naturally available in diffusion models due to their reverse-time SDE formulation.
In contrast, flow-matching is typically defined by a deterministic ODE: $\mathrm{d}x_t = v_t\,\mathrm{d}t$, which cannot provide such conditional distributions because it lacks the sampling randomness.
To bridge this gap, we reinterpret the flow-matching sampling dynamics through an equivalent velocity-based SDE, enabling the same DPO construction.
A detailed derivation can be found in Appendix~\ref{app.1}.

\textbf{From Velocity Field to a Sampling SDE.}
Although flow-matching is usually defined through a deterministic ODE, the forward dynamics of both diffusion and flow-matching models can be viewed as instances of stochastic interpolation~\cite{albergo2022building,ma2024sit}. From this perspective, we consider the following velocity-based SDE, which shares the same time-marginals $p_t(x)$ as the corresponding reverse-time SDE in Eq.~\ref{eq.revers_sde}:
\begin{equation}
\mathrm{d}x_t = \Bigl(v_t - \frac{g^2(t)}{2}\nabla_x \log p_t(x_t)\Bigr)\,\mathrm{d}t + g(t)\,\mathrm{d}\bar{w}_t.
\label{eq.vel_score_sde}
\end{equation}
For rectified flow, the score function admits an expression in terms of the velocity field:
$\nabla_x \log p_t(x_t) = -\frac{x_t}{t} - \frac{1-t}{t}\,v_t.$
Substituting this into Eq.~\ref{eq.vel_score_sde} yields an equivalent SDE that depends only on the velocity field:
\begin{equation}
\mathrm{d}x_t = \Bigl(v_t + \frac{g^2(t)}{2t}\bigl(x_t+(1-t)v_t\bigr)\Bigr)\,\mathrm{d}t + g(t)\,\mathrm{d}\bar{w}_t.
\label{eq.vel-sde}
\end{equation}
With this formulation, flow-matching admits an SDE representation with the same time-marginals. Discretizing this SDE yields tractable one-step Gaussian conditional distributions and enables stochastic sampling.

\textbf{DPO Objective for Flow-Matching.}
Applying Euler–Maruyama discretization to Eq.~\ref{eq.vel-sde} with $\Delta t=-1$ yields a one-step conditional distribution that induces a Markov transition with a closed-form Gaussian expression:
\begin{align}
    q(x_{t-1}\mid x_t)
    &= \mathcal{N}\!\Bigl(
        x_{t-1};
        \mu(x_t,t),\,
        g^2(t)I
    \Bigr),\nonumber \\
    \mu(x_t,t)
    = x_t &- \bigl[v_t + \frac{g^2(t)}{2t}(x_t+(1-t)v_t)\bigr].
    \label{eq.true-cond-sde}
\end{align}
We can reuse the same DPO derivation as in Diffusion-DPO.
In particular, since both the policy transition $p_\theta(x_{t-1}\mid x_t)$ and the reference transition $p_{\text{ref}}(x_{t-1}\mid x_t)$ share the same covariance $g^2(t)I$,
the KL terms reduce to squared distances between the corresponding means, which further become squared errors on the velocity predictions.
As a result, the DPO loss for flow-matching can be written as:
\begin{align}
    L_2(\theta) 
    & = -\,\mathbb{E}\Bigl[
        \log \sigma\Bigl(
          - \frac{\beta T}{2g^2(t)} (1+\frac{g^2(t)(1-t)}{2t})^2 \bigl(\nonumber \\
    &\quad \|v^w - v_\theta(x_t^w,t,c)\|^2_2
    - \|v^w - v_\text{ref}(x_t^w,t,c)\|^2_2 \nonumber \\
    &- (\|v^l - v_\theta(x_t^l,t,c)\|^2_2
    - \|v^l - v_\text{ref}(x_t^l,t,c)\|^2_2)      
    \bigr)
    \Bigr),
\end{align}
which shares the same expectation as $L_1(\theta)$ in Eq.~\ref{eq.diffdpo}. We also derive the DPO objective of score-matching~\cite{song2019generative} in Appendix~\ref{app.1}.

\textbf{Unified DPO Objective.}
Comparing $L_1$ and $L_2$, both objectives share the same structure:
a logistic loss on the difference between the policy and reference squared errors on a target $y$.
Therefore, the unified DPO objective is
\begin{align}
\mathcal{L}(\theta)
=
&-\mathbb{E}_{\substack{
    (x^w_0, x^l_0, c) \sim \mathcal{D},\; t \sim \mathcal{U}(0,T),\\
    x^w_t \sim q(x^w_t \mid x^w_0),\; x^l_t \sim q(x^l_t \mid x^l_0)
}}\log\sigma\Bigl(
        -\beta T \lambda(t)\bigl(  \nonumber \\
&\quad \|y^w - y_\theta(x_t^w,t,c)\|^2_2
- \|y^w - y_\text{ref}(x_t^w,t,c)\|^2_2 \nonumber \\
&- (\|y^l - y_\theta(x_t^l,t,c)\|^2_2
- \|y^l - y_\text{ref}(x_t^l,t,c)\|^2_2)      
        \bigr)
    \Bigr),
\end{align}
where \(\lambda(t)\) is a function that only depends on \(t\) (constant in practice~\cite{wallace2024diffusion,ho2020denoising});
\(y\), \(y_\theta\), and \(y_{\text{ref}}\) are the prediction target, policy model output, and reference model output, respectively.

\subsection{Linear Diffusion and Flow-Matching DPO}
\label{sec.linear}
\begin{figure}[t]
    \centering
    \includegraphics[width=0.9\linewidth]{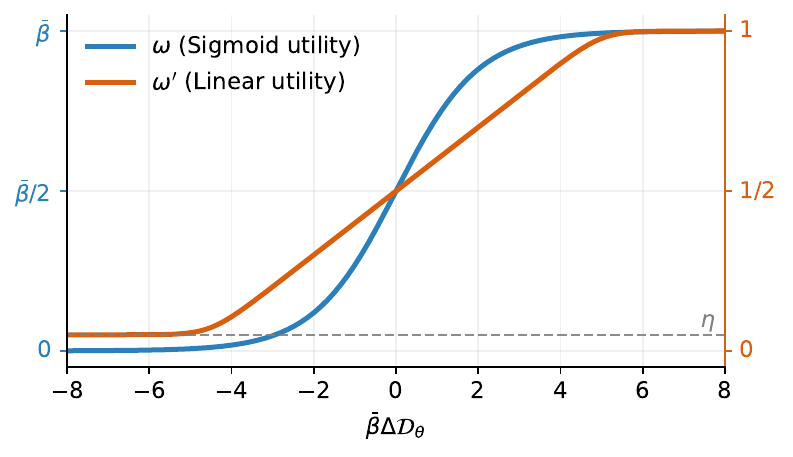}
    \vspace{0.1em}
    \begin{minipage}{0.50\linewidth}
        \centering
        \includegraphics[width=\linewidth]{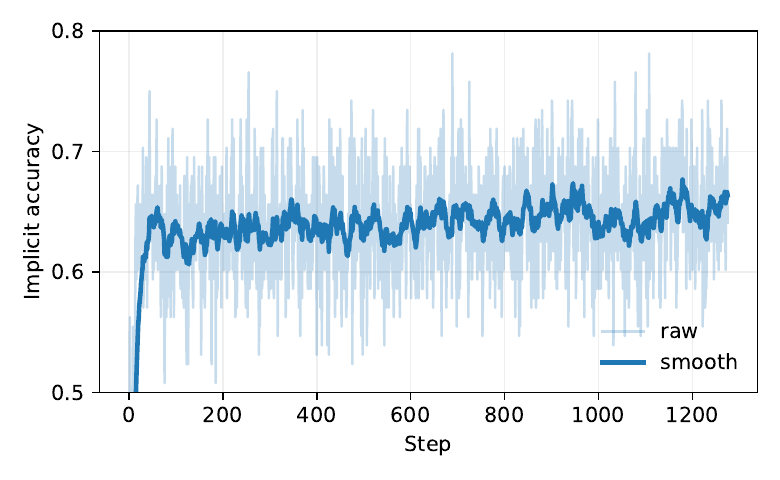}\\
        \vspace{-0.2em}
        {\footnotesize \textit{(a) Sigmoid utility}}
    \end{minipage}\hfill
    \begin{minipage}{0.50\linewidth}
        \centering
        \includegraphics[width=\linewidth]{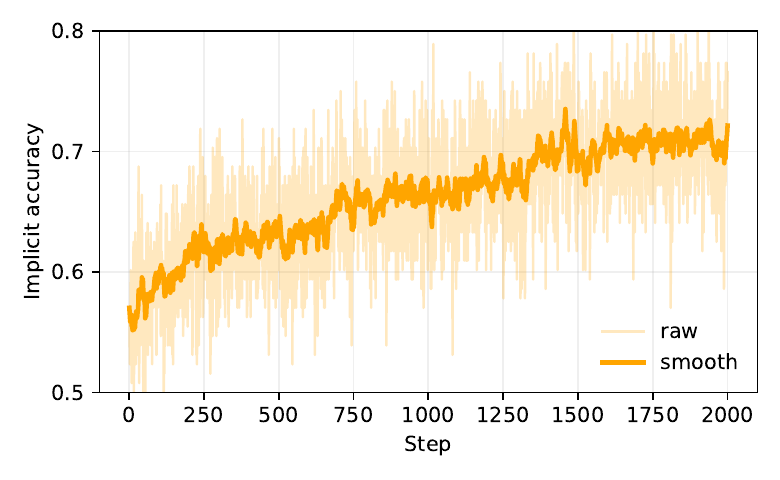}\\
        \vspace{-0.2em}
        {\footnotesize \textit{(b) Linear utility}}
    \end{minipage}
    \caption{Curves of the original sigmoid utility in Diffusion-DPO and our proposed linear utility function (top). (a) and (b) show the implicit accuracy during training with the two utility functions, respectively (Bottom). }
    \label{fig.uitlity_acc}
    \vspace{-0.2in}
\end{figure}

\textbf{Analyze $\mathcal{L}(\theta)$ from a Gradient Perspective.}
We define:
\begin{align}
\mathcal{D}_\theta(x_t, c) &:= \|y - y_\theta(x_t, t, c)\|_2^2 - \|y - y_{\text{ref}}(x_t, t, c)\|_2^2, \nonumber \\
\Delta \mathcal{D}_\theta&:= \mathcal{D}_\theta(x_t^w, c) - \mathcal{D}_\theta(x_t^l, c),\,\,\bar{\beta} := \beta T \lambda(t).
\end{align}
By taking derivatives, the corresponding gradient of $\mathcal{L}(\theta)$ is given as follows (see Appendix~\ref{app.gradient} for details):
\begin{align}
\label{eq:gradient}
\nabla_\theta &\mathcal{L}(\theta) =  \mathbb{E}_{\substack{
    (x^w_0, x^l_0, c) \sim \mathcal{D},\; t \sim \mathcal{U}(0,T),\\
    x^w_t \sim q(x^w_t \mid x^w_0),\; x^l_t \sim q(x^l_t \mid x^l_0)}}\Bigl[
\bar\beta\sigma\bigl(\bar\beta\Delta \mathcal{D}_\theta\bigr) \nonumber \\
&\nabla_\theta (\|y^w-y_\theta(x_t^w,t,c)\|^2_2- \|y^l-y_\theta(x_t^l,t,c)\|^2_2)\Bigr].
\end{align}
Eq. \ref{eq:gradient} shows that the optimization of $\mathcal{L}(\theta)$ can be interpreted as a form of weighted supervised fine-tuning: it performs gradient descent SFT on the winning sample \(x^w\) and gradient ascent SFT on the losing sample \(x^l\). The update step size is dynamically modulated by the weighting function $\omega(\Delta \mathcal{D}_\theta) := \bar{\beta}\,\sigma\bigl(\bar{\beta}\,\Delta \mathcal{D}_\theta\bigr)$ (see Fig.~\ref{fig.uitlity_acc}, top, blue).

Although the sigmoid-based  weighted optimization is validated for the classification/ranking NLP tasks, it does not seamlessly transfer to diffusion/flow-matching generative modeling, where training is fundamentally \(\ell_2\)-regression-based:
\textbf{(1) \(\bar\beta\) is too large compared with typical values in NLP.}  With \(T{=}1000\), $\bar{\beta}$ (e.g. 2500) is far beyond the NLP range, causing highly volatile \(\omega\) and overly large updates. To maintain training stability, existing methods require tiny learning rates (e.g., $10^{-8}$) and gradient clipping. 
\textbf{(2) Coupling between "temperature" and update magnitude complicates hyperparameter tuning.} The same \(\bar\beta\) simultaneously controls the discrimination temperature (how quickly preferences become separated) and the overall gradient scale, creating a direct trade-off and making it difficult to achieve both robustness and effectiveness.
\textbf{(3) Mismatch with the optimization needs of regression-based generative modeling.} The objective of  preference optimization in NLP is to rapidly increase the probability gap or logit margin between chosen and rejected samples, and it is acceptable for gradients decay quickly once preferences are separated. In contrast, training of generative models require long-horizon, broadly supported, and stable small-step updates to continuously refine fine-grained generation quality across many samples. Sigmoid gating tends to become ``hard'' early, leading to apparently fast convergence but insufficient late-stage refinement.

\textbf{From Sigmoid to Linear Utility Function.}
To construct a simple, robust, and efficient objective better suited to generative tasks, we redesign the weighting function \(\omega(\cdot)\) as follows:
\textbf{(1)} We first remove \(\bar{\beta}\) from outside the weighting function and restrict its role to modulating the preference gap. The gradient weight then becomes
\begin{equation}
\omega'(\Delta\mathcal{D}_\theta)=\sigma(\bar{\beta}\Delta\mathcal{D}_\theta).
\end{equation}
This change bounds the range of \(\omega\), eliminating extreme gradients caused by overly large \(\bar{\beta}\). Consequently, we no longer need to adjust the learning rate for different \(\bar{\beta}\), and we can directly reuse the SFT learning rate.
\textbf{(2)} Following Diffusion-KTO~\cite{li2024aligning}, we refer to \(\sigma(\cdot)\) in \(\omega'\)—which maps the loss difference to the overall weight—as the utility function. To mitigate the sharp gradient variation around zero, we replace the sigmoid with a smoother linear utility $u_{\text{linear}}(x)=0.2x+0.5$.
Moreover, we clip the upper bound to 1 and the lower bound to a small constant \(\eta\) to make the model continue to make small but persistent updates after preferences are roughly separated:
\begin{equation}
\omega'(\Delta\mathcal{D}_\theta)=\mathrm{clip}\bigl(u_{\text{linear}}(\bar{\beta}\Delta\mathcal{D}_\theta),\,\eta,\,1\bigr),
\end{equation}
where \(\operatorname{clip}(x,a,b)\) clips \(x\) to \([a,b]\) (see Figure~\ref{fig.uitlity_acc}, top, red).

We can then derive a loss whose gradient matches the improved update above; we term it \textbf{Linear-DPO}:
\begin{align}
\mathcal{L}_{\text{Linear-DPO}}(\theta) &= \mathbb{E}_{\substack{
    (x^w_0, x^l_0, c) \sim \mathcal{D},\; t \sim \mathcal{U}(0,T),\\
    \scriptscriptstyle x^w_t \sim q(x^w_t \mid x^w_0),\; x^l_t \sim q(x^l_t \mid x^l_0)}}\Bigl[ 
    \mathrm{sg}\,\bigl( \omega'
    (\Delta\mathcal{D}_\theta)\bigr)\nonumber \\
    (\|y^w-y&_\theta(x_t^w,t,c)\|^2_2- \|y^l-y_\theta(x_t^l,t,c)\|^2_2) \Bigr],
\end{align}
where \(\mathrm{sg}\,(x)\) denotes the stop-gradient operation.

\begin{algorithm}[t]
\footnotesize
\caption{Linear-DPO with EMA Reference Update}
\label{alg:ema-ref1}

\begin{minipage}{\linewidth}
\begin{algorithmic}
    \setlength{\baselineskip}{1.1em}
    \STATE \textbf{Input:} Dataset $\mathcal{D}=\{(x_0^w,x_0^l,c)\}$; initial parameters $\theta_0$; learning rate $\alpha$; EMA decay $\gamma$, total training timesteps $T$.
    
    \vspace{0.2em}
    \STATE \textbf{Output:} Optimized policy $\theta$ and EMA reference $\theta_{\mathrm{ref}}$.
    
    \vspace{0.2em}
    \STATE \textbf{Init:} $\theta \leftarrow \theta_0$; $\theta_{\mathrm{ref}} 
    \leftarrow \theta_0$ 
    \STATE \textcolor{commentgrey}{\fontsize{7.5pt}{8pt}// $\triangleright$ \texttt{Initialize policy and reference model equally.}}
    
    \vspace{0.2em}
    \WHILE{\textit{not converged}}
        \STATE $(x_0^w, x_0^l, c) \sim \mathcal{D}$ 
        \STATE \textcolor{commentgrey}{\fontsize{7.8pt}{8pt}// $\triangleright$ \texttt{Sample a preference pair and caption.}}
        
        \STATE $t \sim \mathcal{U}(0,T), x_t^w \sim q(x_t^w|x_0^w), x_t^l \sim q(x_t^l|x_0^l)$ 
        \STATE \textcolor{commentgrey}{\fontsize{7.8pt}{8pt}// $\triangleright$ \texttt{Sample timestep and add forward noise.}}
        
        \STATE $\hat{y}^w\leftarrow y_{\theta}(x_t^w,t,c), \hat{y}^l\leftarrow y_{\theta}(x_t^l,t,c)$ 
        \STATE \textcolor{commentgrey}{\fontsize{7.8pt}{8pt}// $\triangleright$ \texttt{Policy model predictions.}}
        
        \STATE $\hat{y}_{\mathrm{ref}}^w\leftarrow y_\mathrm{ref}(x_t^w,t,c), \hat{y}_{\mathrm{ref}}^l\leftarrow y_{\mathrm{ref}}(x_t^l,t,c)$ 
        \STATE \textcolor{commentgrey}{\fontsize{7.8pt}{8pt}// $\triangleright$ \texttt{Reference model predictions (no gradient).}}
        
        \STATE $\mathcal{L} \leftarrow \mathcal{L}_{\mathrm{Linear\text{-}DPO}}(\theta, \theta_{\mathrm{ref}}; \hat{y}^w, \hat{y}^l, \hat{y}_{\mathrm{ref}}^w, \hat{y}_{\mathrm{ref}}^l)$ 
        \STATE \textcolor{commentgrey}{\fontsize{7.8pt}{8pt}// $\triangleright$ \texttt{Compute final loss as per Eq.(10).}}
        
        \STATE $\theta \leftarrow \theta - \alpha \nabla_{\theta} \mathcal{L}$ 
        \STATE \textcolor{commentgrey}{\fontsize{7.8pt}{8pt}// $\triangleright$ \texttt{Update policy via gradient descent.}}
        
        \STATE $\theta_{\mathrm{ref}} \leftarrow \gamma \theta_{\mathrm{ref}} + (1-\gamma) \theta$ 
        \STATE \textcolor{commentgrey}{\fontsize{7.8pt}{8pt}// $\triangleright$ \texttt{Smoothly update reference via EMA.}}
    \ENDWHILE
\end{algorithmic}
\end{minipage}%
\end{algorithm}

\fontfamily{ptm}\selectfont
\begin{figure*}[t]
  \centering
  \begin{minipage}{0.88\textwidth}
  \centering
  \begin{minipage}{0.20\textwidth}
  \begin{minipage}{0.90\textwidth}
    \centering \small \textbf{Prompt}
  \end{minipage}
  \end{minipage}%
  \begin{minipage}{0.133\textwidth} \centering \small \textbf{SD1.5} \end{minipage}%
  \begin{minipage}{0.133\textwidth} \centering \small \textbf{SFT} \end{minipage}%
  \begin{minipage}{0.133\textwidth} \centering \small \textbf{Diff-DPO} \end{minipage}%
  \begin{minipage}{0.133\textwidth} \centering \small \textbf{Diff-KTO} \end{minipage}%
  \begin{minipage}{0.133\textwidth} \centering \small \textbf{DSPO} \end{minipage}%
  \begin{minipage}{0.133\textwidth} \centering \small \textbf{Linear-DPO} \end{minipage}
  
  \vspace{0.2em}

  \begin{minipage}{0.20\textwidth}
  \begin{minipage}{0.90\textwidth}
    \centering \scriptsize \raggedright {\textbf{A lemon wearing a suit and tie, full body portrait.}}
  \end{minipage}
  \end{minipage}%
  \begin{minipage}{0.133\textwidth}\includegraphics[width=\linewidth]{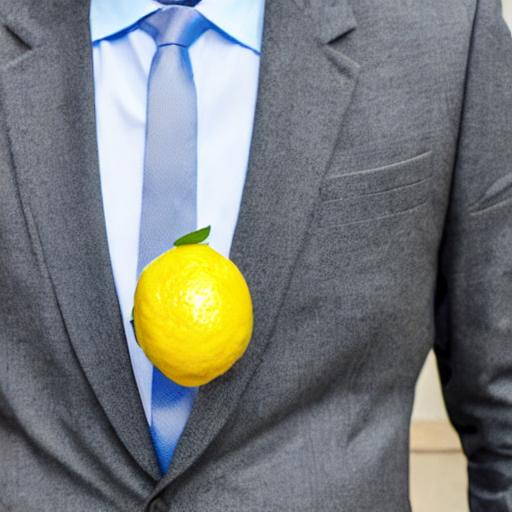}\end{minipage}%
  \begin{minipage}{0.133\textwidth}\includegraphics[width=\linewidth]{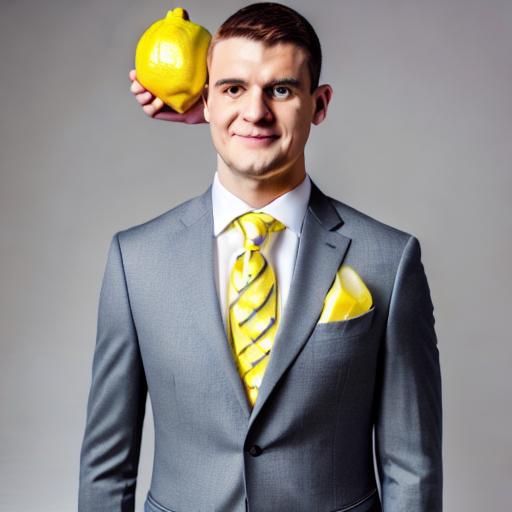}\end{minipage}%
  \begin{minipage}{0.133\textwidth}\includegraphics[width=\linewidth]{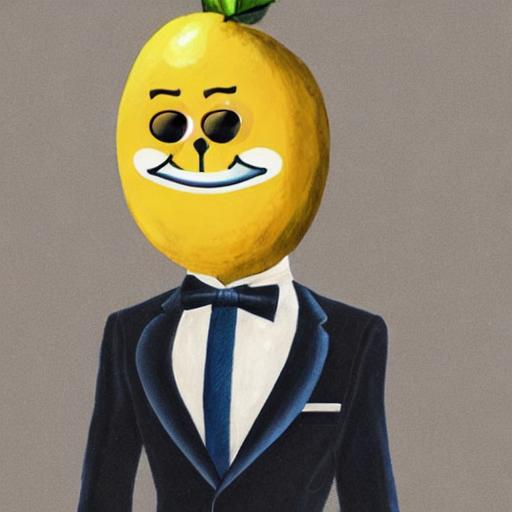}\end{minipage}%
  \begin{minipage}{0.133\textwidth}\includegraphics[width=\linewidth]{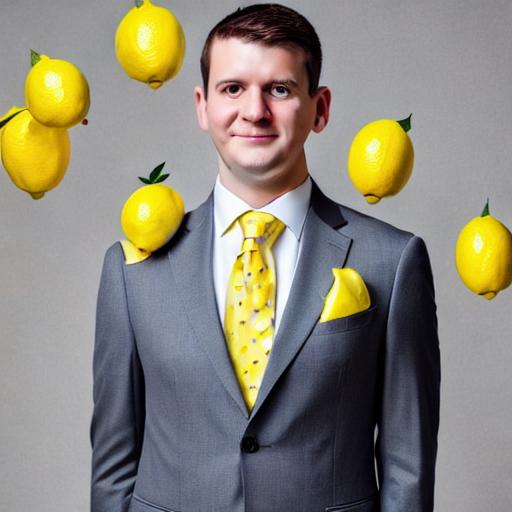}\end{minipage}%
  \begin{minipage}{0.133\textwidth}\includegraphics[width=\linewidth]{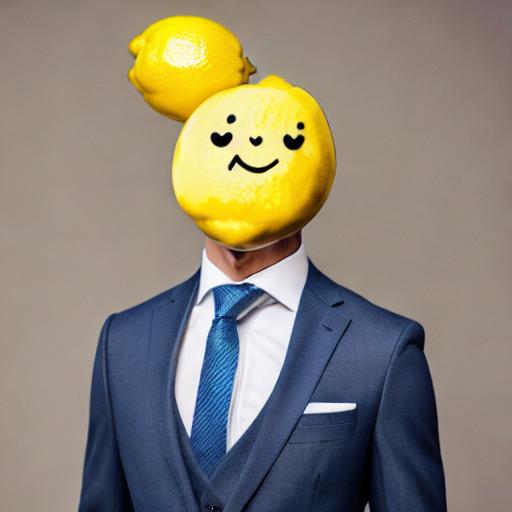}\end{minipage}%
  \begin{minipage}{0.133\textwidth}\includegraphics[width=\linewidth]{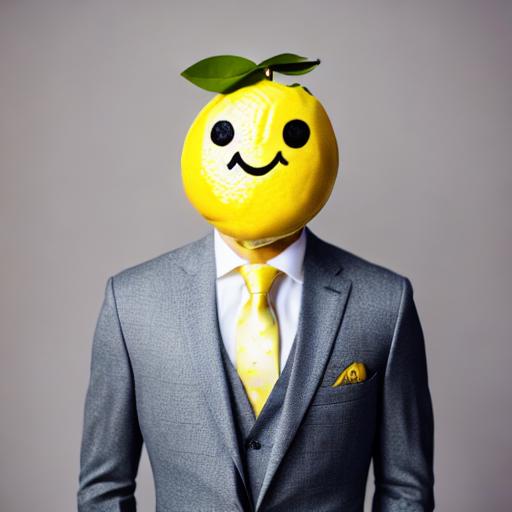}\end{minipage}
  
  \begin{minipage}{0.20\textwidth}
  \begin{minipage}{0.90\textwidth}
    \centering \scriptsize \raggedright {\textbf{A hybrid creature concept painting of a zebra-striped unicorn with bunny ears and a colorful mane.}}
  \end{minipage}
  \end{minipage}%
  \begin{minipage}{0.133\textwidth}\includegraphics[width=\linewidth]{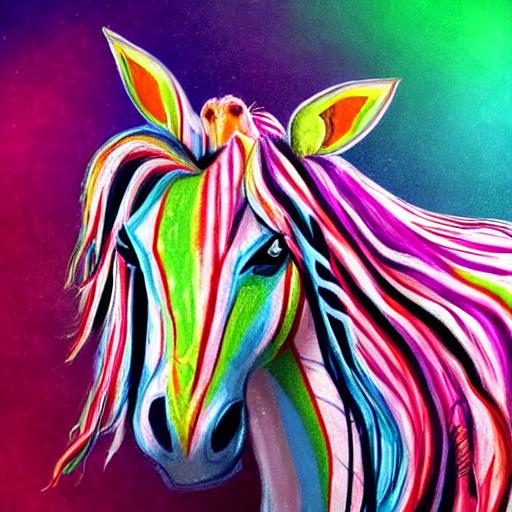}\end{minipage}%
  \begin{minipage}{0.133\textwidth}\includegraphics[width=\linewidth]{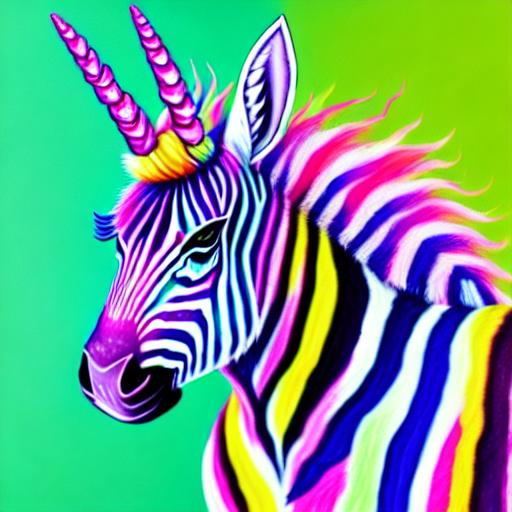}\end{minipage}%
  \begin{minipage}{0.133\textwidth}\includegraphics[width=\linewidth]{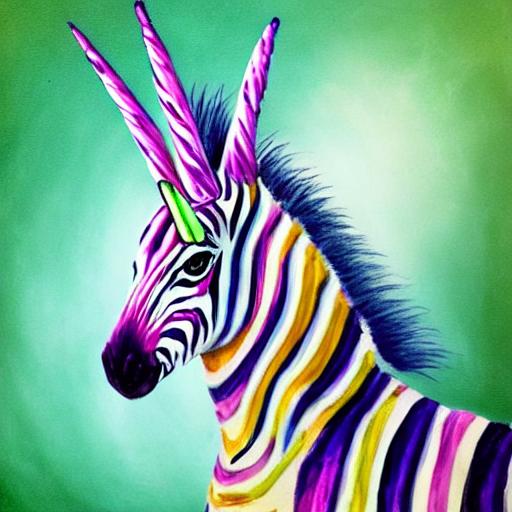}\end{minipage}%
  \begin{minipage}{0.133\textwidth}\includegraphics[width=\linewidth]{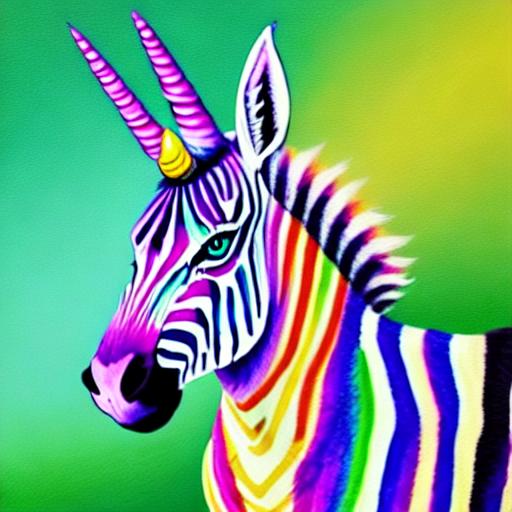}\end{minipage}%
  \begin{minipage}{0.133\textwidth}\includegraphics[width=\linewidth]{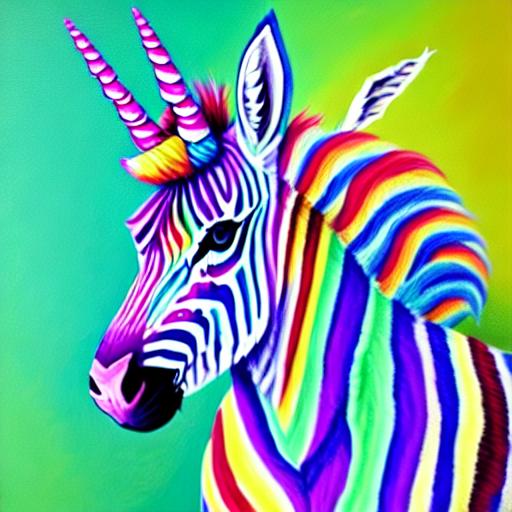}\end{minipage}%
  \begin{minipage}{0.133\textwidth}\includegraphics[width=\linewidth]{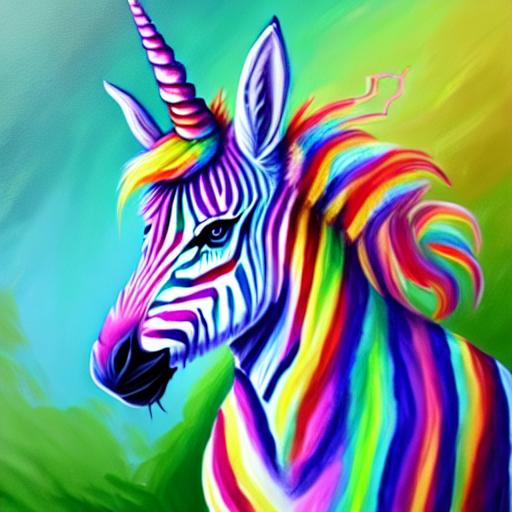}\end{minipage}

  \begin{minipage}{0.20\textwidth}
  \begin{minipage}{0.90\textwidth}
    \centering \scriptsize \raggedright {\textbf{A painting depicting a snowy winter scene featuring a river, a small house on a hill, and a dreamy cloudy sky.}}
  \end{minipage}
  \end{minipage}%
  \begin{minipage}{0.133\textwidth}\includegraphics[width=\linewidth]{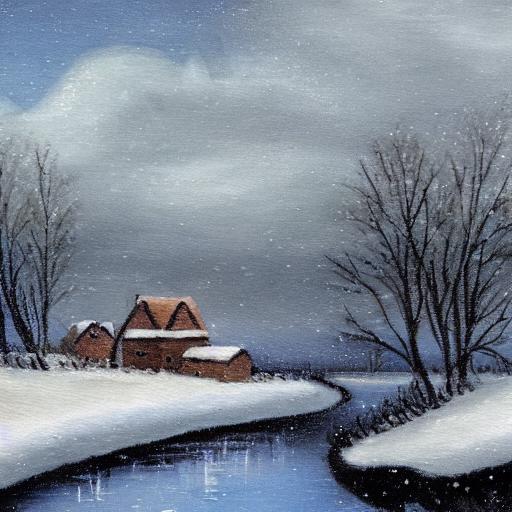}\end{minipage}%
  \begin{minipage}{0.133\textwidth}\includegraphics[width=\linewidth]{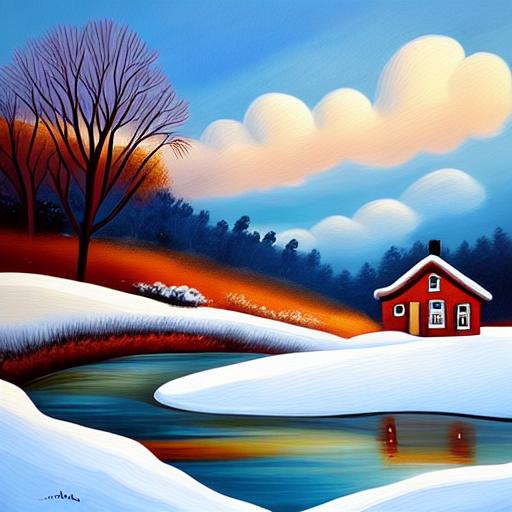}\end{minipage}%
  \begin{minipage}{0.133\textwidth}\includegraphics[width=\linewidth]{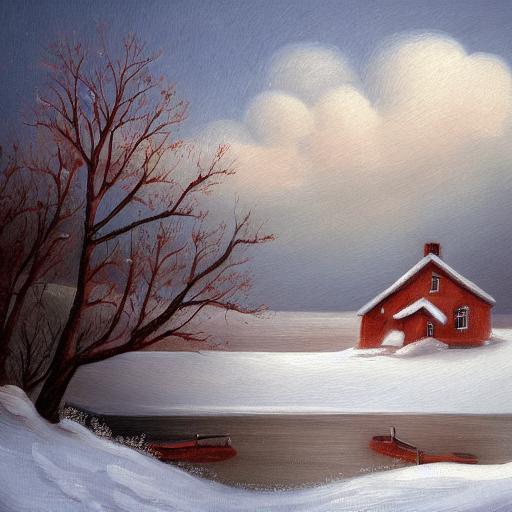}\end{minipage}%
  \begin{minipage}{0.133\textwidth}\includegraphics[width=\linewidth]{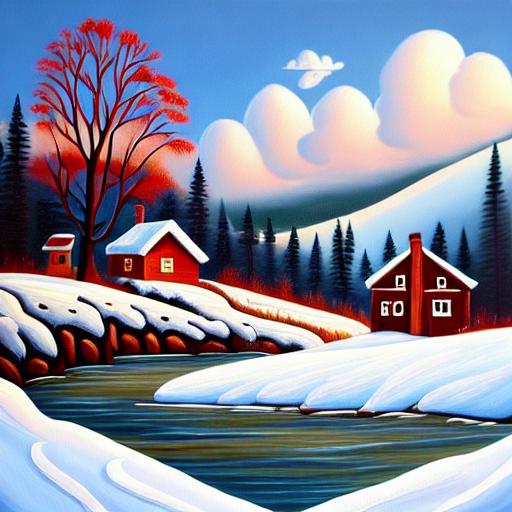}\end{minipage}%
  \begin{minipage}{0.133\textwidth}\includegraphics[width=\linewidth]{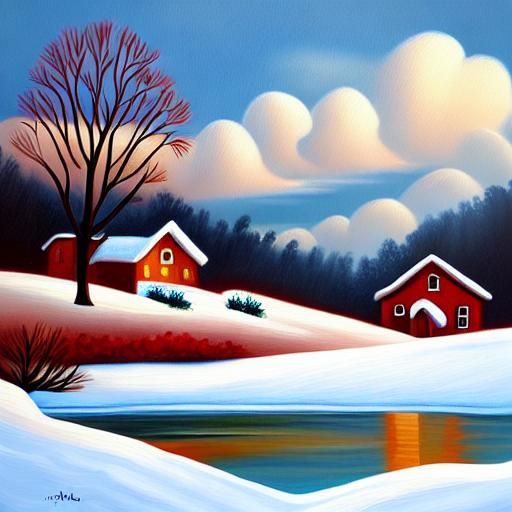}\end{minipage}%
  \begin{minipage}{0.133\textwidth}\includegraphics[width=\linewidth]{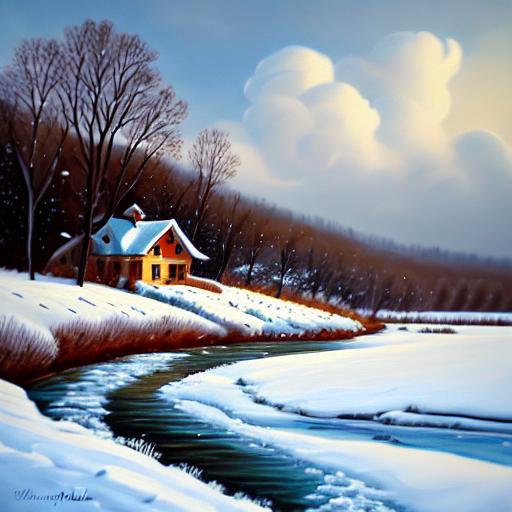}\end{minipage}

  \begin{minipage}{0.20\textwidth}
  \begin{minipage}{0.90\textwidth}
    \centering \scriptsize \raggedright {\textbf{Geometric, colorful creature painted with rough brushstrokes on an abstract background by Pavel Lizano (2018).}}
  \end{minipage}
  \end{minipage}%
  \begin{minipage}{0.133\textwidth}\includegraphics[width=\linewidth]{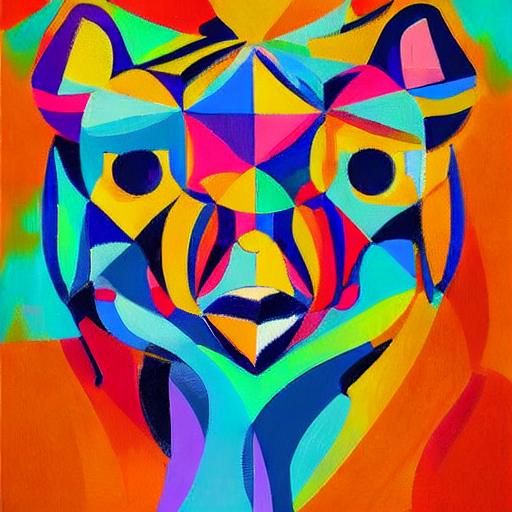}\end{minipage}%
  \begin{minipage}{0.133\textwidth}\includegraphics[width=\linewidth]{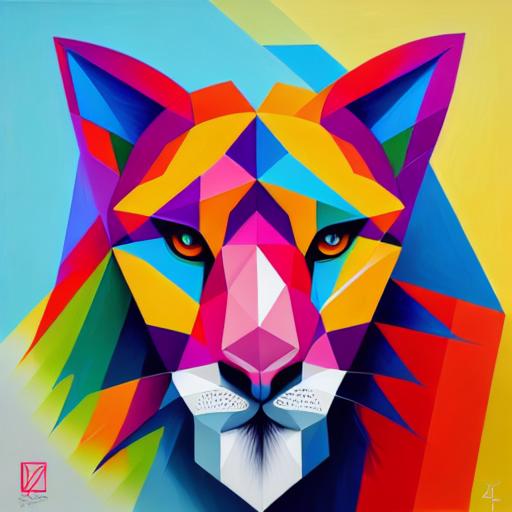}\end{minipage}%
  \begin{minipage}{0.133\textwidth}\includegraphics[width=\linewidth]{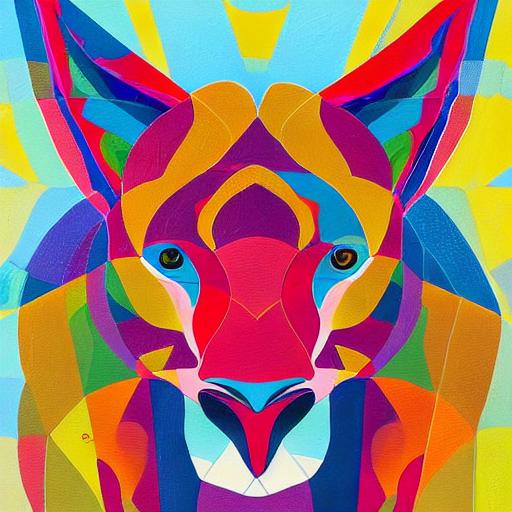}\end{minipage}%
  \begin{minipage}{0.133\textwidth}\includegraphics[width=\linewidth]{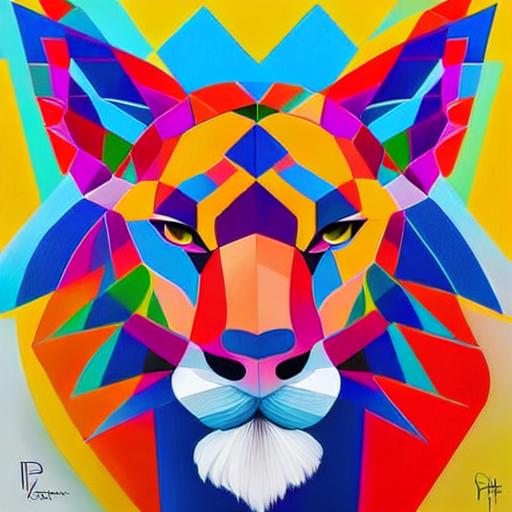}\end{minipage}%
  \begin{minipage}{0.133\textwidth}\includegraphics[width=\linewidth]{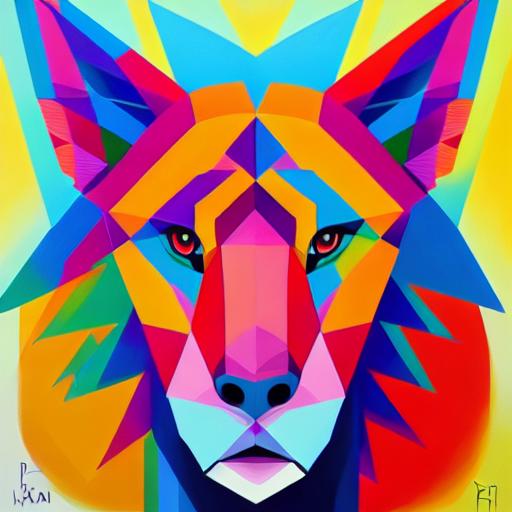}\end{minipage}%
  \begin{minipage}{0.133\textwidth}\includegraphics[width=\linewidth]{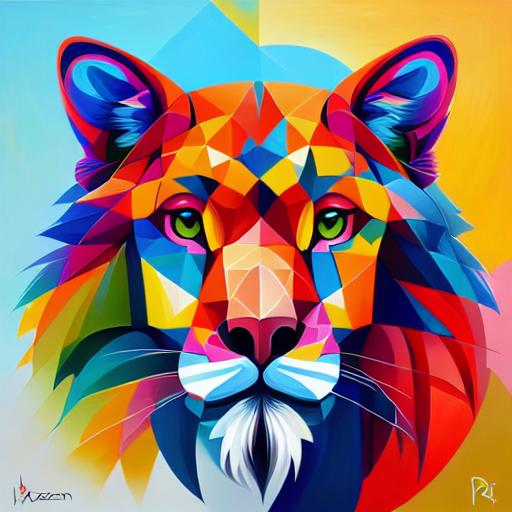}\end{minipage}
  
  \end{minipage}
  \caption{Qualitative comparison on SD1.5 model of images generated by each methods using the same prompt are presented. The proposed Linear-DPO substantially improves the visual quality and text–image alignment compared to other methods.}
  \label{fig.sd15}
  \vspace{-0.15in}
\end{figure*}

\textbf{EMA Reference Model.}
In RLHF, a frozen base model is commonly used as a reference, together with a KL penalty to prevent the policy from drifting too far. DPO for generative models typically follows the same practice. However, once the policy substantially outperforms the reference, a fixed reference can hinder further improvement.
In practice, the reference is often either updated abruptly (e.g., periodically replaced with the latest policy) or dropped entirely with a constant reference~\cite{meng2024simpo}, which may destabilize training.
In Linear-DPO, we instead update the reference smoothly by maintaining an exponential moving average (EMA) of the policy throughout training, as summarized in Algorithm~\ref{alg:ema-ref1}.

\section{Experiments}
\subsection{Experimental Setup}
\textbf{Models and Datasets.}
Following previous work, we use Stable Diffusion v1.5 (SD1.5)~\cite{rombach2022high}
and Stable Diffusion XL-1.0 (SDXL)~\cite{podell2023sdxl} as our diffusion models.
For flow-matching, we select Stable Diffusion 3-Medium (SD3-M)~\cite{esser2024scaling},
which adopts the MMDiT architecture~\cite{esser2024scaling} and is trained with rectified flow~\cite{liu2022flow}.

Pick-a-Pic v2~\cite{kirstain2023pick} contains 851,293 human-preference image pairs generated by early foundation models such as SD1.5 and SDXL, and it is the standard dataset used in previous works~\cite{wallace2024diffusion, li2024aligning}. However, its relatively low resolution and limited visual quality may underutilize more capable models such as SD3-M. Therefore, for flow-matching models, we use a high-quality subset of HPDv3~\cite{ma2025hpsv3}, containing images generated by state-of-the-art models such as SD3-M and FLUX.1-dev~\cite{flux2024}. Further details are available in Appendix~\ref{app.hpsv3}.

\textbf{Baselines.}
For diffusion models (SD1.5 and SDXL), we compare Linear-DPO against the original model, SFT, Diffusion-DPO~\cite{wallace2024diffusion}, Diffusion-KTO~\cite{li2024aligning}, DSPO~\cite{zhu2025dspo}, MaPO~\cite{hong2024margin}, and DMPO~\cite{li2025divergence}.
Since no preference-optimization method is designed for flow-matching, for SD3-M we compare only against the original model, SFT, and Diffusion-DPO.
Implementation details are available in Appendix~\ref{app.imple}.

\textbf{Evaluation.}
For evaluation prompts, we use prompts from PartiPrompt~\cite{yu2022scaling} as well as test prompts from Pick-a-Pic v2~\cite{kirstain2023pick} and HPDv2~\cite{wu2023human}, which contain 1,632, 500, and 400 prompts, respectively.
For quantitative results, we report the average scores under  PickScore~\cite{kirstain2023pick}, HPSv2~\cite{wu2023human}, HPSv3~\cite{ma2025hpsv3}, LAION Aesthetics Score~\cite{Schuhmann2023LAION}, CLIP~\cite{radford2021learning}, and Image Reward~\cite{xu2023imagereward} over images generated by each baseline.

\subsection{Main Results}
\label{sec.results}
\begin{table*}[thbp]
\caption{Quantitative results of reward scores for various methods on SD1.5 across three validation datasets. $*$: We retrain DSPO using the official code as its weights are unavailable. $\dagger$: For DMPO, we adopt the data reported in its original paper since neither the code nor the weights are accessible. Bold indicates the best performance, and underlining indicates our method is the second-best.}
\vspace{-1.0em}
\label{tab.sd15}
\begin{center}
\small
\begin{tabular}{l|l|cccccc}
\toprule
\textbf{Dataset} & \textbf{Method} & \textbf{PickScore} & \textbf{HPSv2}& \textbf{Aesthetics} &  \textbf{CLIP} & \textbf{Image Reward} & \textbf{HPSv3} \\
\midrule
\multirow{7}{*}{Pick-a-Pic v2} & SD1.5 & 0.2075 &0.2677&5.4461&0.3325&0.1009&3.7102\\
& SFT&0.2137&0.2778&5.7147&0.3430&0.6349&4.7256\\
& Diffusion-DPO & 0.2119& 0.2723& 5.5564& 0.3411& 0.3070&4.3751\\
& Diffusion-KTO & 0.2136& 0.2780& 5.6744& 0.3451& 0.6675&4.8341\\
& $\text{DSPO}^*$& 0.2138& 0.2784& 5.7007& 0.3442& 0.6942& 4.8768 \\
& $\text{DMPO}^\dagger$ & 0.2165& 0.2705& 5.6304& 0.3453& 0.5412&-\\
\rowcolor{gray!15}\cellcolor{white}& \textbf{Linear-DPO} & \textbf{0.2177} & \textbf{0.2806} & \textbf{5.7875}& \textbf{0.3489}&\textbf{0.8098} & \textbf{5.2523}\\
\midrule
\multirow{7}{*}{PartiPrompt}& SD1.5 & 0.2151&0.2749&5.3638&0.3330&0.2308&3.8698\\
& SFT&0.2185&0.2833&5.6115&0.3406&0.6465&5.3270\\
& Diffusion-DPO & 0.2176& 0.2783& 5.4391& 0.3380& 0.3784& 4.5727\\
& Diffusion-KTO & 0.2181& 0.2826& 5.5678& 0.3396& 0.6422 & 5.2354\\
&$\text{DSPO}^*$&0.2184&0.2831&5.5957&0.3388&0.6438&5.2742 \\
& $\text{DMPO}^\dagger$ & 0.2205& 0.2758& 5.5438& \textbf{0.3483}& 0.6614& -\\
\rowcolor{gray!15}\cellcolor{white}& \textbf{Linear-DPO} & \textbf{0.2209} & \textbf{0.2853} & \textbf{5.6641}&\underline{0.3456}& \textbf{0.7896} &\textbf{5.7795}\\ 
\midrule
\multirow{7}{*}{HPDv2} & SD1.5 &0.2103&0.2716&5.5838&0.3545&0.1040&5.3861 \\
& SFT & 0.2176&0.2838&5.8344&0.3638&0.7356&10.0600\\
& Diffusion-DPO & 0.2144& 0.2754& 5.7042& 0.3600& 0.2946& 7.1220\\
& Diffusion-KTO & 0.2170& 0.2840& 5.8154& 0.3661& 0.7028&10.1747\\
&$\text{DSPO}^*$&0.2175&0.2833&5.8404&0.3646&0.7540&10.0996 \\
& $\text{DMPO}^\dagger$ & 0.2195& 0.2768& 5.7997& 0.3629& 0.6350&- \\
\rowcolor{gray!15}\cellcolor{white}&\textbf{Linear-DPO} & \textbf{0.2213} & \textbf{0.2866} & \textbf{5.9236}&\textbf{0.3688}& \textbf{0.8494} &\textbf{11.0940}
\\
\bottomrule
\end{tabular}
\end{center}
\vskip -0.15in
\end{table*}
\begin{table}[t]
\caption{Quantitative comparison of three representative reward scores on SDXL. DPO is short for Diffusion-DPO.}
\label{tab.sdxl_short}
\vspace{-1.0em}
\begin{center}
\begin{footnotesize}
\begin{tabular}{l|l|ccc}
\toprule
\textbf{Dataset} & \textbf{Method} & \textbf{PickScore} & \textbf{HPSv2} & \textbf{HPSv3} \\
\midrule
\multirow{5}{*}{Pick-a-Pic v2} & SDXL & 0.2229 &0.2805&6.6575\\
& SFT& 0.2171&0.2768&6.0146\\
& DPO & 0.2271&0.2868&7.1570\\
&MaPO& 0.2232& 0.2830&6.8585\\
\rowcolor{gray!15}\cellcolor{white}& \textbf{Ours} & \textbf{0.2283}&\textbf{0.2884}&\textbf{7.1615}\\
\midrule
\multirow{5}{*}{PartiPrompt}& SDXL & 0.2266&0.2839&6.4288\\
& SFT&0.2214&0.2813&5.8542\\
& DPO & 0.2295&0.2893&\textbf{6.9971}\\
&MaPO&0.2268&0.2864&6.6485\\
\rowcolor{gray!15}\cellcolor{white}& \textbf{Ours} &\textbf{0.2297}&\textbf{0.2906}&\underline{6.9953} \\ 
\midrule
\multirow{5}{*}{HPDv2} & SDXL&0.2288&0.2849&10.6086\\
& SFT &0.2219&0.2824&10.2284 \\
& DPO & 0.2325& 0.2907& 11.2931\\
& MaPO& 0.2295& 0.2889&11.1285 \\
\rowcolor{gray!15}\cellcolor{white}&\textbf{Ours} & \textbf{0.2332}&\textbf{0.2924}&\textbf{11.3601}\\
\bottomrule
\end{tabular}
\end{footnotesize}
\end{center}
\vskip -0.15in
\end{table}

\textbf{Quantitative Results.}
As presented in Table~\ref{tab.sd15}, Linear-DPO shows substantial improvements over the original SD1.5 and SFT. Compared to other DPO-based baselines, Linear-DPO also exhibits clear advantages, achieving state-of-the-art results in aesthetics (Aesthetics), text--image alignment (CLIP), and human preference alignment (PickScore, HPSv2/v3, and Image Reward). These results verify that Linear-DPO enables diffusion models to align with human preferences more accurately.

Table~\ref{tab.sdxl_short} presents results on the larger and more powerful SDXL model at a higher resolution of $1024\times1024$. It can be observed that Linear-DPO maintains its superior performance. Furthermore, although the training data contains a large number of images generated by weaker models such as SD1.4/1.5, Linear-DPO still yields significant performance improvements. In contrast, traditional SFT degrades model performance.

\begin{figure}[bthp]
    \centering
    \includegraphics[width=\linewidth]{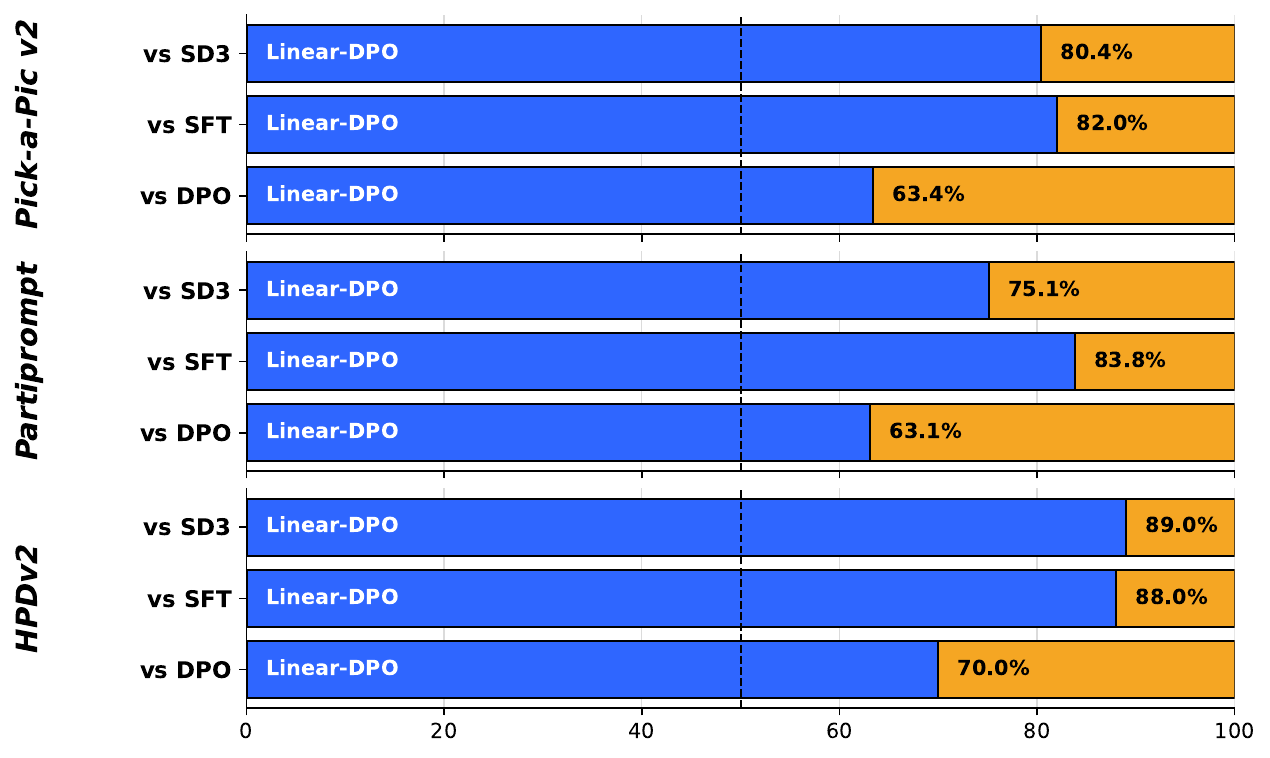}
    \caption{Win ratios of Linear-DPO vs. other methods on SD3-M across three validation datasets, based on automated evaluations using the HPSv3 score.}
    \label{fig.win}
    \vskip -0.15in
\end{figure}

For rectified-flow-based SD3-M, Figure~\ref{fig.win} shows Linear-DPO's competitive win rates across three evaluation datasets, as measured by HPSv3. These results demonstrate that our method generalizes effectively from diffusion to flow-matching models. Detailed scores and additional results are provided in Appendix~\ref{app.quantitative}.

\begin{figure}[t]
  \centering
  \begin{minipage}{0.48\textwidth}
  \centering
  \begin{minipage}{0.28\textwidth}
  \begin{minipage}{0.90\textwidth}
    \centering \scriptsize \textbf{Prompt}
  \end{minipage}
  \end{minipage}%
  \begin{minipage}{0.18\textwidth} \centering \scriptsize \textbf{SD3} \end{minipage}%
  \begin{minipage}{0.18\textwidth} \centering \scriptsize \textbf{SFT} \end{minipage}%
  \begin{minipage}{0.18\textwidth} \centering \scriptsize \textbf{Diff-DPO} \end{minipage}%
  \begin{minipage}{0.18\textwidth} \centering \scriptsize 
  \textbf{Linear-DPO} \end{minipage}%
  
  \begin{minipage}{0.28\textwidth}
  \begin{minipage}{0.90\textwidth}
    \centering \scriptsize \raggedright {Beefy cowboy, tucked in shirt}
  \end{minipage}
  \end{minipage}%
  \begin{minipage}{0.18\textwidth}\includegraphics[width=\linewidth]{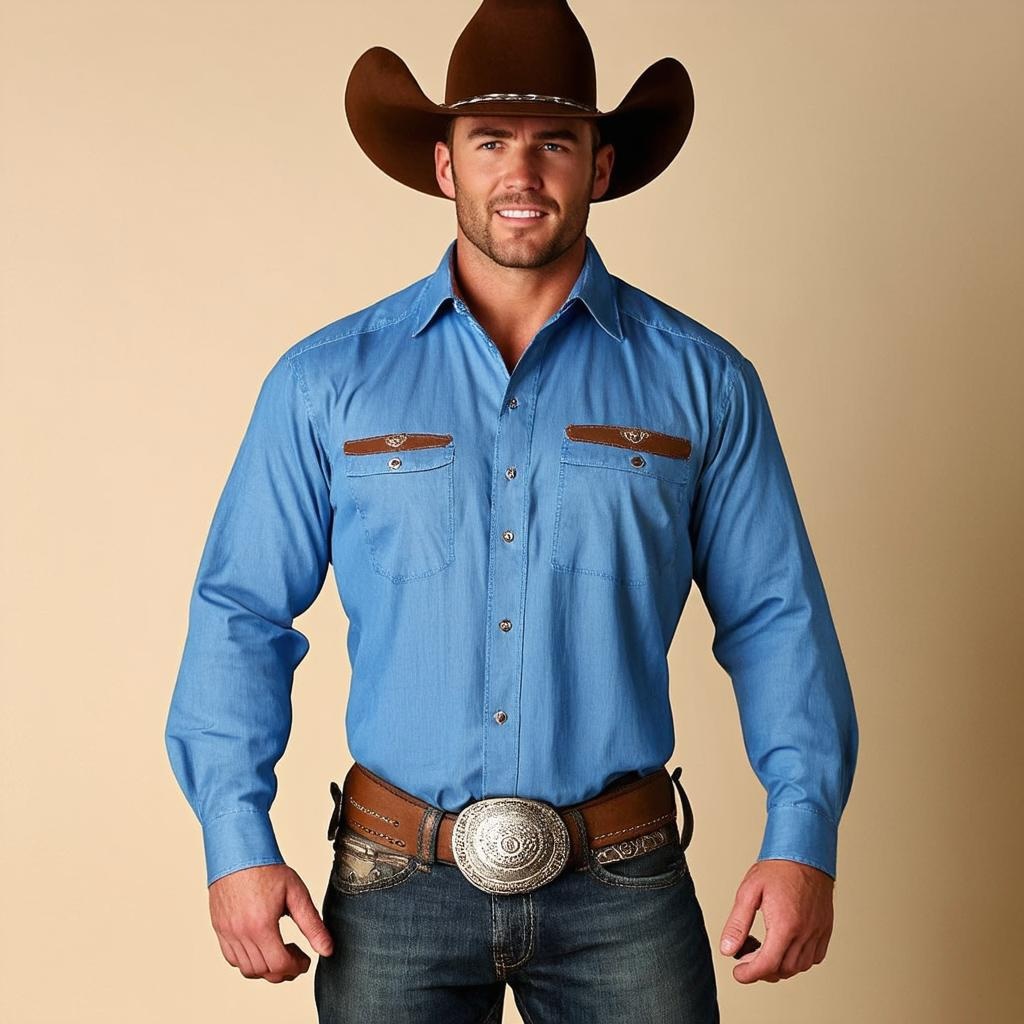}\end{minipage}%
  \begin{minipage}{0.18\textwidth}\includegraphics[width=\linewidth]{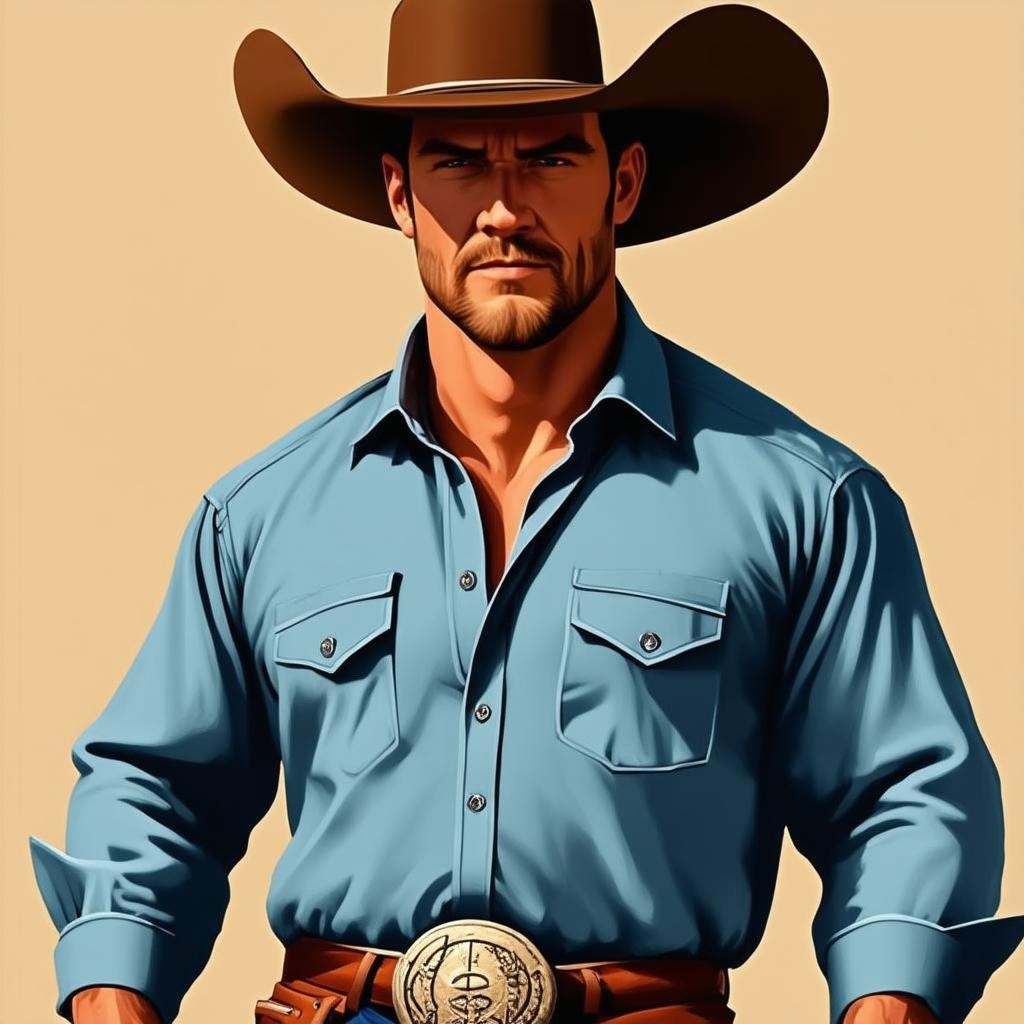}\end{minipage}%
  \begin{minipage}{0.18\textwidth}\includegraphics[width=\linewidth]{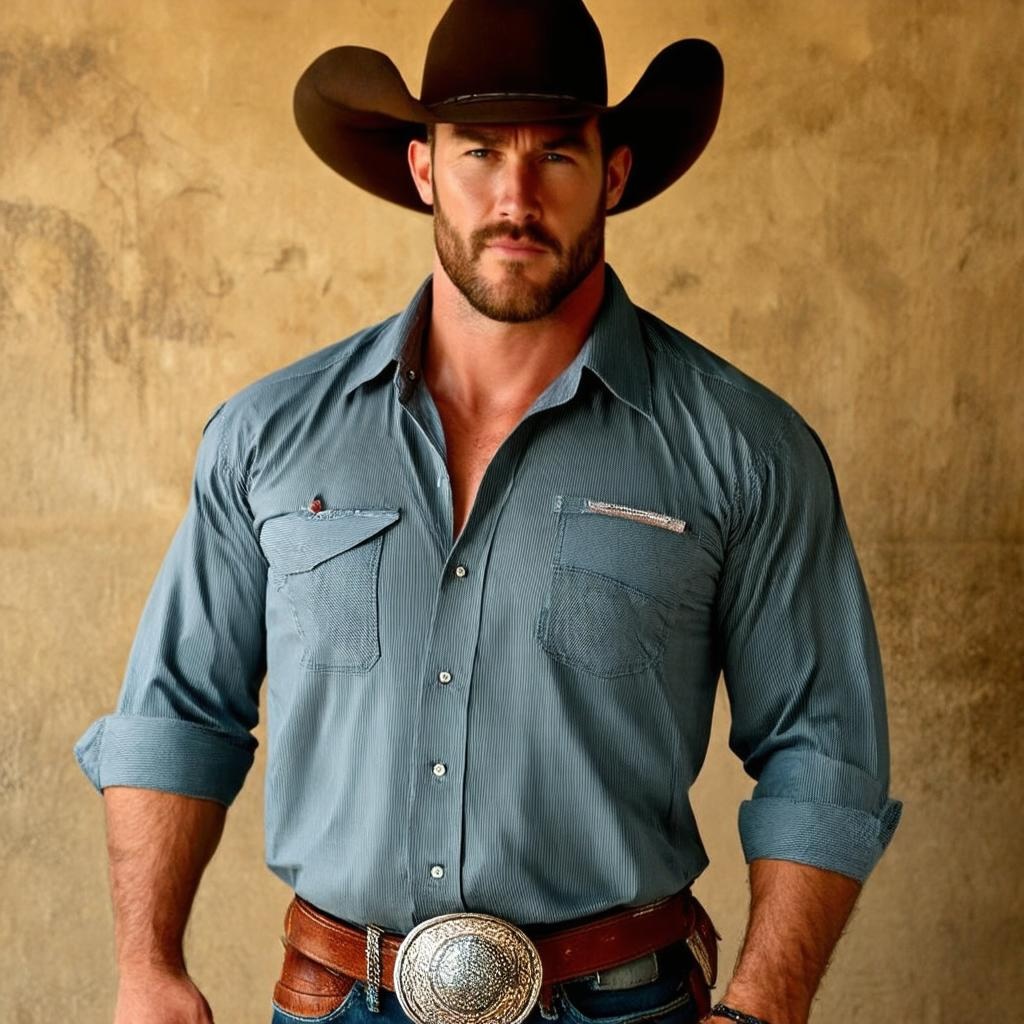}\end{minipage}%
  \begin{minipage}{0.18\textwidth}\includegraphics[width=\linewidth]{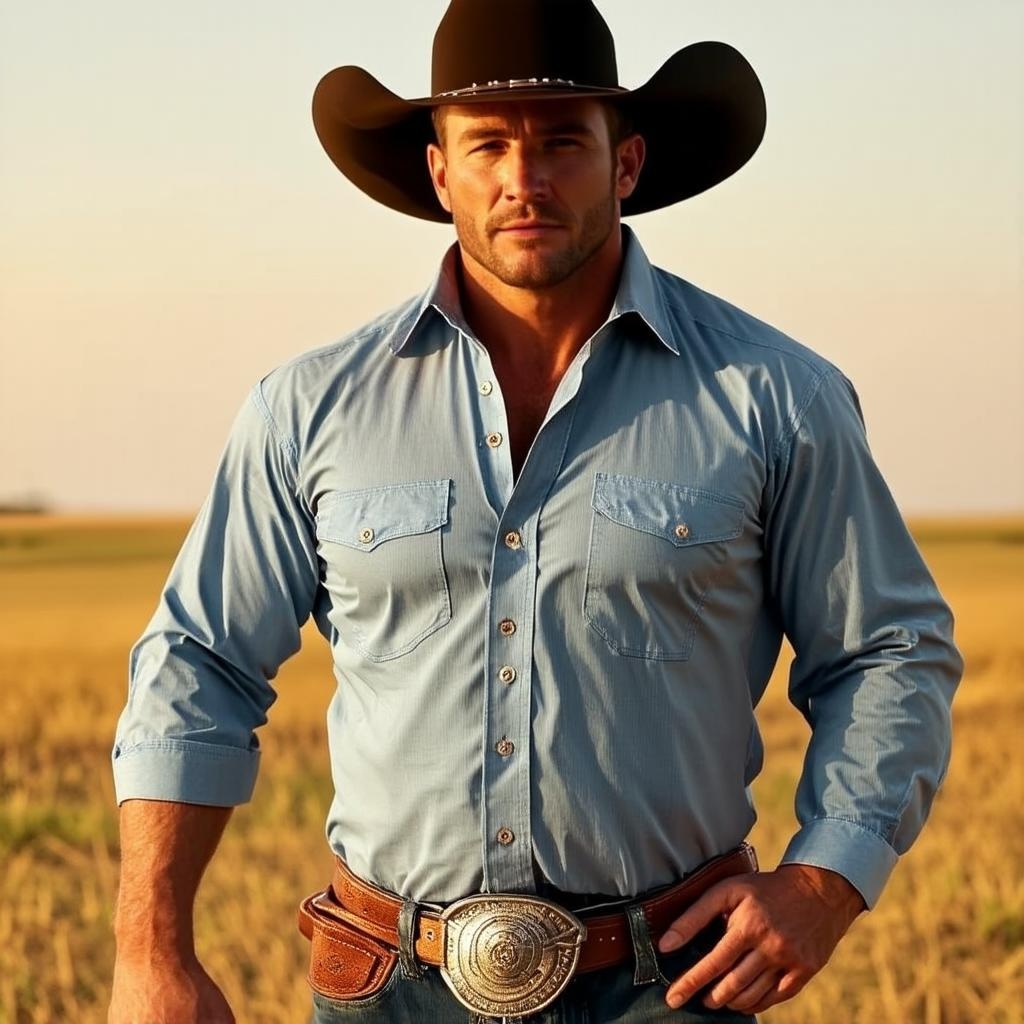}\end{minipage}%

  \begin{minipage}{0.28\textwidth}
  \begin{minipage}{0.90\textwidth}
    \centering \tiny \raggedright {A mushroom on a cloud with ghosts all around it flying and then spikes on the sides}
  \end{minipage}
  \end{minipage}%
  \begin{minipage}
  {0.18\textwidth}\includegraphics[width=\linewidth]{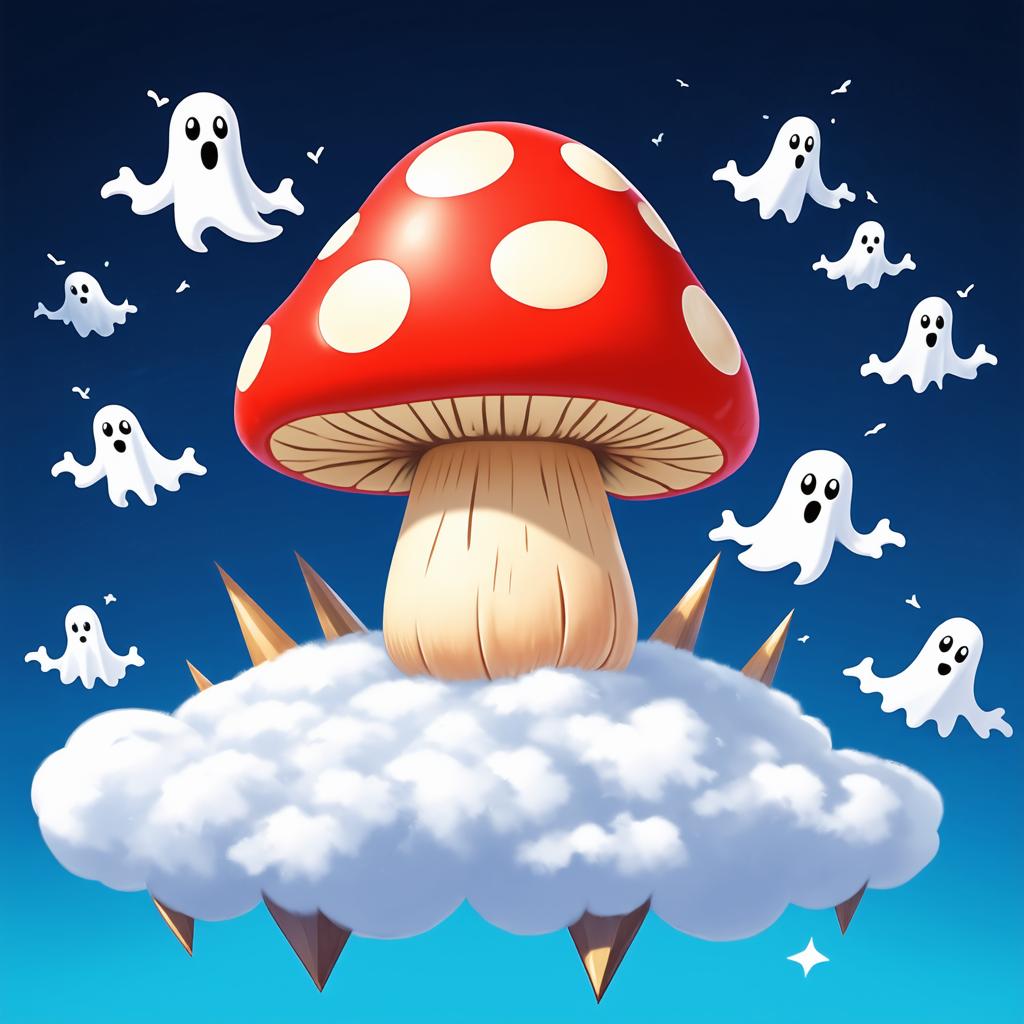}\end{minipage}%
  \begin{minipage}{0.18\textwidth}\includegraphics[width=\linewidth]{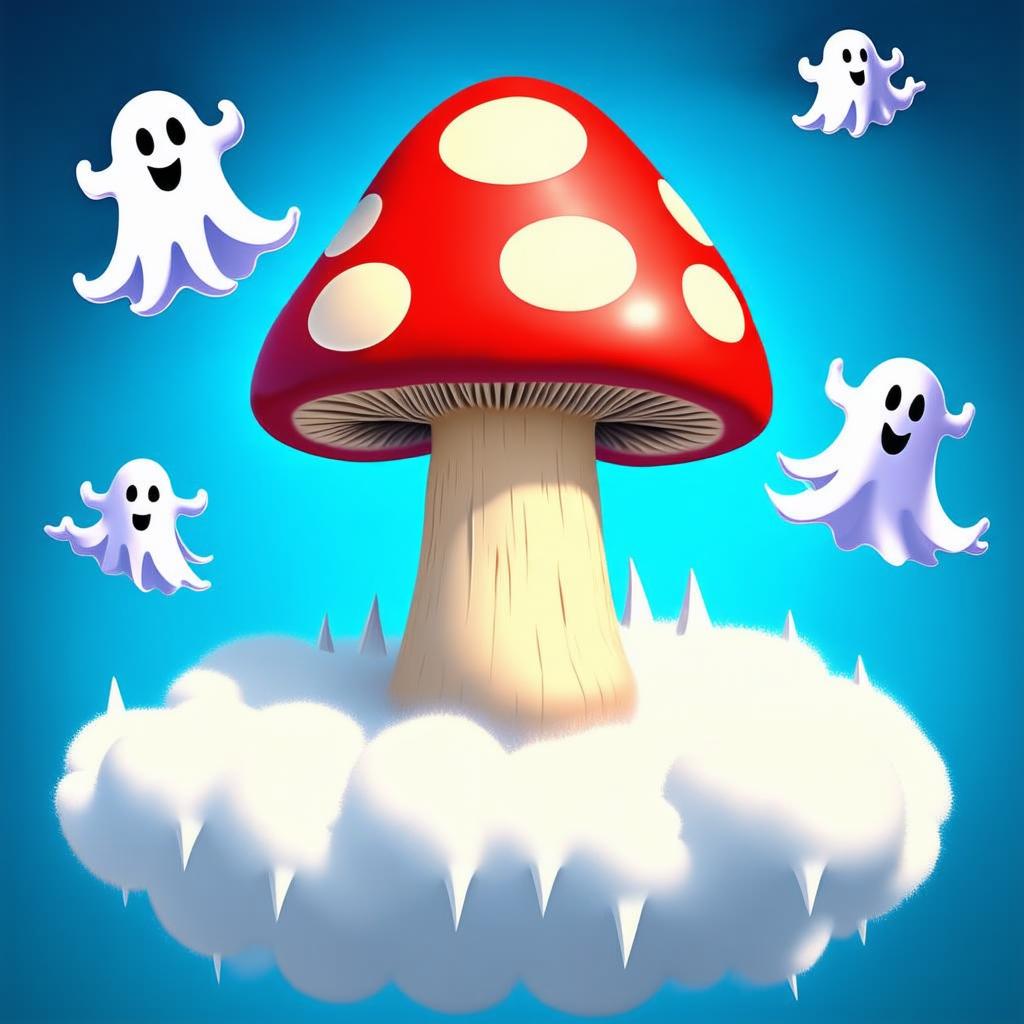}\end{minipage}%
  \begin{minipage}{0.18\textwidth}\includegraphics[width=\linewidth]{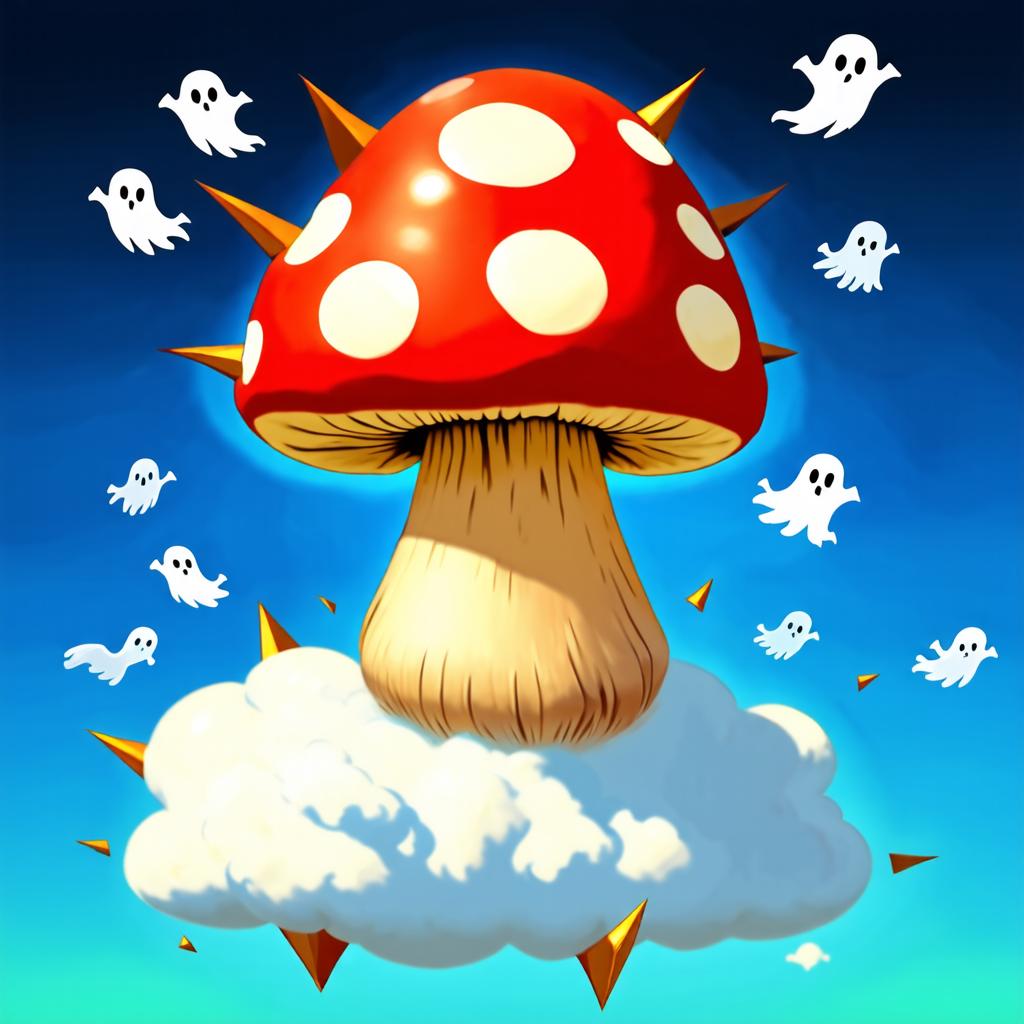}\end{minipage}%
  \begin{minipage}{0.18\textwidth}\includegraphics[width=\linewidth]{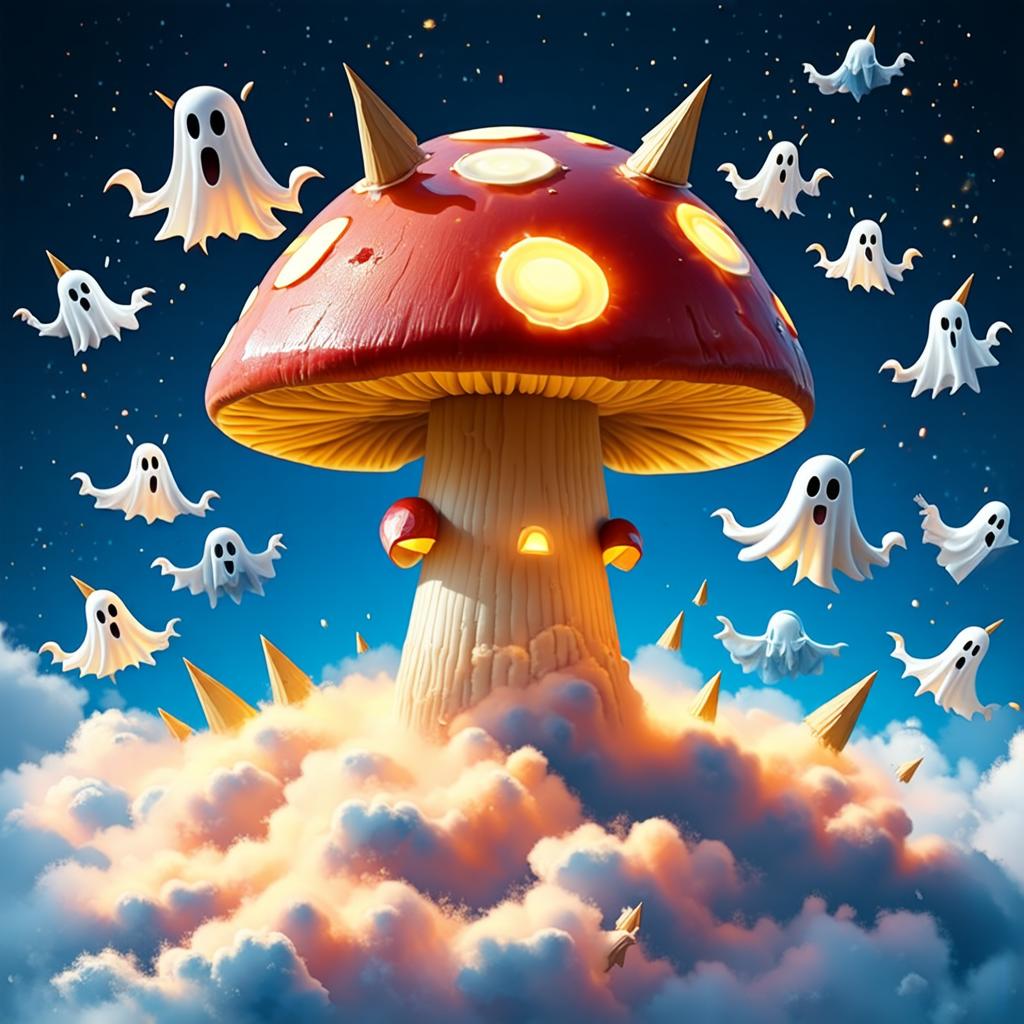}\end{minipage}%

  \begin{minipage}{0.28\textwidth}
  \begin{minipage}{0.90\textwidth}
    \centering \tiny \raggedright {Digital painting of a lush natural scene on an alien planet with colourful, weird vegetation, cliffs, and water by Gerald Brom.}
  \end{minipage}
  \end{minipage}%
  \begin{minipage}{0.18\textwidth}\includegraphics[width=\linewidth]{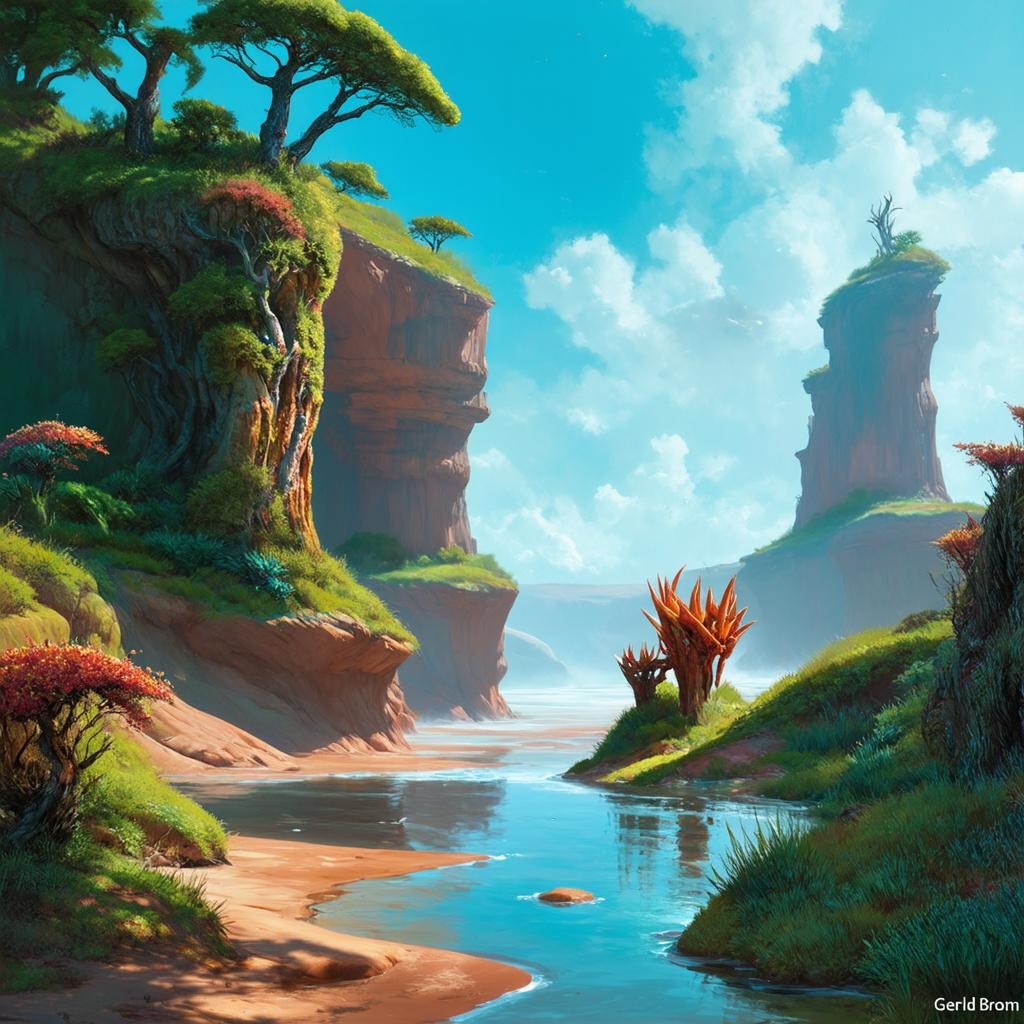}\end{minipage}%
  \begin{minipage}{0.18\textwidth}\includegraphics[width=\linewidth]{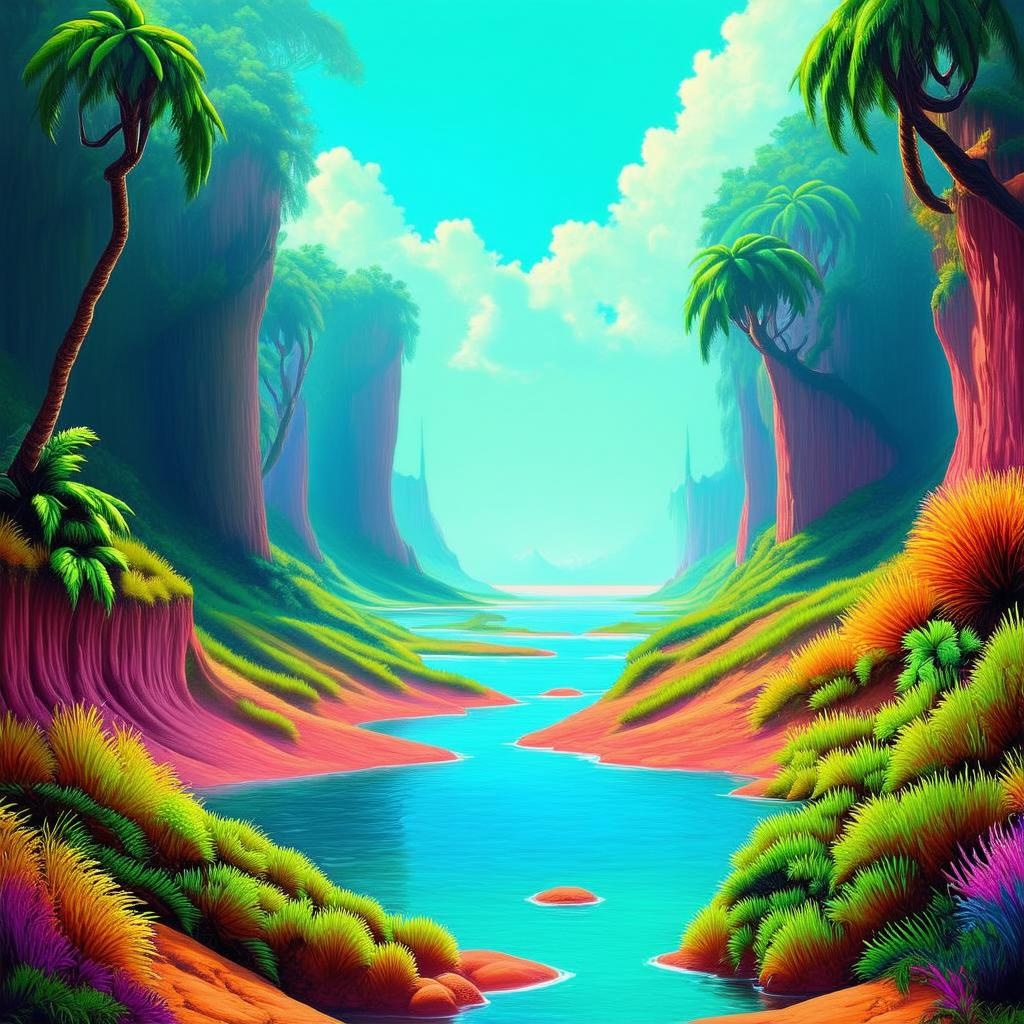}\end{minipage}%
  \begin{minipage}{0.18\textwidth}\includegraphics[width=\linewidth]{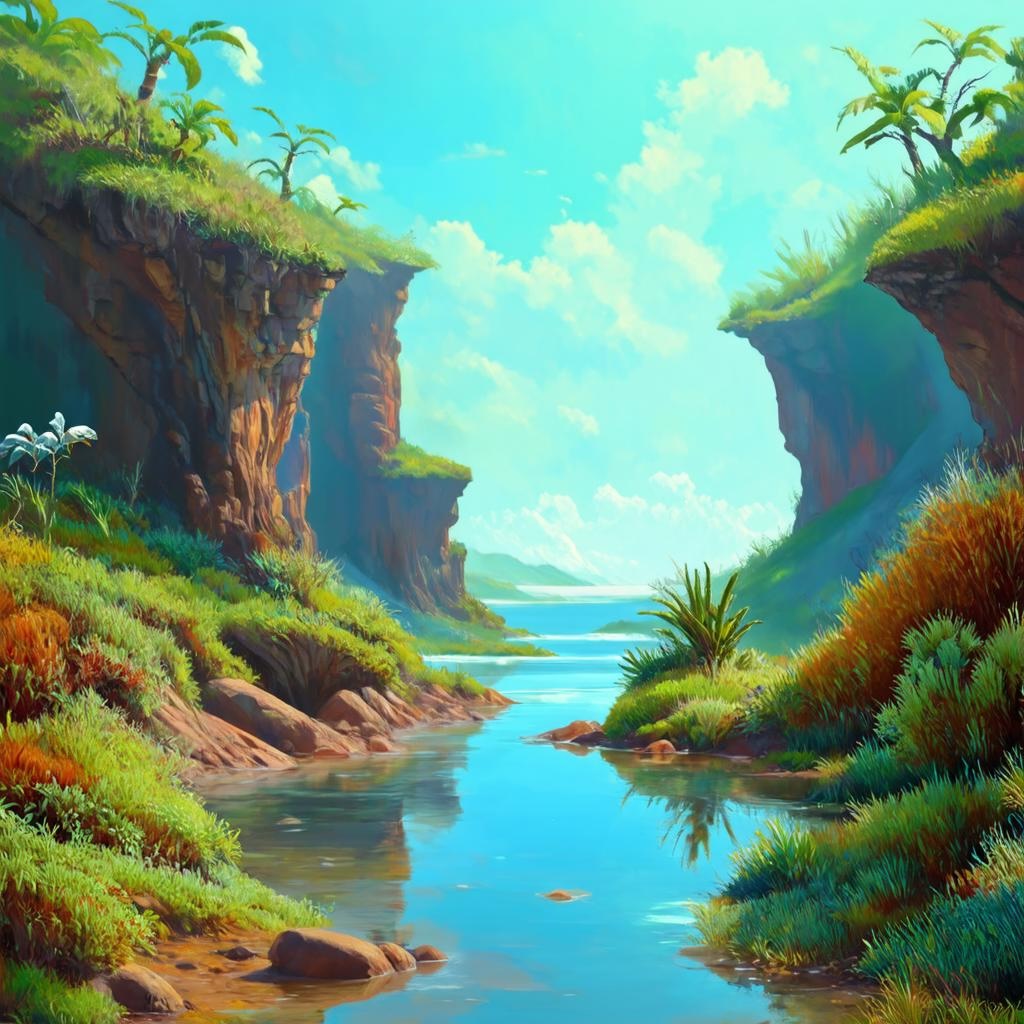}\end{minipage}%
  \begin{minipage}{0.18\textwidth}\includegraphics[width=\linewidth]{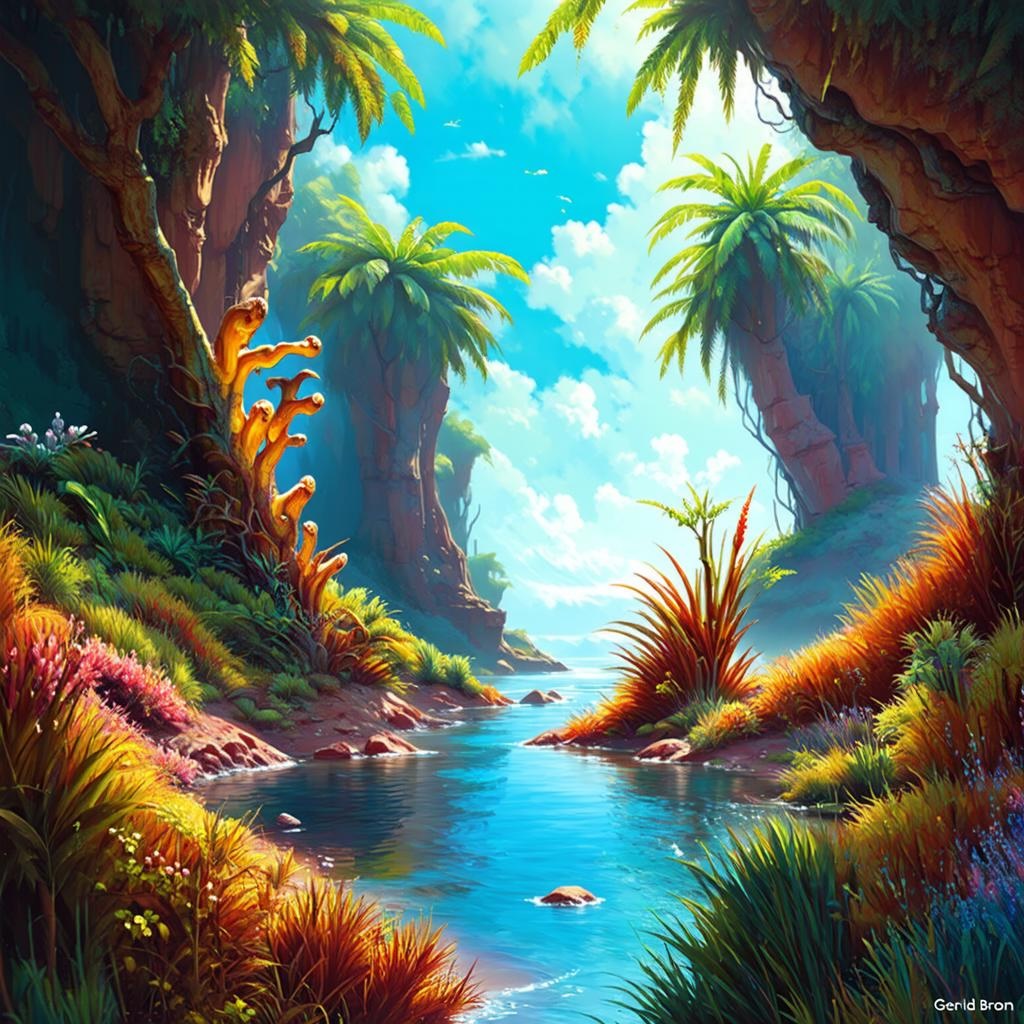}\end{minipage}%

  \begin{minipage}{0.28\textwidth}
  \begin{minipage}{0.90\textwidth}
    \centering \tiny \raggedright {a portrait of a statue of anubis with a crown and wearing a yellow t-shirt that has a space shuttle drawn on it}
  \end{minipage}
  \end{minipage}%
  \begin{minipage}{0.18\textwidth}\includegraphics[width=\linewidth]{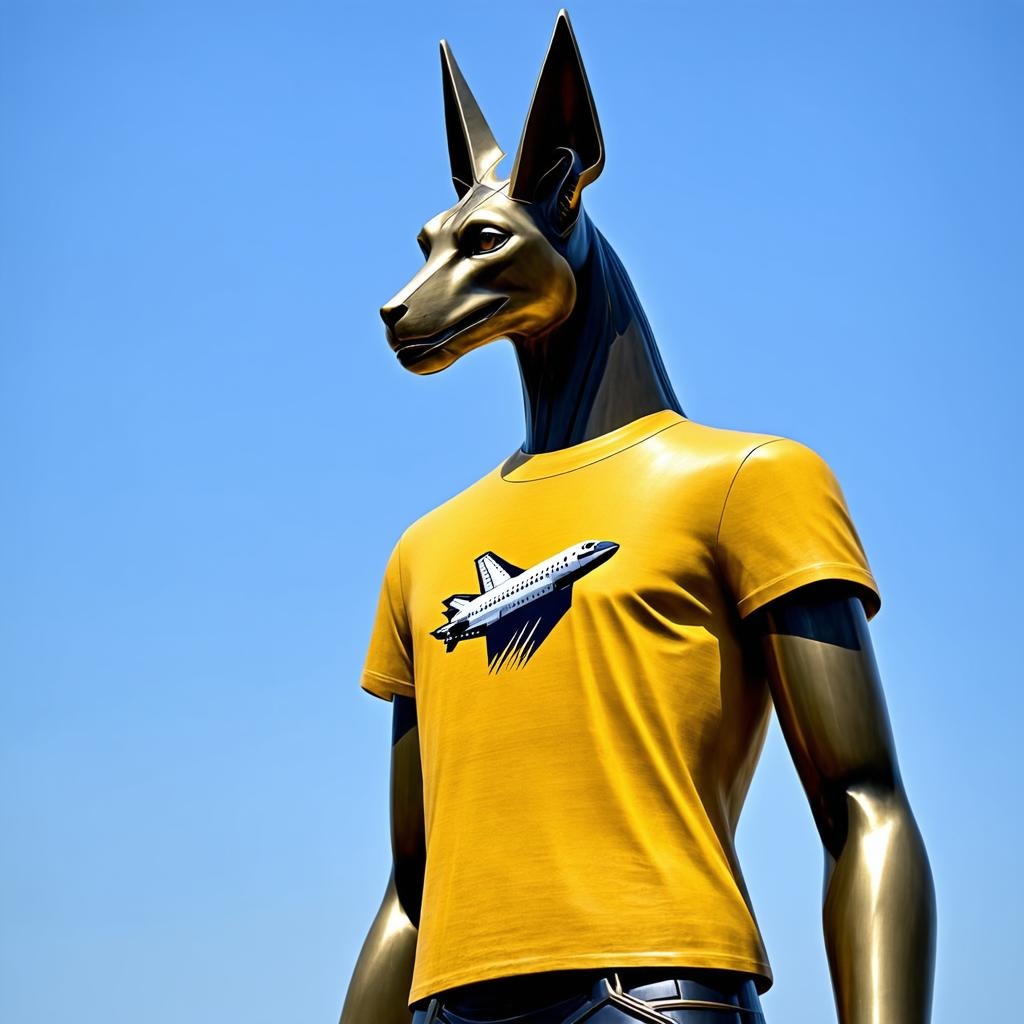}\end{minipage}%
  \begin{minipage}{0.18\textwidth}\includegraphics[width=\linewidth]{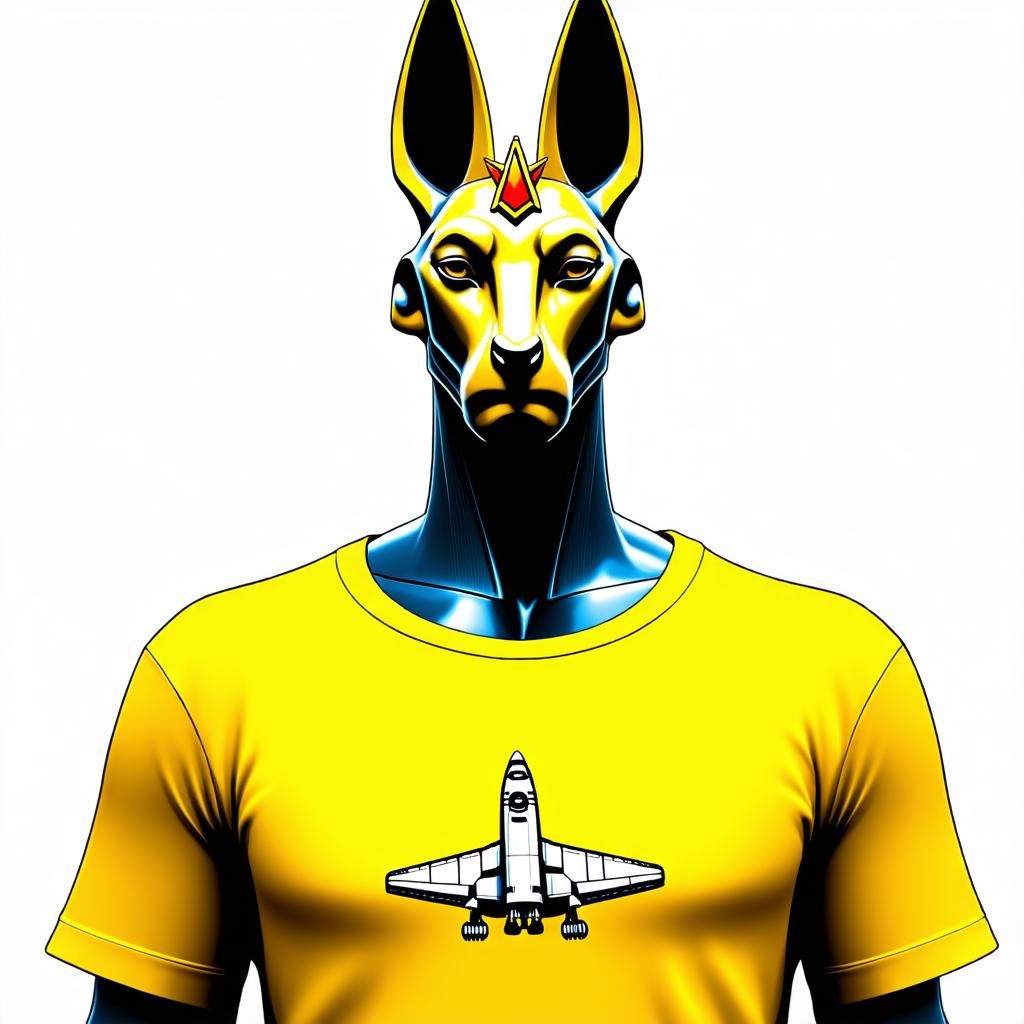}\end{minipage}%
  \begin{minipage}{0.18\textwidth}\includegraphics[width=\linewidth]{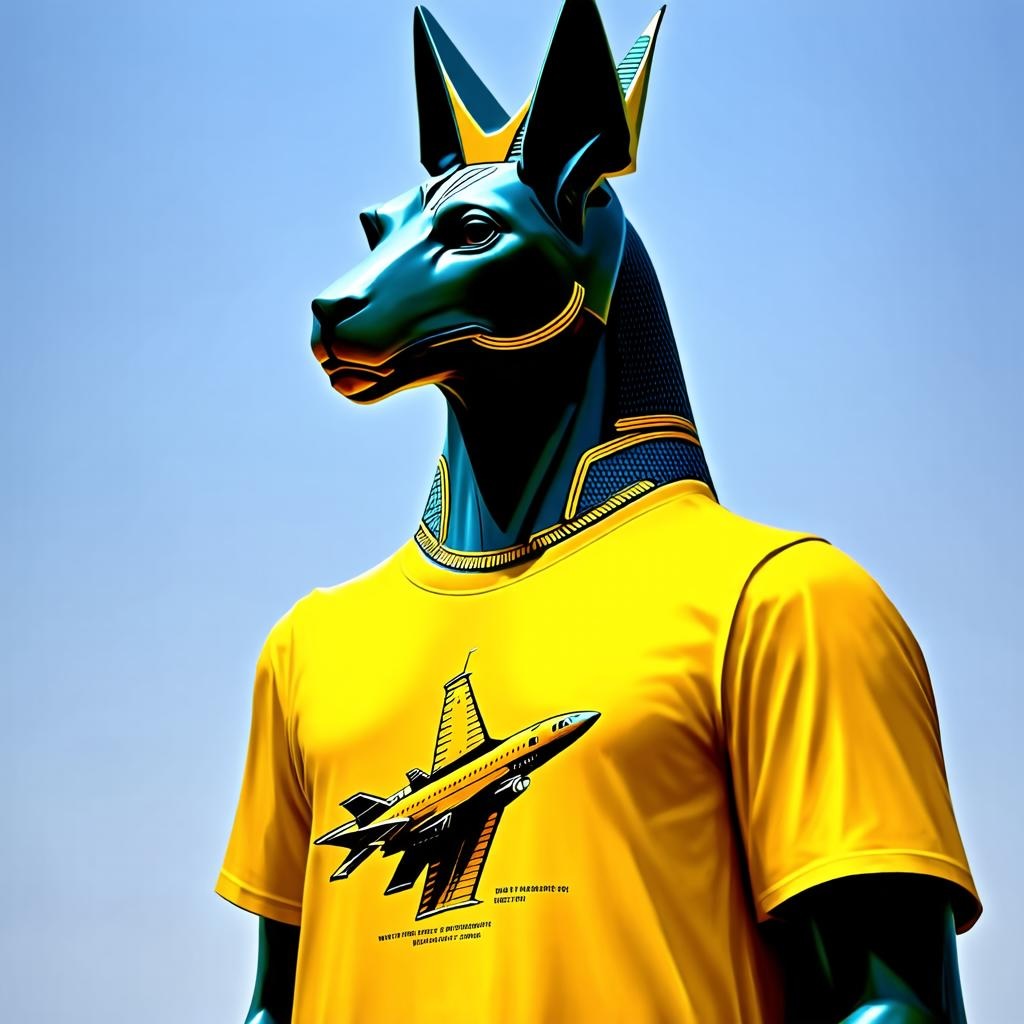}\end{minipage}%
  \begin{minipage}{0.18\textwidth}\includegraphics[width=\linewidth]{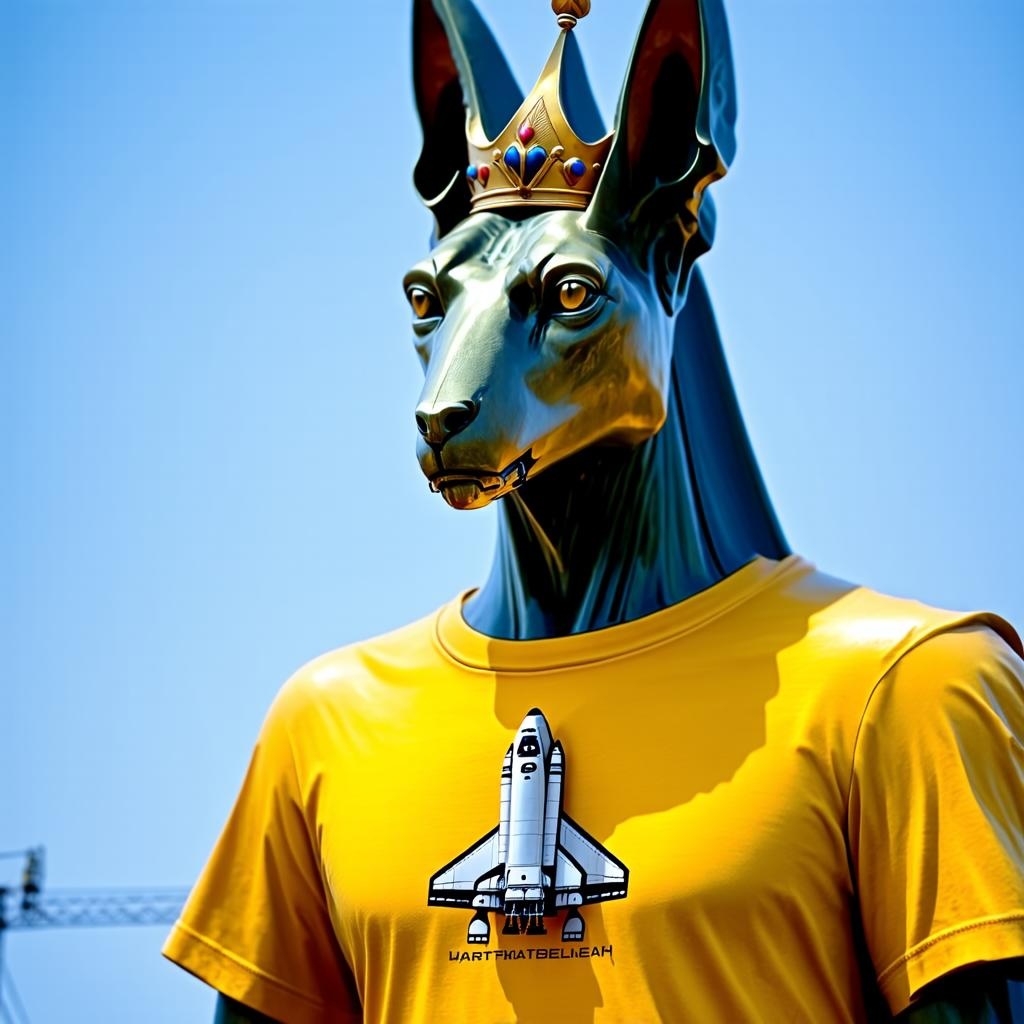}\end{minipage}%
  \end{minipage}
  \caption{Qualitative results of generated images by different methods on SD3-M. Even for the high-capacity SD3-M, Linear-DPO still yields notable improvements in overall human preference alignment. Diff-DPO denotes Diffusion-DPO for short.}
  \label{fig.sd3_short}
  \vskip -0.2in
\end{figure}

\textbf{Qualitative Results.}
Figure~\ref{fig.sd15} and Figure~\ref{fig.sd3_short} present qualitative comparisons of images generated by Linear-DPO and other baselines under identical prompts.

As illustrated in Figure~\ref{fig.sd15}, Linear-DPO yields substantial improvements for SD1.5, whose limited capacity often leads to structural warping and poor adherence to complex prompts. Our method mitigates these distortions and significantly strengthens the model's ability to follow intricate textual instructions, resulting in better text--image alignment and images with richer details.

A similar trend is observed in Figure~\ref{fig.sd3_short} for the more advanced SD3-M model. While SD3-M already exhibits strong performance with few structural failures, Linear-DPO further improves aesthetic quality to better align with human preferences. Specifically, it refines stylistic coherence, color harmony, and background detail, providing a noticeable boost in visual quality even for this high-capacity base model.

\fontfamily{ptm}\selectfont

\subsection{Ablation Study}
\begin{figure}[tbhp]
\centering
\begin{minipage}{0.49\textwidth}
\begin{minipage}{0.495\textwidth}
\centering
\includegraphics[width=\linewidth]{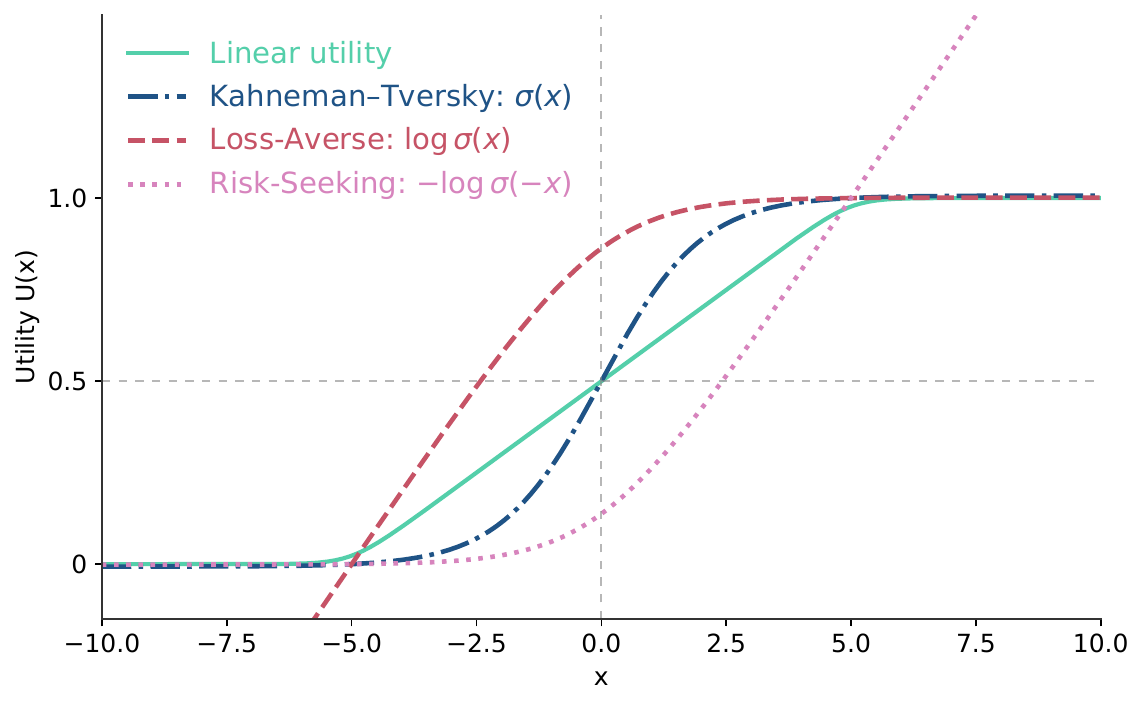}\\
{\small (a) Utility functions}
\end{minipage}
\begin{minipage}{0.495\textwidth}
\centering
\includegraphics[width=\linewidth]{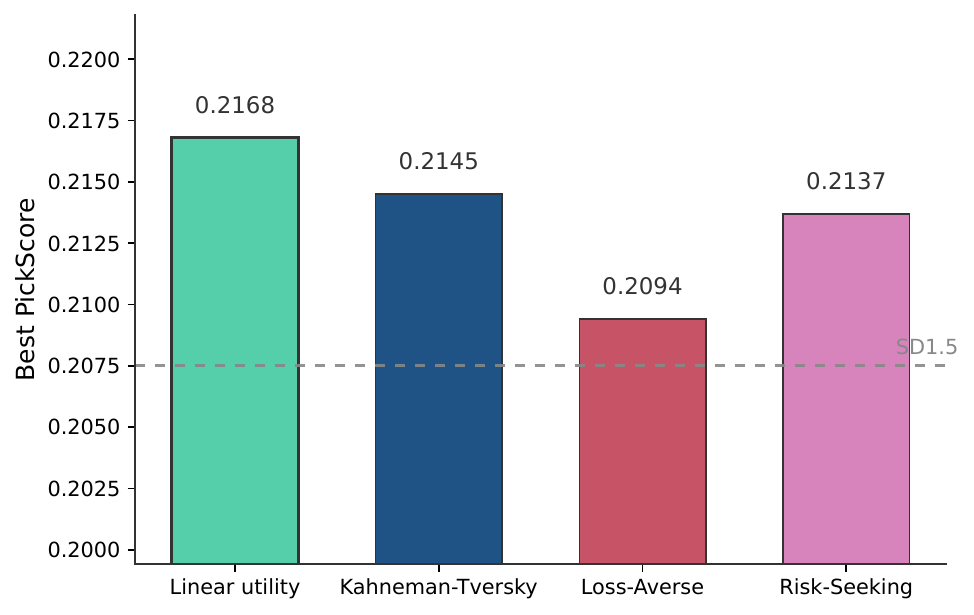}\\
{\small (b) PickScores}
\end{minipage}
\caption{Normalized utility curves (a) and corresponding PickScore performance (b) Our linear utility achieves the superior performance over other utility functions mentioned in Diffusion-KTO.}
\label{fig.utility}
\end{minipage}
\vskip -0.1in
\end{figure}

\textbf{Choice of Utility Function.}
We explore several alternative utility functions discussed in Diffusion-KTO~\cite{li2024aligning}, including Loss-Averse ($\log \sigma(x)$), Risk-Seeking ($-\log \sigma(-x)$), and the Kahneman--Tversky~\cite{tversky1992advances} form ($\sigma(x)$). To make their effective ranges and input domains as comparable as possible, we apply the same normalization \(\frac{U(x)-U(-5)}{U(5)-U(-5)}\) to each utility function \(U(x)\) and then clip the result to \([0,1]\). The resulting normalized curves are shown in Figure~\ref{fig.utility}(a).
We conduct experiments on a subset of Pick-a-Pic v2 and report the best PickScore achieved by each utility function. As shown in Figure~\ref{fig.utility}(b), the Kahneman--Tversky utility achieves higher PickScore than the other two asymmetric variants, which is consistent with the results in Diffusion-KTO; beyond these three forms, our linear utility achieves the highest PickScore. Additional analysis of the utility functions is provided in Appendix~\ref{app.utility_extra}.

\begin{figure}
    \centering
    \includegraphics[width=0.9\linewidth]{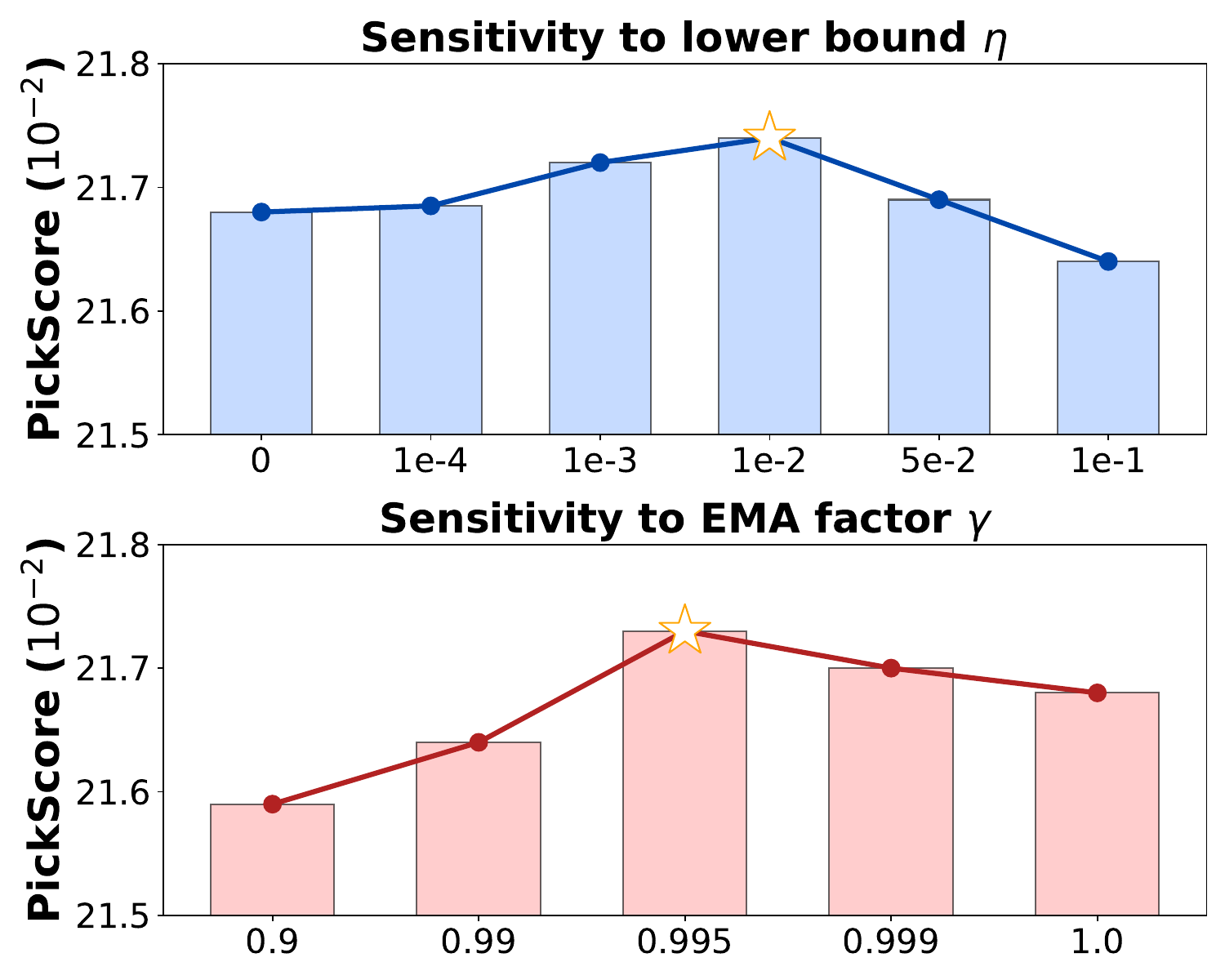}
    \caption{PickScores under different $\eta$ (top) and  $\gamma$ (bottom).}
    \label{fig.hyperparameter}
    \vskip -0.2in
\end{figure}

\textbf{Effect of \(\eta\) in the Linear Utility Function.}
To prevent premature stagnation that may cause the model to overlook further refinement of fine-grained details, we set the lower clipping bound of the linear utility to a small constant \(\eta\) instead of 0. We then study the effect of different \(\eta\) values on Linear-DPO. As shown in Figure~\ref{fig.hyperparameter}, using a small \(\eta\) yields consistent gains, with the best performance achieved at \(\eta=1\times10^{-2}\), demonstrating the effectiveness of the clipping mechanism. However, when \(\eta\) is too large, it leads to over-optimization and consequently hurts performance.

\textbf{Effect of the EMA Reference Model.}
In Section~\ref{sec.linear}, we discuss using an EMA copy of the policy model as the reference model to enable smoother and more sustained optimization. To validate the effectiveness of the EMA reference and select an appropriate decay factor \(\gamma\), we compare the PickScore achieved by a fixed reference model (\(\gamma=1\)) with EMA references using \(\gamma \in \{0.9, 0.99, 0.995, 0.999\}\). As shown in Figure~\ref{fig.hyperparameter}, updating the reference too aggressively (\(\gamma=0.9, 0.99\)) hurts performance compared to a fixed reference. The best results are obtained at \(\gamma=0.995\), while further increasing \(\gamma\) yields diminishing returns.
\section{Conclusion}
In this paper, we bridge the theoretical gap between diffusion and flow matching in direct preference optimization (DPO) by deriving a generalized objective from a unified SDE perspective. Through a rigorous analysis of the gradient dynamics, we identify a "pseudo-convergence" trap in standard DPO when applied to regression-based generative tasks. To address this issue, we introduce Linear-DPO, a method that uses a linear utility function and an EMA-updated reference model to improve alignment performance for generative models. Across models ranging from SD1.5 to SD3-M, we show that Linear-DPO overcomes key optimization hurdles of prior methods, achieving better preference alignment and higher visual quality. Overall, our results establish a versatile framework that supports both superiority and scalability, offering a principled path for future preference learning in large-scale generative models.
\nocite{langley00}

\bibliography{example_paper}
\bibliographystyle{icml2026}

\newpage
\appendix
\onecolumn
\section{Detailed derivation of a unified DPO for diffusion and flow-matching}
\label{app.1}
We start from the unified perturbation (noising) parameterization
\begin{equation}
x_t \;=\; \alpha_t x_0 + \sigma_t \epsilon,\qquad \epsilon\sim\mathcal N(0,I),
\label{eq:unified-perturb}
\end{equation}
which can be viewed as a \emph{stochastic interpolation}~\cite{albergo2023stochastic, ma2024sit} between the clean sample $x_0$ and Gaussian noise $\epsilon$.
This form encompasses both diffusion-based models and flow-matching models under different choices of
$(\alpha_t,\sigma_t)$~\cite{karras2022elucidating}.

\paragraph{Forward SDE.}
We consider a stochastic differential equation (SDE) as the forward noising process
\begin{equation}
    d x_t = f(t)\,x_t\,dt + g(t)\,d w_t,\quad t\in[0,T],
    \label{eq:forward-sde}
\end{equation}
where $w_t$ is a standard Wiener process and $f(t)$, $g(t)>0$ are given time-dependent coefficients. The process is initialized at $x_0 \sim p_0(x)$, the unknown data distribution, and we denote by $p_t(x)$ the resulting marginal distribution of $x_t$.
When \eqref{eq:forward-sde} is initialized with $x_0\sim p_0(x)$, it admits the closed-form marginal
\begin{equation}
x_t \mid x_0 \;\sim\; \mathcal N\!\bigl(\alpha_t x_0,\; \sigma_t^2 I\bigr),
\label{eq:cond-gaussian}
\end{equation}
i.e., it matches the perturbation form in \eqref{eq:unified-perturb}, with
\begin{equation}
\alpha_t = \exp\!\Bigl(\int_0^t f(s)\,ds\Bigr),
\qquad
\sigma_t^2 = \int_0^t \exp\!\Bigl(2\!\int_s^t f(u)\,du\Bigr)\, g^2(s)\,ds.
\label{eq:alpha-sigma-from-fg}
\end{equation}
Equivalently, $(\alpha_t,\sigma_t)$ satisfy the ODEs
\begin{equation}
\dot \alpha_t = f(t)\alpha_t,\qquad
\frac{d}{dt}\sigma_t^2 = 2 f(t)\sigma_t^2 + g^2(t),
\label{eq:alpha-sigma-ode}
\end{equation}
with $\alpha_0=1$ and $\sigma_0=0$. These identities provide the numerical correspondence between
the interpolation coefficients $(\alpha_t,\sigma_t)$ and the SDE coefficients $(f(t),g(t))$.

\paragraph{Reverse-time SDE.}
Under mild regularity conditions, the corresponding ideal reverse-time SDE \cite{anderson1982reverse,song2020score} is
\begin{equation}
    d x_t
    =
    \bigl[
      f(t)\,x_t - g^2(t)\,\nabla_{x_t}\log p_t(x_t)
    \bigr]\,dt
    + g(t)\,d\bar w_t,
    \label{eq:reverse-sde}
\end{equation}
where $dt<0$ indicates integration in reverse time, $\bar w_t$ is a reverse-time Wiener process, and $\nabla_{x_t}\log p_t(x_t)$ is the score of the true marginal $p_t$. In principle, if one had access to the exact score, solving \eqref{eq:reverse-sde} backward from the terminal noise distribution $p_T$ would exactly recover samples from $p_0$.

\paragraph{One-step Euler--Maruyama discretization.}
To obtain a one-step conditional distribution, we discretize the reverse-time SDE in \eqref{eq:reverse-sde} using the Euler--Maruyama scheme. For a small step size $\Delta t>0$, a backward step from $t$ to $t-\Delta t$ gives
\begin{equation}
    x_{t-\Delta t} 
    \approx x_t
    + \bigl[
        f(t)\,x_t - g^2(t)\,\nabla_{x_t}\log p_t(x_t)
      \bigr]\,(-\Delta t)
    + g(t)\sqrt{\Delta t}\,\epsilon,
    \quad \epsilon\sim\mathcal{N}(0,I).
    \label{eq:reverse-step-dt}
\end{equation}
For notational simplicity, we absorb $\Delta t$ into the time-dependent coefficients and write a single backward step from $t$ to $t-1$ as
\begin{equation}
    x_{t-1}
    = x_t
      - \bigl[
          f(t)\,x_t - g^2(t)\,\nabla_{x_t}\log p_t(x_t)
        \bigr]
      + g(t)\,\epsilon,
    \quad
    \epsilon\sim\mathcal{N}(0,I).
    \label{eq:reverse-step}
\end{equation}
Conditioned on $x_t$, this update is Gaussian. The corresponding one-step reverse conditional distribution under the true process can be written as
\begin{equation}
    q(x_{t-1}\mid x_t)
    = \mathcal{N}\!\Bigl(
        x_{t-1};
        \mu^\star(x_t,t),\,
        \Sigma_t
    \Bigr),
    \quad
    \mu^\star(x_t,t)
    := x_t - \bigl[f(t)\,x_t - g^2(t)\,\nabla_{x_t}\log p_t(x_t)\bigr],
    \label{eq:true-cond-sde}
\end{equation}
where, in the simplest case, $\Sigma_t = g^2(t)I$. Here $q(x_{t-1}\mid x_t)$ denotes the ideal one-step conditional distribution implied by the Euler discretization, assuming access to the true score $\nabla_{x_t}\log p_t(x_t)$.

\paragraph{Unified Gaussian conditional form.}
More broadly, many SDE-based generative models (including VP/VE diffusions, score-based SDEs, and stochastic-interpolant-based flows) lead to a one-step reverse conditional distribution of the form
\begin{equation}
    q(x_{t-1}\mid x_t)
    = \mathcal{N}\!\bigl(
        x_{t-1};
        \mu^\star(x_t,t),\,
        \Sigma_t
    \bigr),
    \label{eq:true-unified-cond}
\end{equation}
where the mean $\mu^\star(x_t,t)$ is determined by the forward SDE together with the true score, and $\Sigma_t$ is a positive-definite covariance (e.g., $\Sigma_t=g^2(t)I$ or a more general time-dependent covariance). In what follows, we use \eqref{eq:true-unified-cond} as a unified description of the ideal one-step reverse dynamics.

In practice, $\mu^\star$ is intractable, so we approximate the one-step conditional distribution with a parametric policy model and a reference model. The policy model with parameters $\theta$ is defined as
\begin{equation}
    p_\theta(x_{t-1}\mid x_t,c)
    = \mathcal{N}\!\bigl(
        x_{t-1};
        \mu_\theta(x_t,t,c),\,
        \Sigma_t
    \bigr),
    \label{eq:model-unified-cond}
\end{equation}
and the reference model as
\begin{equation}
    p_{\mathrm{ref}}(x_{t-1}\mid x_t,c)
    = \mathcal{N}\!\bigl(
        x_{t-1};
        \mu_{\mathrm{ref}}(x_t,t,c),\,
        \Sigma_t
    \bigr).
    \label{eq:ref-unified-cond}
\end{equation}
We assume that $q$, $p_\theta$, and $p_{\mathrm{ref}}$ share the same covariance $\Sigma_t$ at each time $t$; the modeling choices are fully captured by the mean functions $\mu^\star$, $\mu_\theta$, and $\mu_{\mathrm{ref}}$ (e.g., the parameterization as a denoiser, score, or velocity). Based on this unified conditional form, we now derive our DPO objective.

\paragraph{Unified DPO for the SDE.}
Direct Preference Optimization (DPO)~\cite{rafailov2023direct} was originally proposed in NLP to learn a conditional policy $p_\theta(x\mid c)$ from pairwise preferences. Given a condition $c$ (e.g., an instruction) and a preferred/dispreferred pair $(x^w,x^l)$, the DPO loss is
\begin{align}
\mathcal{L}_{\text{DPO}}(\theta)
=
-\mathbb{E}_{(c,x^w,x^l)\sim\mathcal D}
\log\sigma\!\Big(
\beta \log\frac{p_\theta(x^w\mid c)}{p_{\mathrm{ref}}(x^w\mid c)}
-\beta \log\frac{p_\theta(x^l\mid c)}{p_{\mathrm{ref}}(x^l\mid c)}
\Big),
\label{eq:dpo-original}
\end{align}
where $p_{\mathrm{ref}}$ is a fixed reference model and $\beta$ controls the strength of the preference signal.

For diffusion/flow generative models, however, $p_\theta(x_0\mid c)$ is not tractable because it is defined implicitly through a latent trajectory $x_{1:T}$ and requires marginalizing over all intermediate states:
\begin{equation}
p_\theta(x_0\mid c)=\int p_\theta(x_{0:T}\mid c)\,dx_{1:T}.
\label{eq:intractable-marginal}
\end{equation}

Following the treatment in Diffusion-DPO~\cite{wallace2024diffusion}, we introduce the latent chain $x_{1:T}$ and rewrite the log-ratio using the joint distribution over trajectories and approximate true posterior $p_\theta(x_{1:T}\mid x_0,c)$ with the forward noising process $q(x_{1:T}\mid x_0)$:

\begin{align}
L(\theta)
&= - \log \sigma\!\left(
    \beta \,
    \mathbb{E}_{x_{1:T}^w \sim q(x_{1:T}\mid x_0^w),\,x_{1:T}^l \sim q(x_{1:T}\mid x_0^l)}
    \biggl[
        \log \frac{p_\theta(x_{0:T}^w)}{p_{\mathrm{ref}}(x_{0:T}^w)}
        - \log \frac{p_\theta(x_{0:T}^l)}{p_{\mathrm{ref}}(x_{0:T}^l)}
    \biggr]
\right) \nonumber\\
&= - \log \sigma\!\left(
    \beta \,
    \mathbb{E}_{x_{1:T}^w \sim q(x_{1:T}\mid x_0^w),\,x_{1:T}^l \sim q(x_{1:T}\mid x_0^l)}
    \biggl[
        \sum_{t=1}^{T}
        \log \frac{p_\theta(x_{t-1}^w \mid x_t^w)}{p_{\mathrm{ref}}(x_{t-1}^w \mid x_t^w)}
        -
        \log \frac{p_\theta(x_{t-1}^l \mid x_t^l)}{p_{\mathrm{ref}}(x_{t-1}^l \mid x_t^l)}
    \biggr]
\right) \nonumber\\
&= - \log \sigma\!\left(
    \beta \,
    \mathbb{E}_{x_{1:T}^w \sim q(x_{1:T}\mid x_0^w),\,x_{1:T}^l \sim q(x_{1:T}\mid x_0^l)}
    \biggl[
        T \,\mathbb{E}_{t}
        \Bigl[
            \log \frac{p_\theta(x_{t-1}^w \mid x_t^w)}{p_{\mathrm{ref}}(x_{t-1}^w \mid x_t^w)}
            -
            \log \frac{p_\theta(x_{t-1}^l \mid x_t^l)}{p_{\mathrm{ref}}(x_{t-1}^l \mid x_t^l)}
        \Bigr]
    \biggr]
\right) \nonumber\\
&= - \log \sigma\!\left(
    \beta T \,
    \mathbb{E}_{t}
    \mathbb{E}_{x_{t-1,t}^w \sim q(x_{t-1,t}\mid x_0^w),\,x_{t-1,t}^l \sim q(x_{t-1,t}\mid x_0^l)}
    \Bigl[
        \log \frac{p_\theta(x_{t-1}^w \mid x_t^w)}{p_{\mathrm{ref}}(x_{t-1}^w \mid x_t^w)}
        -
        \log \frac{p_\theta(x_{t-1}^l \mid x_t^l)}{p_{\mathrm{ref}}(x_{t-1}^l \mid x_t^l)}
    \Bigr]
\right) \nonumber\\
&= - \log \sigma\!\left(
    \beta T \,
    \mathbb{E}_{t,\,x_t^w \sim q(x_t\mid x_0^w),\,x_t^l \sim q(x_t\mid x_0^l)}
\right. \nonumber\\
&\hspace{5em}\left.
    \mathbb{E}_{x_{t-1}^w \sim q(x_{t-1}\mid x_t^w,x_0^w),\,x_{t-1}^l \sim q(x_{t-1}\mid x_t^l,x_0^l)}
    \Bigl[
        \log \frac{p_\theta(x_{t-1}^w \mid x_t^w)}{p_{\mathrm{ref}}(x_{t-1}^w \mid x_t^w)}
        -
        \log \frac{p_\theta(x_{t-1}^l \mid x_t^l)}{p_{\mathrm{ref}}(x_{t-1}^l \mid x_t^l)}
    \Bigr]
\right).
\end{align}
For brevity, we omit the conditioning variable $c$ from all distributions.

Using Jensen's inequality and the concavity of the function $u\mapsto \log\sigma(u)$, we obtain the following upper bound on $L(\theta)$ (equivalently, a lower bound on the corresponding DPO objective):
\begin{align}
L(\theta) \leq\;
& - \mathbb{E}_{t,\,x_t^w \sim q(x_t\mid x_0^w),\,x_t^l \sim q(x_t\mid x_0^l)}
\log \sigma\!\Biggl(
    \beta T \nonumber\\
&\qquad
    \mathbb{E}_{x_{t-1}^w \sim q(x_{t-1}\mid x_t^w,x_0^w),\,x_{t-1}^l \sim q(x_{t-1}\mid x_t^l,x_0^l)}
    \Bigl[
        \log \frac{p_\theta(x_{t-1}^w \mid x_t^w)}{p_{\mathrm{ref}}(x_{t-1}^w \mid x_t^w)}
        -
        \log \frac{p_\theta(x_{t-1}^l \mid x_t^l)}{p_{\mathrm{ref}}(x_{t-1}^l \mid x_t^l)}
    \Bigr]
\Biggr) \nonumber\\
=\;
& - \mathbb{E}_{t,\,x_t^w \sim q(x_t\mid x_0^w),\,x_t^l \sim q(x_t\mid x_0^l)}
\log \sigma\!\Biggl(
    \beta T \nonumber\\
&\qquad
    \mathbb{E}_{x_{t-1}^w \sim q(x_{t-1}\mid x_t^w,x_0^w)}
    \Bigl[
        \log \frac{p_\theta(x_{t-1}^w \mid x_t^w)}{p_{\mathrm{ref}}(x_{t-1}^w \mid x_t^w)}
    \Bigr]
    -
    \mathbb{E}_{x_{t-1}^l \sim q(x_{t-1}\mid x_t^l,x_0^l)}
    \Bigl[
        \log \frac{p_\theta(x_{t-1}^l \mid x_t^l)}{p_{\mathrm{ref}}(x_{t-1}^l \mid x_t^l)}
    \Bigr]
\Biggr) \nonumber\\
=\;
& - \mathbb{E}_{t,\,x_t^w \sim q(x_t\mid x_0^w),\,x_t^l \sim q(x_t\mid x_0^l)}
\log \sigma\!\Biggl(
    \beta T \nonumber\\
&\qquad
    \mathbb{E}_{x_{t-1}^w \sim q(\cdot\mid x_t^w,x_0^w)}\!\bigl[\log p_\theta(x_{t-1}^w \mid x_t^w)\bigr]
    -
    \mathbb{E}_{x_{t-1}^w \sim q(\cdot\mid x_t^w,x_0^w)}\!\bigl[\log p_{\mathrm{ref}}(x_{t-1}^w \mid x_t^w)\bigr]
    \nonumber\\
&\qquad
    -
    \mathbb{E}_{x_{t-1}^l \sim q(\cdot\mid x_t^l,x_0^l)}\!\bigl[\log p_\theta(x_{t-1}^l \mid x_t^l)\bigr]
    +
    \mathbb{E}_{x_{t-1}^l \sim q(\cdot\mid x_t^l,x_0^l)}\!\bigl[\log p_{\mathrm{ref}}(x_{t-1}^l \mid x_t^l)\bigr]
\Biggr).
\end{align}

We then use the identity
\begin{equation}
\mathbb{E}_q[\log p]
= -\mathbb{D}_{\mathrm{KL}}(q\|p) + \mathbb{E}_q[\log q],
\label{eq:ekl-identity}
\end{equation}
to rewrite the expected log-likelihood terms in terms of KL divergences. The $\mathbb{E}_q[\log q]$ terms cancel, yielding
\begin{align}
L(\theta) \leq\;
& - \mathbb{E}_{t,\,x_t^w \sim q(x_t\mid x_0^w),\,x_t^l \sim q(x_t\mid x_0^l)}
\log \sigma\!\Biggl(
    - \beta T \nonumber\\
&\qquad
\mathbb{D}_{\mathrm{KL}}\!\Bigl(q(x_{t-1}^w\!\mid x_t^w,x_0^w)\,\big\|\,p_\theta(x_{t-1}^w\!\mid x_t^w)\Bigr)
+
\mathbb{D}_{\mathrm{KL}}\!\Bigl(q(x_{t-1}^w\!\mid x_t^w,x_0^w)\,\big\|\,p_{\mathrm{ref}}(x_{t-1}^w\!\mid x_t^w)\Bigr)
\nonumber\\
&\qquad
-
\mathbb{D}_{\mathrm{KL}}\!\Bigl(q(x_{t-1}^l\!\mid x_t^l,x_0^l)\,\big\|\,p_\theta(x_{t-1}^l\!\mid x_t^l)\Bigr)
-
\mathbb{D}_{\mathrm{KL}}\!\Bigl(q(x_{t-1}^l\!\mid x_t^l,x_0^l)\,\big\|\,p_{\mathrm{ref}}(x_{t-1}^l\!\mid x_t^l)\Bigr)
\Biggr).
\label{eq:l1-kl-form}
\end{align}
Since $q$, $p_\theta$, and $p_{\mathrm{ref}}$ are Gaussians with the same covariance $\Sigma_t$ at time $t$,
\begin{equation}
q(x_{t-1}\!\mid x_t,x_0)=\mathcal{N}\!\bigl(\mu^\star(x_t,t),\Sigma_t\bigr),\quad
p_\theta(x_{t-1}\!\mid x_t)=\mathcal{N}\!\bigl(\mu_\theta(x_t,t),\Sigma_t\bigr),\quad
p_{\mathrm{ref}}(x_{t-1}\!\mid x_t)=\mathcal{N}\!\bigl(\mu_{\mathrm{ref}}(x_t,t),\Sigma_t\bigr),
\end{equation}
the KL divergence has a closed form:
\begin{align}
\mathbb{D}_{\mathrm{KL}}\!\Bigl(q(x_{t-1}\!\mid x_t,x_0)\,\big\|\,p_\theta(x_{t-1}\!\mid x_t)\Bigr)
&= \frac{1}{2}\,(\mu^\star-\mu_\theta)^\top \Sigma_t^{-1}(\mu^\star-\mu_\theta) \nonumber\\
&= \frac{1}{2}\,\bigl\|\mu^\star(x_t,t)-\mu_\theta(x_t,t)\bigr\|^2_{\Sigma_t^{-1}}.
\label{eq:unified-kl-gauss}
\end{align}
In the isotropic case $\Sigma_t=g^2(t)I$, this further reduces to
\begin{equation}
\mathbb{D}_{\mathrm{KL}}\!\Bigl(q\,\big\|\,p_\theta\Bigr)
=\frac{1}{2g^2(t)}\,
\bigl\|\mu^\star(x_t,t)-\mu_\theta(x_t,t)\bigr\|_2^2.
\label{eq:unified-kl-gauss-iso}
\end{equation}
Similarly,
\begin{equation}
\mathbb{D}_{\mathrm{KL}}\!\Bigl(q(x_{t-1}\!\mid x_t,x_0)\,\big\|\,p_{\mathrm{ref}}(x_{t-1}\!\mid x_t)\Bigr)
=\frac{1}{2g^2(t)}\,
\bigl\|\mu^\star(x_t,t)-\mu_{\mathrm{ref}}(x_t,t)\bigr\|_2^2,
\label{eq:unified-kl-ref}
\end{equation}
where the last equality again assumes $\Sigma_t=g^2(t)I$.

Substituting \eqref{eq:unified-kl-gauss-iso}--\eqref{eq:unified-kl-ref} into the KL-based form of the objective yields the unified SDE-based DPO loss:
\begin{align}
    \mathcal{L}_{\mathrm{DPO\text{-}SDE}}(\theta)
    = & -\,\mathbb{E}_{t,\,x_t^w \sim q(x_t\mid x_0^w),\,x_t^l \sim q(x_t\mid x_0^l)}\Bigl[
        \log \sigma\bigl(
          - \frac{\beta T}{2g^2(t)} \cdot \nonumber \\
          &\|\mu^\star(x_t^w,t) - \mu_\theta(x_t^w,t)\|^2_2 - 
          \|\mu^\star(x_t^w,t) - \mu_\text{ref}(x_t^w,t)\|^2_2 \nonumber \\
          &-(\|\mu^\star(x_t^l,t) - \mu_\theta(x_t^l,t)\|^2_2 - 
          \|\mu^\star(x_t^l,t) - \mu_\text{ref}(x_t^l,t)\|^2_2)
        \bigr)
      \Bigr],
    \label{eq:unified-dpo-sde-final}
\end{align}
we optionally include the factor $T$ to account for the total number of time steps, as is common in diffusion-based DPO formulations. Eq.~\eqref{eq:unified-dpo-sde-final} depends only on the squared discrepancy between the ideal mean $\mu^\star$ and the model/reference means $\mu_\theta$ and $\mu_{\mathrm{ref}}$ under the shared covariance $\Sigma_t$, providing a unified DPO formulation for SDE-based generative models whose one-step reverse conditionals are induced by \eqref{eq:forward-sde}--\eqref{eq:reverse-sde}.

\textbf{VE diffusion (Score-based diffusion).}

We first consider the VE setting. Under the reverse-time SDE discretization, the ideal reverse mean and the model reverse mean are
\begin{align}
\mu^\star(x_t,t)
&= x_t - \bigl[f(t)x_t - g^2(t)\,\nabla_{x_t}\log p_t(x_t)\bigr], \nonumber \\
\mu_{\theta}(x_t,t)
&= x_t - \bigl[f(t)x_t - g^2(t)\,s_\theta(x_t,t)\bigr],
\end{align}
where $p_t$ denotes the marginal density of $x_t$ and $s_\theta(x_t,t)$ is the model-predicted score.

Their difference is
\begin{align}
\mu^\star(x_t,t) - \mu_{\theta}(x_t,t)
&=
g^2(t)\,\Bigl(\nabla_{x_t}\log p_t(x_t) - s_\theta(x_t,t)\Bigr).
\label{eq:mu_diff_uncond}
\end{align}

The unconditional score $\nabla_{x_t}\log p_t(x_t)$ is generally intractable. Following \citet{song2019generative}, we replace it with the tractable \emph{conditional} score $\nabla_{x_t}\log p(x_t\mid x_0)$ (which is available under the forward perturbation distribution). For VE diffusion, the forward process satisfies
\begin{equation}
x_t = x_0 + \sigma_t \epsilon,\qquad \epsilon\sim\mathcal N(0,I),
\qquad
p(x_t\mid x_0)=\mathcal N(x_0,\sigma_t^2 I),
\end{equation}
with $\sigma_t^2=\int_0^t g^2(s)\,ds$ (for the standard VE SDE where $f(t)\equiv 0$). Hence,
\begin{equation}
\nabla_{x_t}\log p(x_t\mid x_0)
= -\frac{1}{\sigma_t^2}(x_t-x_0)
= -\frac{1}{\sigma_t}\,\epsilon.
\label{eq:ve-cond-score}
\end{equation}

Substituting \eqref{eq:ve-cond-score} into \eqref{eq:mu_diff_uncond} gives the conditional-mean discrepancy used in training:
\begin{align}
\mu^\star(x_t,t,x_0) - \mu_{\theta}(x_t,t)
&=
g^2(t)\,\Bigl(\nabla_{x_t}\log p(x_t\mid x_0) - s_\theta(x_t,t)\Bigr) \nonumber\\
&=
g^2(t)\,\Bigl(-\frac{1}{\sigma_t}\epsilon - s_\theta(x_t,t)\Bigr).
\label{eq:mu_diff_cond}
\end{align}
Similarly,
\begin{equation}
\mu^\star(x_t,t,x_0) - \mu_{\mathrm{ref}}(x_t,t)
=
g^2(t)\,\Bigl(-\frac{1}{\sigma_t}\epsilon - s_{\mathrm{ref}}(x_t,t)\Bigr).
\label{eq:mu_diff_cond_ref}
\end{equation}

Finally, plugging these expressions into the unified SDE-based DPO objective yields the DPO loss for VE diffusion:
\begin{align}
    \mathcal{L}_{\text{DPO-VE}}(\theta)
    = & -\,\mathbb{E}_{t,\,x_t^w \sim q(x_t\mid x_0^w),\,x_t^l \sim q(x_t\mid x_0^l)}\Bigl[
        \log \sigma\Bigl(
          - \frac{\beta T g^2(t)}{2}\bigl( \nonumber \\
          &\|\frac{1}{\sigma_t}\,\epsilon^w + s_\theta(x_t^w,t)\|^2_2 - 
          \|\frac{1}{\sigma_t}\,\epsilon^w + s_\text{ref}(x_t^w,t)\|^2_2 
          -(\|\frac{1}{\sigma_t}\,\epsilon^l + s_\theta(x_t^l,t)\|^2_2 - 
          \|\frac{1}{\sigma_t}\,\epsilon^l + s_\text{ref}(x_t^l,t)\|^2_2)
        \bigr)
        \Bigr)
      \Bigr].
    \label{eq:unified-dpo-sde-ve}
\end{align}

\textbf{VP diffusion (DDPM/DDIM).}

For VP diffusion, the forward perturbation distribution is Gaussian:
\begin{equation}
q(x_t\mid x_0)=\mathcal N(\alpha_t x_0,\sigma_t^2 I),\qquad
x_t=\alpha_t x_0+\sigma_t\epsilon,\ \ \epsilon\sim\mathcal N(0,I),
\end{equation}
and the corresponding conditional score is
\begin{equation}
\nabla_{x_t}\log q(x_t\mid x_0)
=
-\frac{1}{\sigma_t^2}(x_t-\alpha_t x_0)
=
-\frac{1}{\sigma_t}\epsilon.
\label{eq:vp-cond-score}
\end{equation}

In the common $\epsilon$-parameterization, the model score is written as
\begin{equation}
s_\theta(x_t,t) = -\frac{1}{\sigma_t}\,\epsilon_\theta(x_t,t),
\qquad
s_{\mathrm{ref}}(x_t,t) = -\frac{1}{\sigma_t}\,\epsilon_{\mathrm{ref}}(x_t,t).
\label{eq:vp-score-eps}
\end{equation}

Substituting \eqref{eq:vp-cond-score}--\eqref{eq:vp-score-eps} into the unified VE-style expression
$\mu^\star-\mu_\theta=g^2(t)\bigl(\nabla_{x_t}\log q(x_t\mid x_0)-s_\theta(x_t,t)\bigr)$ yields
\begin{equation}
\mu^\star(x_t,t,x_0)-\mu_\theta(x_t,t)
=
g^2(t)\Bigl(-\tfrac{1}{\sigma_t}\epsilon + \tfrac{1}{\sigma_t}\epsilon_\theta(x_t,t)\Bigr)
=
\frac{g^2(t)}{\sigma_t}\bigl(\epsilon_\theta(x_t,t)-\epsilon\bigr),
\end{equation}
and similarly for the reference model.

Finally, assuming the isotropic covariance $\Sigma_t=g^2(t)I$, the VP-DPO loss becomes
\begin{align}
\mathcal{L}_{\mathrm{DPO\text{-}VP}}(\theta)
= &-\mathbb{E}_{t,\,x_t^w \sim q(x_t\mid x_0^w),\,x_t^l \sim q(x_t\mid x_0^l)}
\Biggl[
\log \sigma\Biggl(
-\frac{\beta T g^2(t)}{2\sigma_t^2}\Bigl(
\|\epsilon^w-\epsilon_\theta(x_t^w,t)\|_2^2
-
\|\epsilon^w-\epsilon_{\mathrm{ref}}(x_t^w,t)\|_2^2 \nonumber\\
&\hspace{9em}
-\bigl(\|\epsilon^l-\epsilon_\theta(x_t^l,t)\|_2^2
-
\|\epsilon^l-\epsilon_{\mathrm{ref}}(x_t^l,t)\|_2^2\bigr)
\Bigr)
\Biggr)
\Biggr].
\label{eq:unified-dpo-vp}
\end{align}
\textbf{Flow-matching.}

We derive \eqref{eq:vel-sde} by matching the marginal evolution of a deterministic flow with that of a stochastic process. Consider a flow model defined by the (deterministic) ODE
\begin{equation}
\mathrm{d}x_t = v_t(x_t)\,\mathrm{d}t,
\label{eq:flow-ode}
\end{equation}
whose marginal density $p_t(x)$ evolves according to the continuity equation
\begin{equation}
\partial_t p_t(x) = -\nabla\cdot\bigl(v_t(x)\,p_t(x)\bigr).
\label{eq:continuity}
\end{equation}
We would like to construct a stochastic process whose marginals match those of \eqref{eq:flow-ode}. To this end, consider a generic SDE with diffusion coefficient $g(t)$:
\begin{equation}
\mathrm{d}x_t = f_t(x_t)\,\mathrm{d}t + g(t)\,\mathrm{d}\bar w_t .
\label{eq:generic-sde}
\end{equation}
Its marginals satisfy the Fokker--Planck equation
\begin{equation}
\partial_t p_t(x) = -\nabla\cdot\bigl(f_t(x)\,p_t(x)\bigr) + \frac{1}{2}\nabla^2\!\bigl(g^2(t)\,p_t(x)\bigr).
\label{eq:fokker-planck}
\end{equation}
Since $g(t)$ is independent of $x$, we have
\begin{align}
\nabla^2\!\bigl(g^2(t)\,p_t(x)\bigr)
&= g^2(t)\,\nabla^2 p_t(x)
= g^2(t)\,\nabla\cdot\bigl(\nabla p_t(x)\bigr) \nonumber\\
&= g^2(t)\,\nabla\cdot\bigl(p_t(x)\,\nabla\log p_t(x)\bigr),
\label{eq:laplace-id}
\end{align}
where we used $\nabla p_t=p_t\nabla\log p_t$. Plugging \eqref{eq:laplace-id} into \eqref{eq:fokker-planck} yields
\begin{equation}
\partial_t p_t(x)
=
-\nabla\cdot\Bigl(\bigl[f_t(x)-\tfrac{g^2(t)}{2}\nabla\log p_t(x)\bigr]\,p_t(x)\Bigr).
\label{eq:fp-rewrite}
\end{equation}
To ensure that the SDE \eqref{eq:generic-sde} has the same marginal evolution as the ODE \eqref{eq:flow-ode}, we impose \eqref{eq:fp-rewrite} to match \eqref{eq:continuity}, i.e.,
\begin{equation}
-\nabla\cdot\Bigl(\bigl[f_t(x)-\tfrac{g^2(t)}{2}\nabla\log p_t(x)\bigr]\,p_t(x)\Bigr)
=
-\nabla\cdot\bigl(v_t(x)\,p_t(x)\bigr).
\label{eq:match-marginals}
\end{equation}
A sufficient condition is to match the vector fields inside the divergence:
\begin{equation}
f_t(x)-\frac{g^2(t)}{2}\nabla\log p_t(x) = v_t(x),
\end{equation}
which gives the drift
\begin{equation}
f_t(x) = v_t(x) + \frac{g^2(t)}{2}\nabla\log p_t(x).
\label{eq:fsde}
\end{equation}
Equivalently, writing the drift directly in terms of the velocity field leads to the velocity-based SDE form
\begin{equation}
\mathrm{d}x_t
=
\Bigl(
v_t(x_t) - \frac{g^2(t)}{2}\nabla_x \log p_t(x_t)
\Bigr)\,\mathrm{d}t
+ g(t)\,\mathrm{d}\bar w_t,
\label{eq:vel-sde}
\end{equation}
where we follow the sign convention adopted in our unified DPO--SDE formulation.

Consider the process in Eq.~\ref{eq:unified-perturb}, let \(\dot{\alpha}_t = \mathrm{d}\alpha_t/\mathrm{d}t\) and \(\dot{\sigma}_t = \mathrm{d}\sigma_t/\mathrm{d}t\),
 the conditional distribution is
\begin{equation}
p_t(x_t\mid x_0)=\mathcal N\!\big(x_t\mid \alpha_t x_0,\ \sigma_t^2 I\big),
\end{equation}
whose conditional score is
\begin{equation}
\nabla_{x_t}\log p_t(x_t\mid x_0)
= -\frac{x_t-\alpha_t x_0}{\sigma_t^2}
= -\frac{1}{\sigma_t}\varepsilon.
\end{equation}
Conditioning on \(x_t\) and taking expectation yields the marginal score–noise identity:
\begin{equation}
\nabla_x\log p_t(x)\Big|_{x=x_t}
=\mathbb E\!\left[\nabla_{x_t}\log p_t(x_t\mid x_0)\mid x_t\right]
= -\frac{1}{\sigma_t}\mathbb E[\varepsilon\mid x_t],
\end{equation}
equivalently,
\begin{equation}
\mathbb E[\varepsilon\mid x_t=x]=-\sigma_t \nabla_x\log p_t(x).
\end{equation}

Next define the velocity field
\begin{equation}
v_t(x_t):=\mathbb E[\dot x_t\mid x_t=x]
=\mathbb E[\dot\alpha_t x_0+\dot\sigma_t\varepsilon\mid x_t=x]
=\dot\alpha_t\,\mathbb E[x_0\mid x_t=x]+\dot\sigma_t\,\mathbb E[\varepsilon\mid x_t=x].
\end{equation}
From \(x_t=\alpha_t x_0+\sigma_t\varepsilon\), we have
\begin{equation}
\mathbb E[x_0\mid x_t=x]
=\mathbb E\!\left[\frac{x_t-\sigma_t\varepsilon}{\alpha_t}\Big|x_t=x\right]
=\frac{x}{\alpha_t}-\frac{\sigma_t}{\alpha_t}\mathbb E[\varepsilon\mid x_t=x].
\end{equation}
Substituting and collecting terms gives
\begin{equation}
v_t(x_t)=\frac{\dot\alpha_t}{\alpha_t}x+\Big(\dot\sigma_t-\frac{\dot\alpha_t\sigma_t}{\alpha_t}\Big)\mathbb E[\varepsilon\mid x_t=x].
\end{equation}
Finally, replacing \(\mathbb E[\varepsilon\mid x_t=x]\) by \(-\sigma_t\nabla_x\log p_t(x)\) yields (with \(\lambda_t\) absorbed into the coefficient)
\begin{equation}
v_t(x_t)=\frac{\dot\alpha_t}{\alpha_t}x
-\sigma_t\Big(\dot\sigma_t-\frac{\dot\alpha_t\sigma_t}{\alpha_t}\Big)\nabla_x\log p_t(x)
\end{equation}
For Rectified Flow~\cite{liu2022flow}, \(\alpha_t = 1-t\) and \(\sigma_t=t\). Then the marginal score can be expressed in terms of the velocity field as
\begin{equation}
\nabla_x\log p_t(x)= -\frac{x}{t}-\frac{1-t}{t}v_t(x).
\label{eq.rf_score}
\end{equation}

Substituting \eqref{eq.rf_score} into \eqref{eq:vel-sde} yields the Rectified-Flow-specific SDE:
\begin{equation}
\mathrm{d}x_t
=
\left[
v_t(x_t)+\frac{g^2(t)}{2t}\bigl(x_t+(1-t)v_t(x_t)\bigr)
\right]\mathrm{d}t
+g(t)\,\mathrm{d}\bar w_t .
\label{eq:sde-rf}
\end{equation}
Therefore, Rectified Flow can be embedded into the SDE framework.

Next, we apply the Euler--Maruyama discretization to \eqref{eq:sde-rf}, which induces a one-step Gaussian Markov transition
\begin{align}
q(x_{t-1}\mid x_t)
&= \mathcal{N}\!\Bigl(
x_{t-1};\,
\mu(x_t,t),\,
g^2(t)I
\Bigr),\nonumber \\
\mu(x_t,t)
&= x_t - \Bigl[v_t(x_t,t)+\frac{g^2(t)}{2t}\bigl(x_t+(1-t)v_t(x_t,t)\bigr)\Bigr].
\label{eq:true-cond-sde}
\end{align}

Substituting this mean $\mu(x_t,t)$ into \eqref{eq:unified-dpo-sde-final}, we obtain
\begin{align}
    L_\text{DPO-RF}(\theta) 
    & = -\mathbb{E}_{\substack{
    (x^w_0, x^l_0, c) \sim \mathcal{D},\; t \sim \mathcal{U}(0,T),\\
    x^w_t \sim q(x^w_t \mid x^w_0),\; x^l_t \sim q(x^l_t \mid x^l_0)
}}\Bigl[
        \log \sigma\Bigl(
          - \frac{\beta T}{2g^2(t)} \Bigl(1+\frac{g^2(t)(1-t)}{2t}\Bigr)^2 \bigl(\nonumber \\
    &\quad \|v^w - v_\theta(x_t^w,t,c)\|^2_2
    - \|v^w - v_\text{ref}(x_t^w,t,c)\|^2_2 - \bigl(\|v^l - v_\theta(x_t^l,t,c)\|^2_2
    - \|v^l - v_\text{ref}(x_t^l,t,c)\|^2_2\bigr)
    \bigr)
    \Bigr)\Bigr],
\end{align}
which is the DPO objective specialized to Rectified Flow.

\section{Deriving the Gradient of the Unified-DPO Objective}
\label{app.gradient}
The loss function of Unified-DPO is:
\begin{equation}
\begin{aligned}
\mathcal{L}(\theta)
&= -\,\mathbb{E}_{(x_0^w,x_0^l,c)\sim\mathcal{D},\; t\sim\mathcal{U}(0,T),\;
      x_t^w\sim q(x_t\mid x_0^w),\; x_t^l\sim q(x_t\mid x_0^l)}
\\
&\qquad
\log \sigma\Bigl(
    -\beta T\,\lambda(t)\bigl(
        \|y^{w} - y_{\theta}(x_t^{w},t,c)\|_2^2
        - \|y^{w} - y_{\mathrm{ref}}(x_t^{w},t,c)\|_2^2
\\
&\qquad\qquad\qquad\qquad
        - (\|y^{l} - y_{\theta}(x_t^{l},t,c)\|_2^2
        - \|y^{l} - y_{\mathrm{ref}}(x_t^{l},t,c)\|_2^2)
    \bigr)
\Bigr).
\end{aligned}
\end{equation}
In short, we define $u = -\beta T\,\lambda(t)\,\bigl(
        \|y^{w} - y_{\theta}(x_t^{w},t,c)\|_2^2
        - \|y^{w} - y_{\mathrm{ref}}(x_t^{w},t,c)\|_2^2
        - (\|y^{l} - y_{\theta}(x_t^{l},t,c)\|_2^2
        - \|y^{l} - y_{\mathrm{ref}}(x_t^{l},t,c)\|_2^2)\bigr)$,
then we derive the gradient of $\mathcal{L}(\theta)$:
\begin{align}
\nabla_\theta\mathcal{L}(\theta)
&= -\nabla_\theta\,\mathbb{E}_{(x_0^w,x_0^l,c)\sim\mathcal{D},\; t\sim\mathcal{U}(0,T),\;
      x_t^w\sim q(x_t\mid x_0^w),\; x_t^l\sim q(x_t\mid x_0^l)}\Bigl[\log \sigma\bigl(u\bigr)\Bigr] \nonumber \\
&= -\,\mathbb{E}_{(x_0^w,x_0^l,c)\sim\mathcal{D},\; t\sim\mathcal{U}(0,T),\; 
      x_t^w\sim q(x_t\mid x_0^w),\; x_t^l\sim q(x_t\mid x_0^l)}\Bigl[\nabla_\theta\log \sigma\bigl(u\bigr)\Bigr]\nonumber \\
&= -\,\mathbb{E}_{(x_0^w,x_0^l,c)\sim\mathcal{D},\; t\sim\mathcal{U}(0,T),\; 
      x_t^w\sim q(x_t\mid x_0^w),\; x_t^l\sim q(x_t\mid x_0^l)}\Bigl[\frac{\sigma'(u)}{\sigma(u)}\nabla_\theta(u)
\Bigr] \nonumber \\
&= -\,\mathbb{E}_{(x_0^w,x_0^l,c)\sim\mathcal{D},\; t\sim\mathcal{U}(0,T),\; 
      x_t^w\sim q(x_t\mid x_0^w),\; x_t^l\sim q(x_t\mid x_0^l)}\Bigl[\frac{\sigma(u)(1-\sigma(u))}{\sigma(u)}\nabla_\theta(u)
\Bigr] \nonumber \\
&= -\,\mathbb{E}_{(x_0^w,x_0^l,c)\sim\mathcal{D},\; t\sim\mathcal{U}(0,T),\; 
      x_t^w\sim q(x_t\mid x_0^w),\; x_t^l\sim q(x_t\mid x_0^l)}\Bigl[\sigma(-u)\nabla_\theta(u)
\Bigr] 
\end{align}

\begin{align}
    \nabla_\theta(u) &= \nabla_\theta \Bigl[-\beta T\,\lambda(t)\Bigl(
        \|y^{w} - y_{\theta}(x_t^{w},t,c)\|_2^2
        - \|y^{w} - y_{\mathrm{ref}}(x_t^{w},t,c)\|_2^2
        - \bigl(\|y^{l} - y_{\theta}(x_t^{l},t,c)\|_2^2
        - \|y^{l} - y_{\mathrm{ref}}(x_t^{l},t,c)\|_2^2\bigr)\Bigr)\Bigr]\nonumber \\
&=  -\beta T\,\lambda(t)\nabla_\theta\Bigl(
        \|y^{w} - y_{\theta}(x_t^{w},t,c)\|_2^2
        - \|y^{w} - y_{\mathrm{ref}}(x_t^{w},t,c)\|_2^2
        - \bigl(\|y^{l} - y_{\theta}(x_t^{l},t,c)\|_2^2
        - \|y^{l} - y_{\mathrm{ref}}(x_t^{l},t,c)\|_2^2\bigr)\Bigr)\nonumber \\
&=  -\beta T\,\lambda(t)\bigl(\nabla_\theta
        \|y^{w} - y_{\theta}(x_t^{w},t,c)\|_2^2
        - \nabla_\theta\|y^{l} - y_{\theta}(x_t^{l},t,c)\|_2^2\bigr)
\end{align}

Finally:
\begin{align}
\nabla_\theta\mathcal{L}(\theta)
&=\mathbb{E}_{(x_0^w,x_0^l,c)\sim\mathcal{D},\; t\sim\mathcal{U}(0,T),\;
      x_t^w\sim q(x_t\mid x_0^w),\; x_t^l\sim q(x_t\mid x_0^l)}\Bigl[\beta T\,\lambda(t)\sigma\bigl(-u\bigr) \nonumber \\
& \quad \bigl(\nabla_\theta
        \|y^{w} - y_{\theta}(x_t^{w},t,c)\|_2^2
        - \nabla_\theta\|y^{l} - y_{\theta}(x_t^{l},t,c)\|_2^2\bigr)
      \Bigr].
\end{align}

\section{HPDv3 Filtered Subset}
\label{app.hpsv3}
Human Preference Dataset v3 (HPDv3)~\cite{ma2025hpsv3} is a wide-spectrum human preference dataset for evaluating and aligning text-to-image models. It contains 1.08M text--image pairs and 1.17M annotated pairwise preference comparisons.

Although HPDv3 expands coverage by including generations from a broad set of state-of-the-art models as well as real-world photographs spanning a wide quality range, it still inherits a substantial portion of legacy diffusion-model data from earlier releases, such as HPDv2~\cite{wu2023human}.

In practice, datasets such as Pick-a-Pic~\cite{kirstain2023pick} and HPDv2 are widely used for alignment studies of diffusion backbones like SD1.5 and SDXL. However, for more capable flow-matching models with an MMDiT backbone, this lower-quality, legacy-dominated data distribution can be suboptimal and may degrade performance. Therefore, we construct a filtered subset of HPDv3 (HPDv3-sub) by retaining only comparisons whose images come from the following sources: \textbf{\textit{FLUX.1-dev~\cite{flux2024}, Kolors~\cite{team2024kolors}, Stable Diffusion 3-Medium~\cite{esser2024scaling}, HunyuanDiT~\cite{li2024hunyuan}, and real images}}, resulting in \textbf{210,008} pairwise preference comparisons.

\section{Implementation Details}
\label{app.imple}
We train Linear-DPO on 8 NVIDIA GPUs using mixed precision (FP16), with a batch size of 1 per GPU and gradient accumulation of 16, yielding an effective batch size of 128. For all models, we use the AdamW~\cite{loshchilov2017decoupled} optimizer with a learning rate of $5\times 10^{-6}$, similar to SFT, and use 200 warmup steps. SD1.5 and SDXL are trained on the full Pick-a-Pic v2 dataset, while SD3-M is trained on the high-quality HPDv3-sub for 2 epochs. We select the best-performing checkpoint for evaluation. For each model, we search $\bar{\beta}$ in the range $\bar{\beta}\in[100,2000]$ and obtain 250 for SD1.5 and 500 for both SDXL and SD3-M.

We use official pretrained checkpoints whenever available. For methods that provide training code but no pretrained weights, we reproduce the models using the authors' implementations for evaluation. For fair comparison, we fix the random seed, use 50 sampling steps, adopt the default classifier-free guidance~\cite{ho2022classifier}, and use the remaining default settings of each model in Diffusers~\cite{von-platen-etal-2022-diffusers} to generate images for evaluation.

\section{Other Ablation Studies}
\subsection{Additional Analysis of Utility Functions}
\label{app.utility_extra}
Figure~\ref{fig.utility}(a) compares four normalized utility functions used to map the margin \(x\) (i.e., the score difference between preferred and dispreferred samples) into a bounded scalar signal. Although all utilities are normalized to \([0,1]\) via \(\frac{U(x)-U(-5)}{U(5)-U(-5)}\) followed by clipping, their local slopes and asymmetries induce markedly different optimization behaviors.

\paragraph{Kahneman--Tversky utility (\(\sigma(x)\)).}
The Kahneman--Tversky form is \emph{symmetric} around \(x=0\) and concentrates most of its gradient near the decision boundary. This is beneficial when many training pairs are near ties, as it focuses updates on ambiguous cases and quickly pushes the model to separate preferred and dispreferred samples by enlarging the margin. Such behavior is particularly well aligned with preference optimization in NLP, where the objective is often to increase the probability gap (or logit margin) between chosen and rejected outputs; once they are clearly separated, a rapid decay of gradients is usually acceptable. On the other hand, the function saturates for large \(|x|\), which reduces gradient contributions from confidently ranked pairs and can lead to premature plateaus in practice, limiting late-stage refinement.

\paragraph{Loss-averse utility (\(\log \sigma(x)\)).}
The loss-averse utility is strongly \emph{asymmetric}: it increases rapidly once \(x\) becomes moderately positive, effectively amplifying updates on already-preferred pairs. While this can speed up short-term gains, it may also induce overly aggressive optimization, increase sensitivity to noisy preferences, and destabilize training (e.g., by encouraging collapse toward a narrow set of high-scoring modes).

\paragraph{Risk-seeking utility (\(-\log \sigma(-x)\)).}
The risk-seeking utility grows slowly for negative to moderately positive margins and increases sharply only for large positive \(x\). This makes the learning signal conservative in early training and can yield slower improvements, since many pairs contribute limited gradients until the model already achieves large margins. Its steeper growth at high \(x\) also makes it more sensitive to large-margin outliers.

\paragraph{Linear utility.}
The linear utility yields a nearly constant slope over the effective margin range, providing a stable and well-conditioned learning signal. Unlike sigmoid-shaped utilities that progressively saturate and concentrate gradients around a narrow margin band, the linear form maintains non-vanishing updates for both moderately and strongly preferred pairs. This is advantageous for generative fine-tuning, where quality improvements are cumulative and distributed: even when a sample is already preferred overall, it may still contain localized defects that require continued correction. By keeping gradient contributions broad and consistent, the linear utility supports sustained refinement throughout training, reduces sensitivity to the evolving margin distribution, and mitigates the tendency toward abrupt, overconfident updates that can harm diversity or stability.

\subsection{Impact of Data Quality}
Training data quality is a critical determinant of performance for text-to-image (T2I) models. In this section, we investigate how training data quality affects different optimization methods, namely SFT, Diffusion-DPO, and our proposed Linear-DPO. We conduct experiments on SD3-M using three datasets of progressively increasing quality (Pick-a-Pic v2, HPDv3, and HPDv3-sub), and report the corresponding HPSv3 scores for each method in Table~\ref{tab.data}. The results show that SFT yields a consistent drop in HPSv3 scores across all three datasets. In contrast, DPO-style methods are markedly more stable: by contrasting preferred and dispreferred pairs, they prioritize learning to discriminate between high- and low-quality outputs rather than simply imitating the training distribution. Notably, as training data quality improves, Linear-DPO exhibits an increasingly pronounced advantage over the other methods.

\begin{table}[htbp]
    \centering
    \small
    \caption{Impact of training data quality on performance: HPSv3 scores of SFT, Diffusion-DPO and Linear-DPO on SD3-M across datasets with increasing quality (Pick-a-Pic v2, HPDv3, HPDv3-Sub).}
    \begin{tabular}{l|ccc}
    \toprule
    & \multicolumn{3}{c}{Dataset} \\
     Method & \textit{Pickapic}& \textit{HPDv3} & \textit{HPDv3-Sub}\\
    \midrule
    SFT & 11.347 \color{orange}$ \downarrow$ & 11.764 \color{orange}$ \downarrow$ &11.992 \color{orange}$ \downarrow$ \\
    Diffusion-DPO&11.592 \color{orange}$ \downarrow$&12.312 \color{green}$ \uparrow$&13.270 \color{green}$ \uparrow$\\
    \textbf{Linear-DPO(Ours)}&11.844 \color{orange}$ \downarrow$&12.460 \color{green}$ \uparrow$&\textbf{13.623} \color{green}$\uparrow$\\ 
    \bottomrule
    \end{tabular}
    \label{tab.data}
\end{table}

\subsection{Effect of different $\bar{\beta}$ in Linear-DPO}

$\beta$ is a critical hyperparameter in DPO, which acts as a scaling factor that regulates the constraint between the policy and reference models, thereby controlling the degree to which the updated policy model deviates from the reference model.  To investigate the impact of \(\bar{\beta}\) on preference alignment and identify its optimal value, we conduct experiments with \(\bar{\beta} \in \{100, 250, 500, 1000, 2000\}\).

We train Linear-DPO on subsets of Pick-a-Pic v2 with different \(\bar{\beta}\) values for both SD1.5 and SDXL, and report PickScore for images generated from prompts in the test split of Pick-a-Pic v2. Table~\ref{tab.beta_sd1.5xl} shows that the optimal \(\bar{\beta}\) is 250 for SD1.5 and 500 for SDXL.

For SD3-M, we conduct experiments on the HPDv3-sub training set and report HPSv3 scores for images generated from prompts in the HPDv2 test set. As shown in Table~\ref{tab.beta_sd3}, different \(\bar{\beta}\) values have a substantial impact on performance, and the best HPSv3 score is achieved at \(\bar{\beta}=500\).

\begin{table*}[htbp]
    \centering
    \caption{Pickscore rewards of Linear-DPO with different $\bar{\beta}$ values on SD1.5 and SDXL trained on Pick-a-Pic v2 subsets and evaluated on unique test split prompts of Pick-a-Pic v2.}
    \begin{tabular}{c|ccccc}
    \toprule
        $\bar{\beta}$ & 100& 250&500&1000&2000  \\
    \midrule
    SD1.5&0.2132&\textbf{0.2164}&0.2141&0.2157&0.2143  \\
    SDXL&0.2248&0.2255&\textbf{0.2270}&0.2259&0.2238 \\
    \bottomrule
    \end{tabular}
    \label{tab.beta_sd1.5xl}
\end{table*}

\begin{table*}[htbp]
    \caption{HPSv3 reward scores of Linear-DPO with different $\bar{\beta}$ values on SD3-M trained on HPDv3-sub training dataset and evaluated on test prompts of HPDv2.}
    \centering
    \begin{tabular}{c|ccccc}
    \toprule
    $\bar{\beta}$&100&250&500&1000&2000  \\
    \midrule
    SD3-M& 12.2513& 12.4935&\textbf{12.7845}&12.6673&12.3891\\
    \bottomrule
    \end{tabular}
    \label{tab.beta_sd3}
\end{table*}
\section{More Quantitative Results}
\label{app.quantitative}
In Section~\ref{sec.results}, we report evaluation scores for each method when fine-tuned on SD1.5, and a subset of evaluation metrics for SDXL. In this section, we present the full set of evaluation metrics for SDXL and SD3-M in Tables~\ref{tab.sdxl} and~\ref{tab.sd3}. Consistent with the qualitative results and the findings in Tables~\ref{tab.sd15} and~\ref{tab.sdxl_short}, Linear-DPO exhibits a clear advantage over competing methods on the stronger, higher-resolution models SDXL and SD3-M.

For SD3-M, Table~\ref{tab.win_ratio} additionally reports the number of wins and the win ratio (shown in \textcolor{blue}{blue}) of Linear-DPO against the original SD3-M, SFT, and Diffusion-DPO. The comparison is performed on images generated from all three validation datasets (2,532 prompts in total), using the scores from each reward model, as well as the aggregated overall win rate. Linear-DPO is highly competitive and achieves a dominant win rate when evaluated by HPSv3.

\begin{table*}[thbp]
\caption{Reward score comparisons of baseline methods on the more powerful diffusion model SDXL evaluated across all six reward models. Against the original SDXL, SFT, Diffusion-DPO and MAPO, our method maintains a leading performance overall.}
\label{tab.sdxl}
\begin{center}
\begin{small}
\begin{tabular}{l|l|cccccc}
\toprule
\textbf{Dataset} & \textbf{Method} & \textbf{Pickscore} & \textbf{HPSv2}& \textbf{Aesthetics} &  \textbf{CLIP} & \textbf{Image Reward} & \textbf{HPSv3} \\
\midrule
\multirow{5}{*}{Pick-a-Pic v2} & SDXL & 0.2229 &0.2805&6.0618&0.3653&0.7637&6.6575\\
& SFT& 0.2171&0.2768&5.6932&0.3670&0.6529&6.0146\\
& Diffusion-DPO & 0.2271&0.2868&6.0370&0.3742&0.9475&7.1570\\
&MaPO& 0.2232& 0.2830& \textbf{6.2070}& 0.3628& 0.8529&6.8585\\
\rowcolor{gray!15}\cellcolor{white}& \textbf{Linear-DPO} & \textbf{0.2283}&\textbf{0.2884}&5.9959&\textbf{0.3777}&\textbf{1.0990}&\textbf{7.1615}\\
\midrule
\multirow{5}{*}{PartiPrompt}& SDXL & 0.2266&0.2839&5.7794&0.3569&0.7665&6.4288\\
& SFT&0.2214&0.2813&5.5763&0.3581&0.7226&5.8542\\
& Diffusion-DPO & 0.2295&0.2893&5.8068&0.3660&1.0788&\textbf{6.9971}\\
&MaPO&0.2268&0.2864&\textbf{5.9328}&0.3572&0.8999&6.6485\\
\rowcolor{gray!15}\cellcolor{white}& \textbf{Linear-DPO} &\textbf{0.2297}&\textbf{0.2906}&\underline{5.8069}&\textbf{0.3667}&\textbf{1.1816}&\underline{6.9953} \\ 
\midrule
\multirow{5}{*}{HPDv2} & SDXL&0.2288&0.2849&6.1320&0.3901&0.8750&10.6086\\
& SFT &0.2219&0.2824&5.8701&0.3880&0.8194&10.2284 \\
& Diffusion-DPO & 0.2325& 0.2907& 6.1544& \textbf{0.3948}& 1.1059&11.2931\\
& MaPO& 0.2295& 0.2889& \textbf{6.2298}& 0.3899& 0.9820&11.1285 \\
\rowcolor{gray!15}\cellcolor{white}&\textbf{Linear-DPO} & \textbf{0.2332}&\textbf{0.2924}&6.1410&\underline{0.3941}&\textbf{1.1876}&\textbf{11.3601}
\\
\bottomrule
\end{tabular}
\end{small}
\end{center}
\vskip -0.1in
\end{table*}
\begin{table*}[thbp]
\caption{Quantitative results of reward scores for SD3-M, SFT, Diffusion-DPO and our proposed Linear-DPO. The results demonstrate that our method is applicable not only to diffusion models but also to flow-matching models.}
\label{tab.sd3}
\begin{center}
\begin{small}
\begin{tabular}{l|l|cccccc}
\toprule
\textbf{Dataset} & \textbf{Method} & \textbf{Pickscore} & \textbf{HPSv2}& \textbf{Aesthetics} &  \textbf{CLIP} & \textbf{Image Reward} & \textbf{HPSv3} \\
\midrule
\multirow{4}{*}{Pick-a-Pic v2} & SD3-M &0.2211&0.2863&5.7624&\textbf{0.3513}&0.9633&7.1674  \\
& SFT&0.2196&0.2850&5.8017&0.3503&\textbf{1.0926}&6.9716\\
& Diffusion-DPO & 0.2227&0.2879&5.8348&0.3509&1.0220&7.7851 \\
\rowcolor{gray!15}\cellcolor{white}& \textbf{Linear-DPO} & \textbf{0.2234}&\textbf{0.2889}&\textbf{5.8824}&0.3498&\underline{1.0811}&\textbf{8.2166}\\
\midrule

\multirow{4}{*}{PartiPrompt}& SD3-M &0.2284&0.2935&5.5725&\textbf{0.3590}&1.1877&7.7376\\
& SFT&0.2259&0.2907&5.5923&0.3548&1.2066&7.1503\\
& Diffusion-DPO &0.2289&0.2938&5.5949&0.3574&1.1893&8.1328 \\
\rowcolor{gray!15}\cellcolor{white}& \textbf{Linear-DPO} &\textbf{0.2295}&\textbf{0.2962}&\textbf{5.6779}&\underline{0.3576}&\textbf{1.2465}&\textbf{8.5791} \\ 
\midrule
\multirow{4}{*}{HPDv2} & SD3-M & 0.2267&0.2929&5.9499&\textbf{0.3666}&1.1721&12.1796\\
& SFT & 0.2240&0.2912&5.8949&0.3606&1.1833&11.9915\\
& Diffusion-DPO & 0.2272& 0.2934&5.9995&0.3654&1.1638&12.8444\\
\rowcolor{gray!15}\cellcolor{white}&\textbf{Linear-DPO} & \textbf{0.2288}&\textbf{0.2967}&\textbf{6.0497}&0.3638&\textbf{1.2338}&\textbf{13.6232}
\\
\bottomrule
\end{tabular}
\end{small}
\end{center}
\vskip -0.1in
\end{table*}
\begin{table*}[hbtp]
    \centering
    \small
    \caption{Statistics on three validation datasets and overall based on automated evaluation using HPSv3 reward scores, showing the number of wins (and \textcolor{blue}{win ratios}) of Linear-DPO against the original SD3-M, SFT, and Diffusion-DPO, where all methods are fine-tuned from SD3-M.}
    \begin{tabular}{c|l|cccccc}
    \toprule
         \textbf{Dataset}& \textbf{Method}& \textbf{PickScore}&\textbf{HPSv2}&\textbf{Aesthetics}&\textbf{CLIP}&\textbf{Image Reward}& \textbf{HPSv3}  \\
         \midrule
         \multirow{3}{*}{\shortstack{Pick-a-Pic v2\\(500)}}&SD3-M&325\,(\textcolor{blue}{65.0\%})&319\,(\textcolor{blue}{63.8\%})&303\,(\textcolor{blue}{60.6\%})&253\,(\textcolor{blue}{50.6\%})&294\,(\textcolor{blue}{58.8\%})&402\,(\textcolor{blue}{80.4\%})\\
         &SFT&358\,(\textcolor{blue}{71.6\%})&322\,(\textcolor{blue}{64.4\%})&300\,(\textcolor{blue}{60.0\%})&240\,(\textcolor{blue}{48.0\%})&254\,(\textcolor{blue}{50.8\%})&410\,(\textcolor{blue}{82.0\%})\\
         &Diffusion-DPO&289\,(\textcolor{blue}{57.8\%})&262\,(\textcolor{blue}{52.4\%})&277\,(\textcolor{blue}{55.4\%})&246\,(\textcolor{blue}{49.2\%})&287\,(\textcolor{blue}{57.4\%})&317\,(\textcolor{blue}{63.4\%})\\
         \midrule
         \multirow{3}{*}{\shortstack{PartiPrompt\\(1632)}}& SD3-M&946\,(\textcolor{blue}{58.0\%})&988\,(\textcolor{blue}{60.5\%})&1028\,(\textcolor{blue}{63.0\%})&810\,(\textcolor{blue}{49.6\%})&994\,(\textcolor{blue}{60.9\%})&1226\,(\textcolor{blue}{75.1\%})\\
         &SFT&1185\,(\textcolor{blue}{72.6\%})&1173\,(\textcolor{blue}{71.9\%})&970\,(\textcolor{blue}{59.4\%})&880\,(\textcolor{blue}{53.9\%})&914\,(\textcolor{blue}{56.0\%})&1368\,(\textcolor{blue}{83.8\%})\\
         &Diffusion-DPO&905\,(\textcolor{blue}{55.5\%})&1006\,(\textcolor{blue}{61.6\%})&982\,(\textcolor{blue}{60.2\%})&829\,(\textcolor{blue}{50.8\%})&927\,(\textcolor{blue}{56.8\%})&1030\,(\textcolor{blue}{63.1\%})\\
         \midrule
         \multirow{3}{*}{\shortstack{HPDv2\\(400)}}&SD3-M&256\,(\textcolor{blue}{64.0\%})&269\,(\textcolor{blue}{67.2\%})&238\,(\textcolor{blue}{59.5\%})&192\,(\textcolor{blue}{48.0\%})&237\,(\textcolor{blue}{59.2\%})&356\,(\textcolor{blue}{89.0\%}) \\
         &SFT&311\,(\textcolor{blue}{77.8\%})&291\,(\textcolor{blue}{72.8\%})&267\,(\textcolor{blue}{66.8\%})&208\,(\textcolor{blue}{52.0\%})&221\,(\textcolor{blue}{55.3\%})&352\,(\textcolor{blue}{88.0\%})\\
         &Diffusion-DPO&242\,(\textcolor{blue}{60.5\%})&256\,(\textcolor{blue}{64.0\%})&212\,(\textcolor{blue}{53.0\%})&195\,(\textcolor{blue}{48.8\%})&217\,(\textcolor{blue}{54.2\%})&280\,(\textcolor{blue}{70.0\%})\\
         \midrule
        \multirow{3}{*}{\shortstack{Overall\\(2532)}}&SD3-M&1527\,(\textcolor{blue}{60.3\%})&1576\,(\textcolor{blue}{62.2\%})&1569\,(\textcolor{blue}{62.0\%})&1255\,(\textcolor{blue}{49.6\%})&1525\,(\textcolor{blue}{60.2\%})&1984\,(\textcolor{blue}{78.4\%})\\
        &SFT&1854\,(\textcolor{blue}{73.2\%})&1786\,(\textcolor{blue}{70.5\%})&1537\,(\textcolor{blue}{60.7\%})&1328\,(\textcolor{blue}{52.4\%})&1389\,(\textcolor{blue}{54.9\%})&2130\,(\textcolor{blue}{84.1\%})\\
        &Diffusion-DPO&1436\,(\textcolor{blue}{56.7\%})&1524\,(\textcolor{blue}{60.2\%})&1471\,(\textcolor{blue}{58.1\%})&1270\,(\textcolor{blue}{50.2\%})&1431\,(\textcolor{blue}{56.5\%})&1627\,(\textcolor{blue}{64.3\%})\\
    \bottomrule
    \end{tabular}
    \label{tab.win_ratio}
\end{table*}

\section{More Qualitative Results}
\begin{figure*}[htbp]
  \centering
  \begin{minipage}{\textwidth}
  \centering
  \begin{minipage}{0.25\textwidth} \centering \small \textbf{Prompt} 
  \end{minipage}%
  \begin{minipage}{0.125\textwidth} \centering \small \textbf{SD1.5} \end{minipage}%
  \begin{minipage}{0.125\textwidth} \centering \small \textbf{SFT} \end{minipage}%
  \begin{minipage}{0.125\textwidth} \centering \small \textbf{Diffusion-DPO} \end{minipage}%
  \begin{minipage}{0.125\textwidth} \centering \small \textbf{Diffusion-KTO} \end{minipage}%
  \begin{minipage}{0.125\textwidth} \centering \small \textbf{DSPO} \end{minipage}%
  \begin{minipage}{0.125\textwidth} \centering \small \textbf{Linear-DPO} \end{minipage}
  
  \vspace{0.2em}
  
  \begin{minipage}{0.25\textwidth}
  \begin{minipage}{0.90\textwidth}
    \centering \scriptsize \raggedright {Magic the gathering, anthro furry knight adventurer, showcase promo full art. Painted impressionist style} 
  \end{minipage}
  \end{minipage}%
  \begin{minipage}{0.125\textwidth}\includegraphics[width=\linewidth]{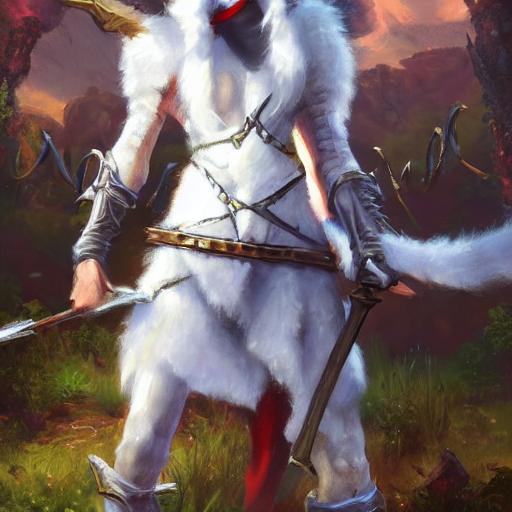}\end{minipage}%
  \begin{minipage}{0.125\textwidth}\includegraphics[width=\linewidth]{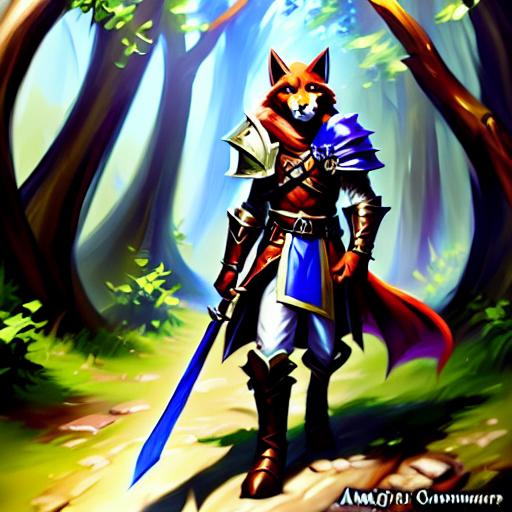}\end{minipage}%
  \begin{minipage}{0.125\textwidth}\includegraphics[width=\linewidth]{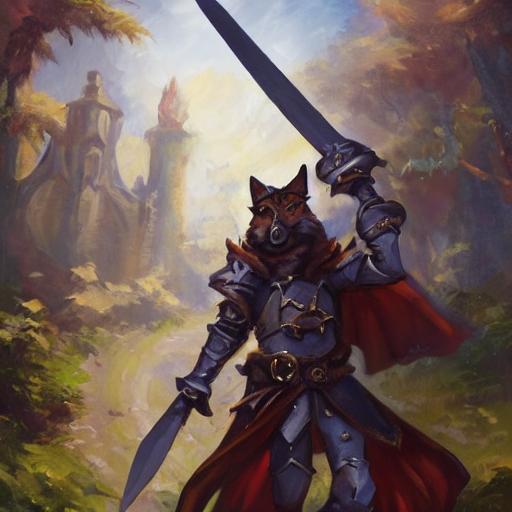}\end{minipage}%
  \begin{minipage}{0.125\textwidth}\includegraphics[width=\linewidth]{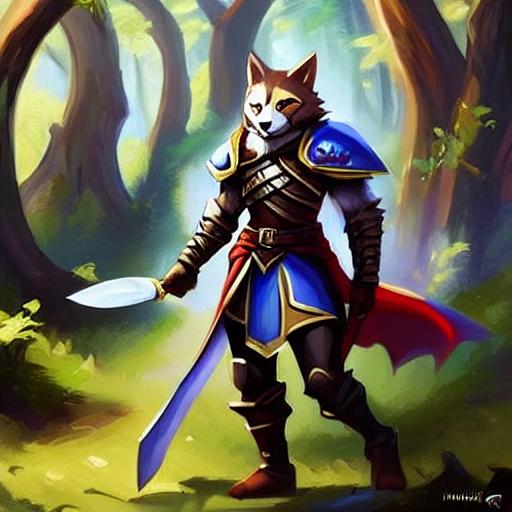}\end{minipage}%
  \begin{minipage}{0.125\textwidth}\includegraphics[width=\linewidth]{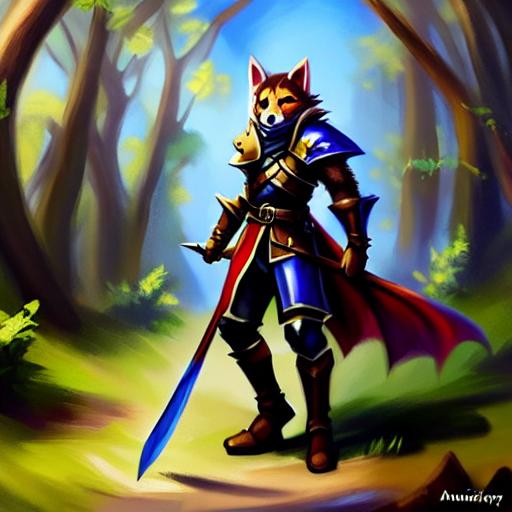}\end{minipage}%
  \begin{minipage}{0.125\textwidth}\includegraphics[width=\linewidth]{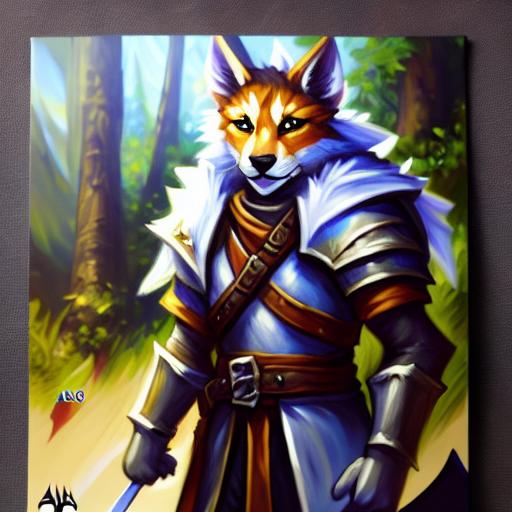}\end{minipage}

  \begin{minipage}{0.25\textwidth}
  \begin{minipage}{0.90\textwidth}
    \centering \scriptsize \raggedright {Photography of an anthropomorphic shark wearing a green velvet gucci suit, fashion photography, kodak gold 200, studio lighting, sharp} 
  \end{minipage}
  \end{minipage}%
  \begin{minipage}{0.125\textwidth}\includegraphics[width=\linewidth]{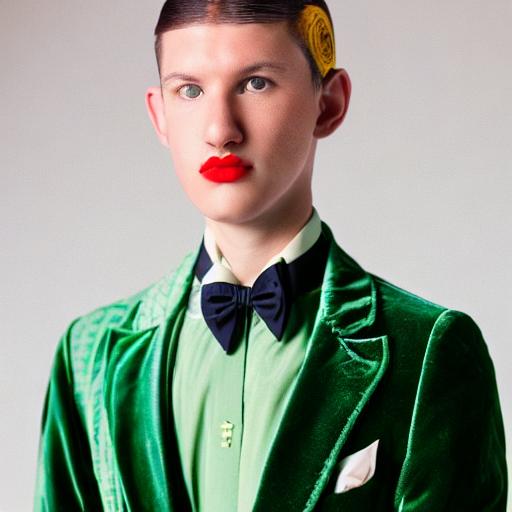}\end{minipage}%
  \begin{minipage}{0.125\textwidth}\includegraphics[width=\linewidth]{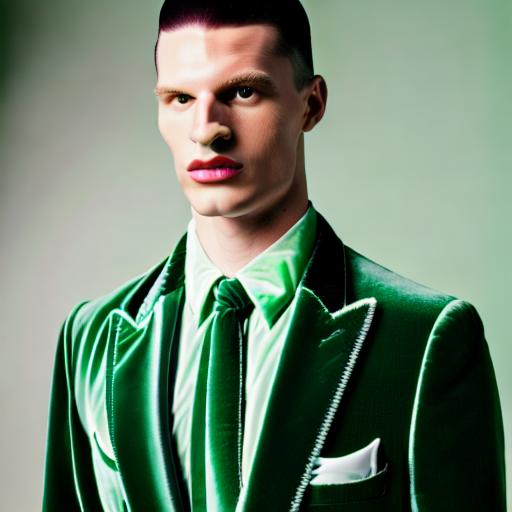}\end{minipage}%
  \begin{minipage}{0.125\textwidth}\includegraphics[width=\linewidth]{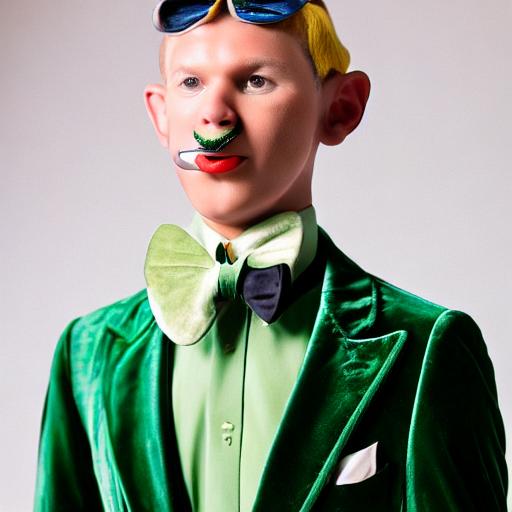}\end{minipage}%
  \begin{minipage}{0.125\textwidth}\includegraphics[width=\linewidth]{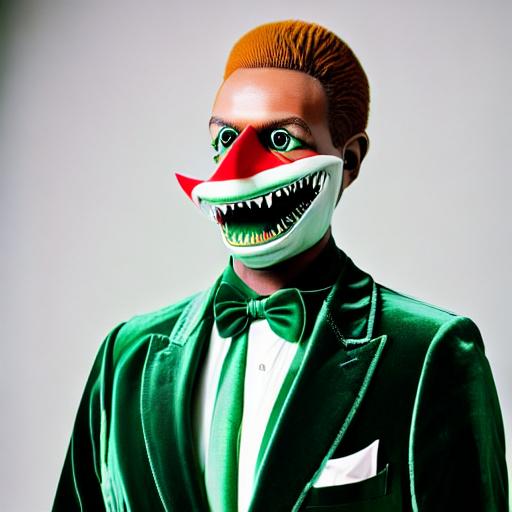}\end{minipage}%
  \begin{minipage}{0.125\textwidth}\includegraphics[width=\linewidth]{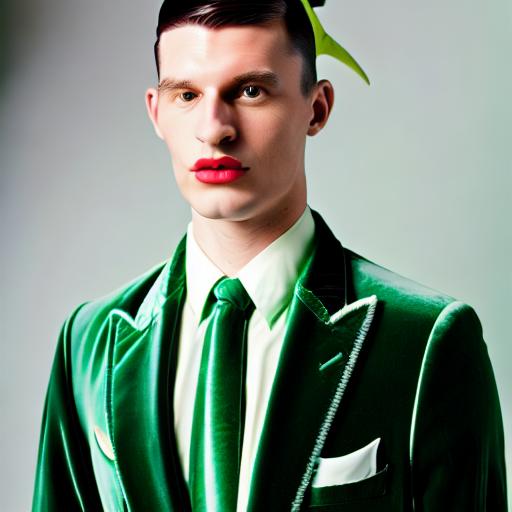}\end{minipage}%
  \begin{minipage}{0.125\textwidth}\includegraphics[width=\linewidth]{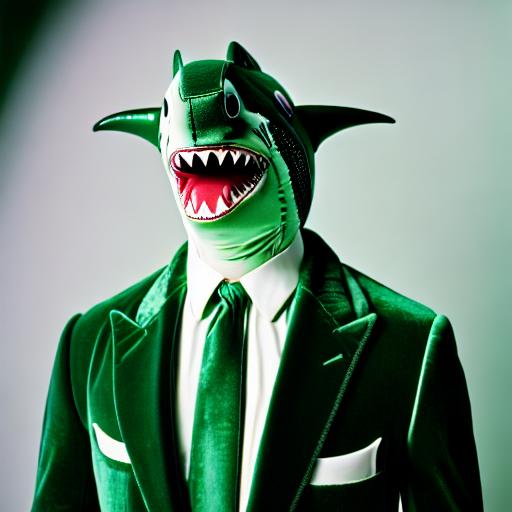}\end{minipage}

  \begin{minipage}{0.25\textwidth}
  \begin{minipage}{0.90\textwidth}
    \centering \scriptsize \raggedright {Big mansion in the daytime} 
  \end{minipage}
  \end{minipage}%
  \begin{minipage}{0.125\textwidth}\includegraphics[width=\linewidth]{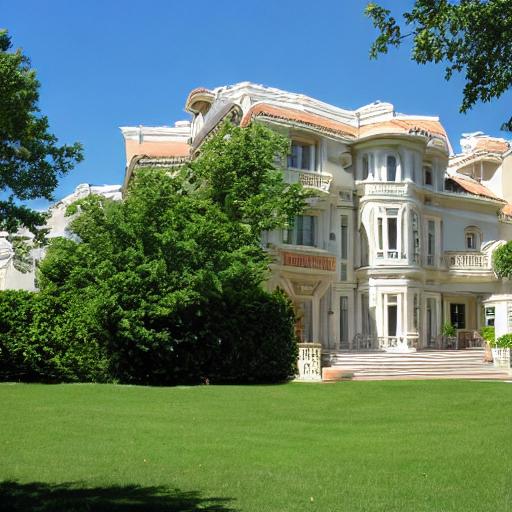}\end{minipage}%
  \begin{minipage}{0.125\textwidth}\includegraphics[width=\linewidth]{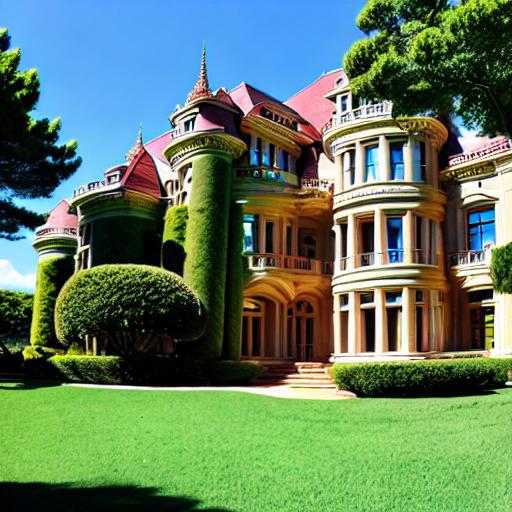}\end{minipage}%
  \begin{minipage}{0.125\textwidth}\includegraphics[width=\linewidth]{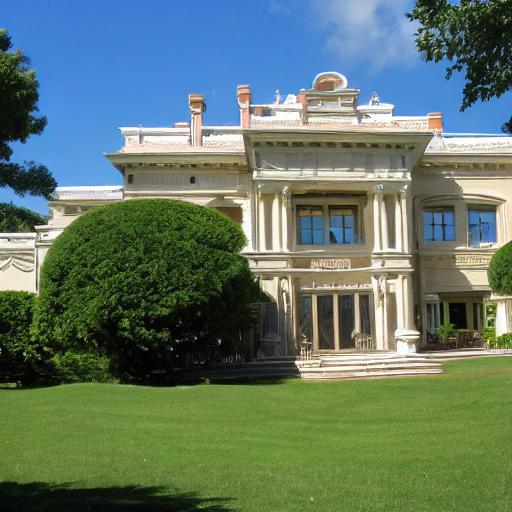}\end{minipage}%
  \begin{minipage}{0.125\textwidth}\includegraphics[width=\linewidth]{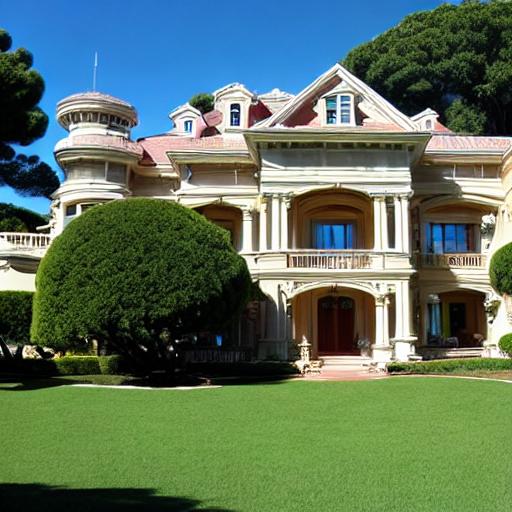}\end{minipage}%
  \begin{minipage}{0.125\textwidth}\includegraphics[width=\linewidth]{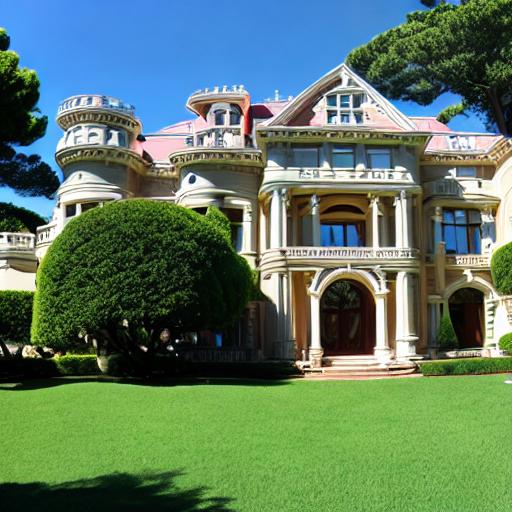}\end{minipage}%
  \begin{minipage}{0.125\textwidth}\includegraphics[width=\linewidth]{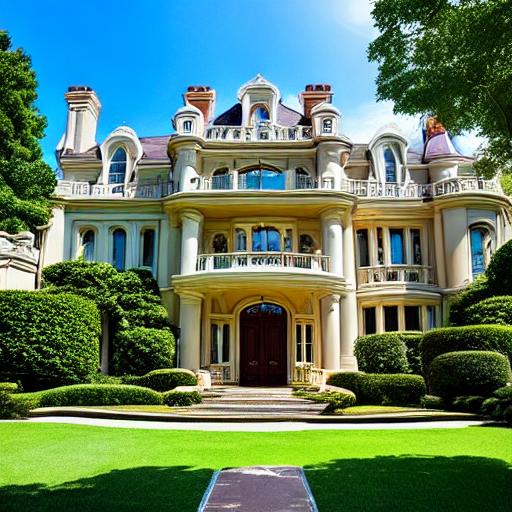}\end{minipage}
  
  \begin{minipage}{0.25\textwidth}
  \begin{minipage}{0.90\textwidth}
    \centering \scriptsize \raggedright {Cute anime girl} 
  \end{minipage}
  \end{minipage}%
  \begin{minipage}{0.125\textwidth}\includegraphics[width=\linewidth]{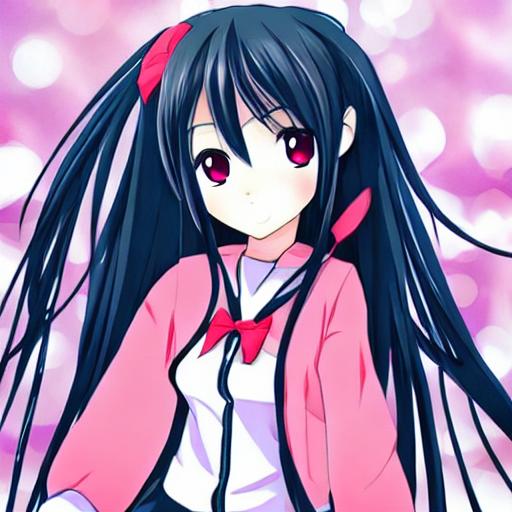}\end{minipage}%
  \begin{minipage}{0.125\textwidth}\includegraphics[width=\linewidth]{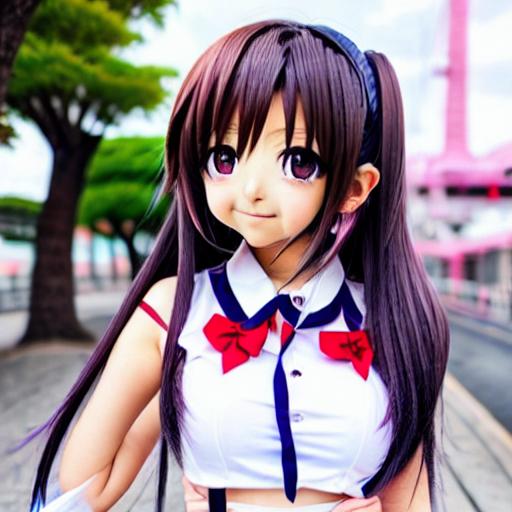}\end{minipage}%
  \begin{minipage}{0.125\textwidth}\includegraphics[width=\linewidth]{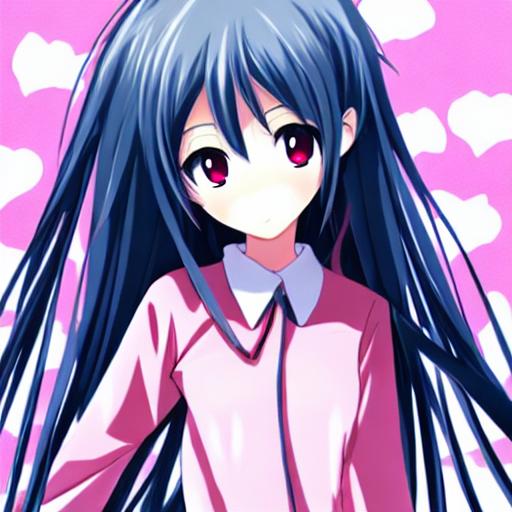}\end{minipage}%
  \begin{minipage}{0.125\textwidth}\includegraphics[width=\linewidth]{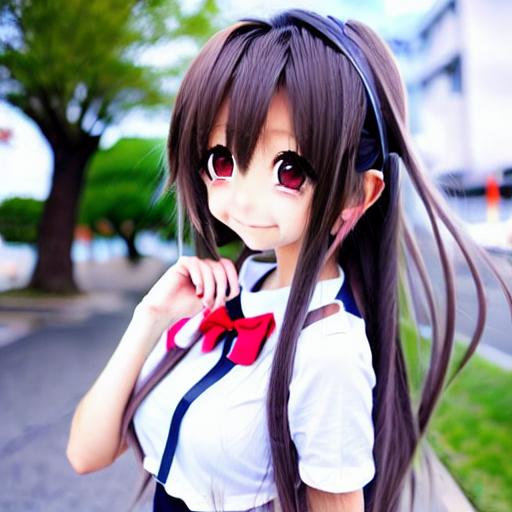}\end{minipage}%
  \begin{minipage}{0.125\textwidth}\includegraphics[width=\linewidth]{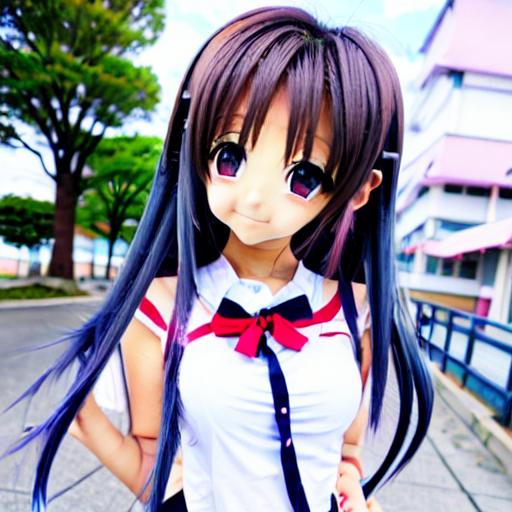}\end{minipage}%
  \begin{minipage}{0.125\textwidth}\includegraphics[width=\linewidth]{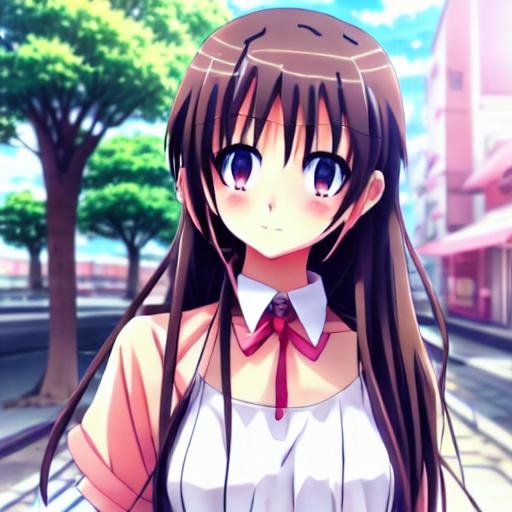}\end{minipage}

  \begin{minipage}{0.25\textwidth}
  \begin{minipage}{0.90\textwidth}
    \centering \scriptsize \raggedright {A panda as a human} 
  \end{minipage}
  \end{minipage}%
  \begin{minipage}{0.125\textwidth}\includegraphics[width=\linewidth]{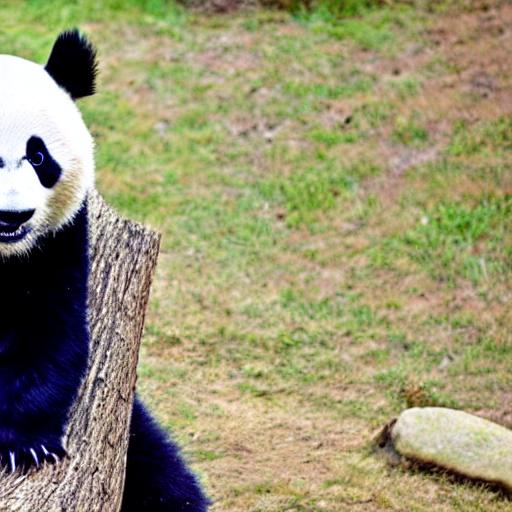}\end{minipage}%
  \begin{minipage}{0.125\textwidth}\includegraphics[width=\linewidth]{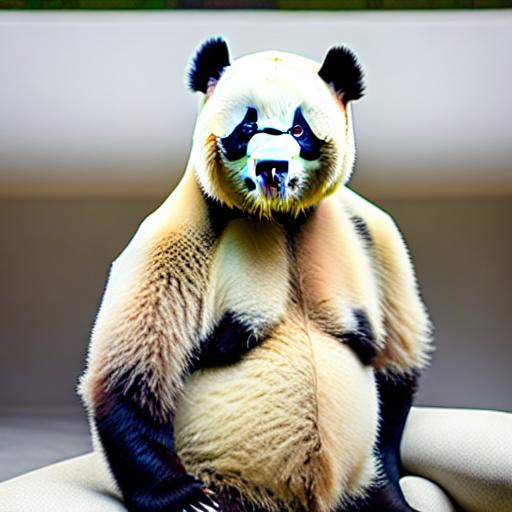}\end{minipage}%
  \begin{minipage}{0.125\textwidth}\includegraphics[width=\linewidth]{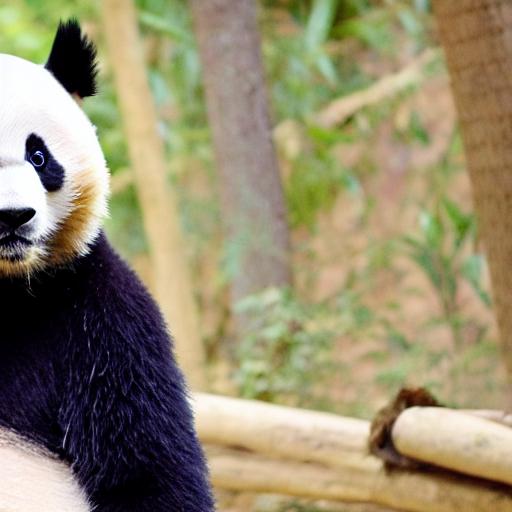}\end{minipage}%
  \begin{minipage}{0.125\textwidth}\includegraphics[width=\linewidth]{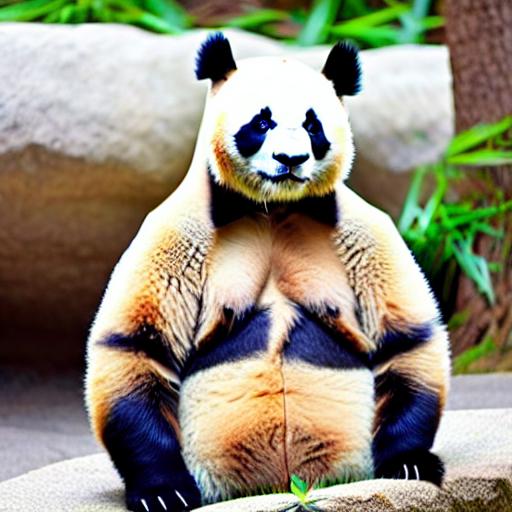}\end{minipage}%
  \begin{minipage}{0.125\textwidth}\includegraphics[width=\linewidth]{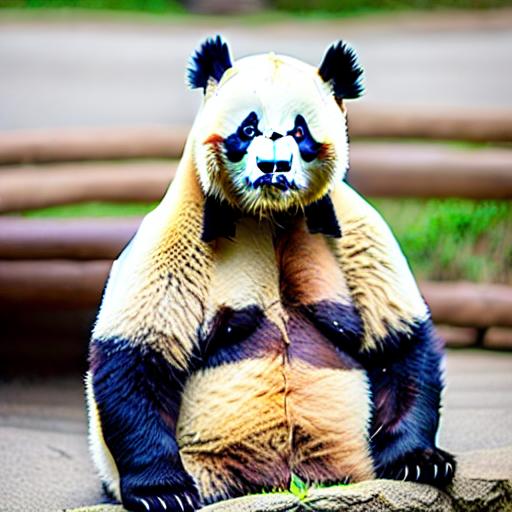}\end{minipage}%
  \begin{minipage}{0.125\textwidth}\includegraphics[width=\linewidth]{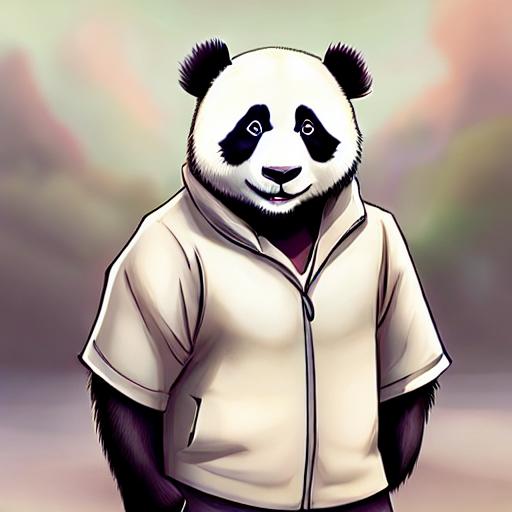}\end{minipage}

  \begin{minipage}{0.25\textwidth}
  \begin{minipage}{0.90\textwidth}
    \centering \scriptsize \raggedright {A raccoon riding an oversized fox through a forest in a furry art anime still.} 
  \end{minipage}
  \end{minipage}%
  \begin{minipage}{0.125\textwidth}\includegraphics[width=\linewidth]{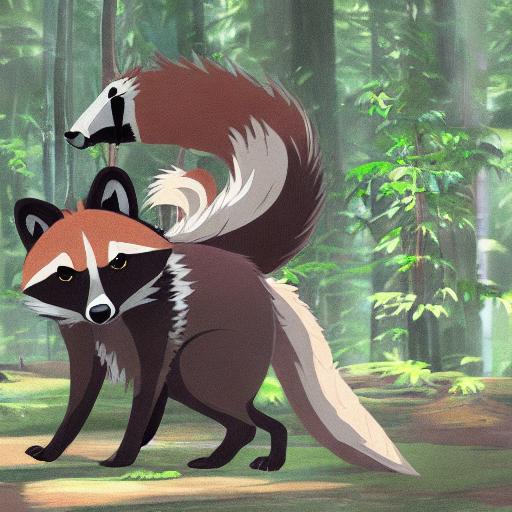}\end{minipage}%
  \begin{minipage}{0.125\textwidth}\includegraphics[width=\linewidth]{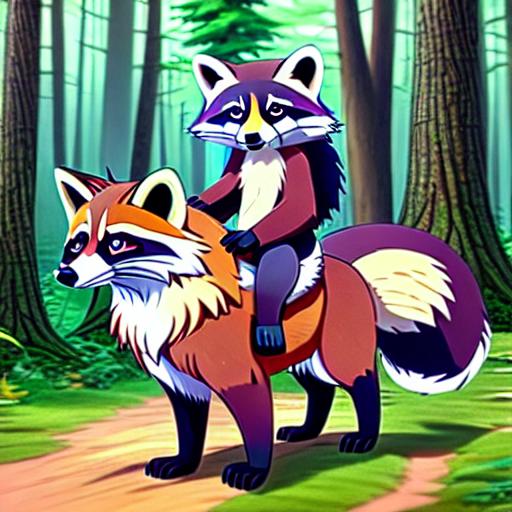}\end{minipage}%
  \begin{minipage}{0.125\textwidth}\includegraphics[width=\linewidth]{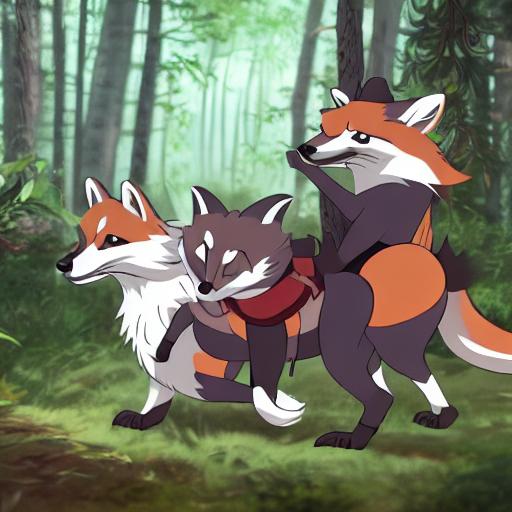}\end{minipage}%
  \begin{minipage}{0.125\textwidth}\includegraphics[width=\linewidth]{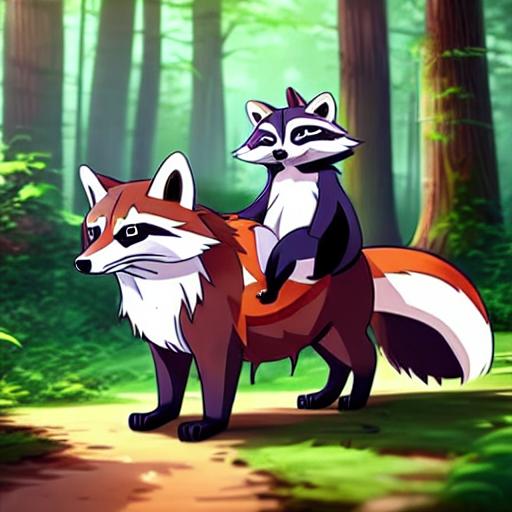}\end{minipage}%
  \begin{minipage}{0.125\textwidth}\includegraphics[width=\linewidth]{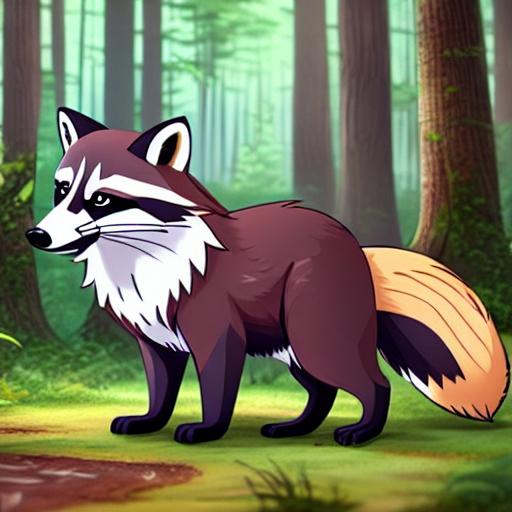}\end{minipage}%
  \begin{minipage}{0.125\textwidth}\includegraphics[width=\linewidth]{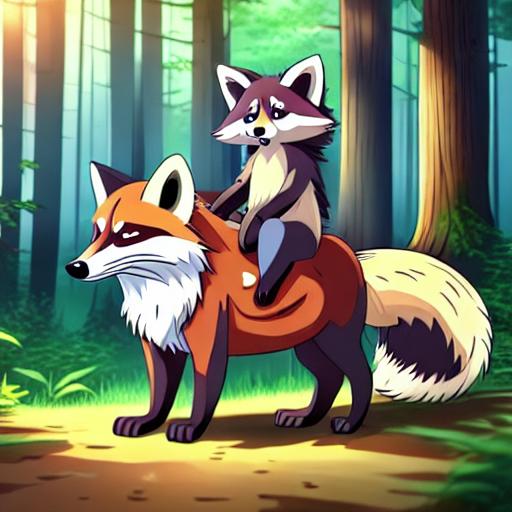}\end{minipage}
  
  \begin{minipage}{0.25\textwidth}
  \begin{minipage}{0.90\textwidth}
    \centering \scriptsize \raggedright {A key visual of a young female swat officer with a neon futuristic gas mask in a cyberpunk setting.} 
  \end{minipage}
  \end{minipage}%
  \begin{minipage}{0.125\textwidth}\includegraphics[width=\linewidth]{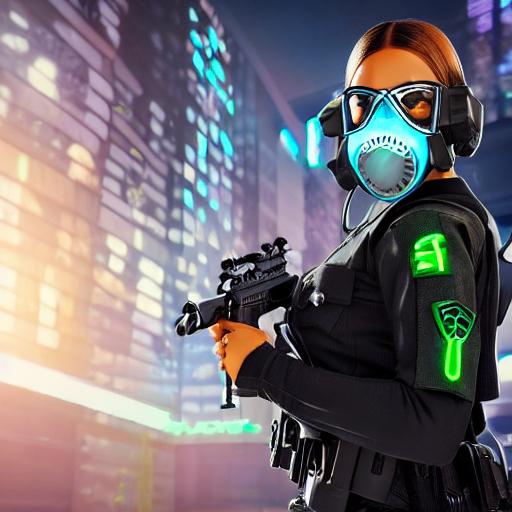}\end{minipage}%
  \begin{minipage}{0.125\textwidth}\includegraphics[width=\linewidth]{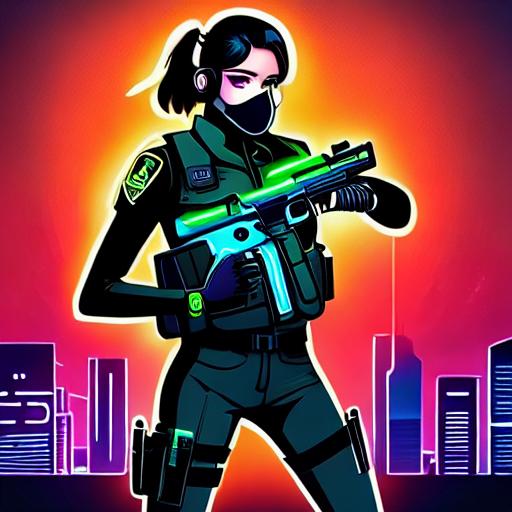}\end{minipage}%
  \begin{minipage}{0.125\textwidth}\includegraphics[width=\linewidth]{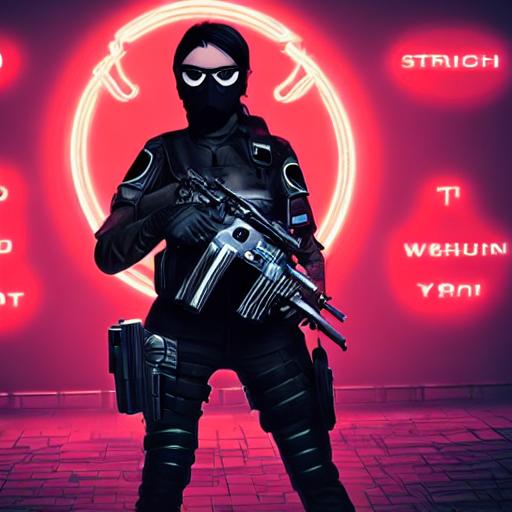}\end{minipage}%
  \begin{minipage}{0.125\textwidth}\includegraphics[width=\linewidth]{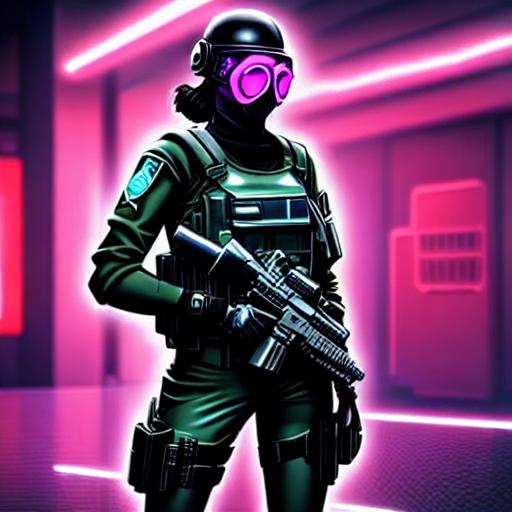}\end{minipage}%
  \begin{minipage}{0.125\textwidth}\includegraphics[width=\linewidth]{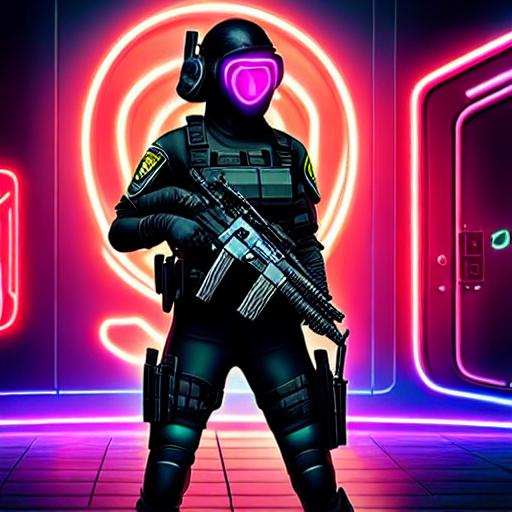}\end{minipage}%
  \begin{minipage}{0.125\textwidth}\includegraphics[width=\linewidth]{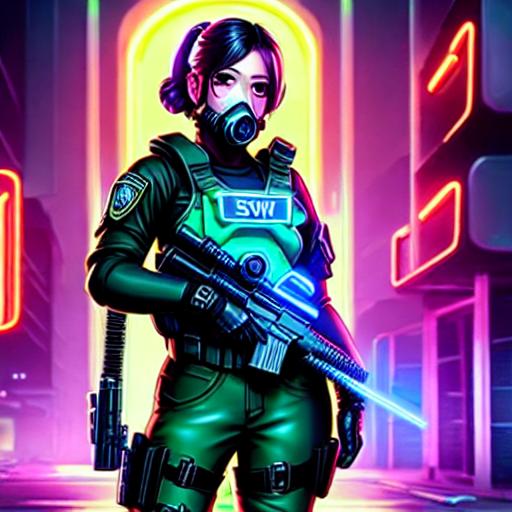}\end{minipage}

  \begin{minipage}{0.25\textwidth}
  \begin{minipage}{0.90\textwidth}
    \centering \scriptsize \raggedright {A portrait painting of a male deer in a suit sitting on a sofa near a window by John Singer Sargent.} 
  \end{minipage}
  \end{minipage}%
  \begin{minipage}{0.125\textwidth}\includegraphics[width=\linewidth]{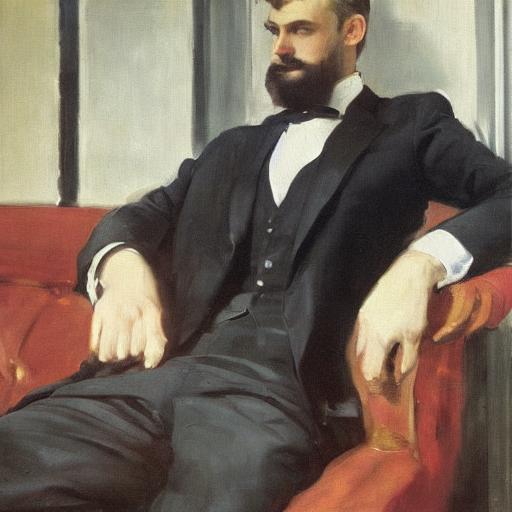}\end{minipage}%
  \begin{minipage}{0.125\textwidth}\includegraphics[width=\linewidth]{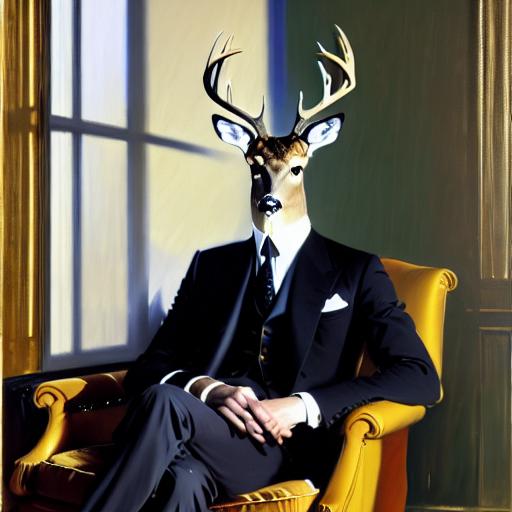}\end{minipage}%
  \begin{minipage}{0.125\textwidth}\includegraphics[width=\linewidth]{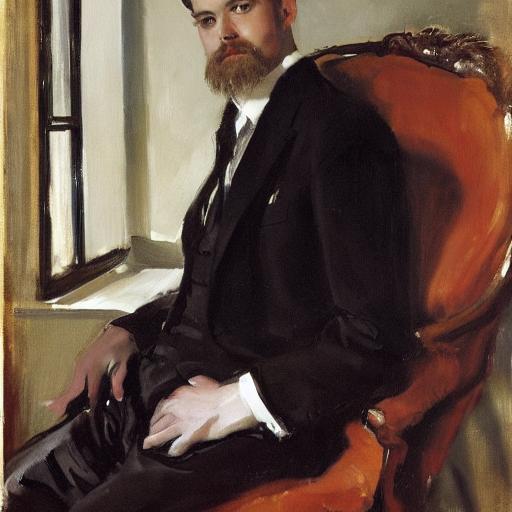}\end{minipage}%
  \begin{minipage}{0.125\textwidth}\includegraphics[width=\linewidth]{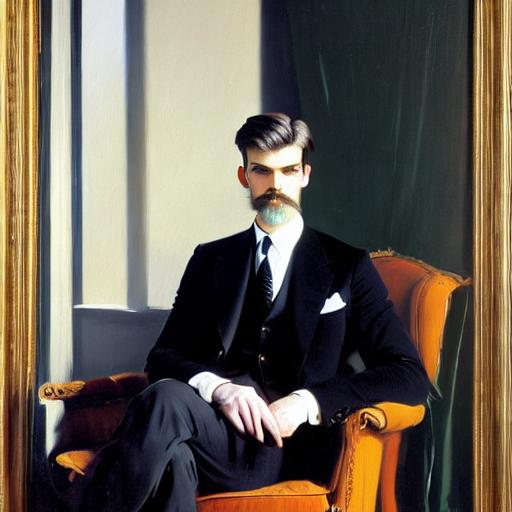}\end{minipage}%
  \begin{minipage}{0.125\textwidth}\includegraphics[width=\linewidth]{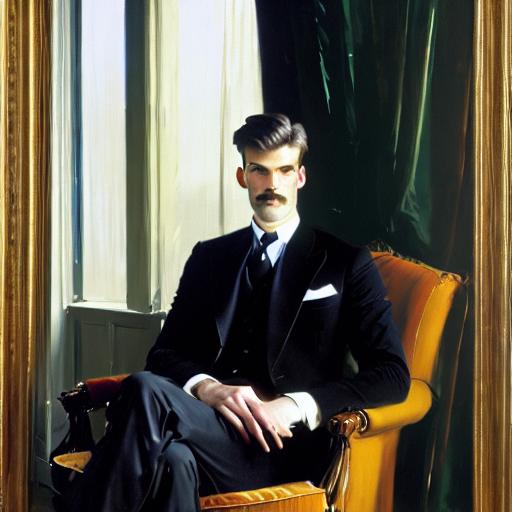}\end{minipage}%
  \begin{minipage}{0.125\textwidth}\includegraphics[width=\linewidth]{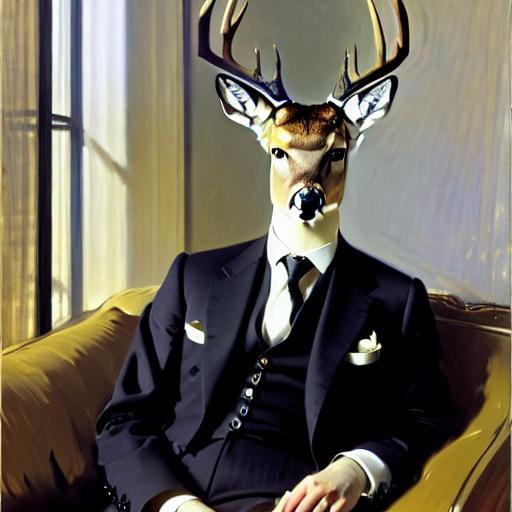}\end{minipage}
  \end{minipage}
  \caption{Image samples generated from SD1.5 fine-tuned with various methods, using validation prompts from Pick-a-Pic v2, HPDv2 and PartiPrompt}
\end{figure*}

\begin{figure*}[htbp]
  \centering
  \begin{minipage}{0.95\textwidth}
  \centering
  \begin{minipage}{0.25\textwidth} \centering \small \textbf{Prompt} 
  \end{minipage}%
  \begin{minipage}{0.15\textwidth} \centering \small \textbf{SDXL} \end{minipage}%
  \begin{minipage}{0.15\textwidth} \centering \small \textbf{SFT} \end{minipage}%
  \begin{minipage}{0.15\textwidth} \centering \small \textbf{Diffusion-DPO} \end{minipage}%
  \begin{minipage}{0.15\textwidth} \centering \small \textbf{MaPO} \end{minipage}%
  \begin{minipage}{0.15\textwidth} \centering \small \textbf{Linear-DPO} \end{minipage}
  
  \vspace{0.2em}
  
  \begin{minipage}{0.25\textwidth}
  \begin{minipage}{0.90\textwidth}
    \centering \scriptsize \raggedright {beautiful portrait of a young woman made of glossy glass skin surrounded with glowing birds} 
  \end{minipage}
  \end{minipage}%
  \begin{minipage}{0.15\textwidth}\includegraphics[width=\linewidth]{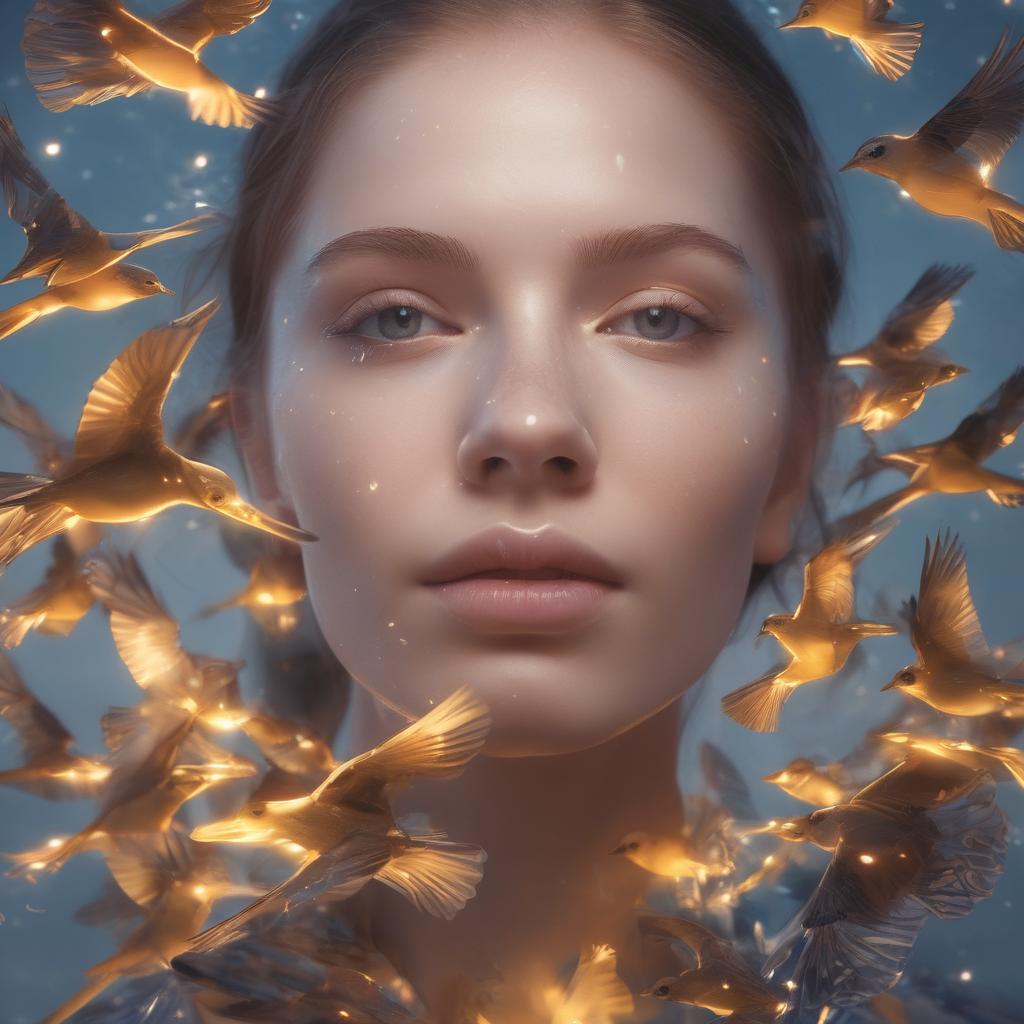}\end{minipage}%
  \begin{minipage}{0.15\textwidth}\includegraphics[width=\linewidth]{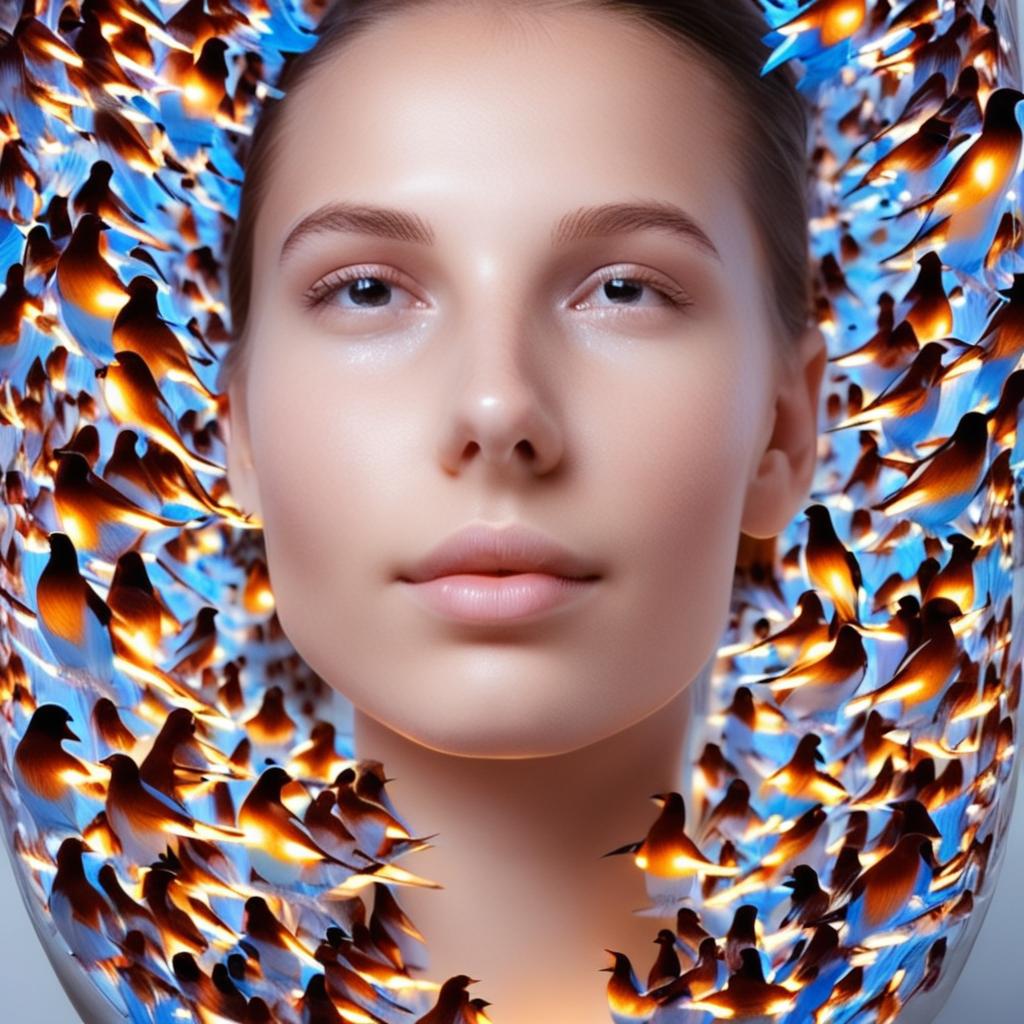}\end{minipage}%
  \begin{minipage}{0.15\textwidth}\includegraphics[width=\linewidth]{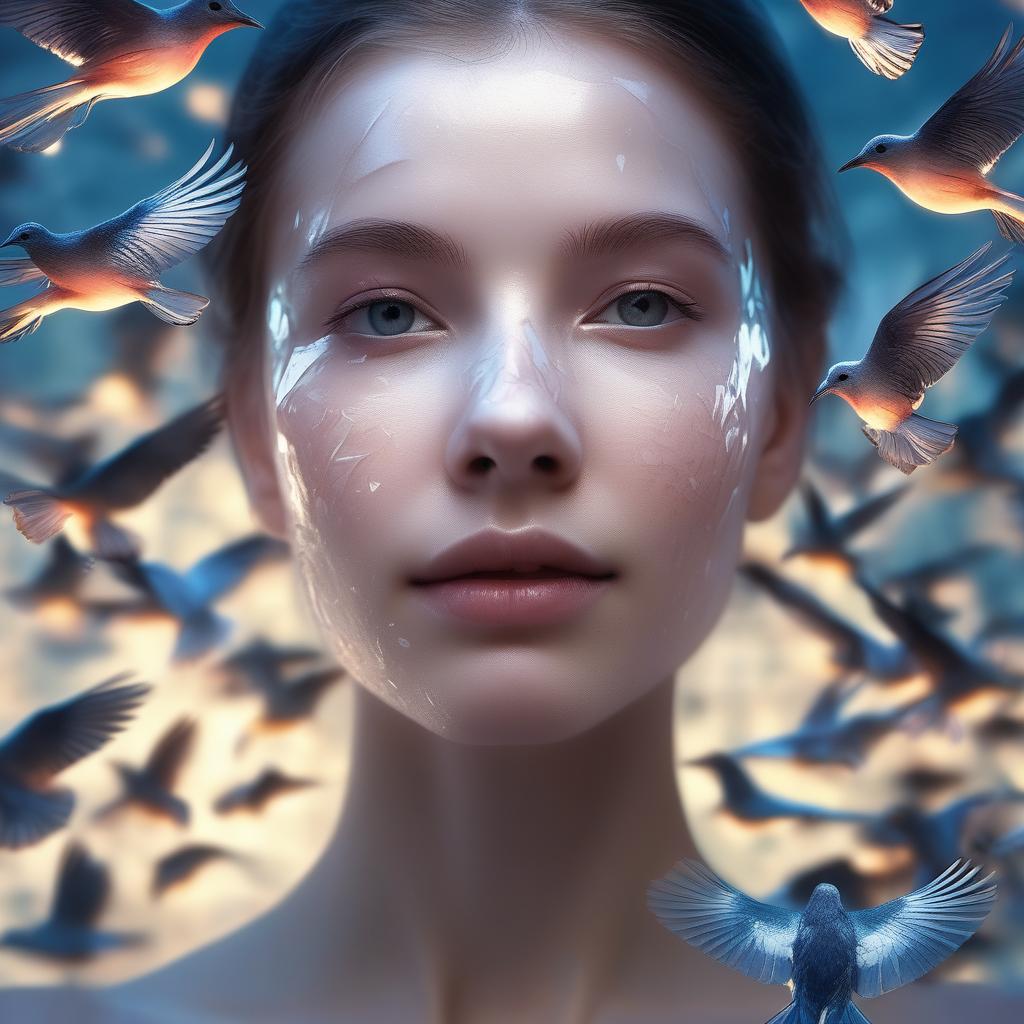}\end{minipage}%
  \begin{minipage}{0.15\textwidth}\includegraphics[width=\linewidth]{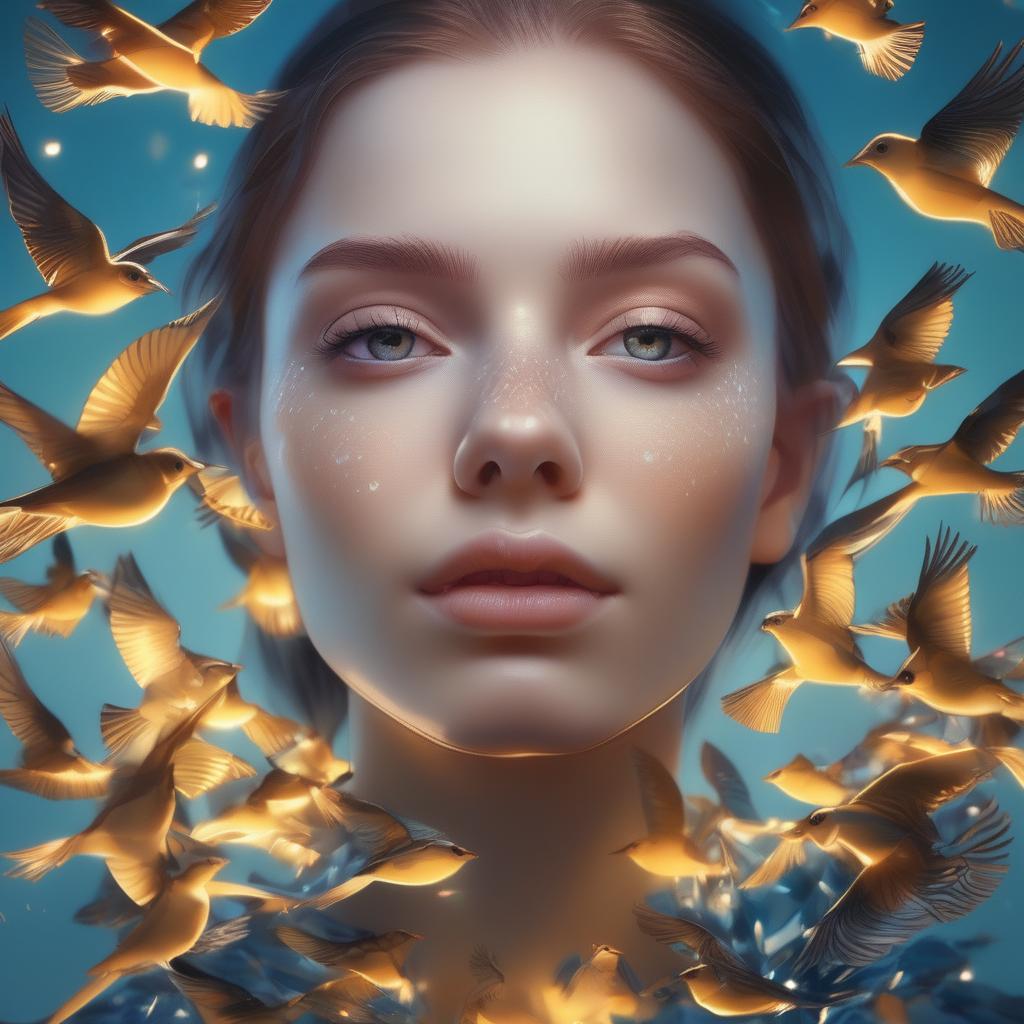}\end{minipage}%
  \begin{minipage}{0.15\textwidth}\includegraphics[width=\linewidth]{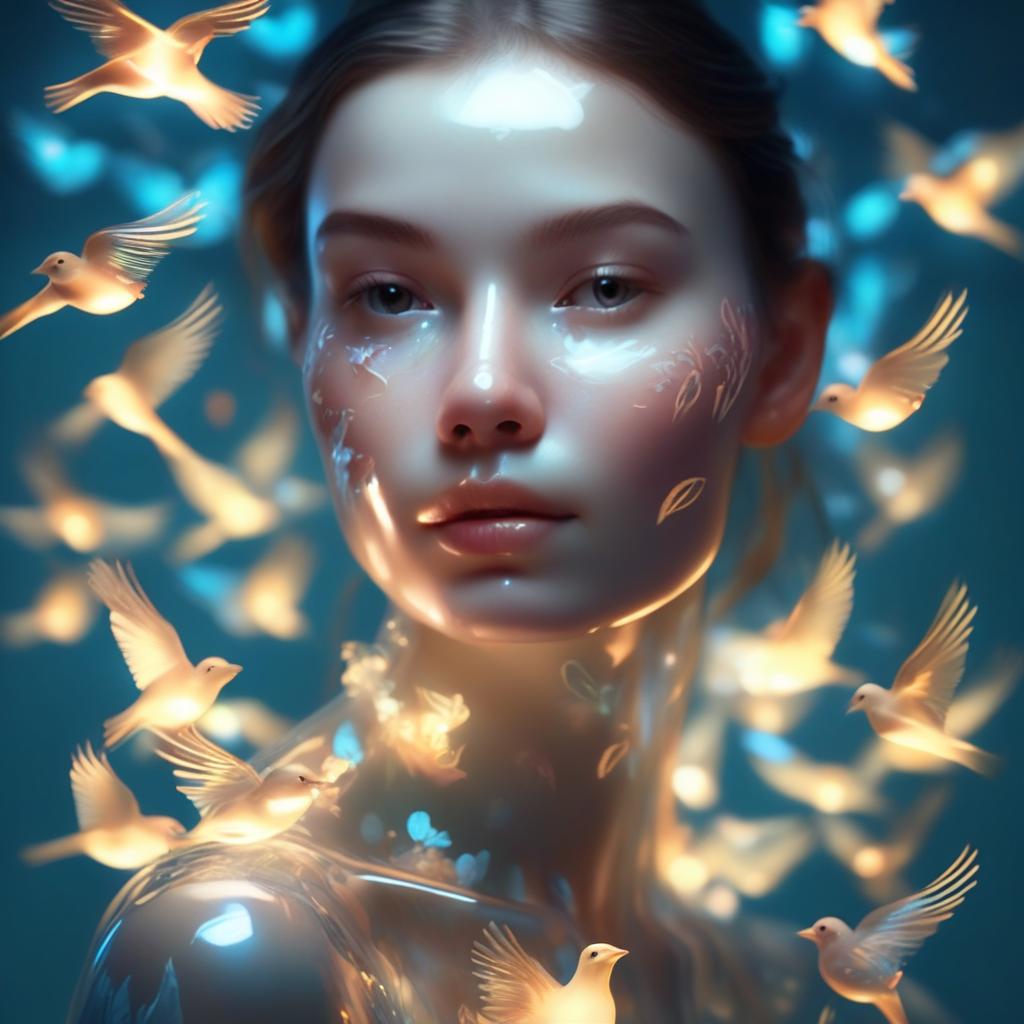}\end{minipage}%
  
  
  \begin{minipage}{0.25\textwidth}
  \begin{minipage}{0.90\textwidth}
    \centering \scriptsize \raggedright {A massive and brightly colored spacecraft in a deserted landscape, depicted in retro 1960s sci-fi art.} 
  \end{minipage}
  \end{minipage}%
  \begin{minipage}{0.15\textwidth}\includegraphics[width=\linewidth]{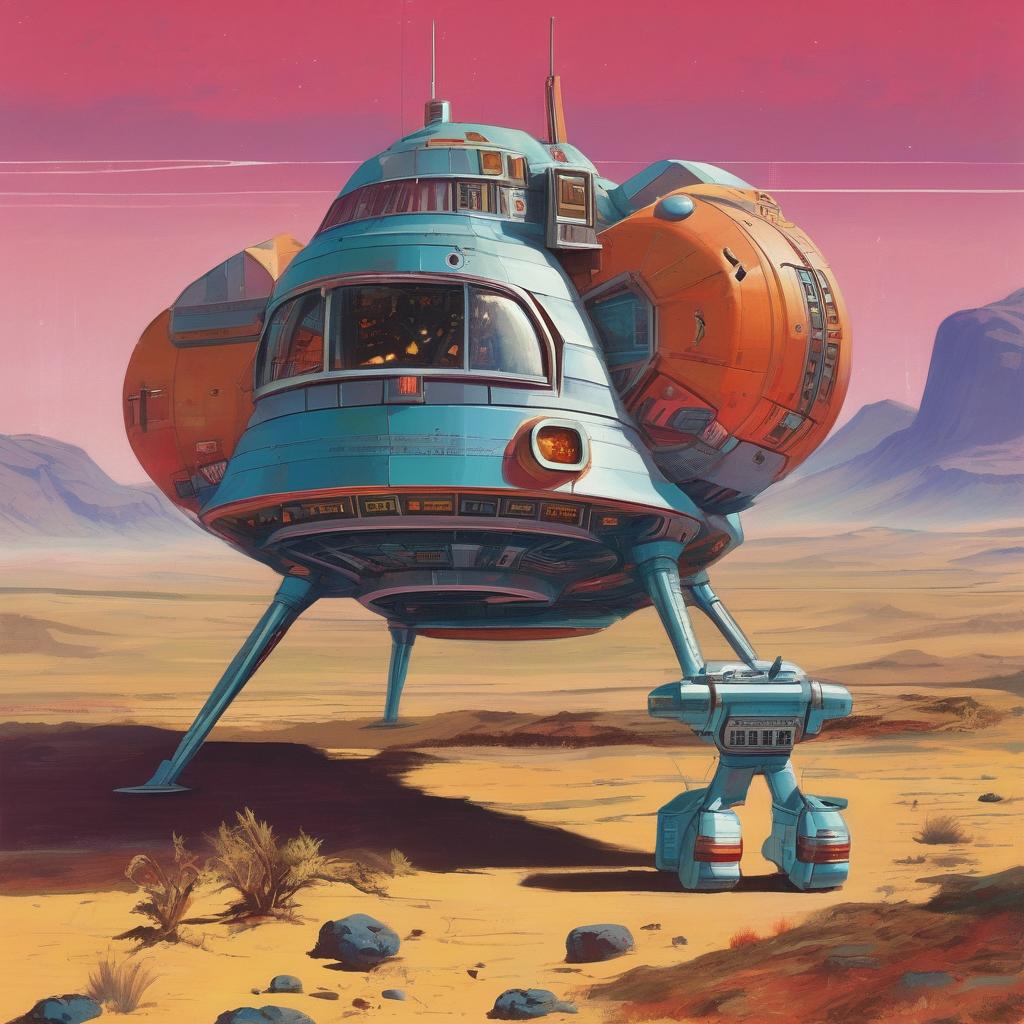}\end{minipage}%
  \begin{minipage}{0.15\textwidth}\includegraphics[width=\linewidth]{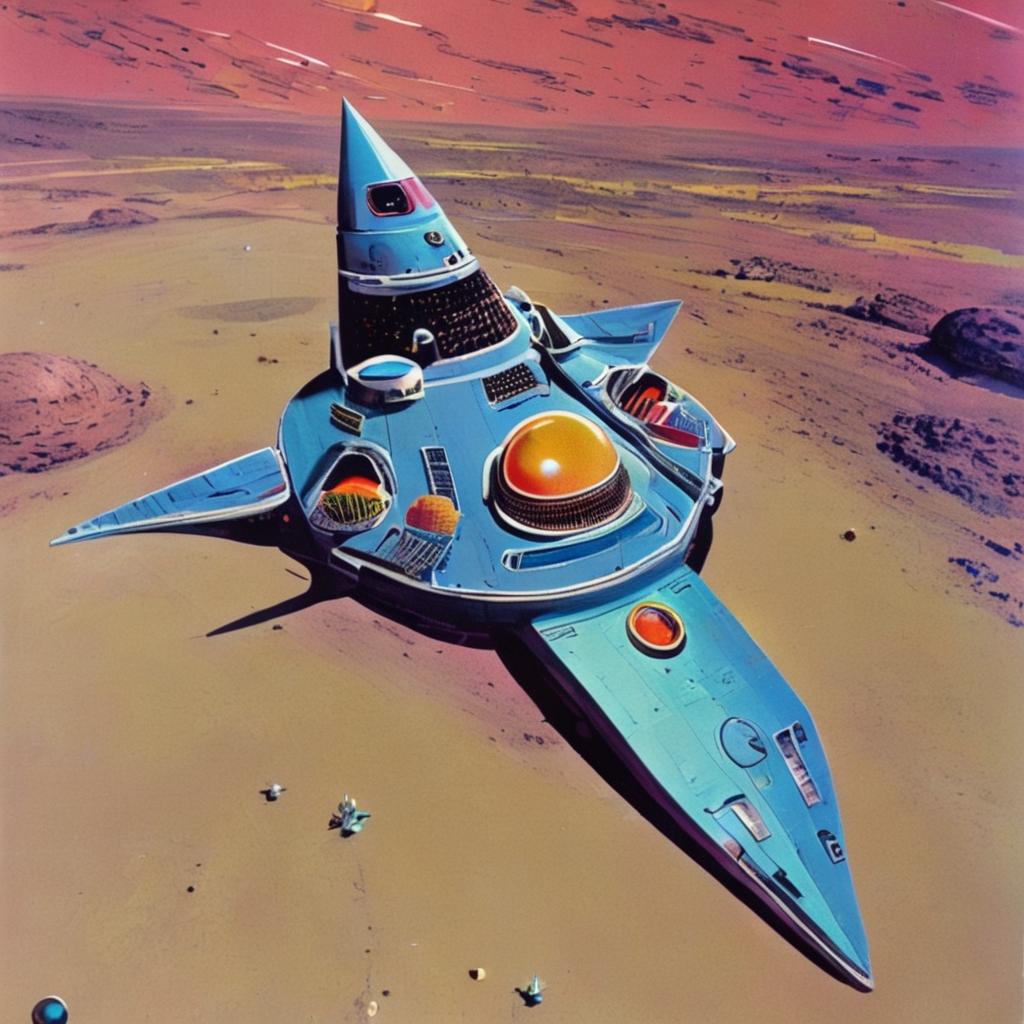}\end{minipage}%
  \begin{minipage}{0.15\textwidth}\includegraphics[width=\linewidth]{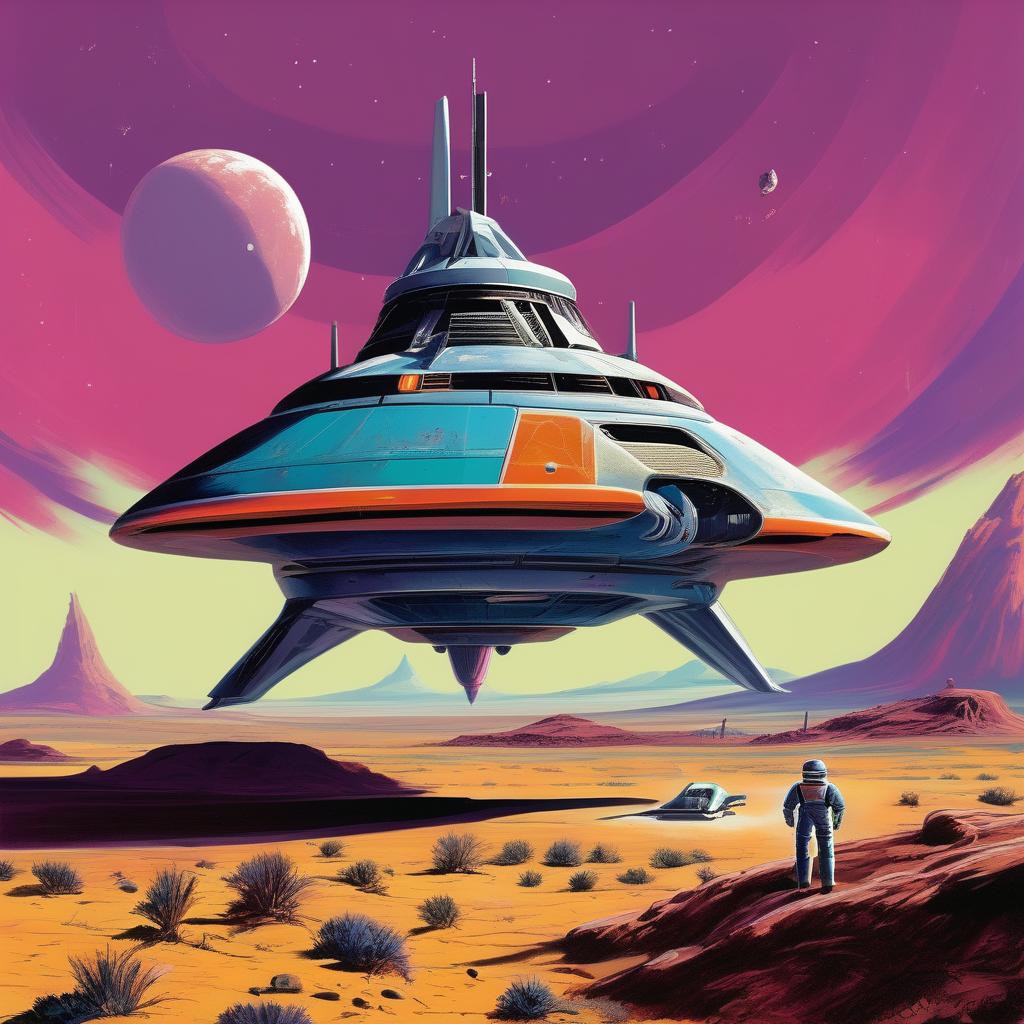}\end{minipage}%
  \begin{minipage}{0.15\textwidth}\includegraphics[width=\linewidth]{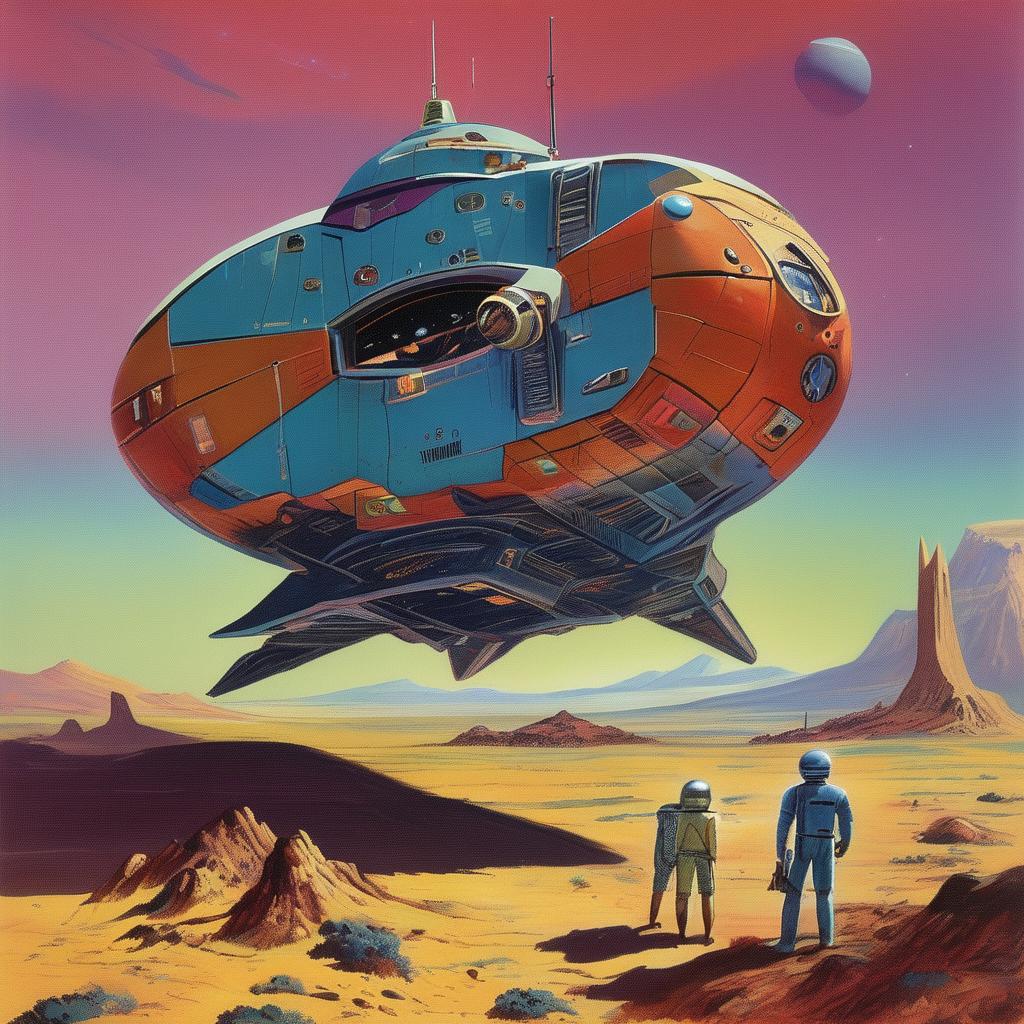}\end{minipage}%
  \begin{minipage}{0.15\textwidth}\includegraphics[width=\linewidth]{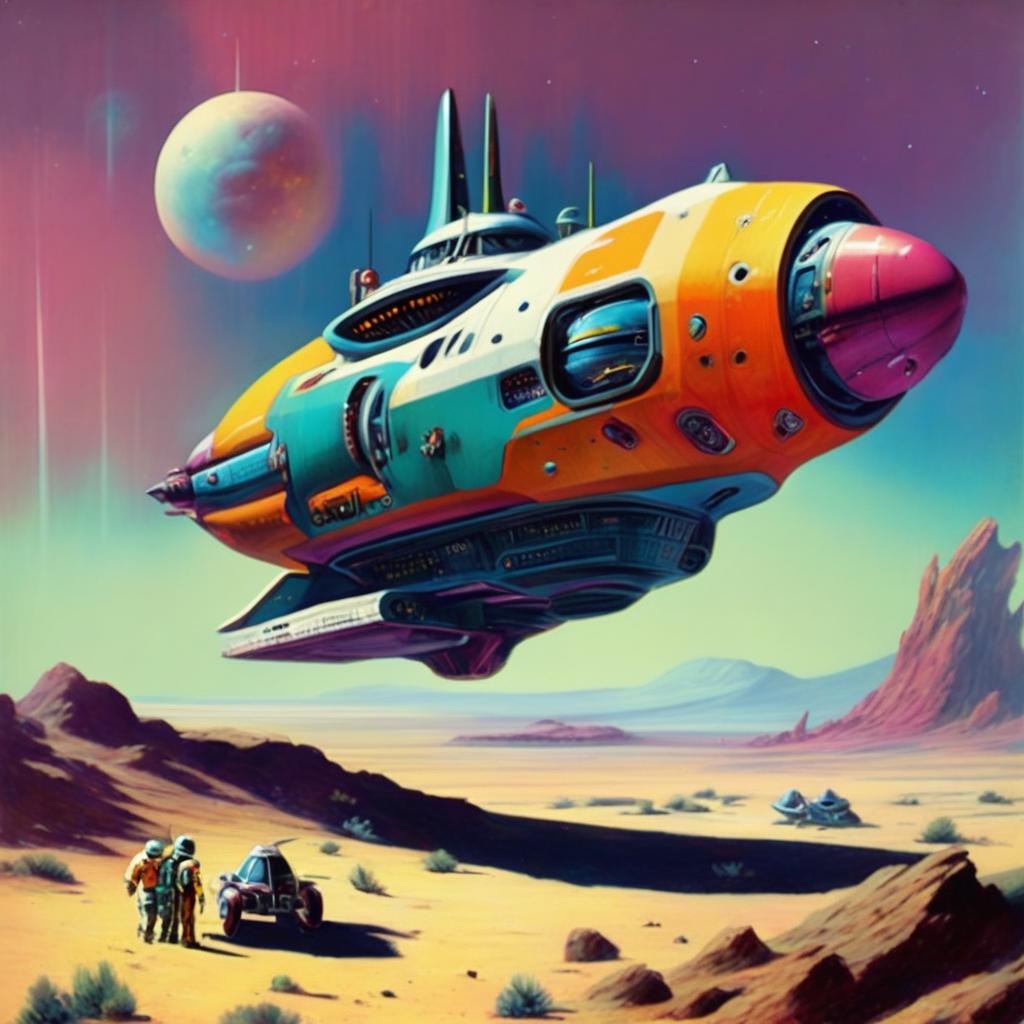}\end{minipage}%

  
  \begin{minipage}{0.25\textwidth}
  \begin{minipage}{0.90\textwidth}
    \centering \scriptsize \raggedright {An abstract painting of the Statue of Liberty} 
  \end{minipage}
  \end{minipage}%
  \begin{minipage}{0.15\textwidth}\includegraphics[width=\linewidth]{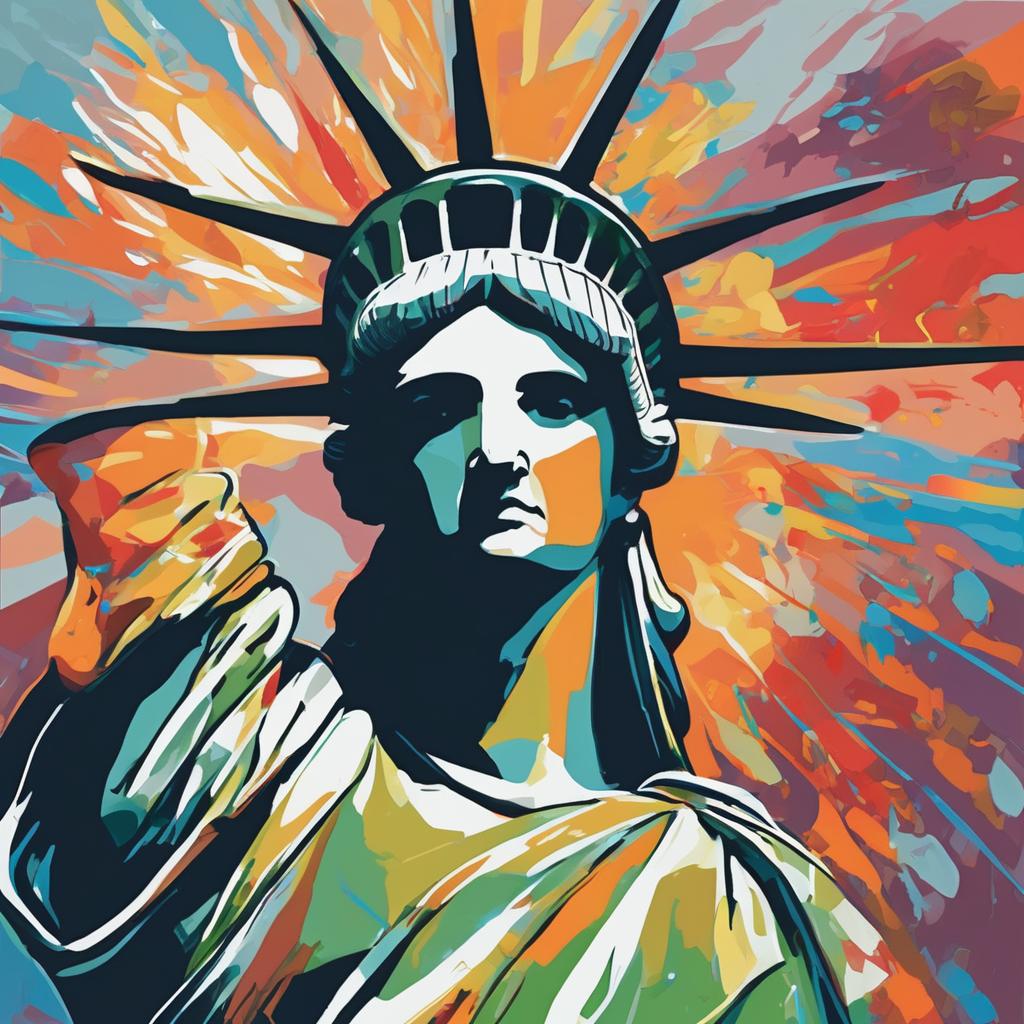}\end{minipage}%
  \begin{minipage}{0.15\textwidth}\includegraphics[width=\linewidth]{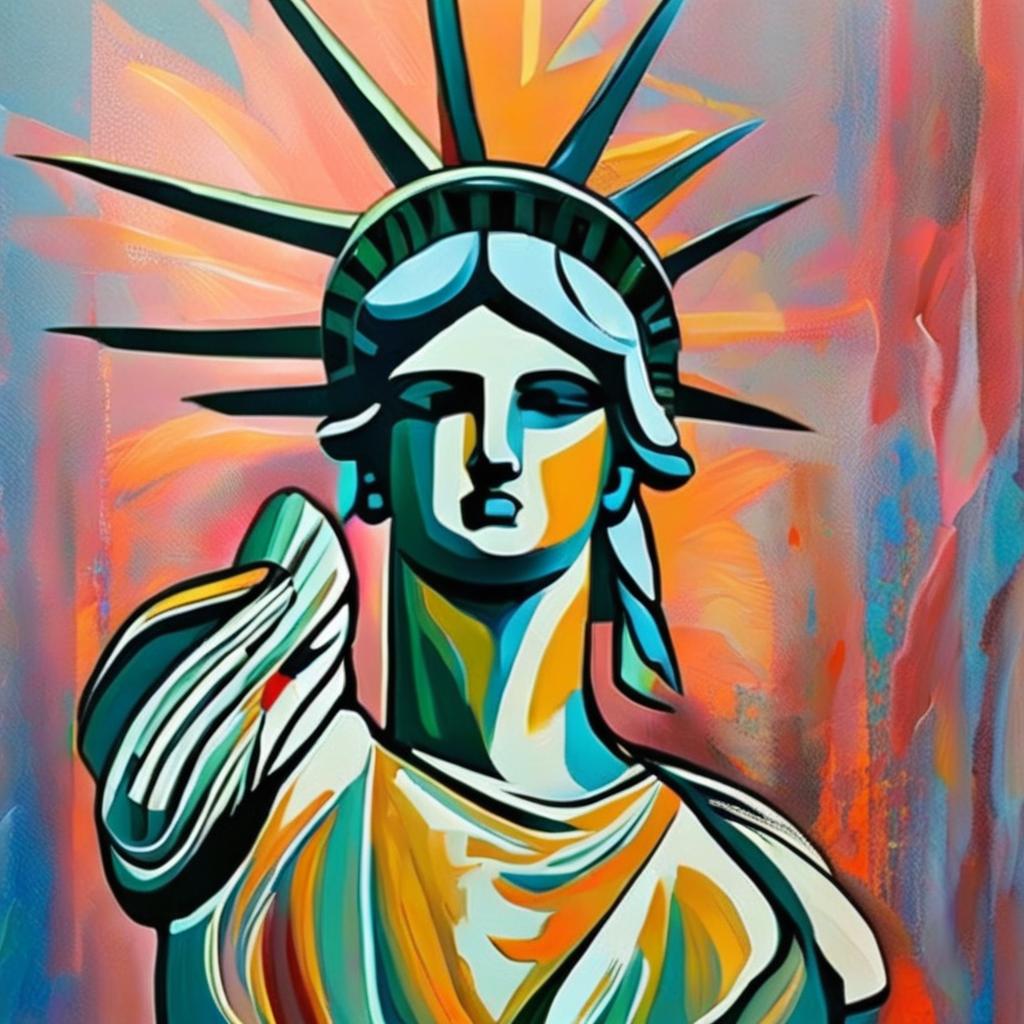}\end{minipage}%
  \begin{minipage}{0.15\textwidth}\includegraphics[width=\linewidth]{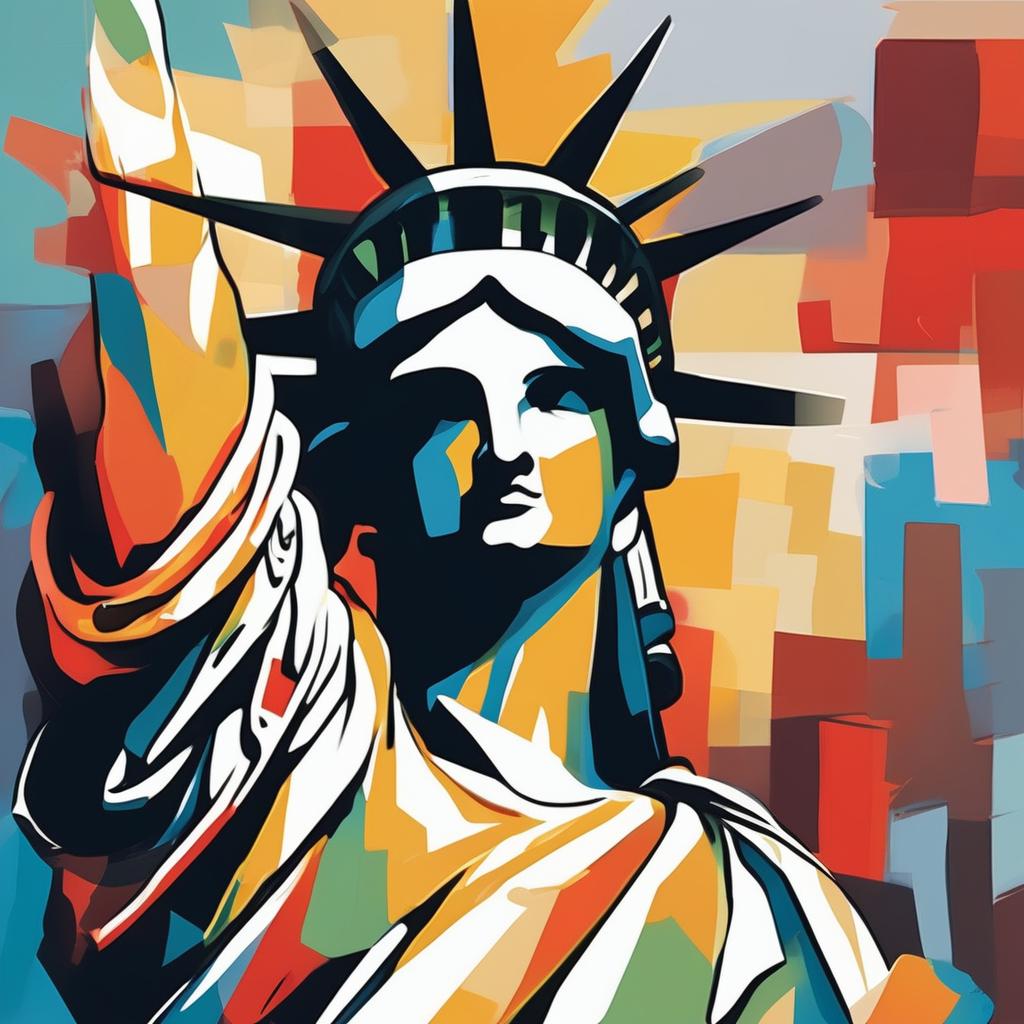}\end{minipage}%
  \begin{minipage}{0.15\textwidth}\includegraphics[width=\linewidth]{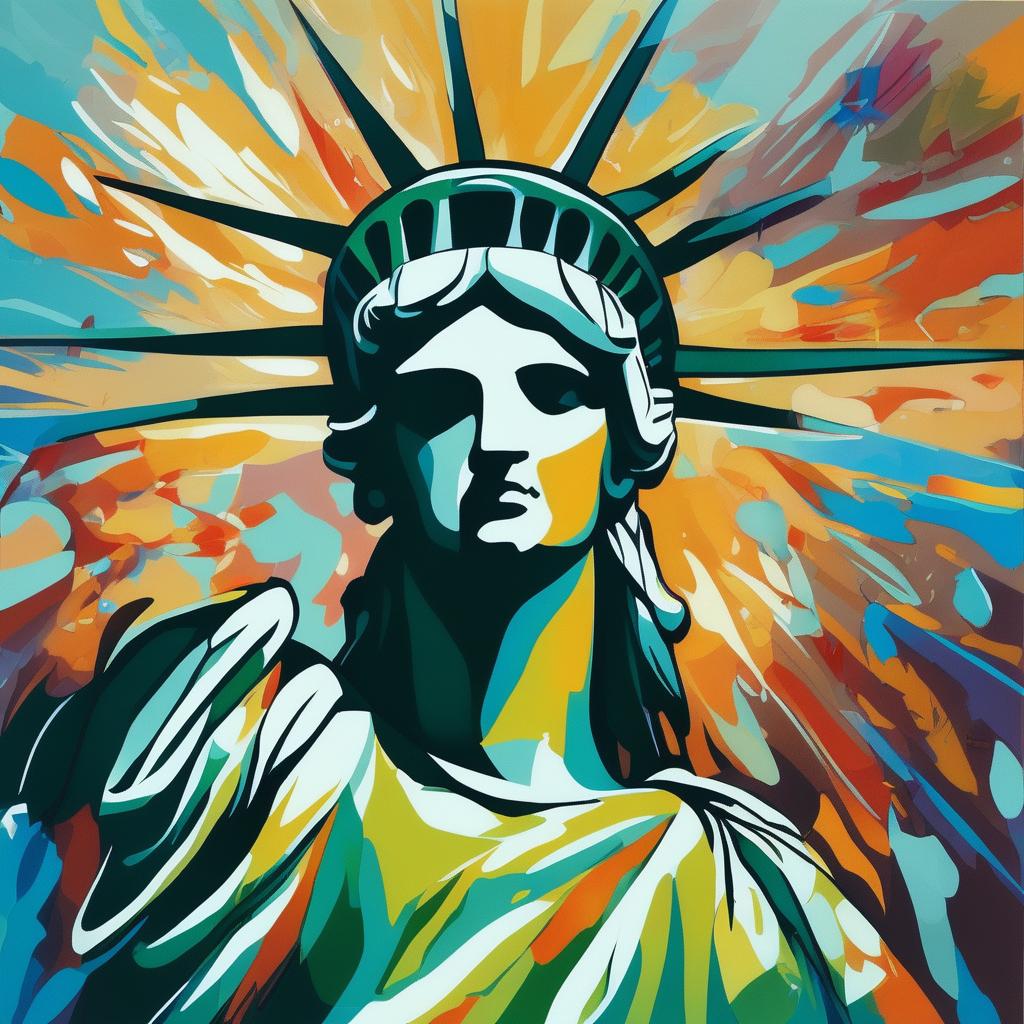}\end{minipage}%
  \begin{minipage}{0.15\textwidth}\includegraphics[width=\linewidth]{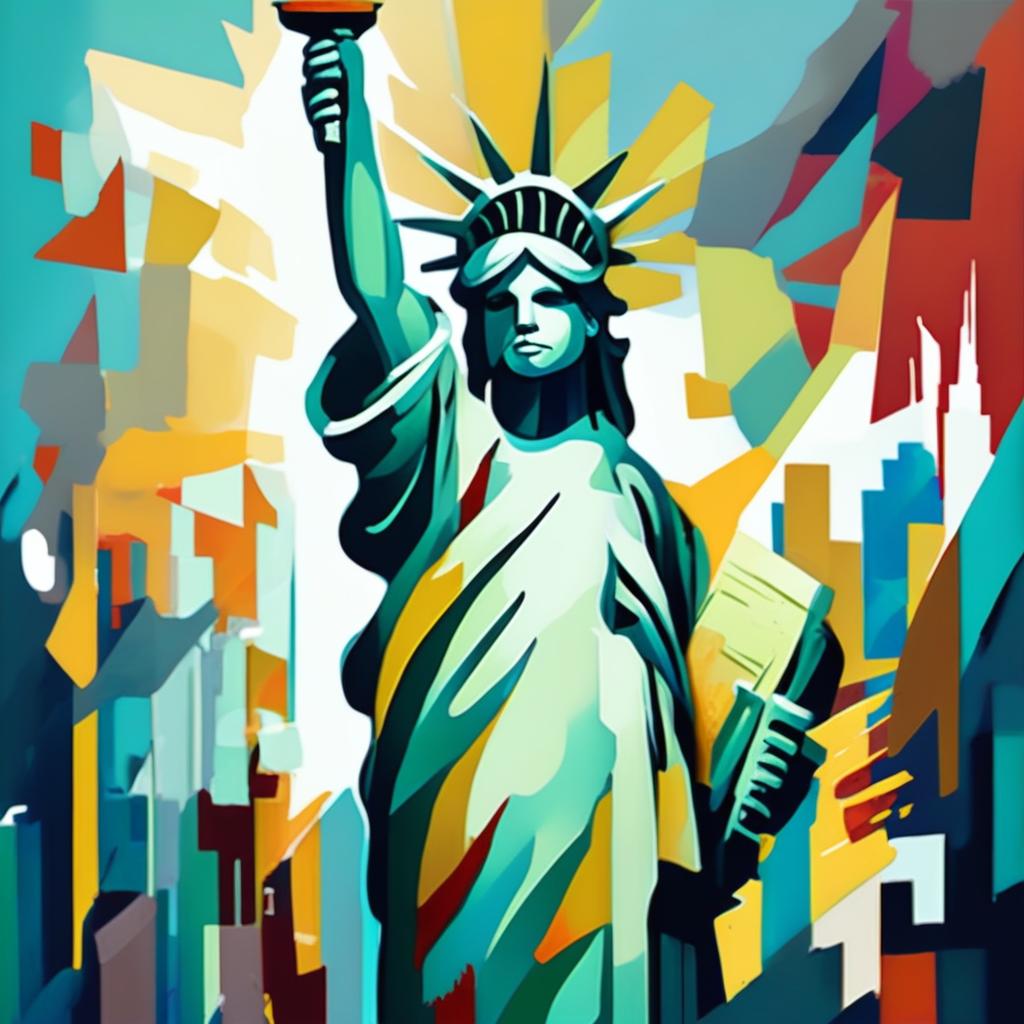}\end{minipage}%

  \begin{minipage}{0.25\textwidth}
  \begin{minipage}{0.90\textwidth}
    \centering \scriptsize \raggedright {the fox in the labyrinth, vivid opulent colors, vector art} 
  \end{minipage}
  \end{minipage}%
  \begin{minipage}{0.15\textwidth}\includegraphics[width=\linewidth]{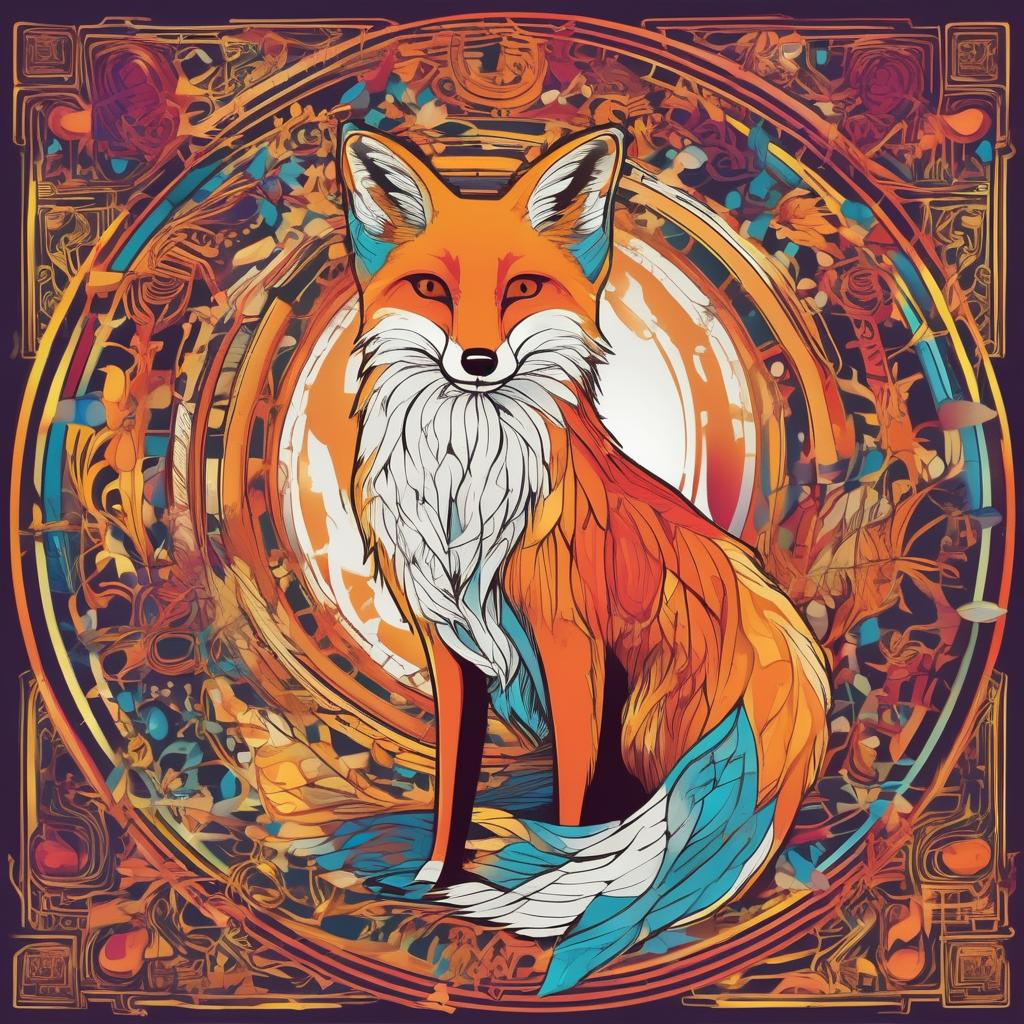}\end{minipage}%
  \begin{minipage}{0.15\textwidth}\includegraphics[width=\linewidth]{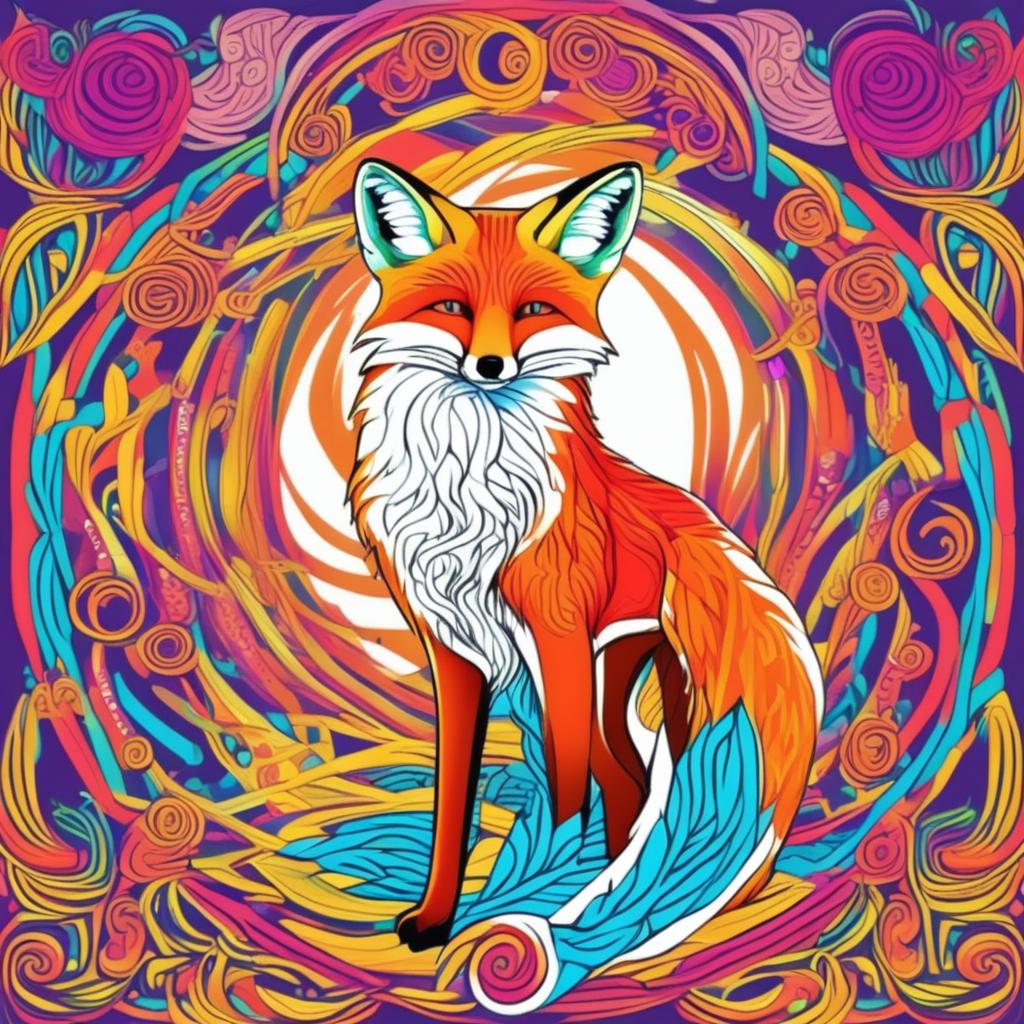}\end{minipage}%
  \begin{minipage}{0.15\textwidth}\includegraphics[width=\linewidth]{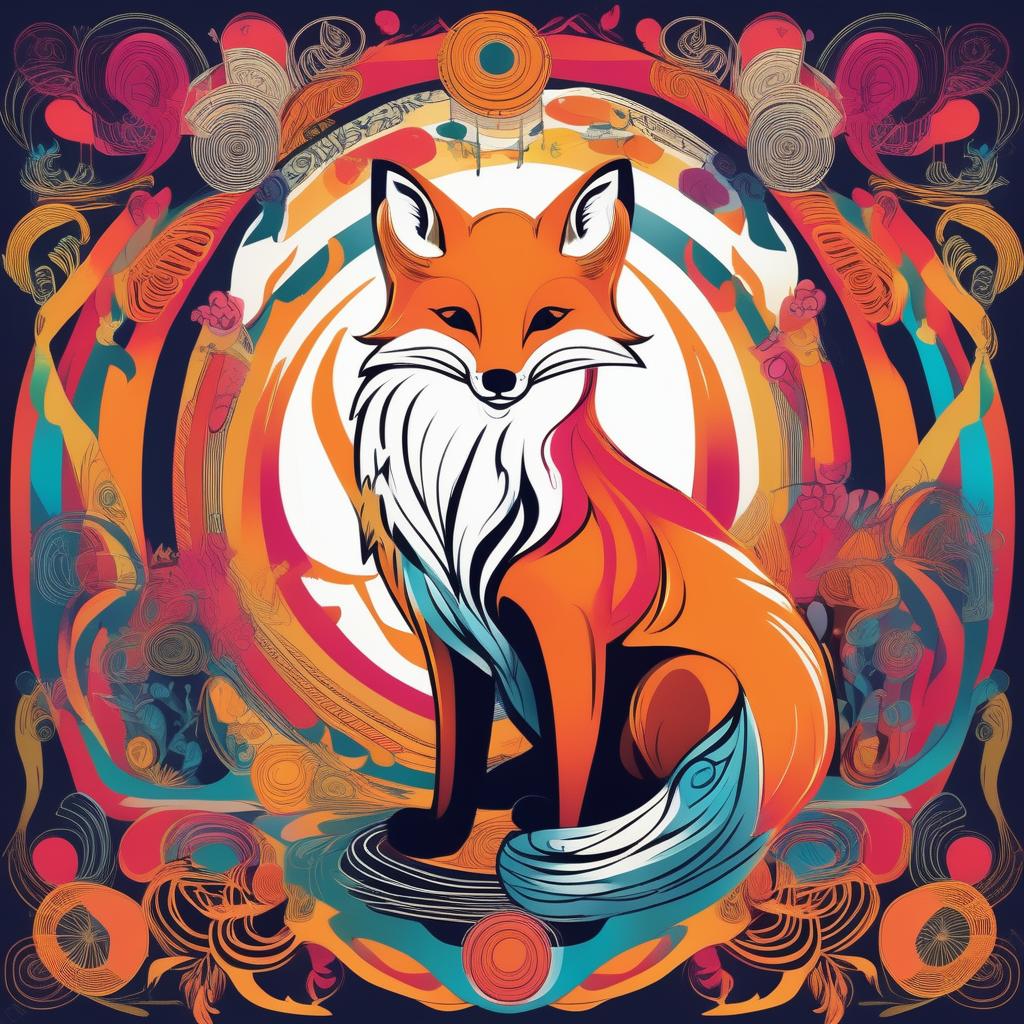}\end{minipage}%
  \begin{minipage}{0.15\textwidth}\includegraphics[width=\linewidth]{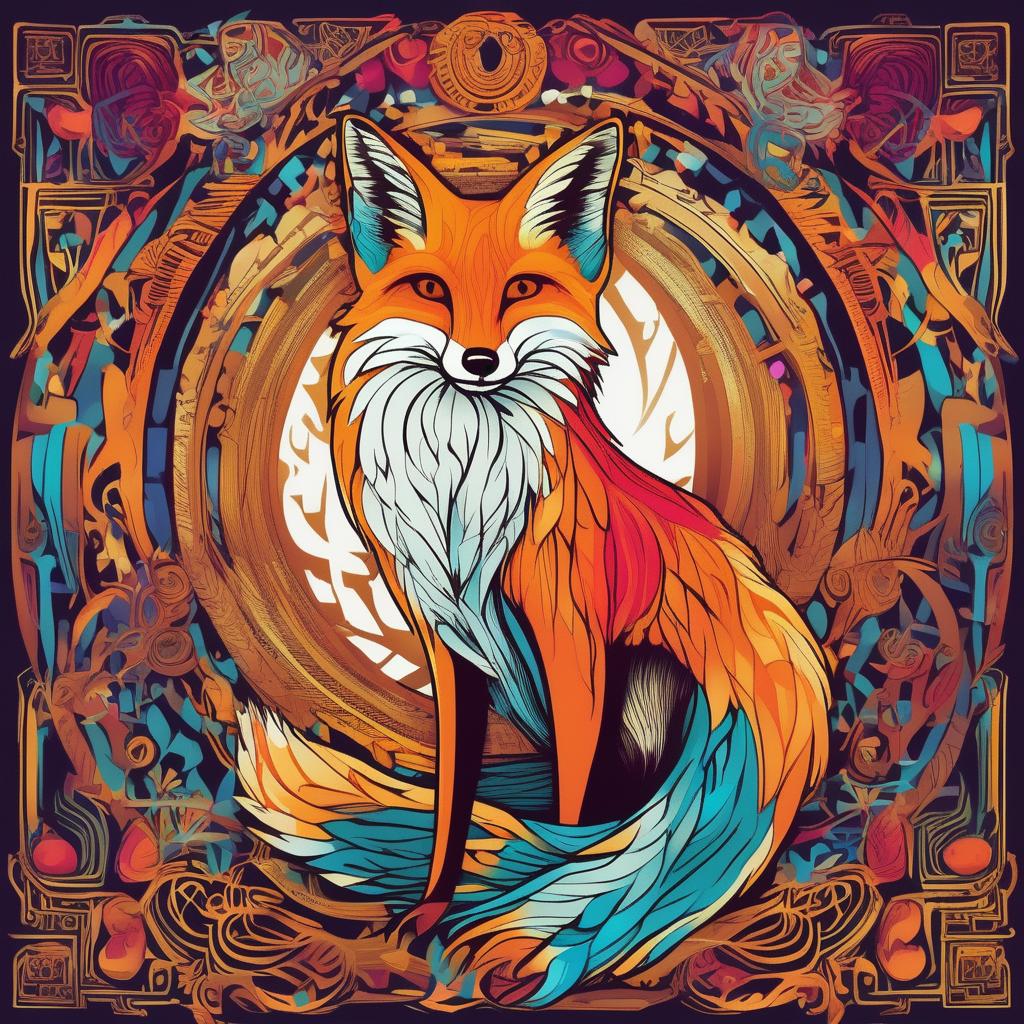}\end{minipage}%
  \begin{minipage}{0.15\textwidth}\includegraphics[width=\linewidth]{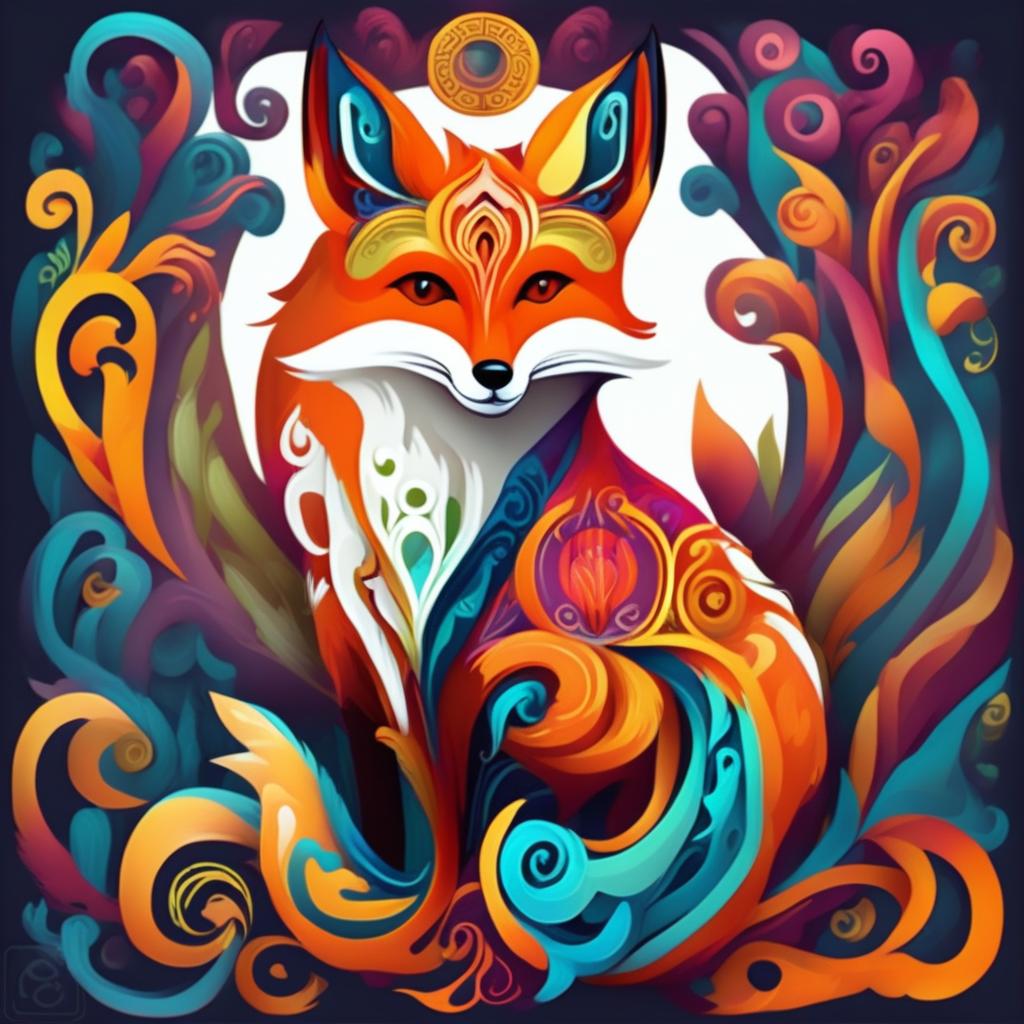}\end{minipage}%

  \begin{minipage}{0.25\textwidth}
  \begin{minipage}{0.90\textwidth}
    \centering \scriptsize \raggedright {a papaya fruit dressed as a sailor.} 
  \end{minipage}
  \end{minipage}%
  \begin{minipage}{0.15\textwidth}\includegraphics[width=\linewidth]{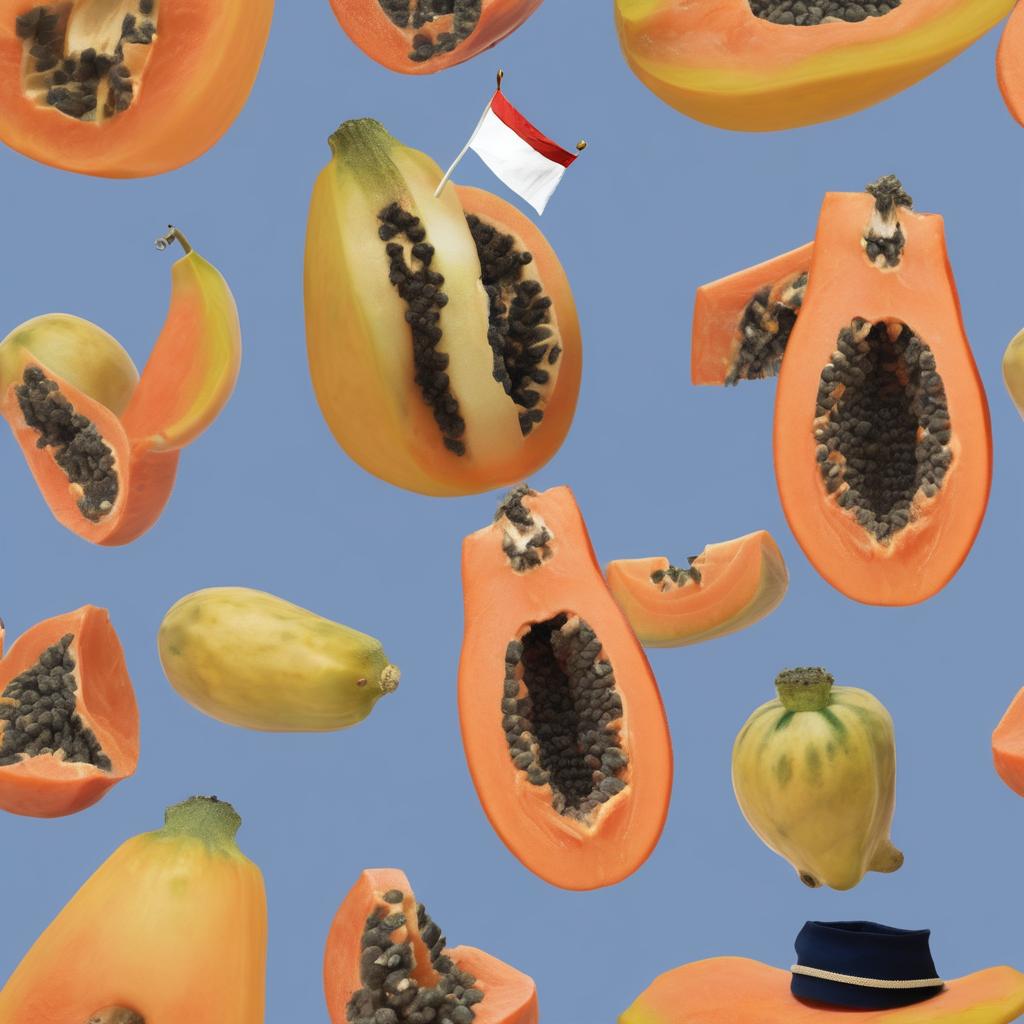}\end{minipage}%
  \begin{minipage}{0.15\textwidth}\includegraphics[width=\linewidth]{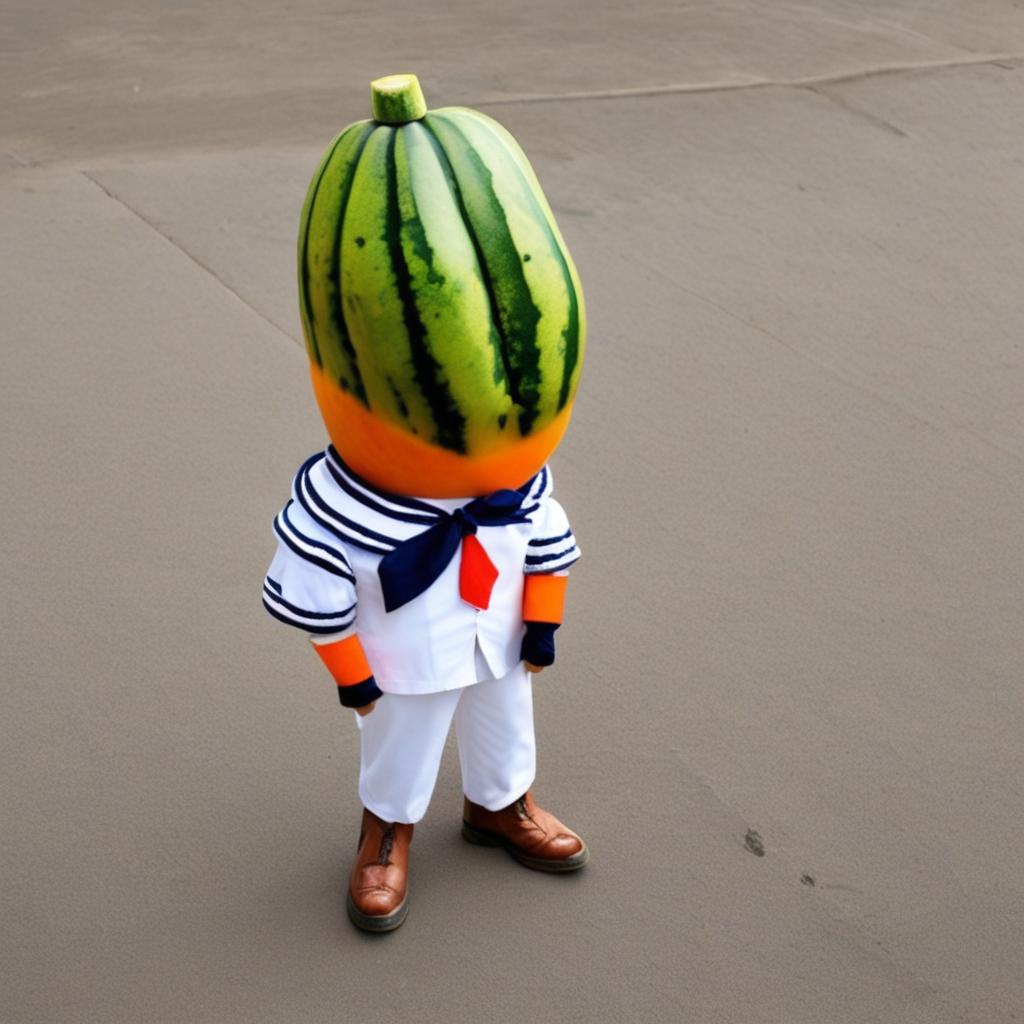}\end{minipage}%
  \begin{minipage}{0.15\textwidth}\includegraphics[width=\linewidth]{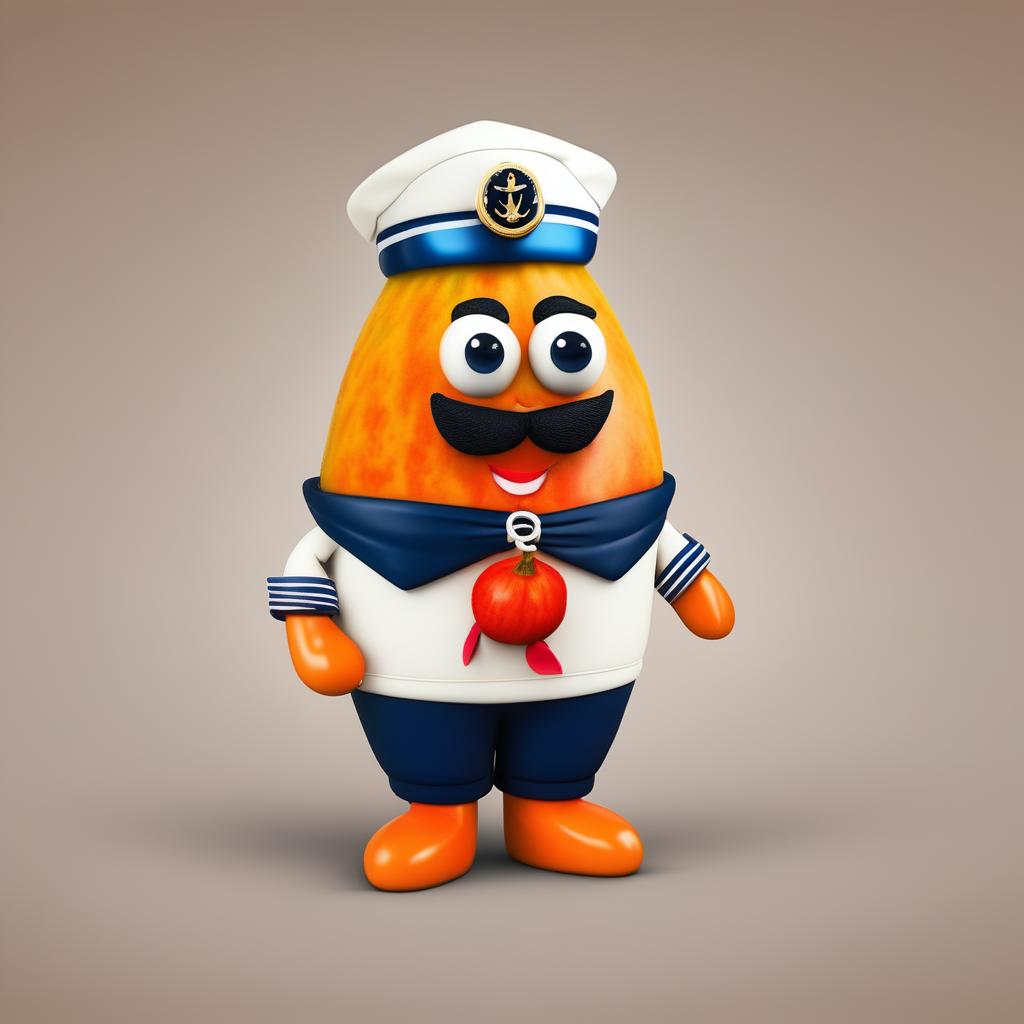}\end{minipage}%
  \begin{minipage}{0.15\textwidth}\includegraphics[width=\linewidth]{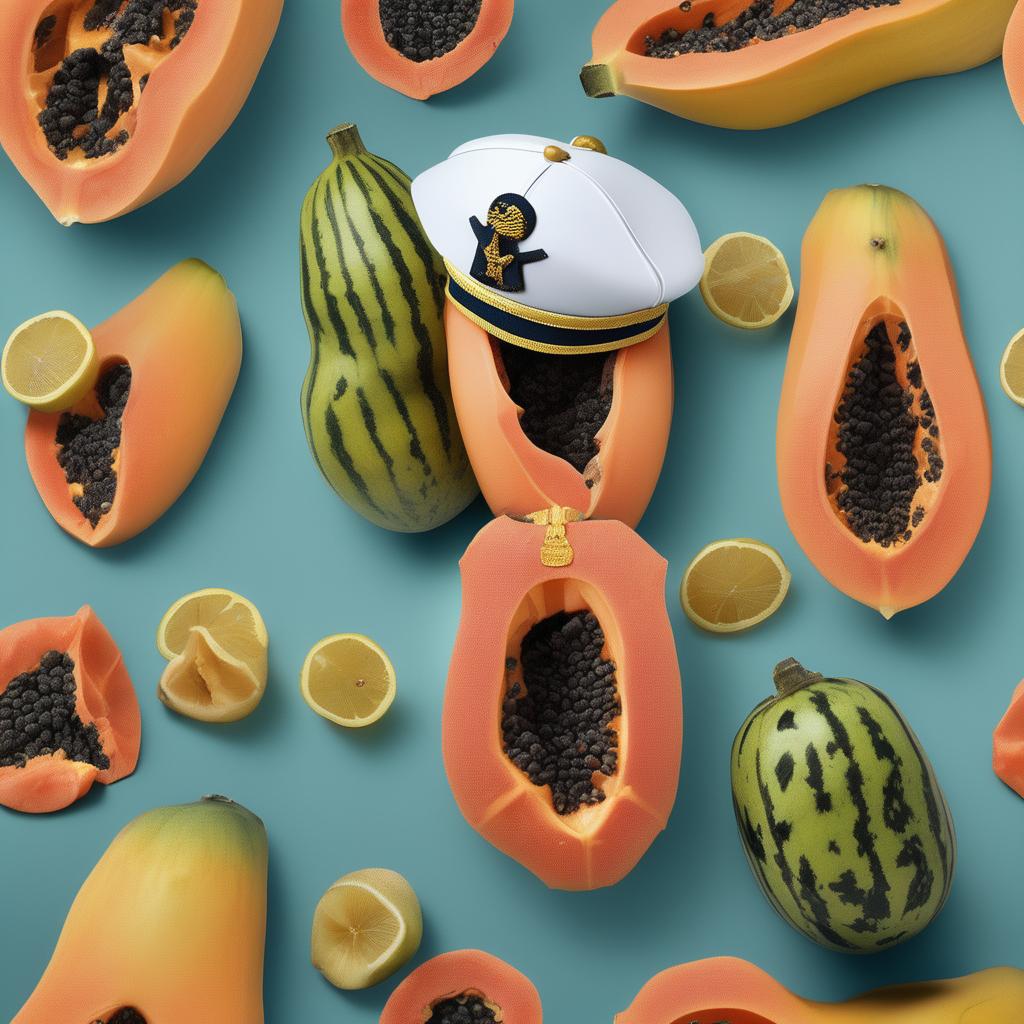}\end{minipage}%
  \begin{minipage}{0.15\textwidth}\includegraphics[width=\linewidth]{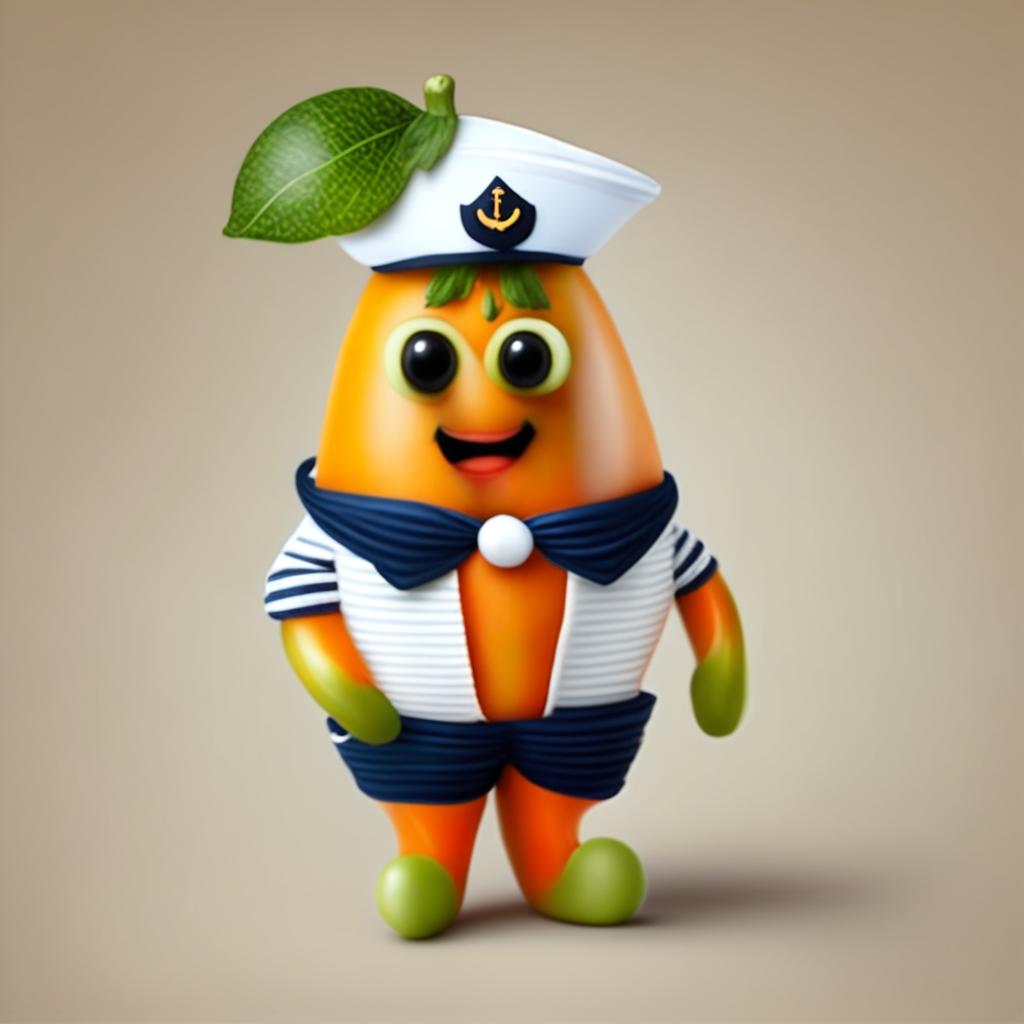}\end{minipage}%

  \begin{minipage}{0.25\textwidth}
  \begin{minipage}{0.90\textwidth}
    \centering \scriptsize \raggedright {A warrior in glowing azure plate armor stands in a doorway to hell sliced by iridescent glass cracks, with crimson clouds and an art deco palace backdrop.} 
  \end{minipage}
  \end{minipage}%
  \begin{minipage}{0.15\textwidth}\includegraphics[width=\linewidth]{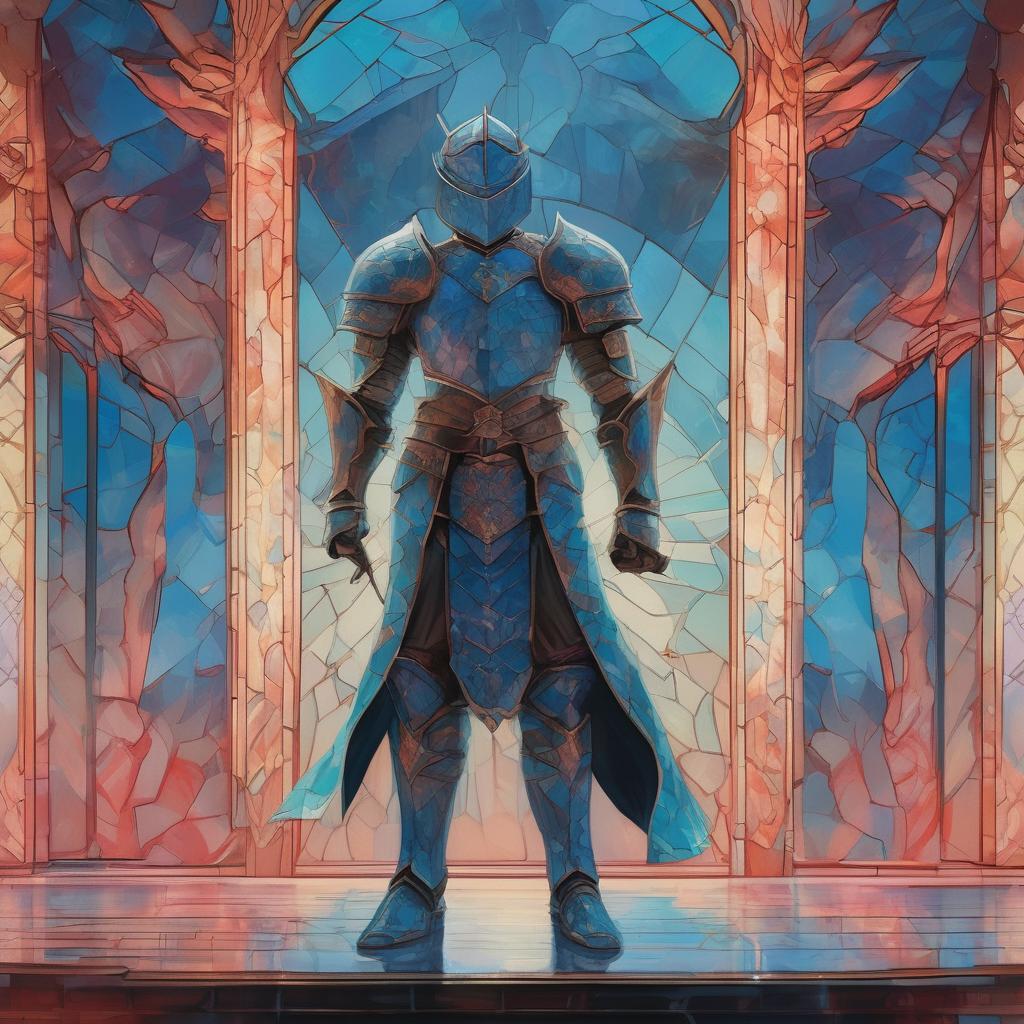}\end{minipage}%
  \begin{minipage}{0.15\textwidth}\includegraphics[width=\linewidth]{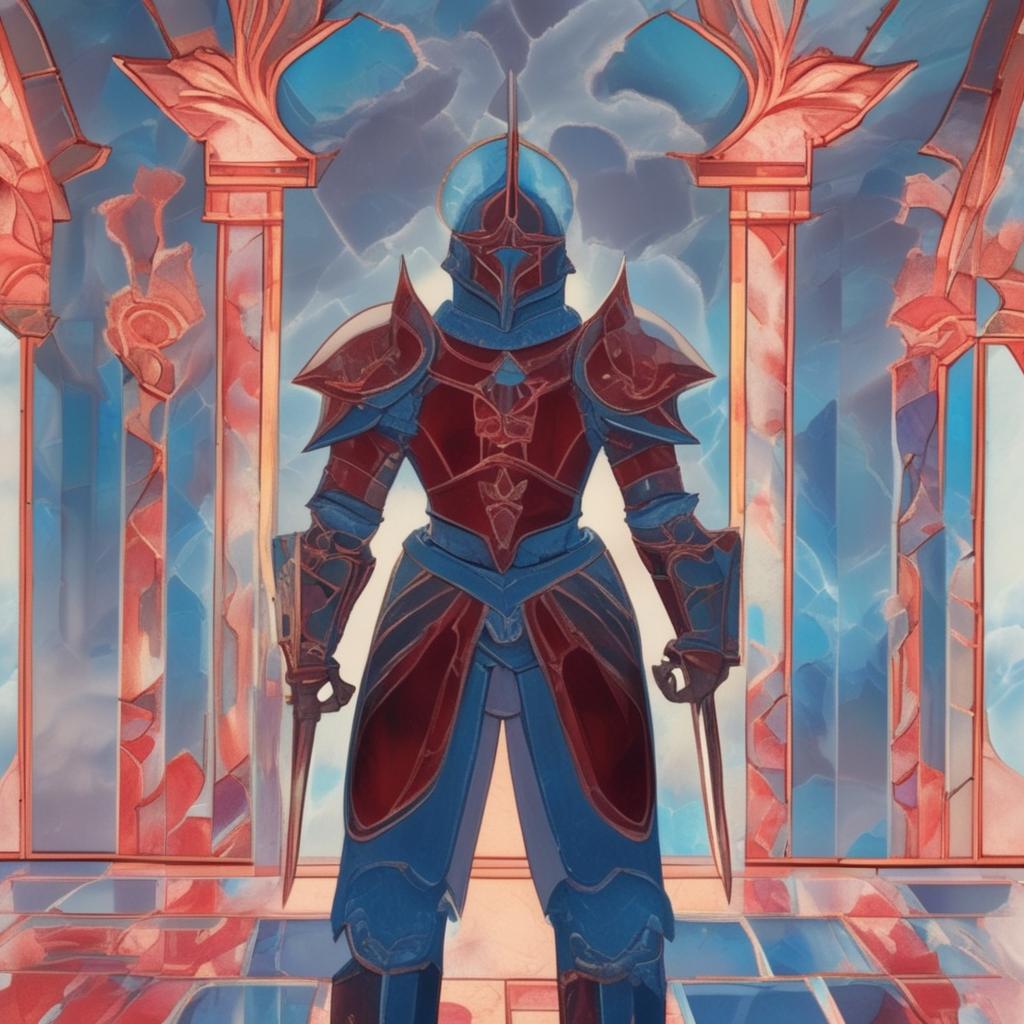}\end{minipage}%
  \begin{minipage}{0.15\textwidth}\includegraphics[width=\linewidth]{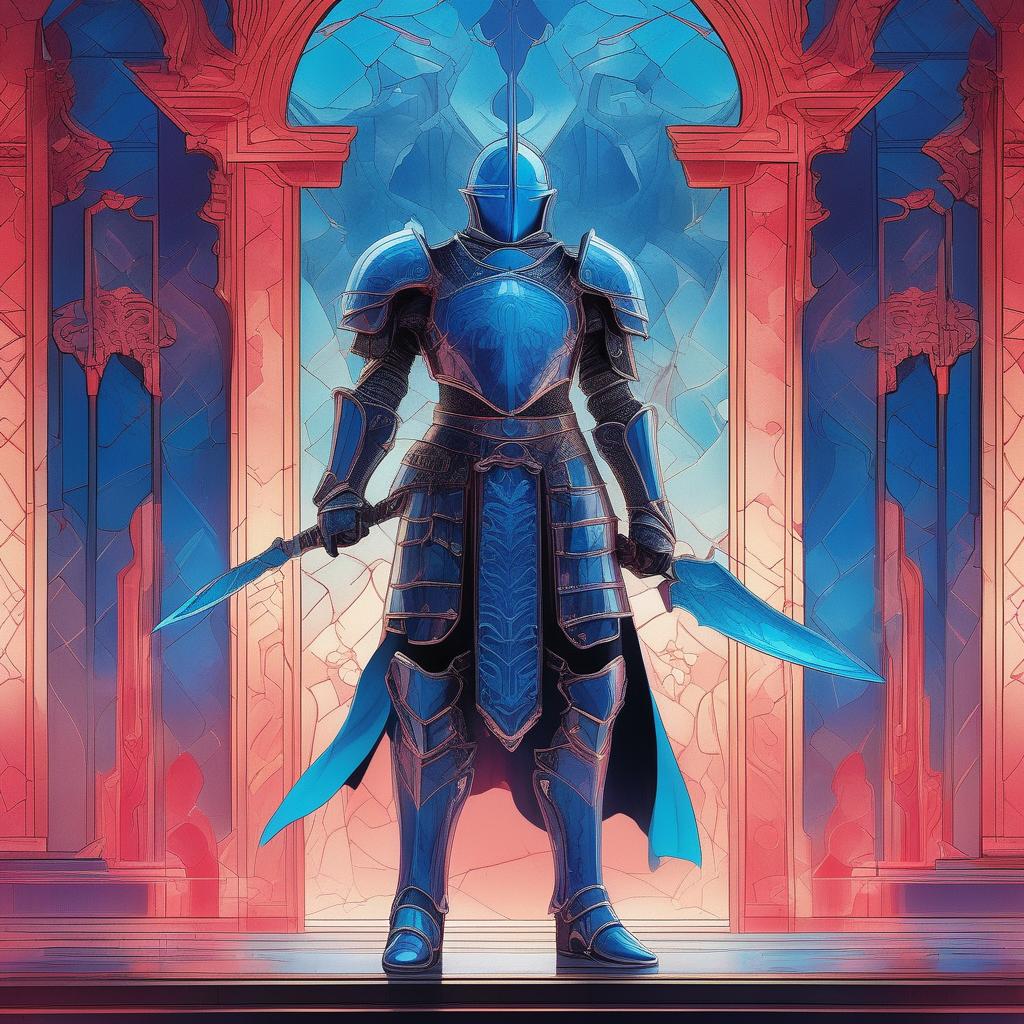}\end{minipage}%
  \begin{minipage}{0.15\textwidth}\includegraphics[width=\linewidth]{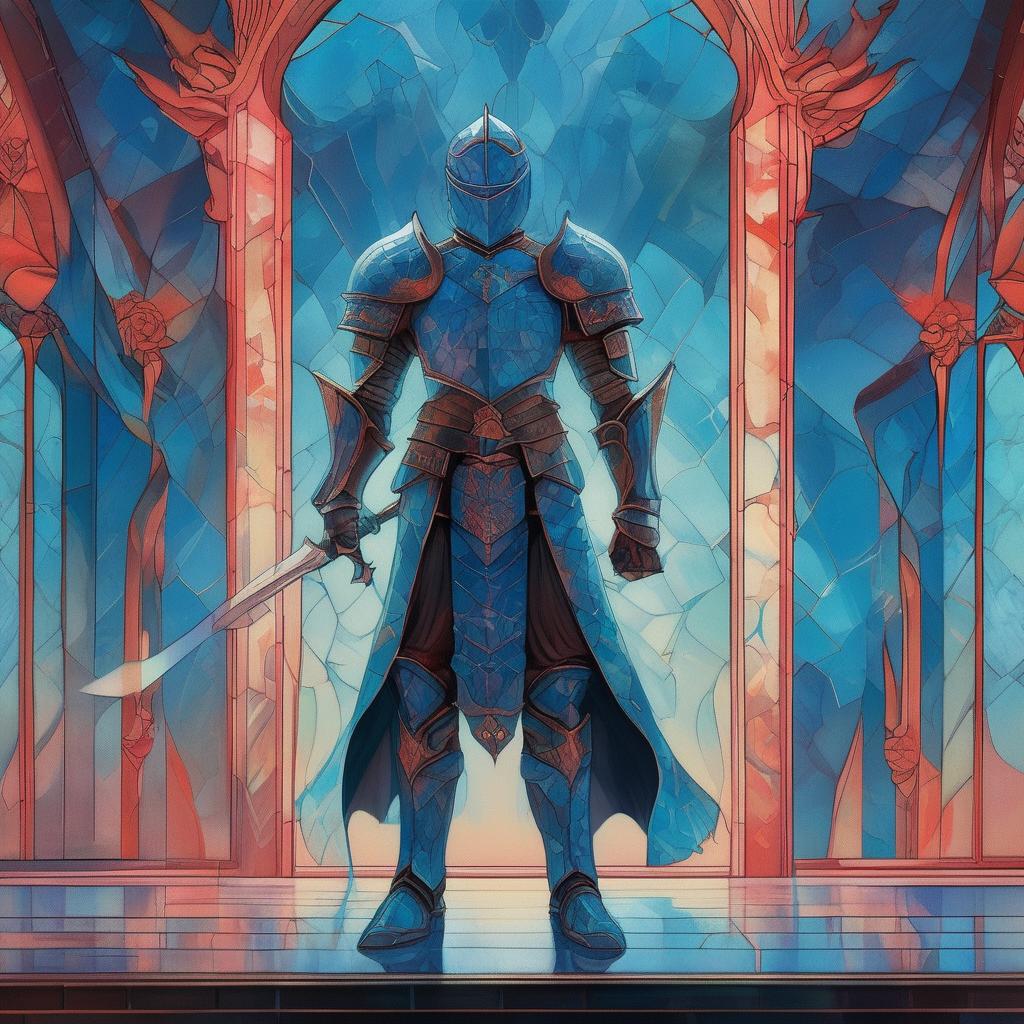}\end{minipage}%
  \begin{minipage}{0.15\textwidth}\includegraphics[width=\linewidth]{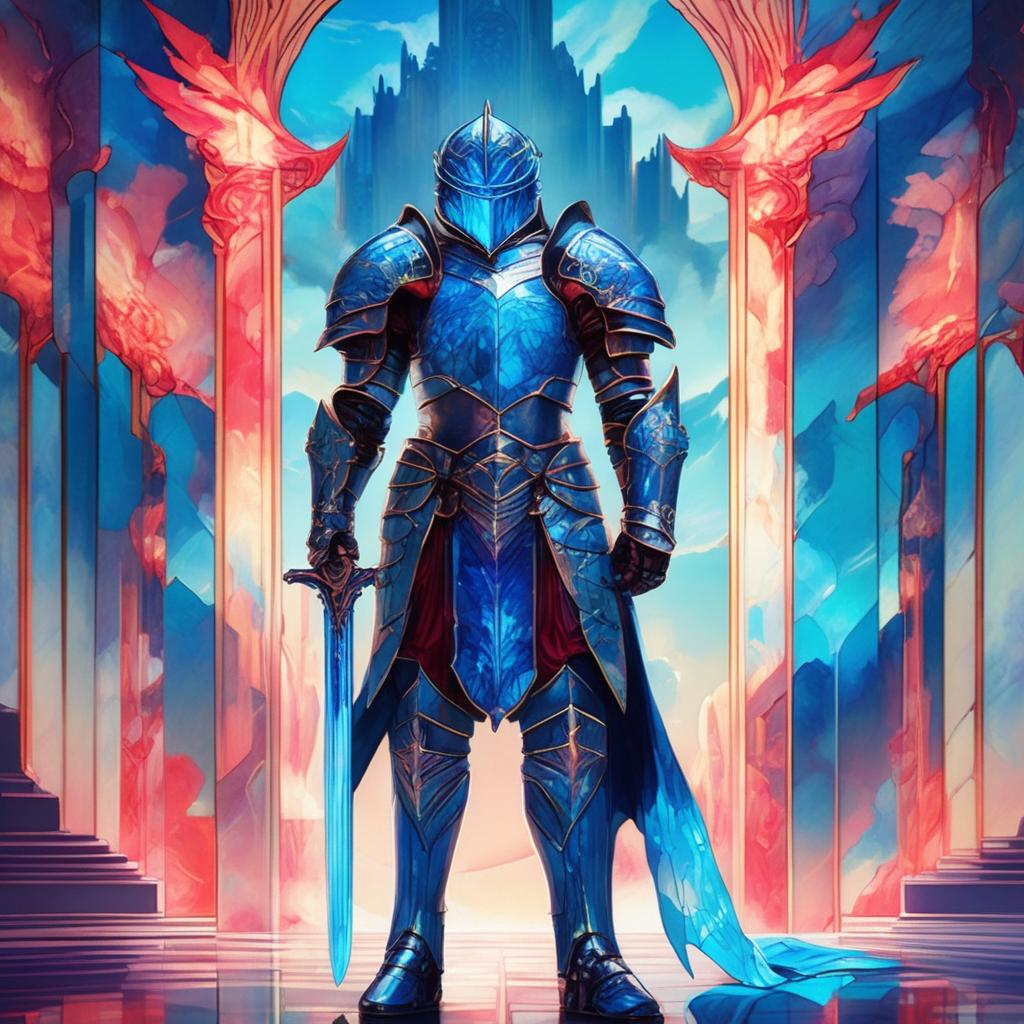}\end{minipage}%

  \begin{minipage}{0.25\textwidth}
  \begin{minipage}{0.90\textwidth}
    \centering \scriptsize \raggedright {The image depicts a stunning supernova within a fantasy artwork on Artstation.} 
  \end{minipage}
  \end{minipage}%
  \begin{minipage}{0.15\textwidth}\includegraphics[width=\linewidth]{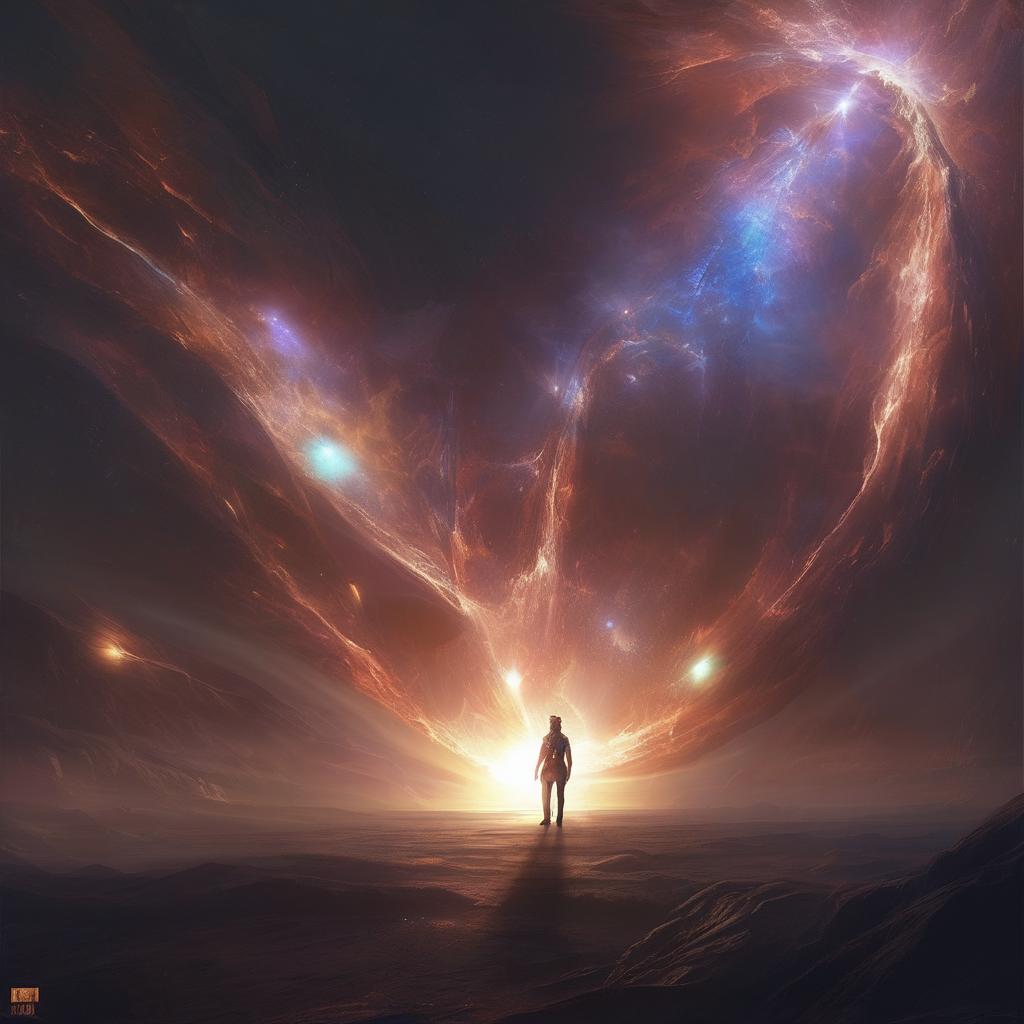}\end{minipage}%
  \begin{minipage}{0.15\textwidth}\includegraphics[width=\linewidth]{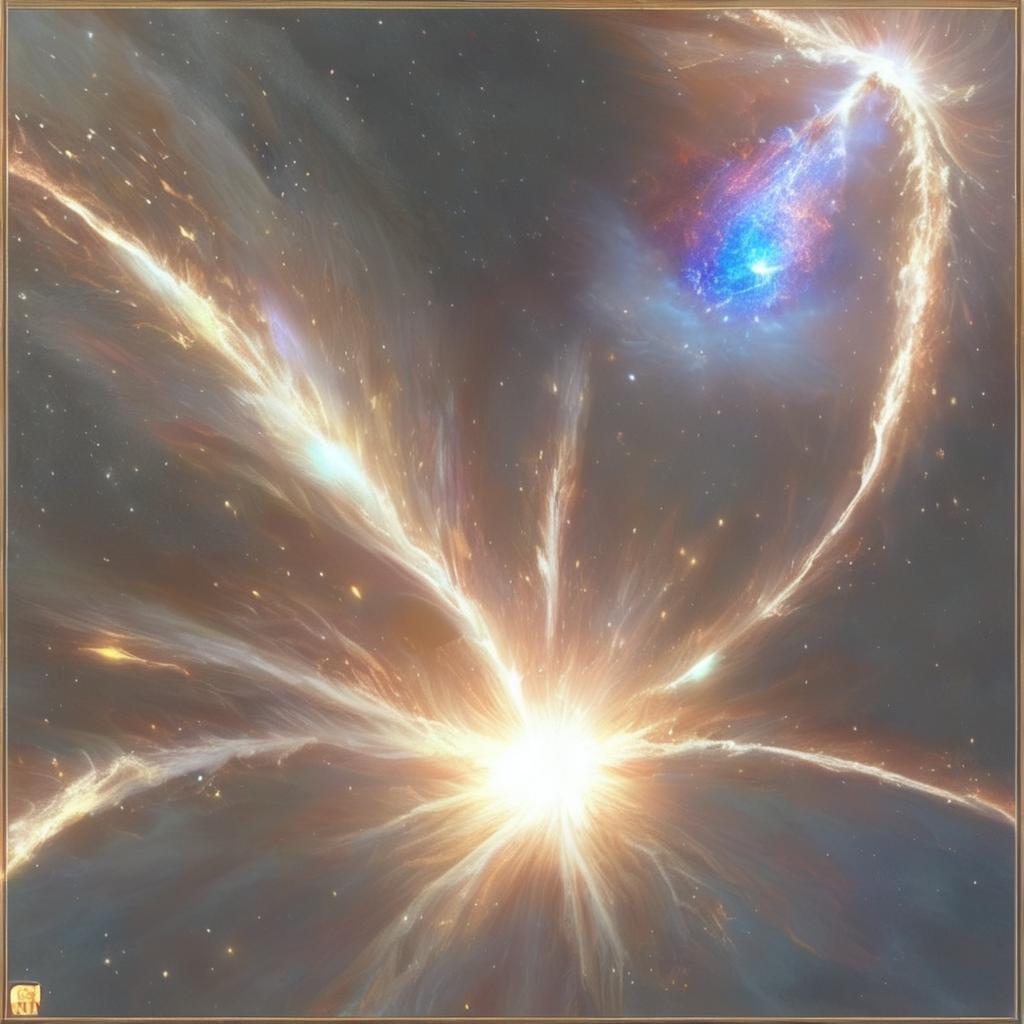}\end{minipage}%
  \begin{minipage}{0.15\textwidth}\includegraphics[width=\linewidth]{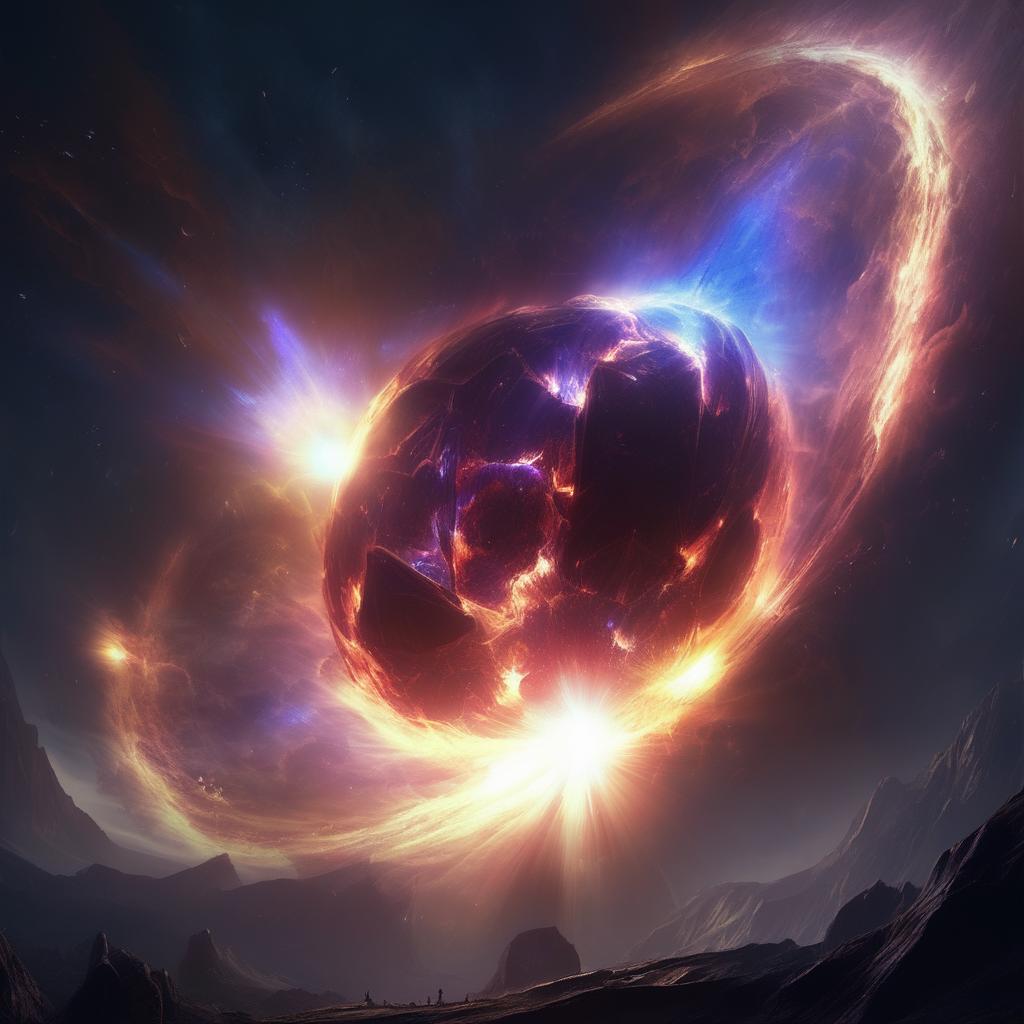}\end{minipage}%
  \begin{minipage}{0.15\textwidth}\includegraphics[width=\linewidth]{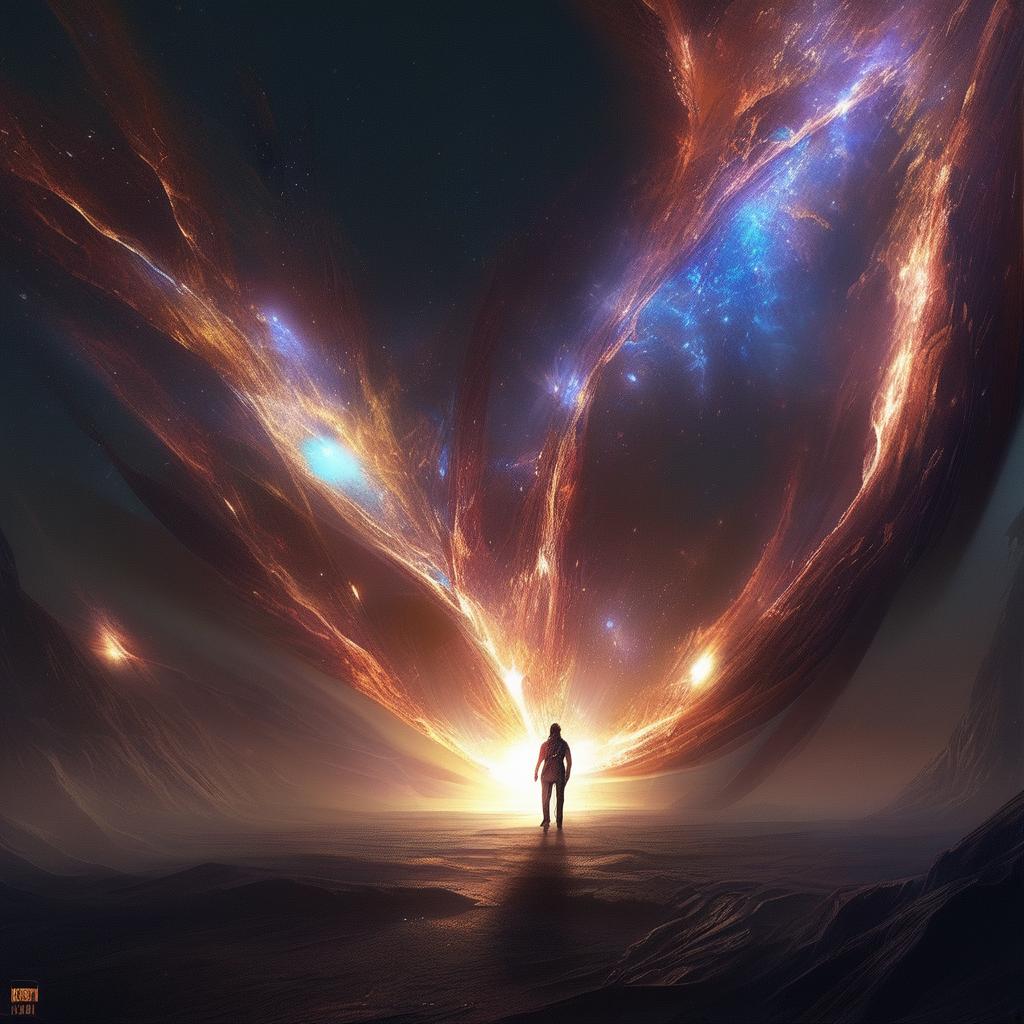}\end{minipage}%
  \begin{minipage}{0.15\textwidth}\includegraphics[width=\linewidth]{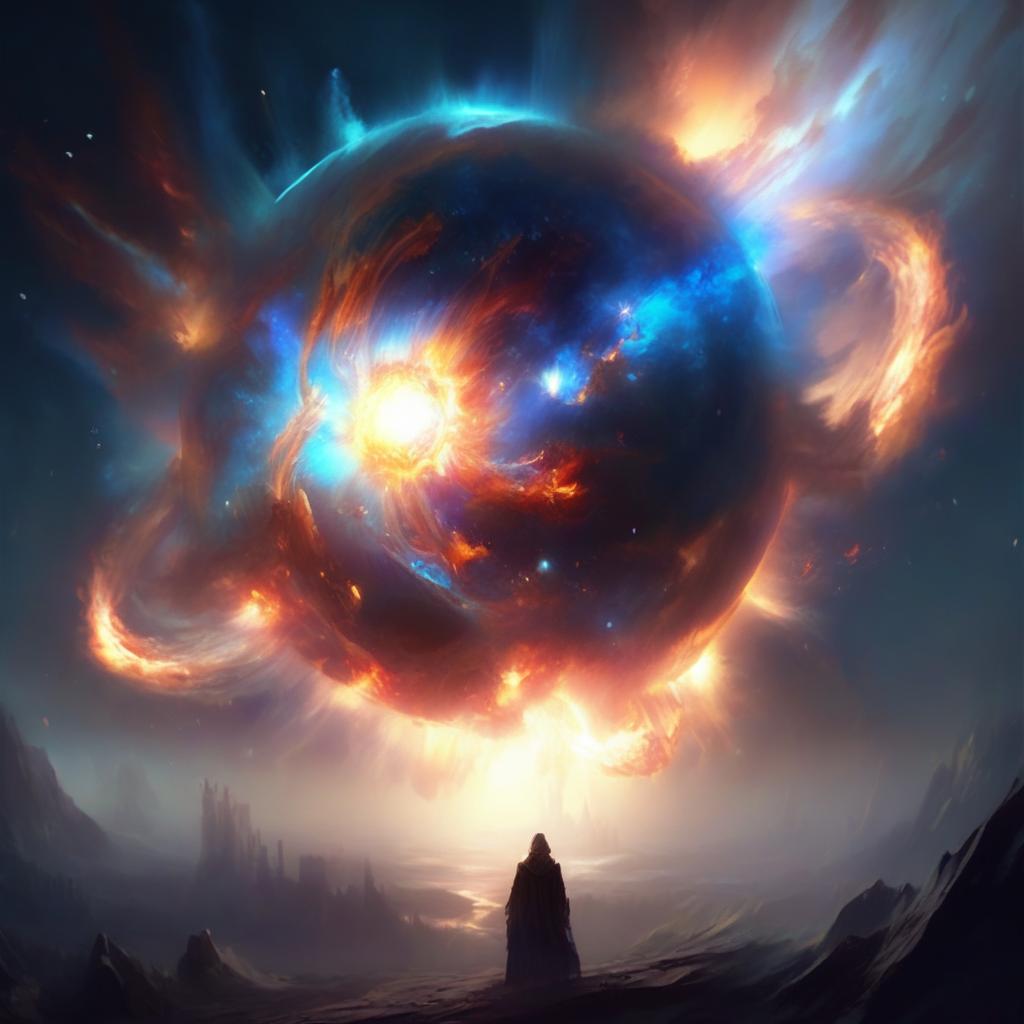}\end{minipage}%

  \begin{minipage}{0.25\textwidth}
  \begin{minipage}{0.90\textwidth}
    \centering \scriptsize \raggedright {motion} 
  \end{minipage}
  \end{minipage}%
  \begin{minipage}{0.15\textwidth}\includegraphics[width=\linewidth]{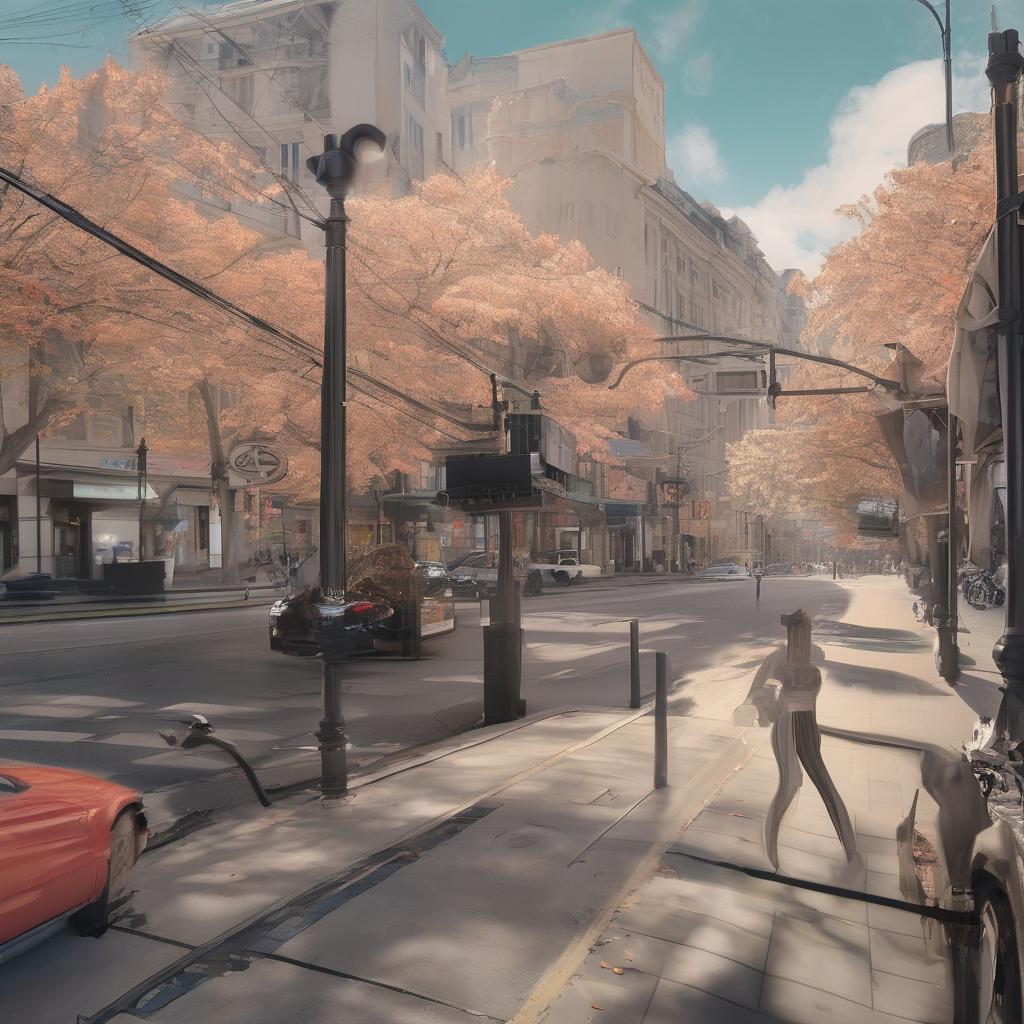}\end{minipage}%
  \begin{minipage}{0.15\textwidth}\includegraphics[width=\linewidth]{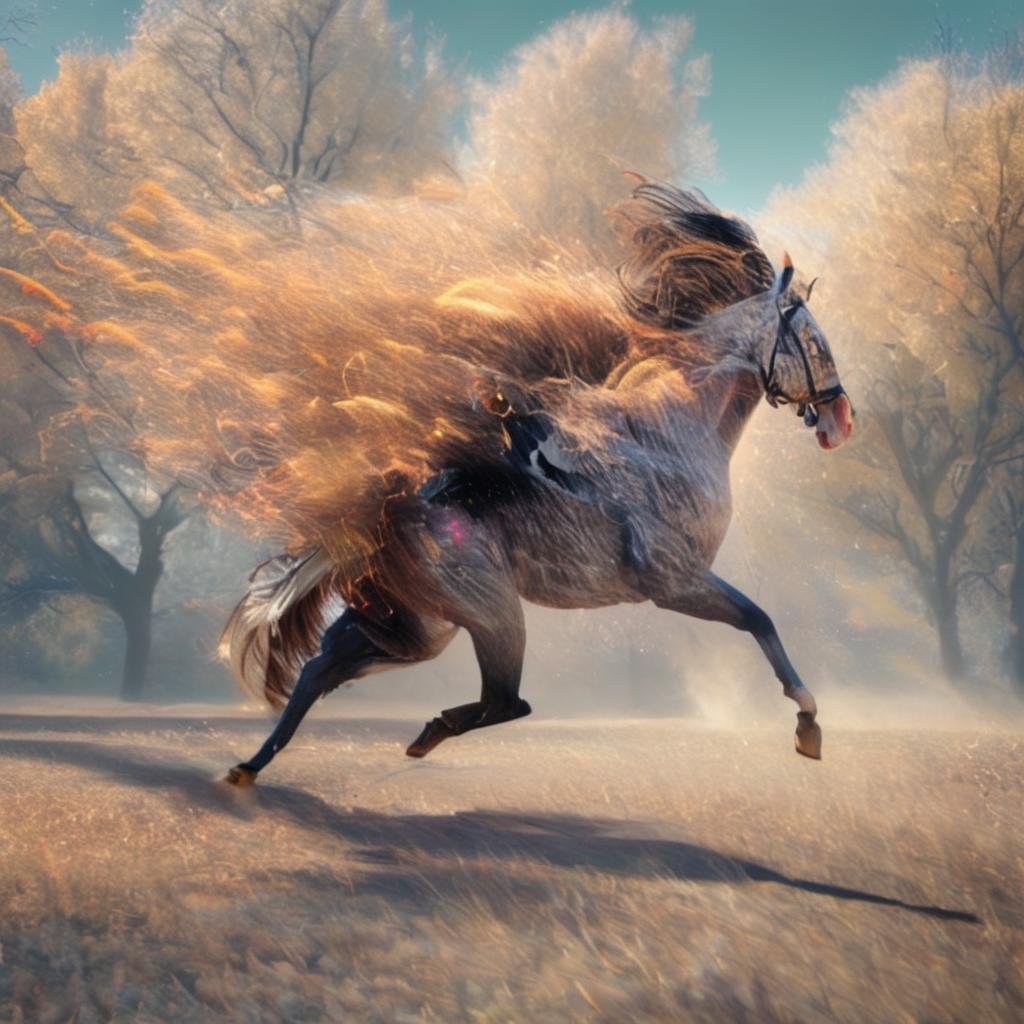}\end{minipage}%
  \begin{minipage}{0.15\textwidth}\includegraphics[width=\linewidth]{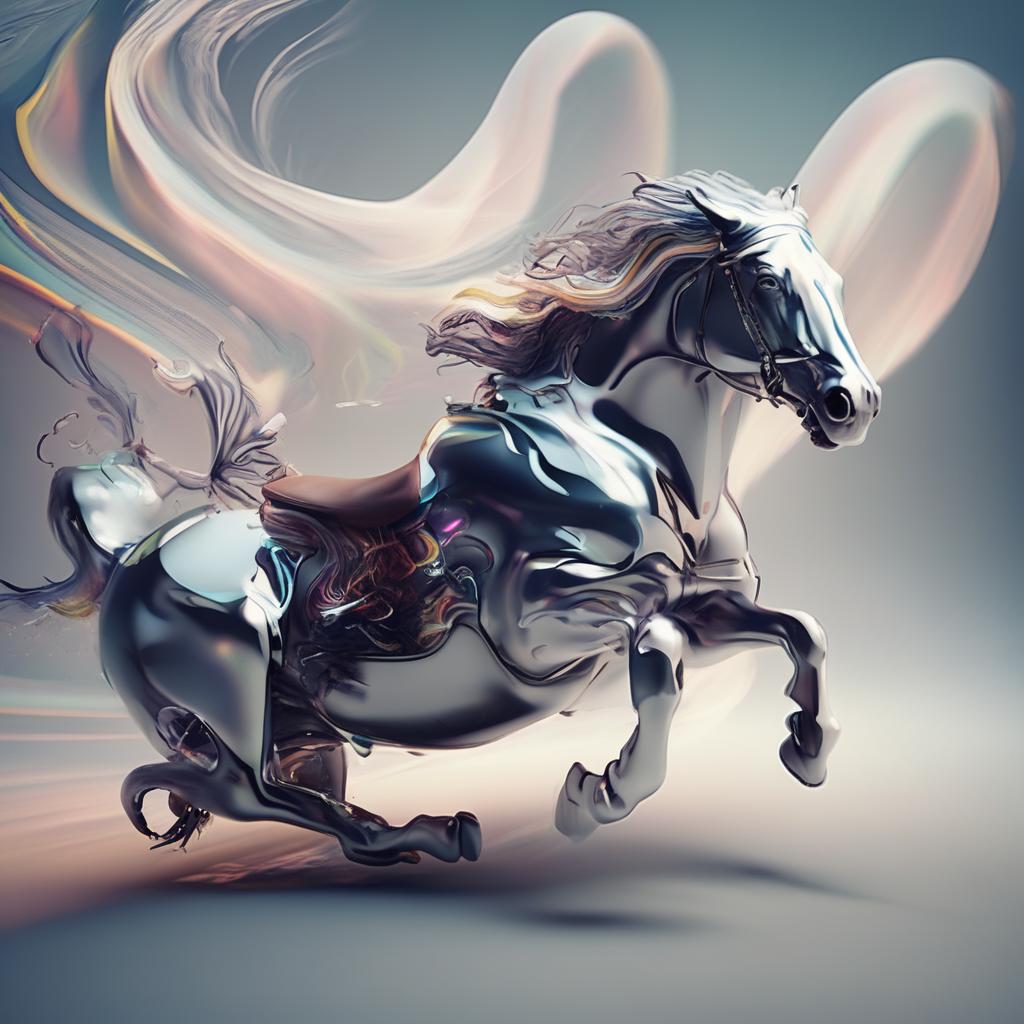}\end{minipage}%
  \begin{minipage}{0.15\textwidth}\includegraphics[width=\linewidth]{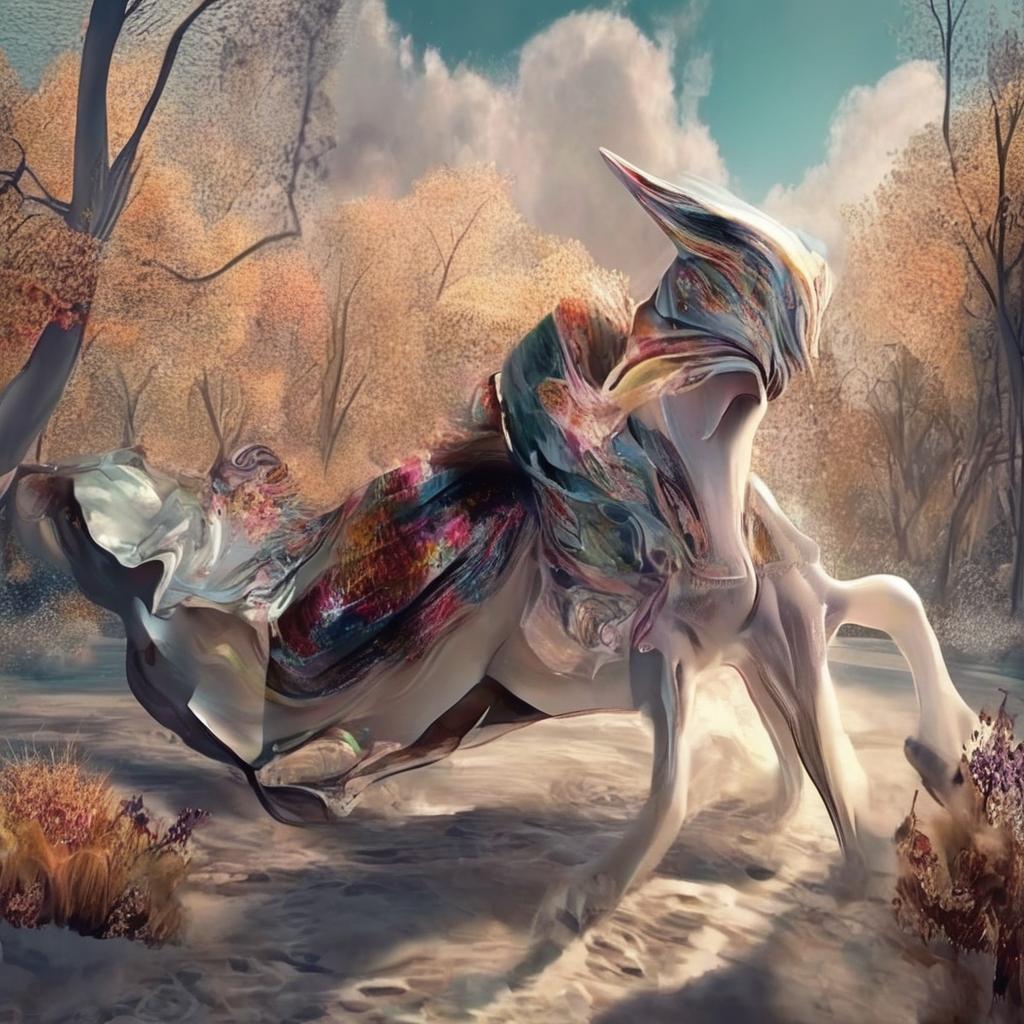}\end{minipage}%
  \begin{minipage}{0.15\textwidth}\includegraphics[width=\linewidth]{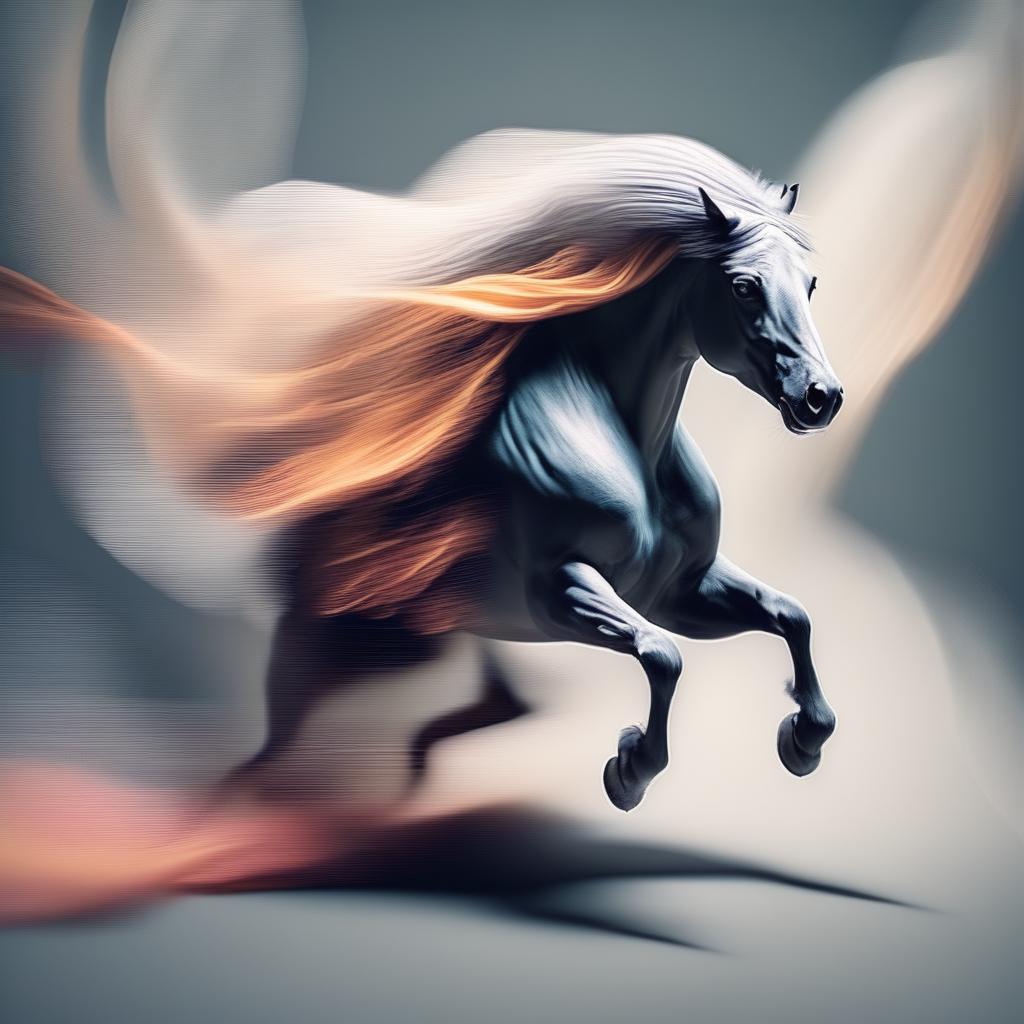}\end{minipage}%

  \end{minipage}
  \caption{Image samples generated from SDXL fine-tuned with various methods, using validation prompts from Pick-a-Pic v2, HPDv2 and PartiPrompt.}
\end{figure*}
\begin{figure*}[htbp]
  \centering
  \begin{minipage}{0.95\textwidth}
  \centering
  \begin{minipage}{0.28\textwidth} \centering \small \textbf{Prompt} 
  \end{minipage}%
  \begin{minipage}{0.18\textwidth} \centering \small \textbf{SD3} \end{minipage}%
  \begin{minipage}{0.18\textwidth} \centering \small \textbf{SFT} \end{minipage}%
  \begin{minipage}{0.18\textwidth} \centering \small \textbf{Diffusion-DPO} \end{minipage}%
  \begin{minipage}{0.18\textwidth} \centering \small \textbf{Linear-DPO} \end{minipage}
  
  \vspace{0.2em}
  
  \begin{minipage}{0.28\textwidth}
  \begin{minipage}{0.90\textwidth}
    \centering \footnotesize \raggedright {A Sign that says: Free Candy!} 
  \end{minipage}
  \end{minipage}%
  \begin{minipage}{0.18\textwidth}\includegraphics[width=\linewidth]{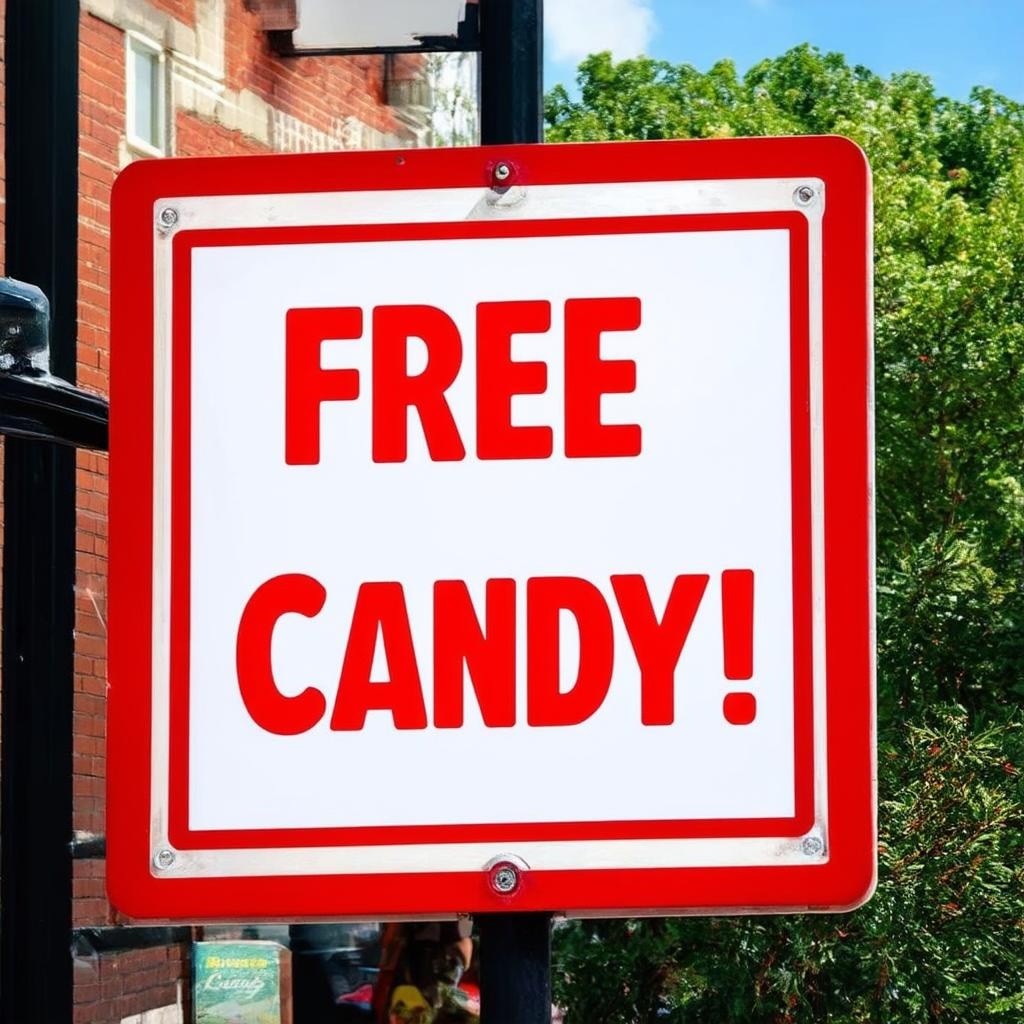}\end{minipage}%
  \begin{minipage}{0.18\textwidth}\includegraphics[width=\linewidth]{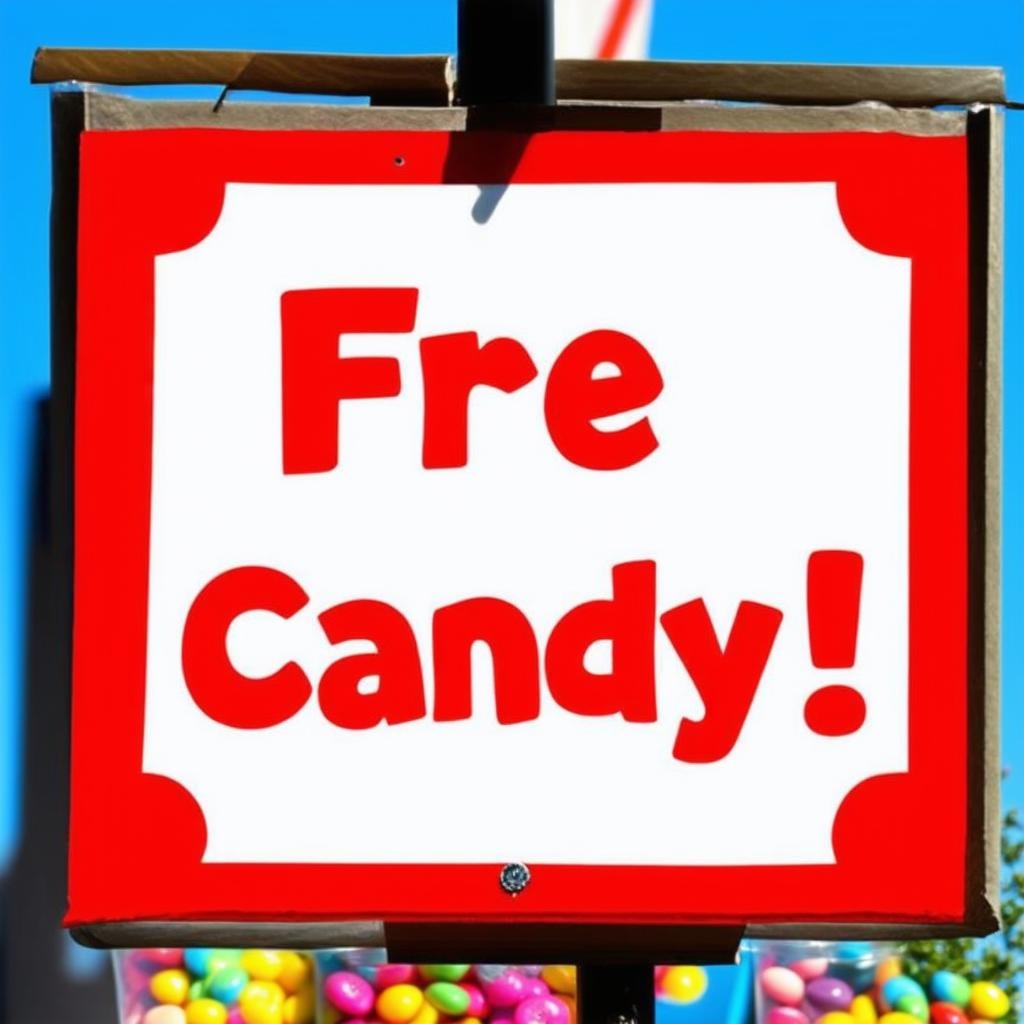}\end{minipage}%
  \begin{minipage}{0.18\textwidth}\includegraphics[width=\linewidth]{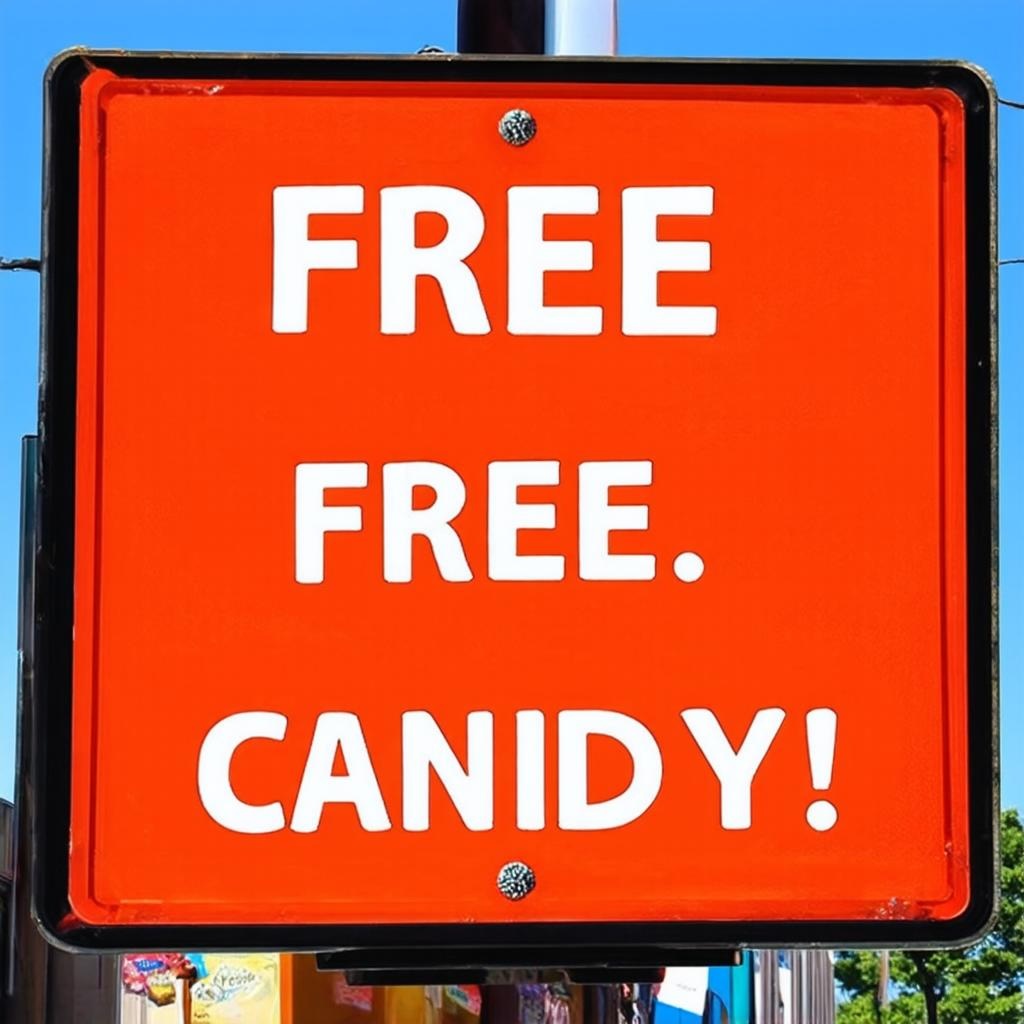}\end{minipage}%
  \begin{minipage}{0.18\textwidth}\includegraphics[width=\linewidth]{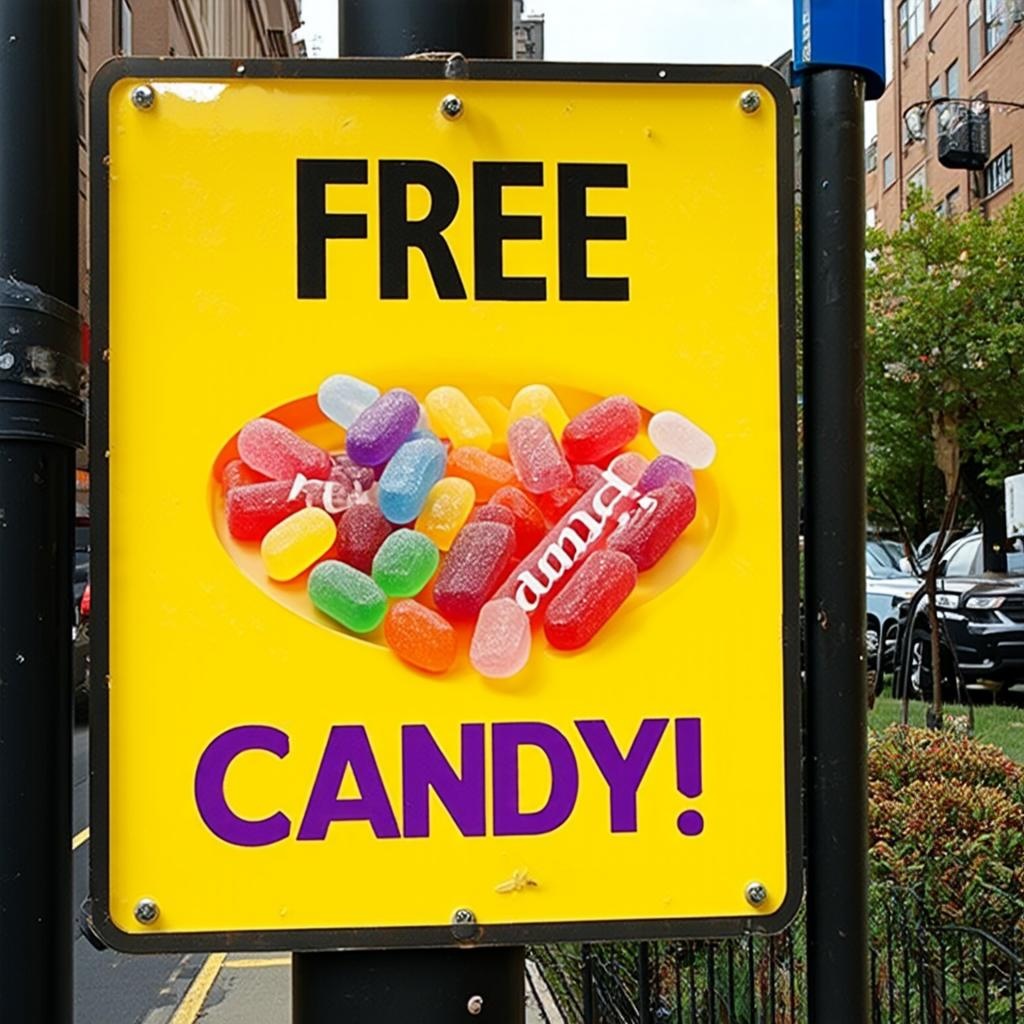}\end{minipage}%

  \begin{minipage}{0.28\textwidth}
  \begin{minipage}{0.90\textwidth}
    \centering \footnotesize \raggedright {pinocchio as superhero} 
  \end{minipage}
  \end{minipage}%
  \begin{minipage}{0.18\textwidth}\includegraphics[width=\linewidth]{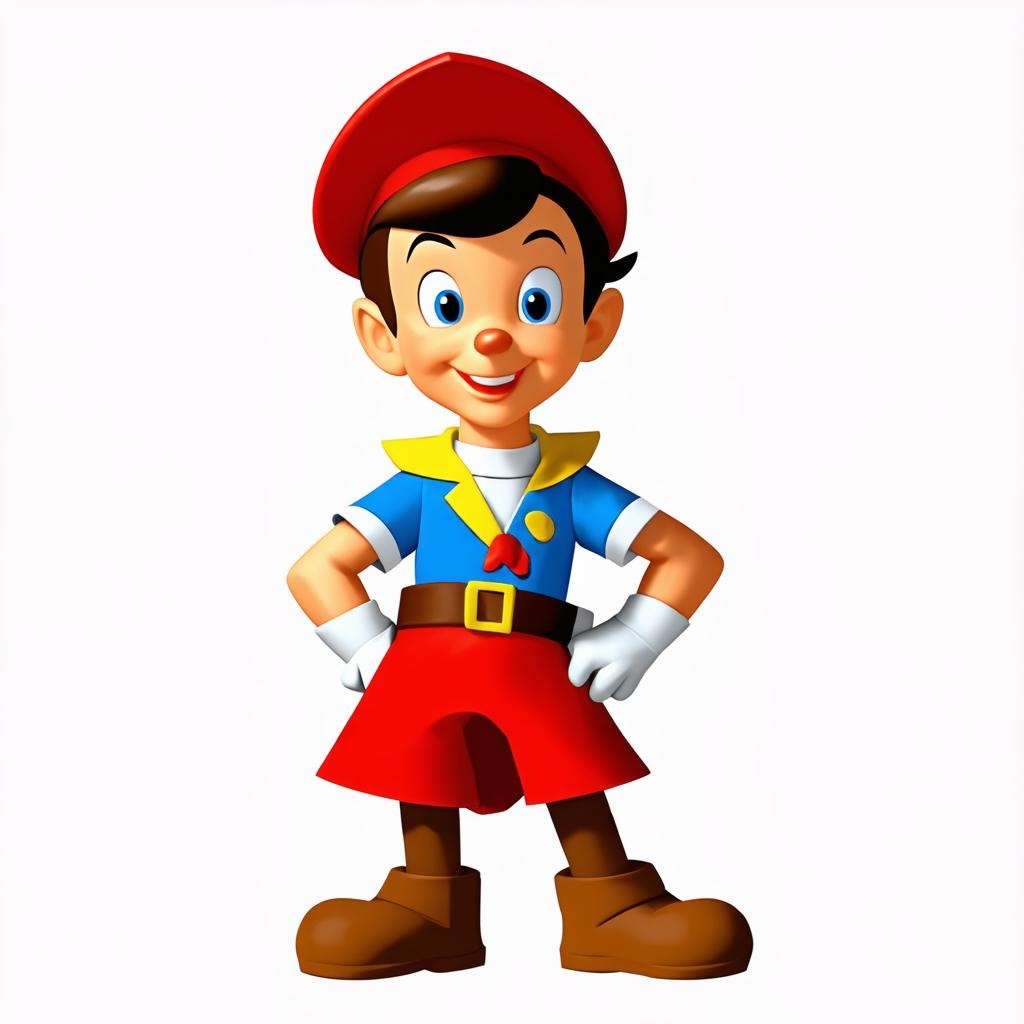}\end{minipage}%
  \begin{minipage}{0.18\textwidth}\includegraphics[width=\linewidth]{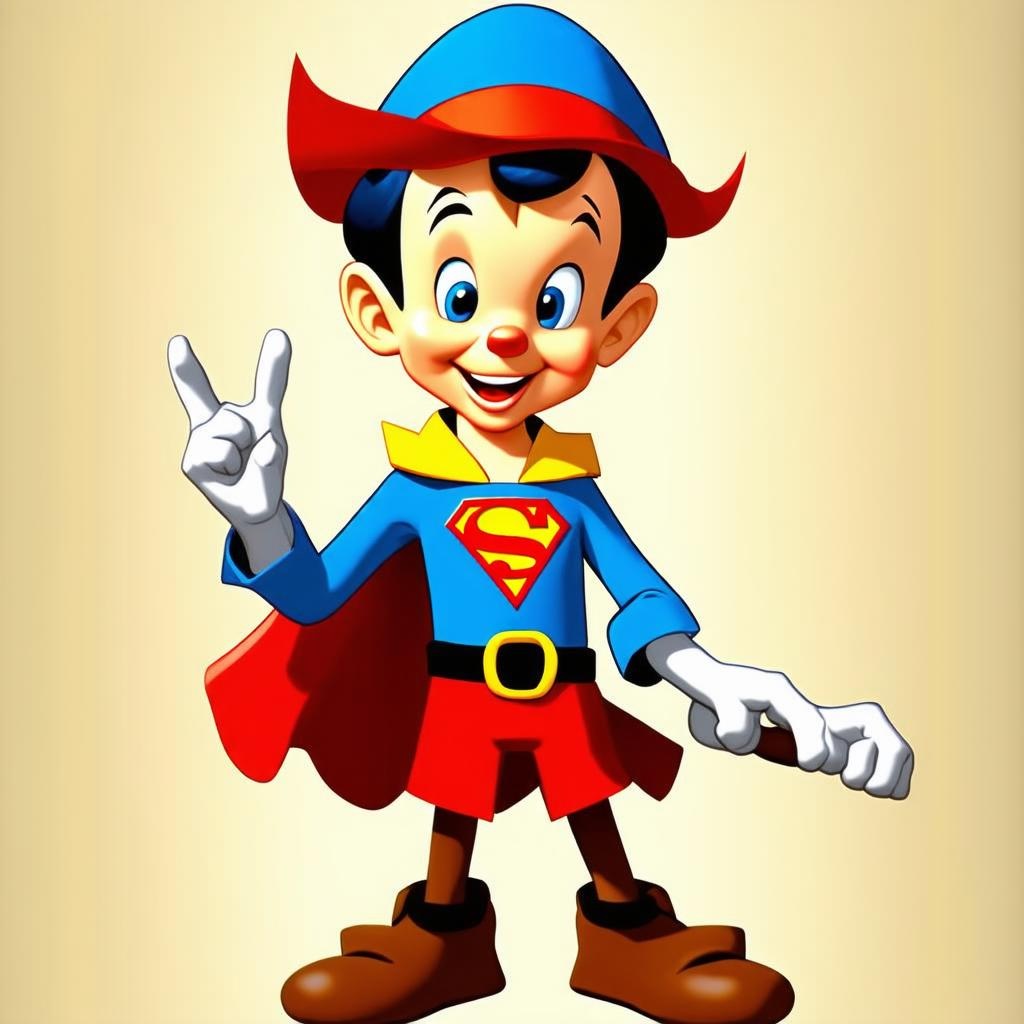}\end{minipage}%
  \begin{minipage}{0.18\textwidth}\includegraphics[width=\linewidth]{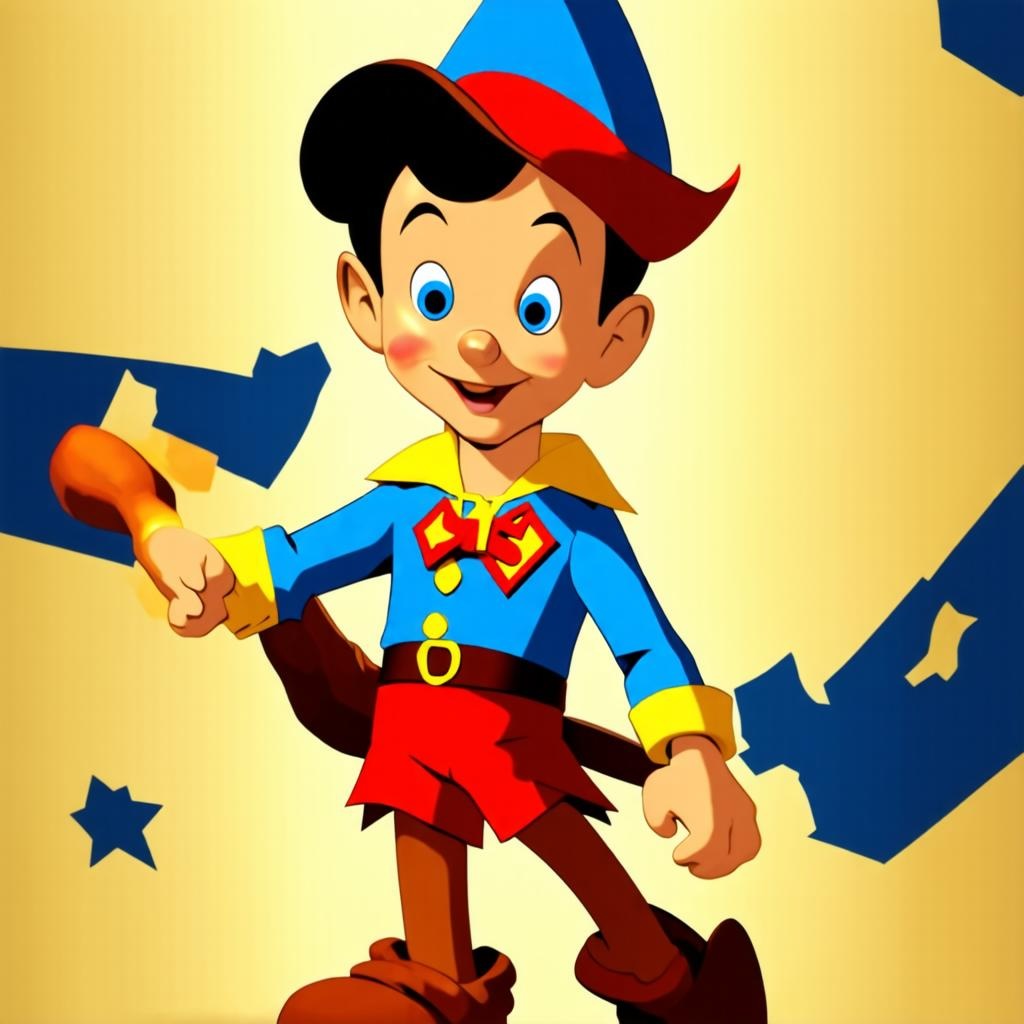}\end{minipage}%
  \begin{minipage}{0.18\textwidth}\includegraphics[width=\linewidth]{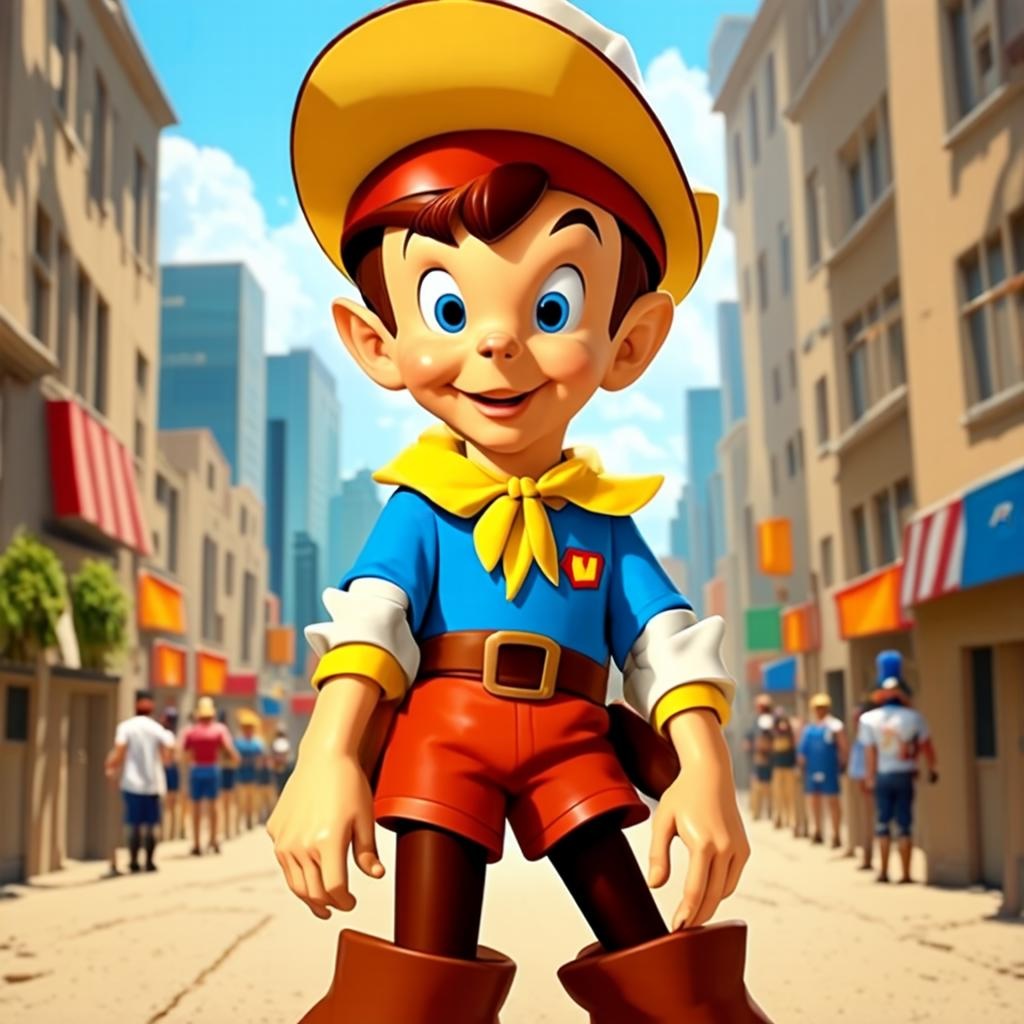}\end{minipage}%
  
  \begin{minipage}{0.28\textwidth}
  \begin{minipage}{0.90\textwidth}
    \centering \footnotesize \raggedright {an impressionist painting of the geyser Old Faithful} 
  \end{minipage}
  \end{minipage}%
  \begin{minipage}{0.18\textwidth}\includegraphics[width=\linewidth]{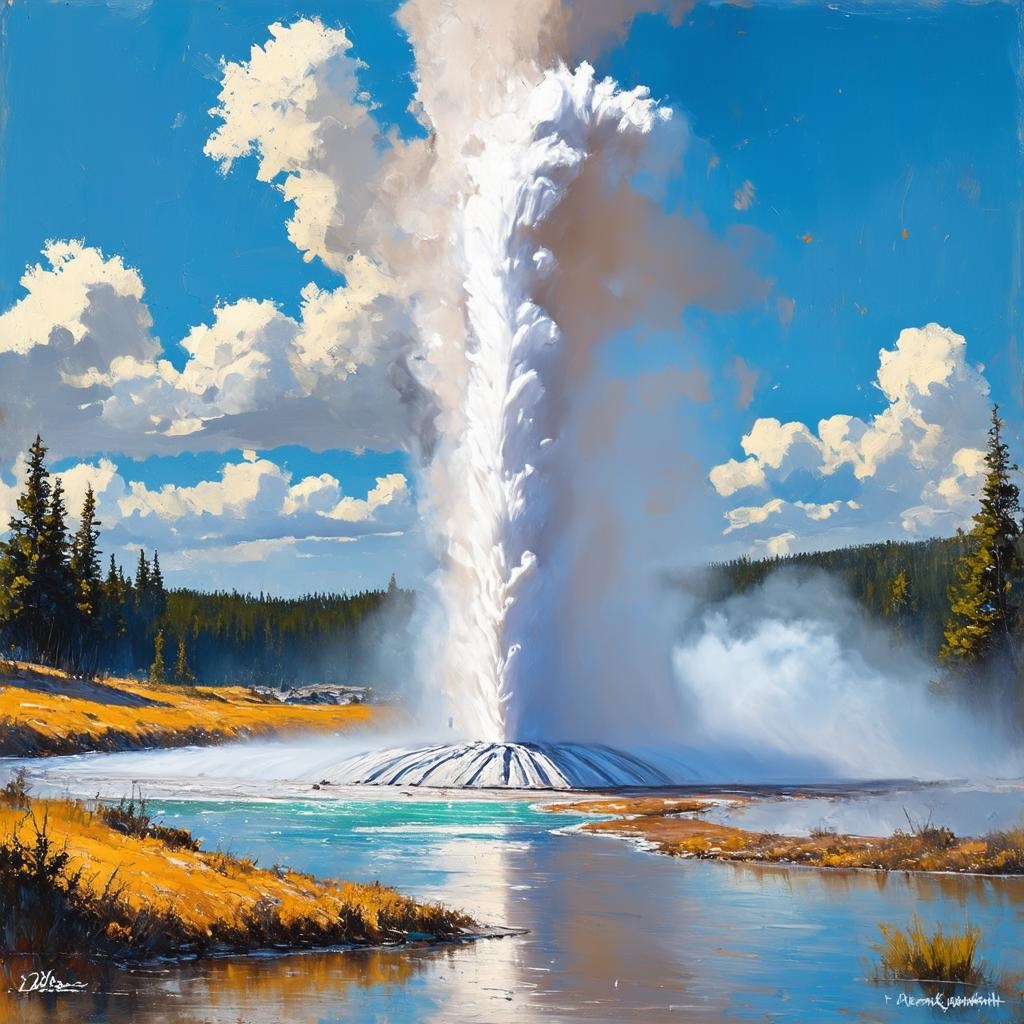}\end{minipage}%
  \begin{minipage}{0.18\textwidth}\includegraphics[width=\linewidth]{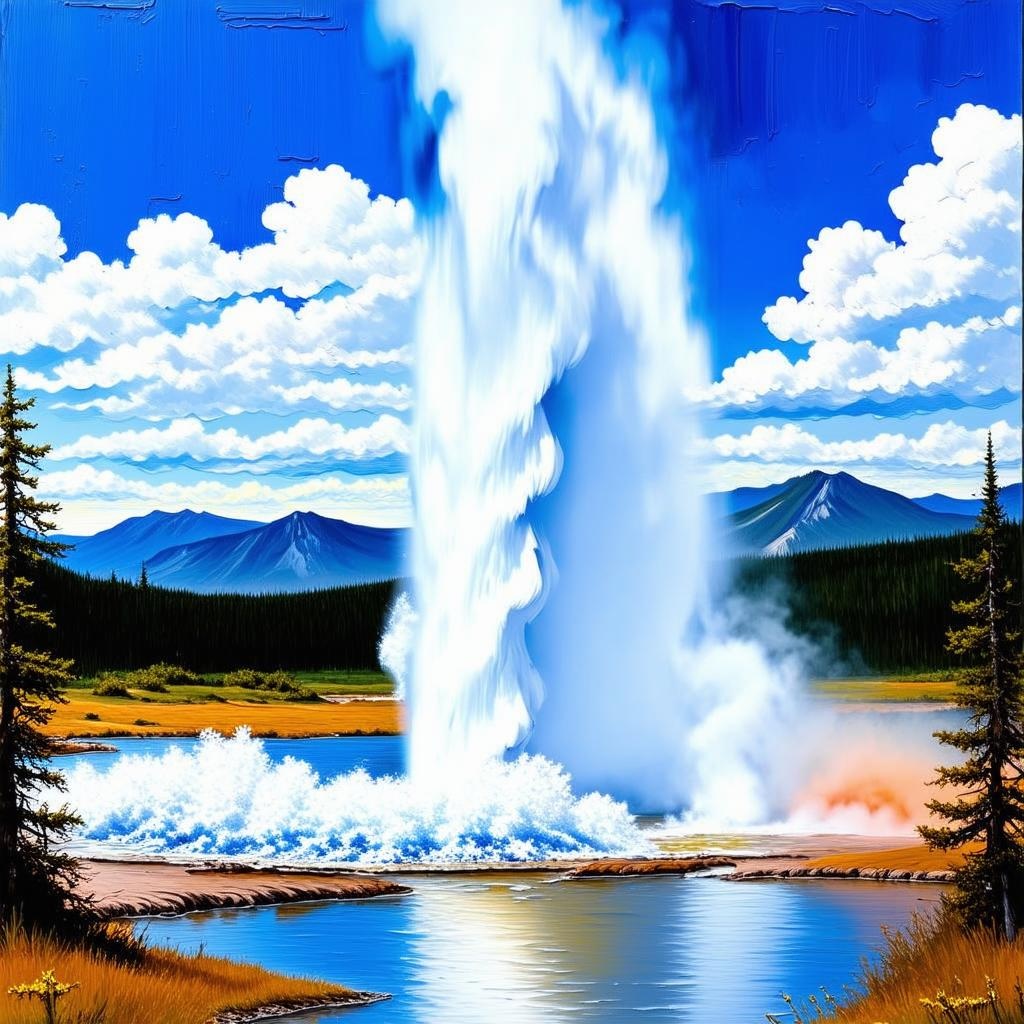}\end{minipage}%
  \begin{minipage}{0.18\textwidth}\includegraphics[width=\linewidth]{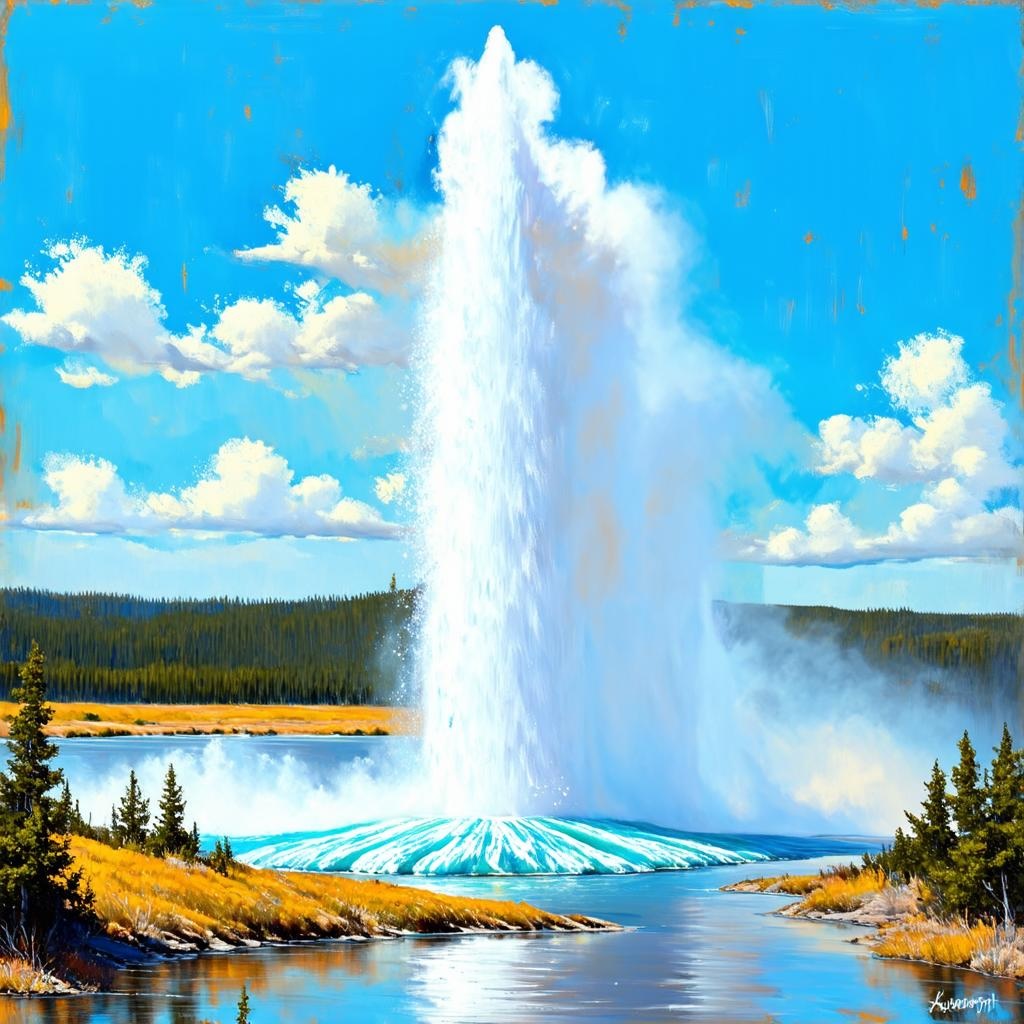}\end{minipage}%
  \begin{minipage}{0.18\textwidth}\includegraphics[width=\linewidth]{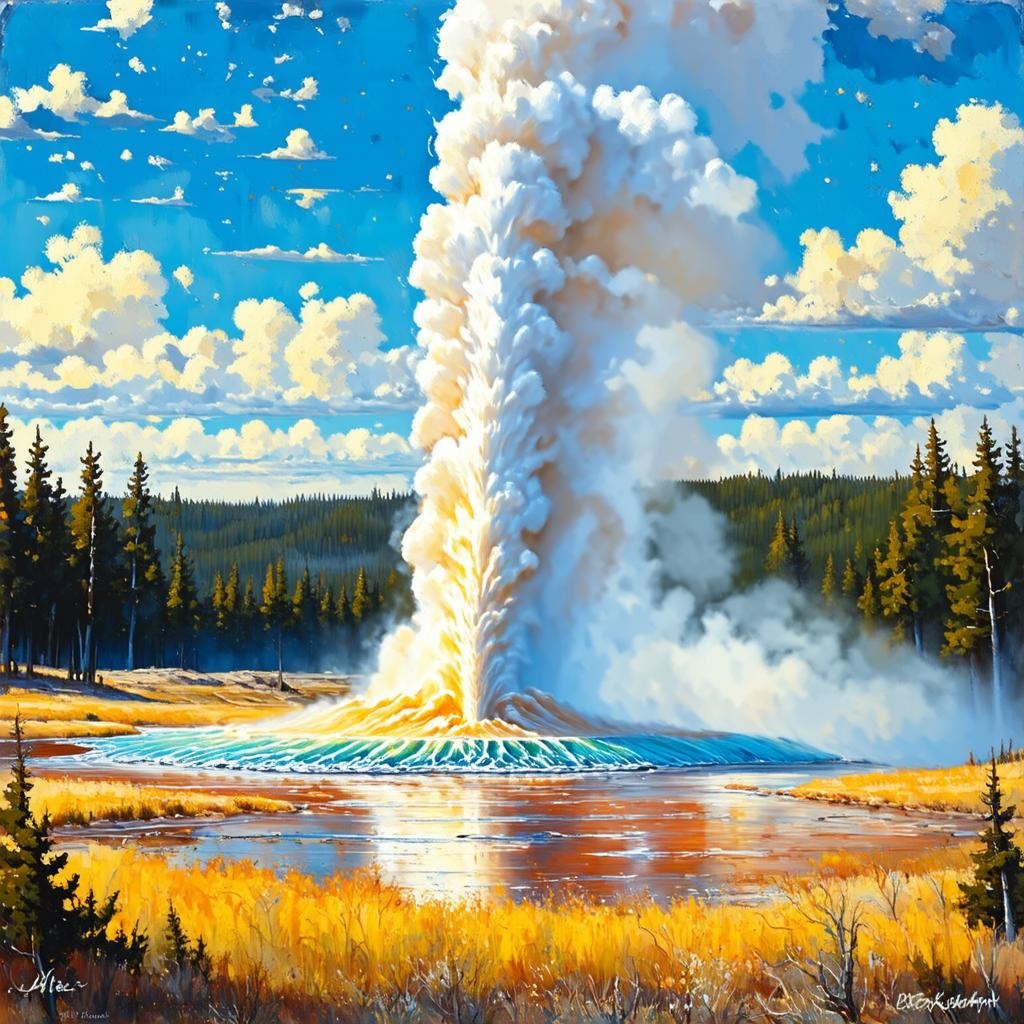}\end{minipage}%
  
  \begin{minipage}{0.28\textwidth}
  \begin{minipage}{0.90\textwidth}
    \centering \footnotesize \raggedright {Anime illustration of a kangaroo holding a sign that says "Starry Night", in front of the Sydney Opera House sitting next to the Eiffel Tower under a blue night sky of roiling energy, exploding yellow stars, and radiating swirls of blu} 
  \end{minipage}
  \end{minipage}%
  \begin{minipage}{0.18\textwidth}\includegraphics[width=\linewidth]{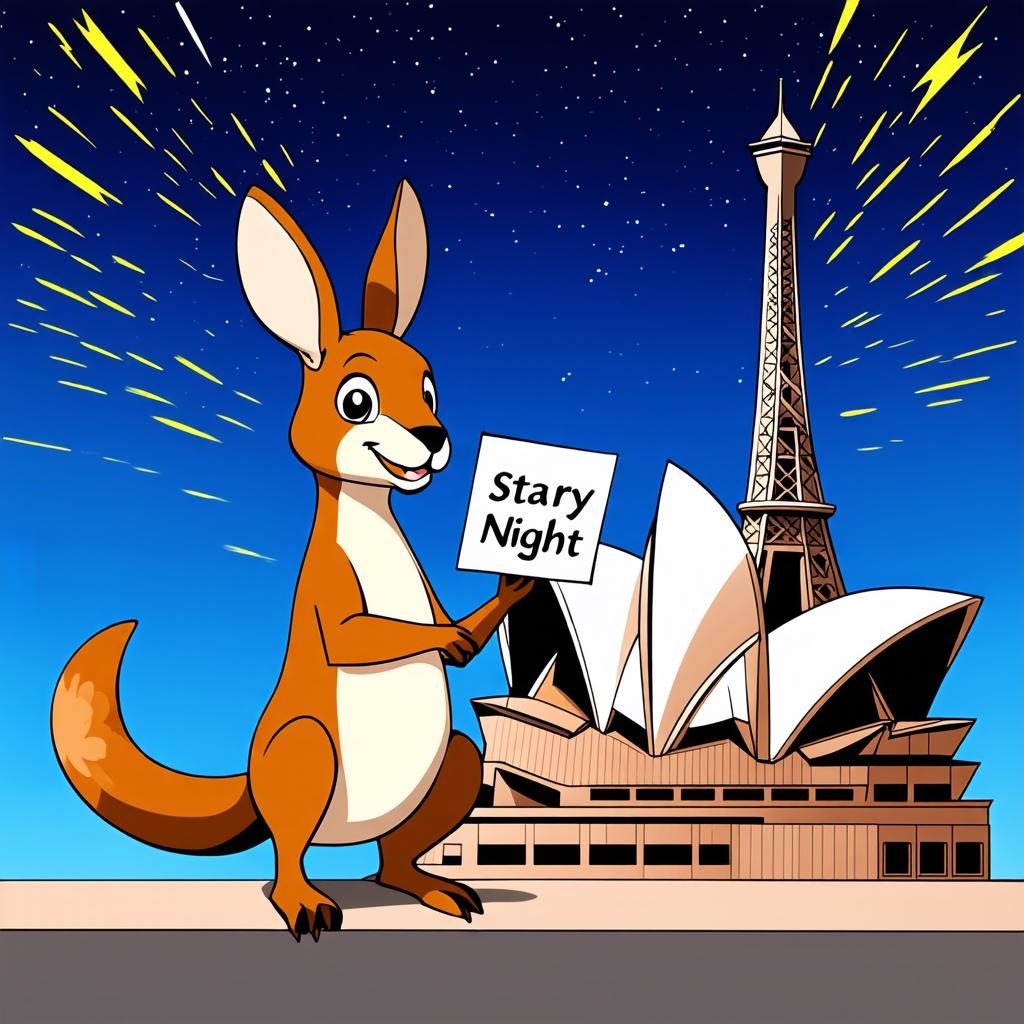}\end{minipage}%
  \begin{minipage}{0.18\textwidth}\includegraphics[width=\linewidth]{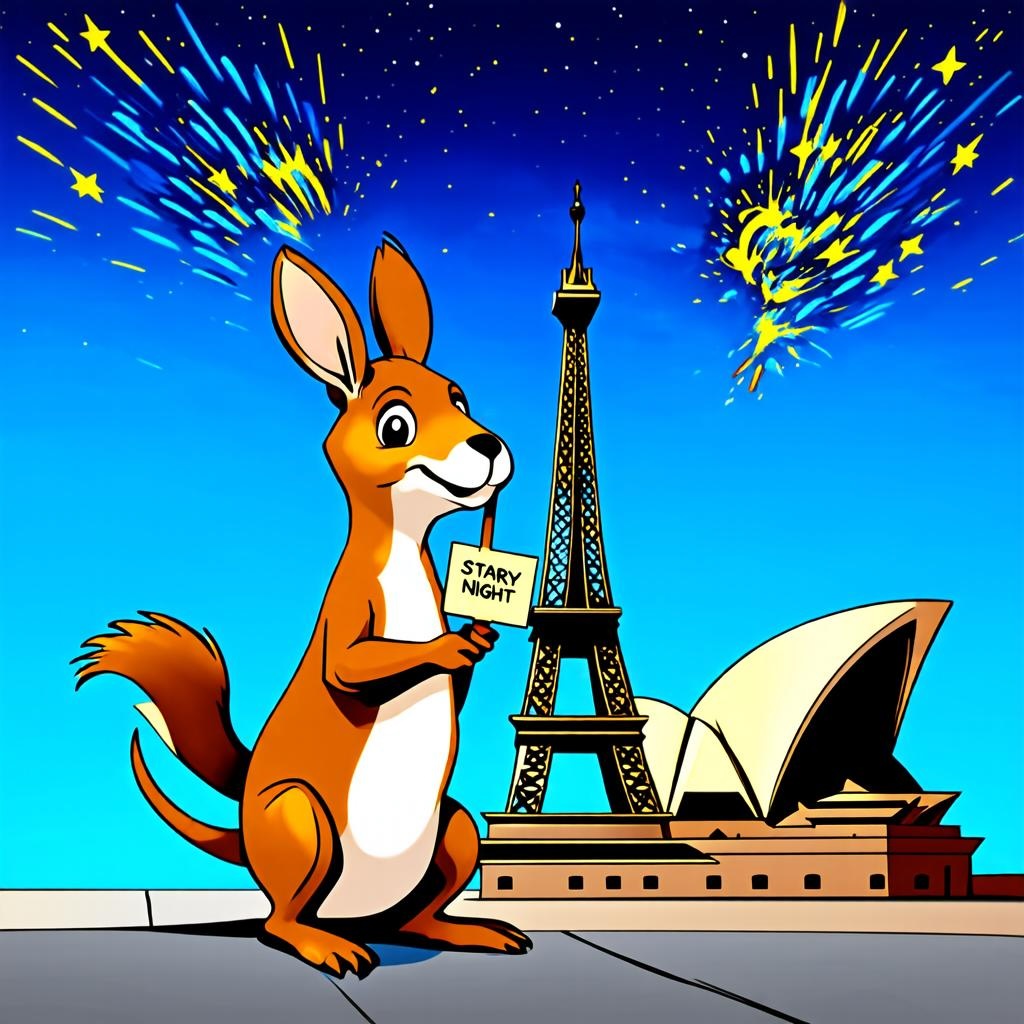}\end{minipage}%
  \begin{minipage}{0.18\textwidth}\includegraphics[width=\linewidth]{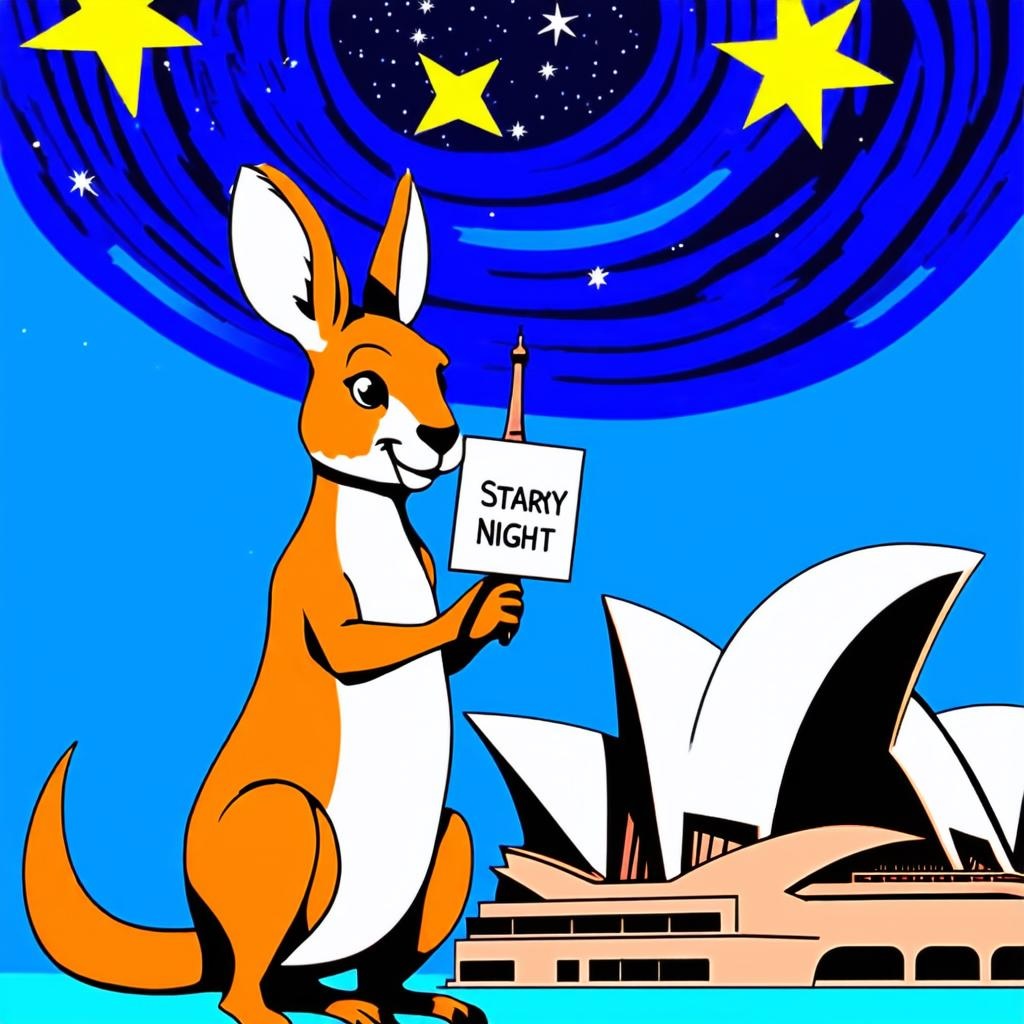}\end{minipage}%
  \begin{minipage}{0.18\textwidth}\includegraphics[width=\linewidth]{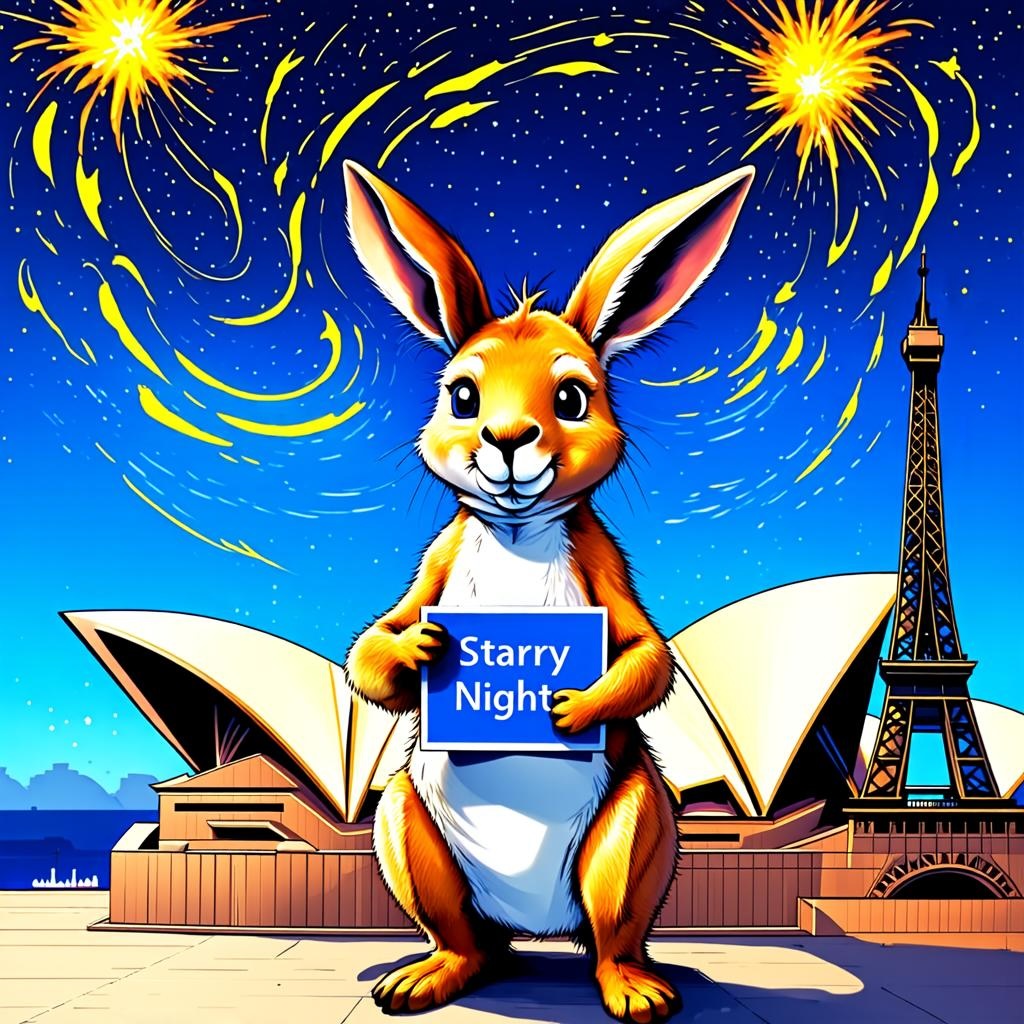}\end{minipage}%
  
  \begin{minipage}{0.28\textwidth}
  \begin{minipage}{0.90\textwidth}
    \centering \footnotesize \raggedright {A nebula forms the shape of a face in this detailed artwork.} 
  \end{minipage}
  \end{minipage}%
  \begin{minipage}{0.18\textwidth}\includegraphics[width=\linewidth]{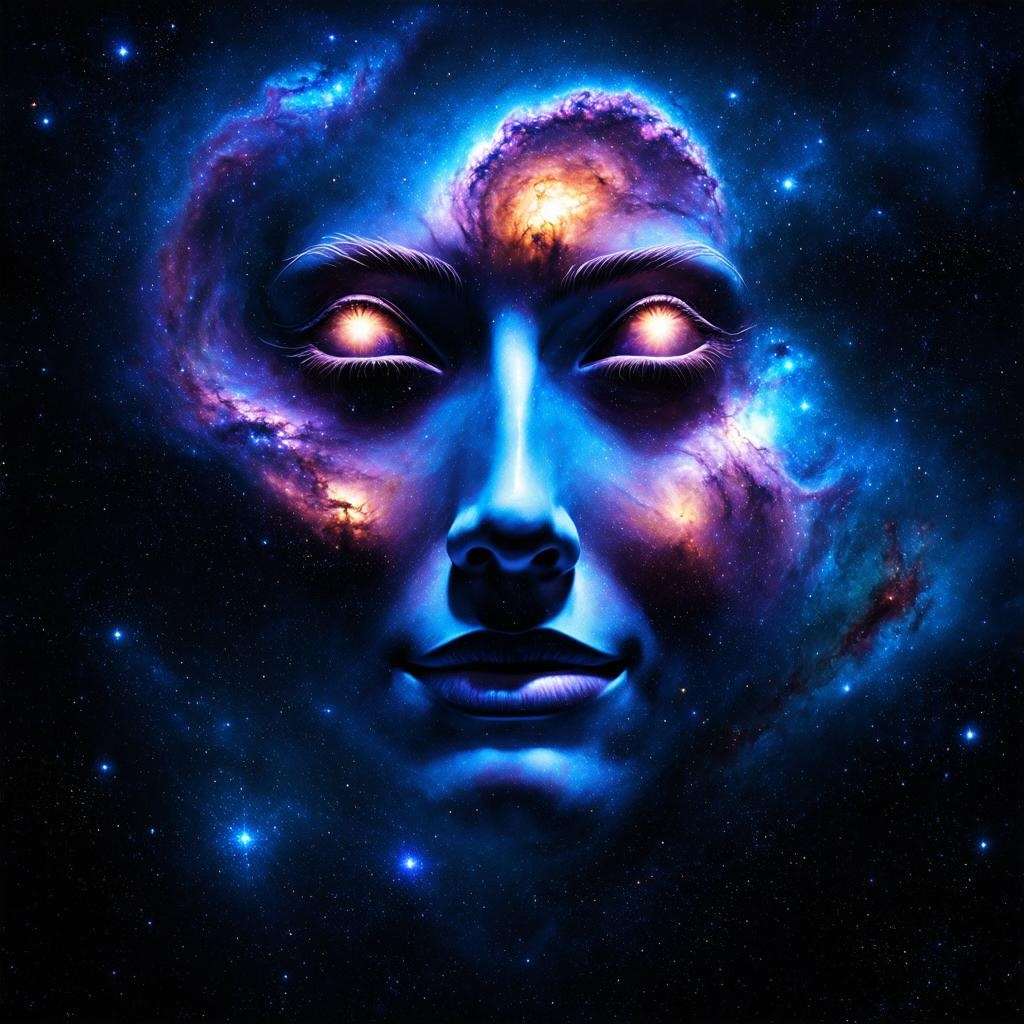}\end{minipage}%
  \begin{minipage}{0.18\textwidth}\includegraphics[width=\linewidth]{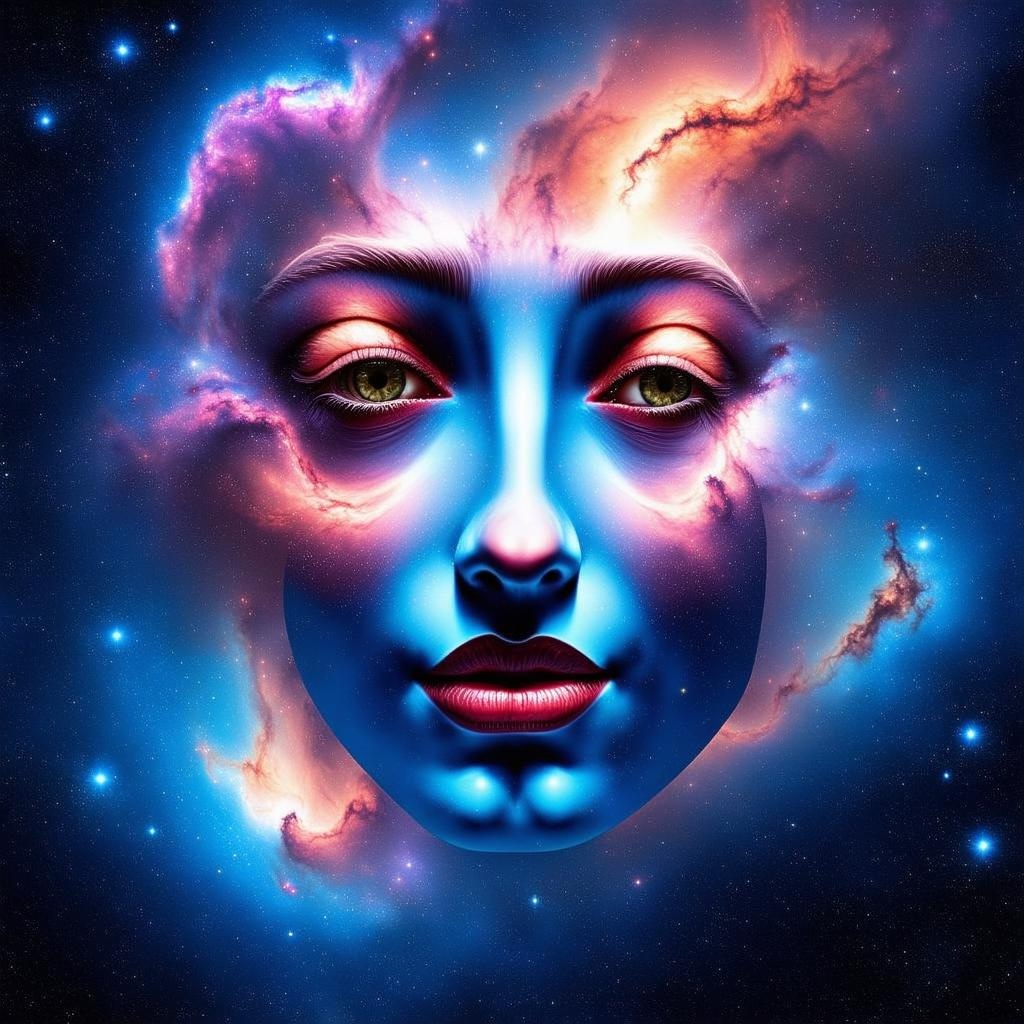}\end{minipage}%
  \begin{minipage}{0.18\textwidth}\includegraphics[width=\linewidth]{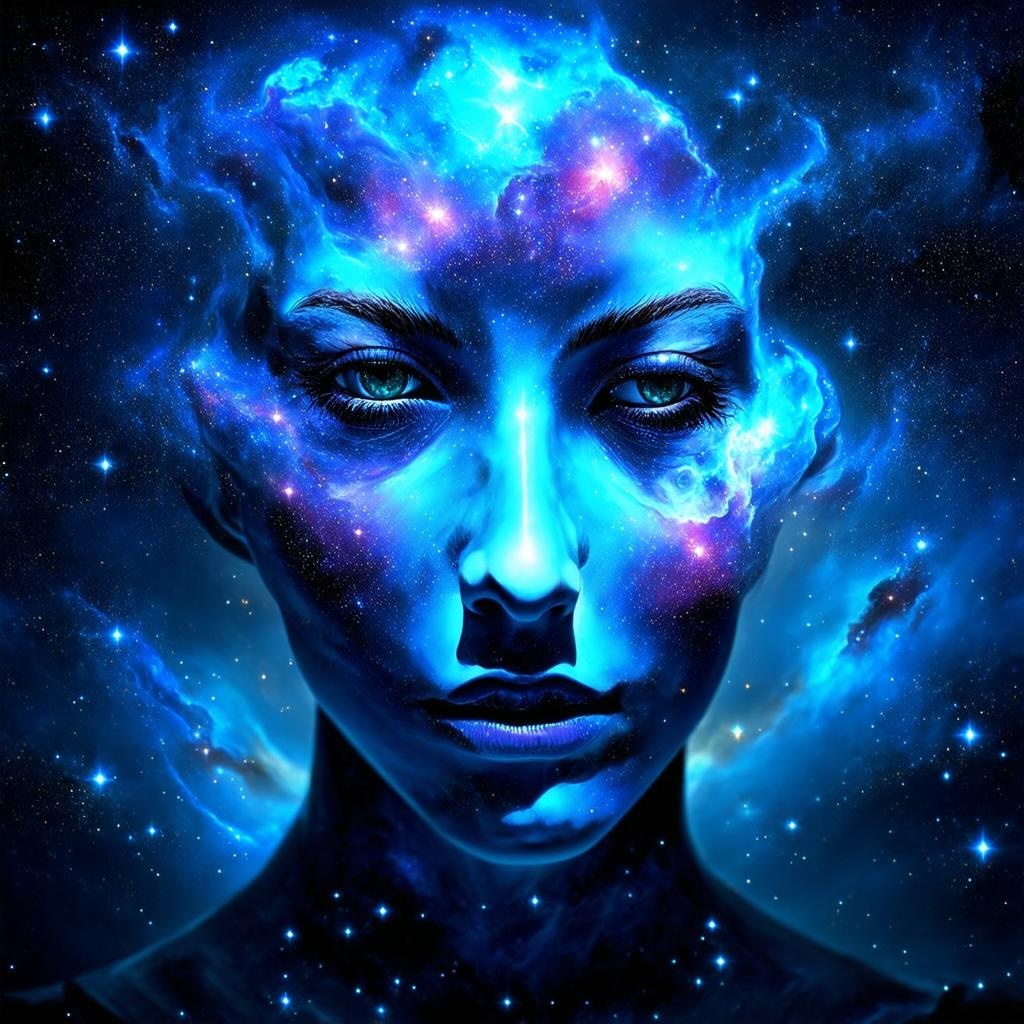}\end{minipage}%
  \begin{minipage}{0.18\textwidth}\includegraphics[width=\linewidth]{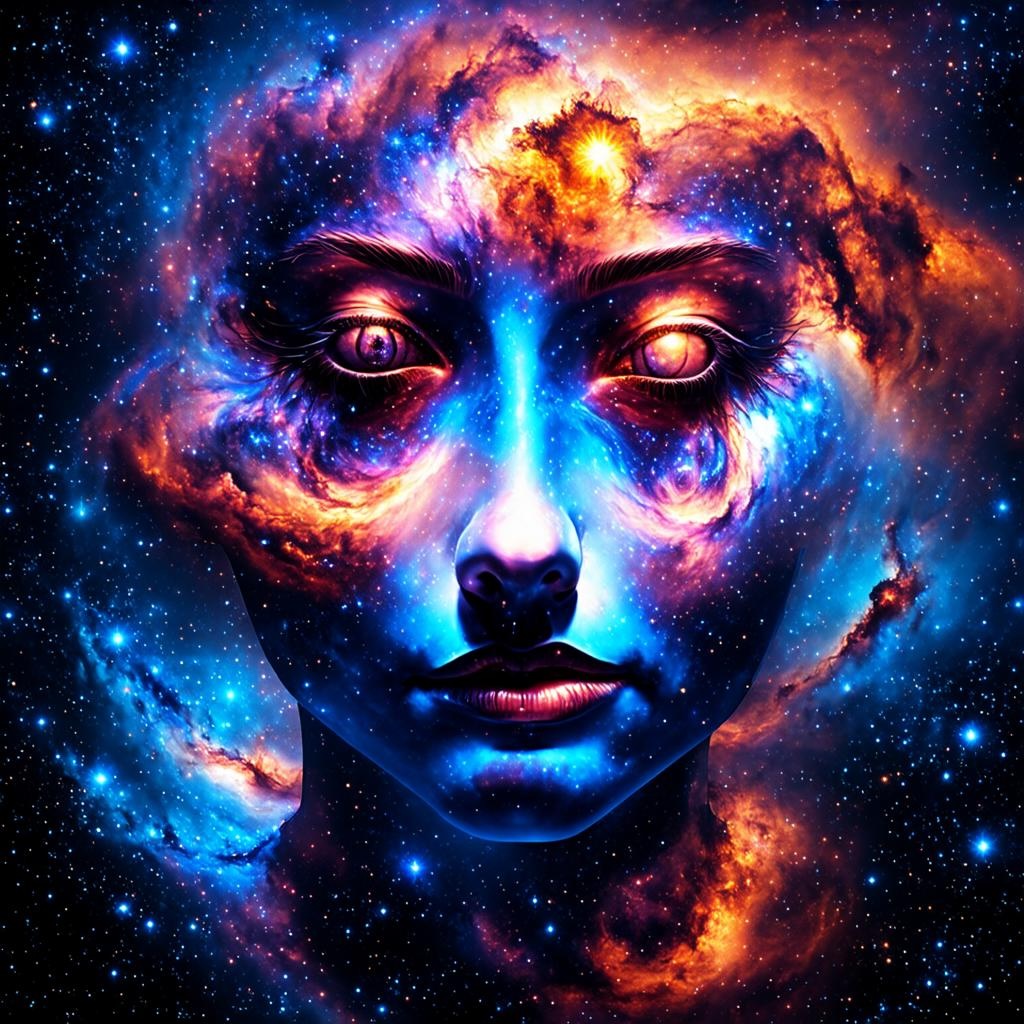}\end{minipage}%

  \begin{minipage}{0.28\textwidth}
  \begin{minipage}{0.90\textwidth}
    \centering \footnotesize \raggedright {The image depicts a portrait of a panda by Petros Afshar.} 
  \end{minipage}
  \end{minipage}%
  \begin{minipage}{0.18\textwidth}\includegraphics[width=\linewidth]{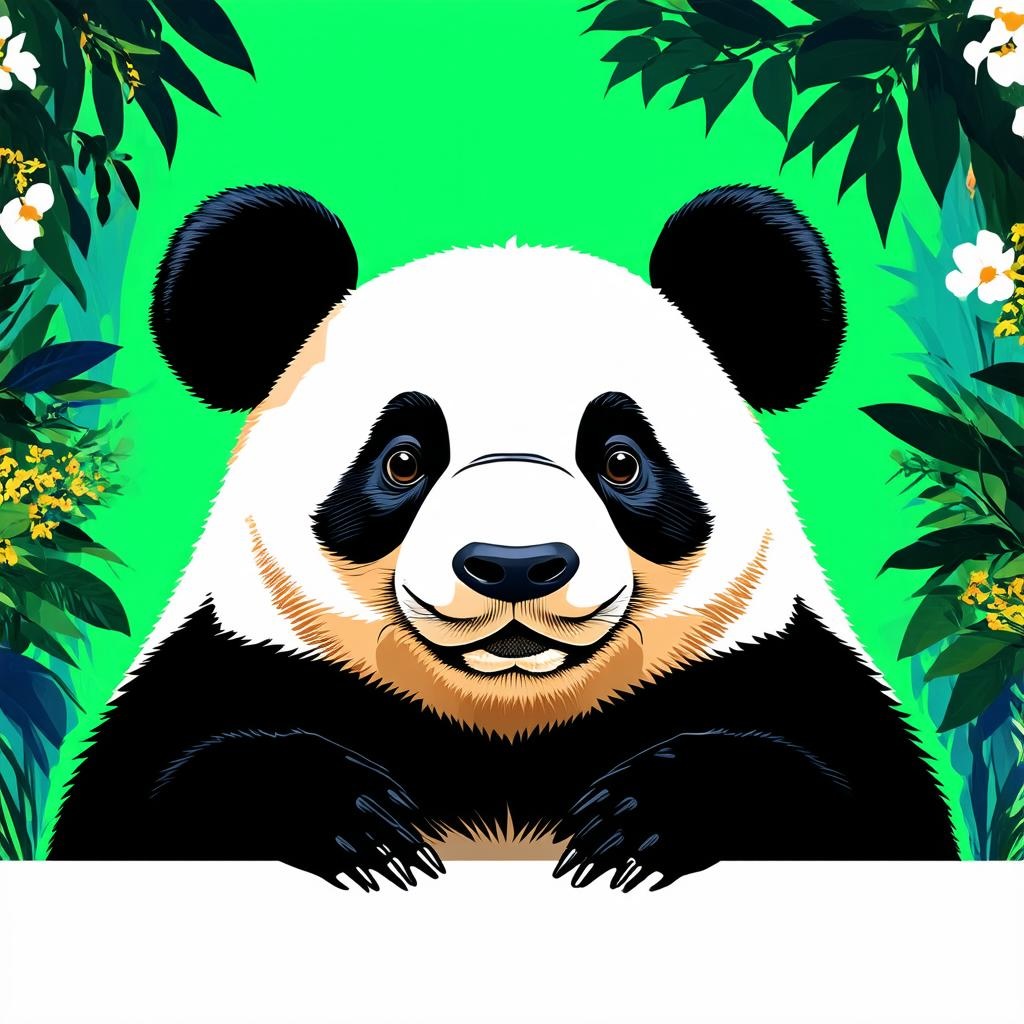}\end{minipage}%
  \begin{minipage}{0.18\textwidth}\includegraphics[width=\linewidth]{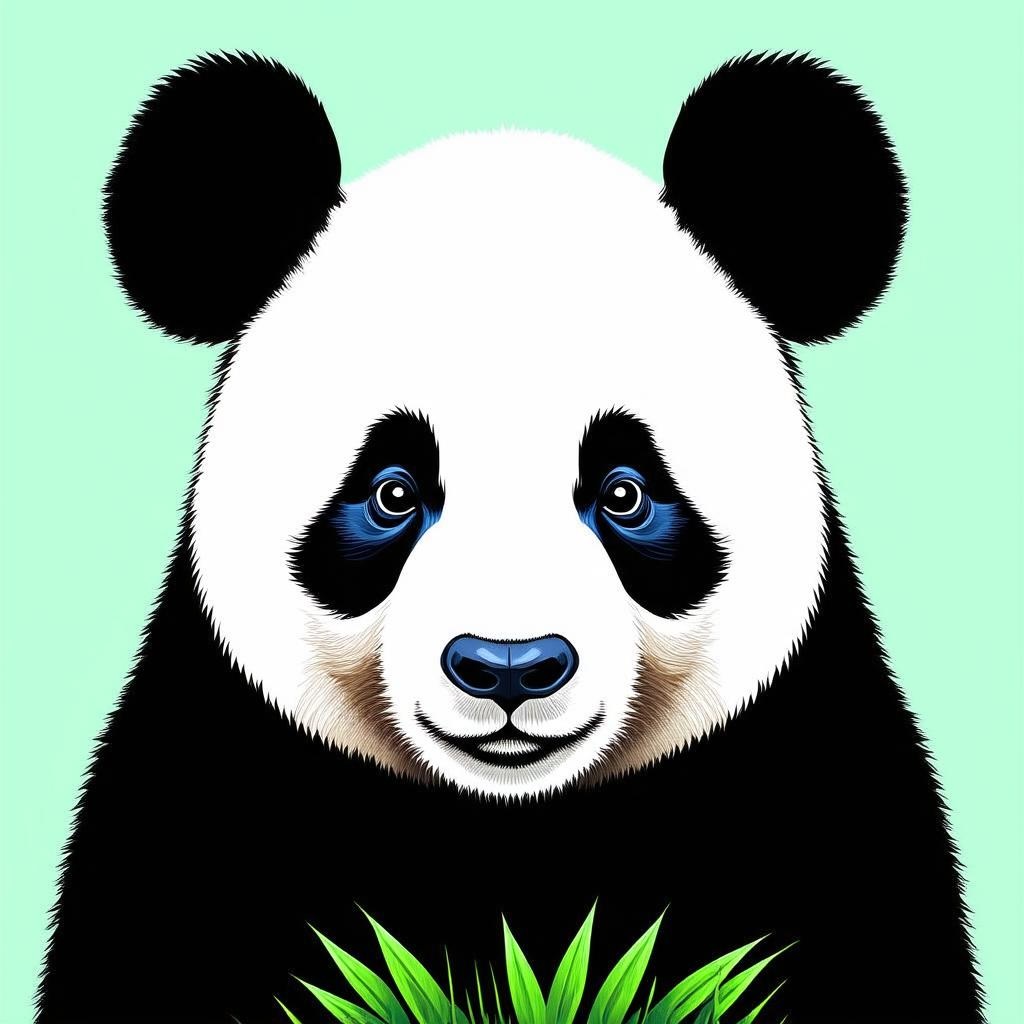}\end{minipage}%
  \begin{minipage}{0.18\textwidth}\includegraphics[width=\linewidth]{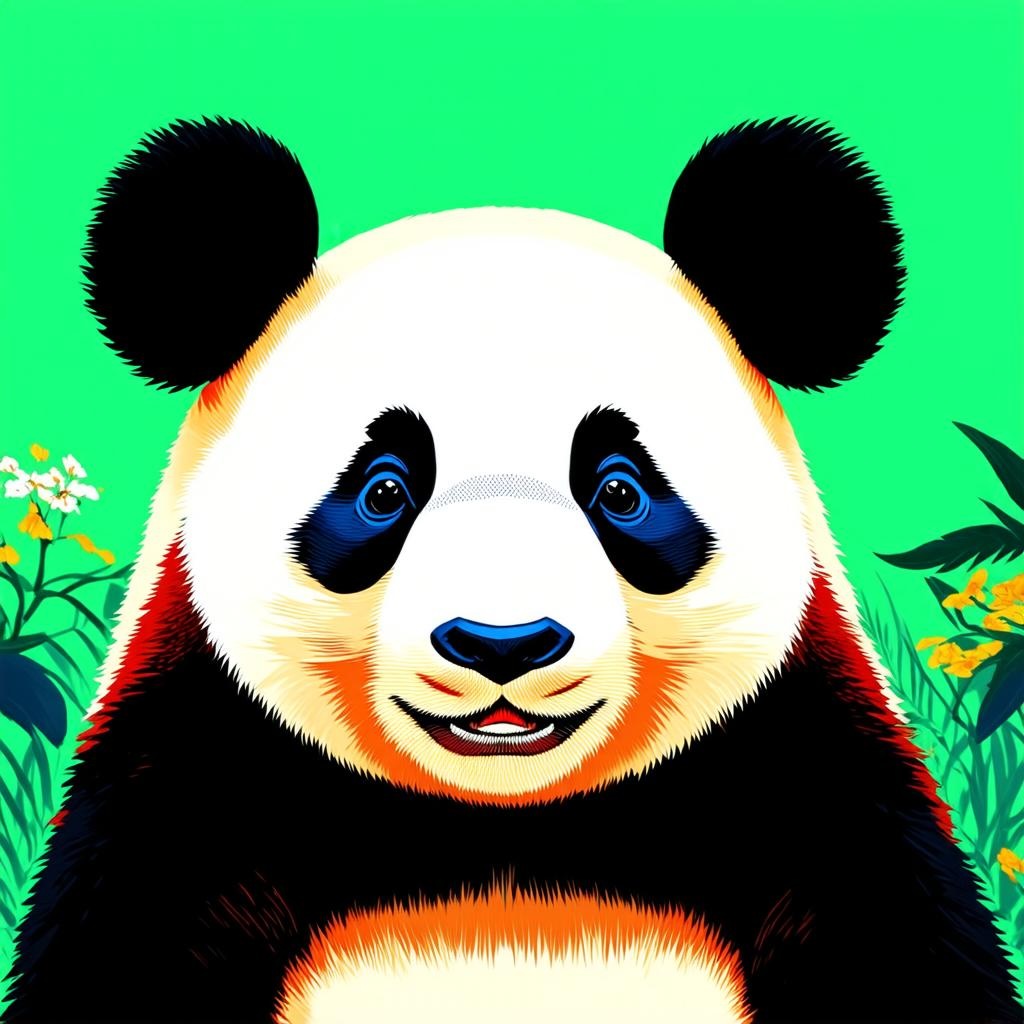}\end{minipage}%
  \begin{minipage}{0.18\textwidth}\includegraphics[width=\linewidth]{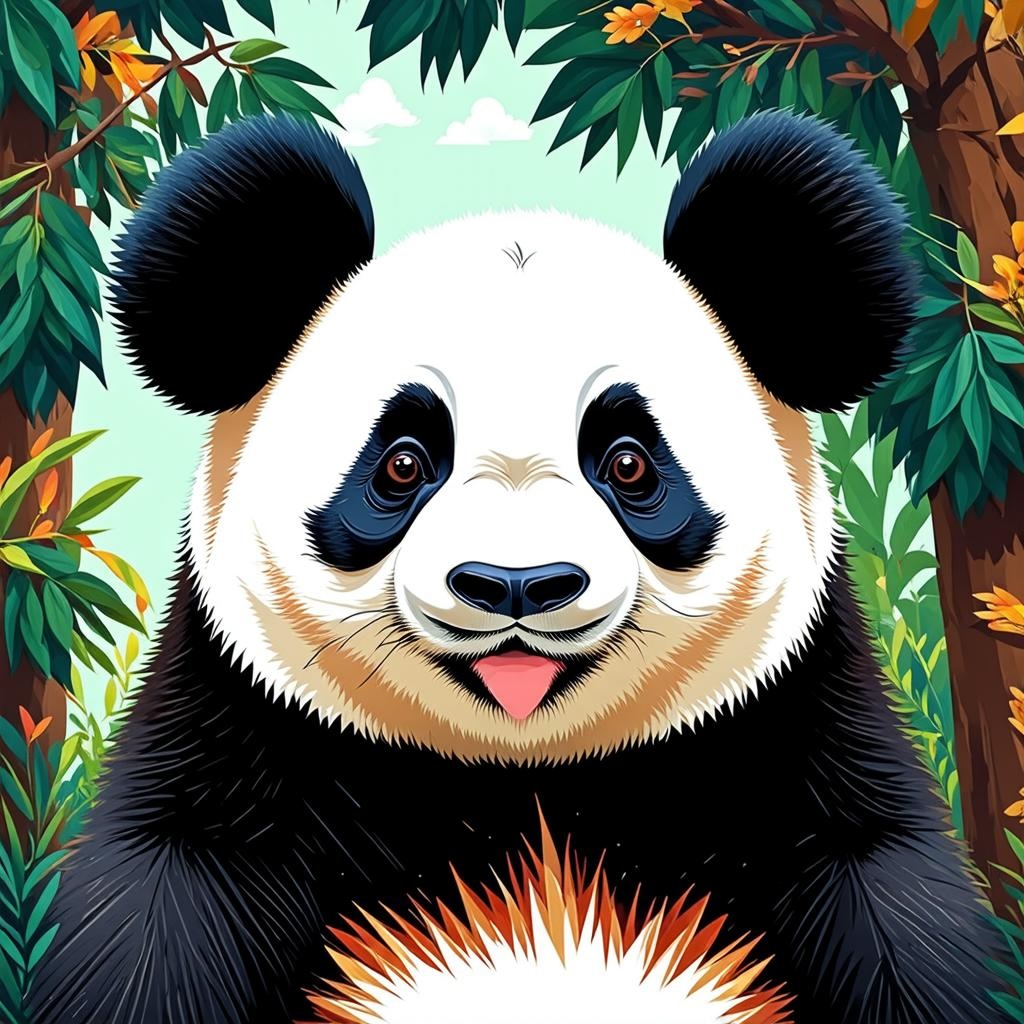}\end{minipage}%
  
  \begin{minipage}{0.28\textwidth}
  \begin{minipage}{0.90\textwidth}
    \centering \footnotesize \raggedright {The image is of a raccoon wearing a Peaky Blinders hat, surrounded by swirling mist and rendered with fine detail.} 
  \end{minipage}
  \end{minipage}%
  \begin{minipage}{0.18\textwidth}\includegraphics[width=\linewidth]{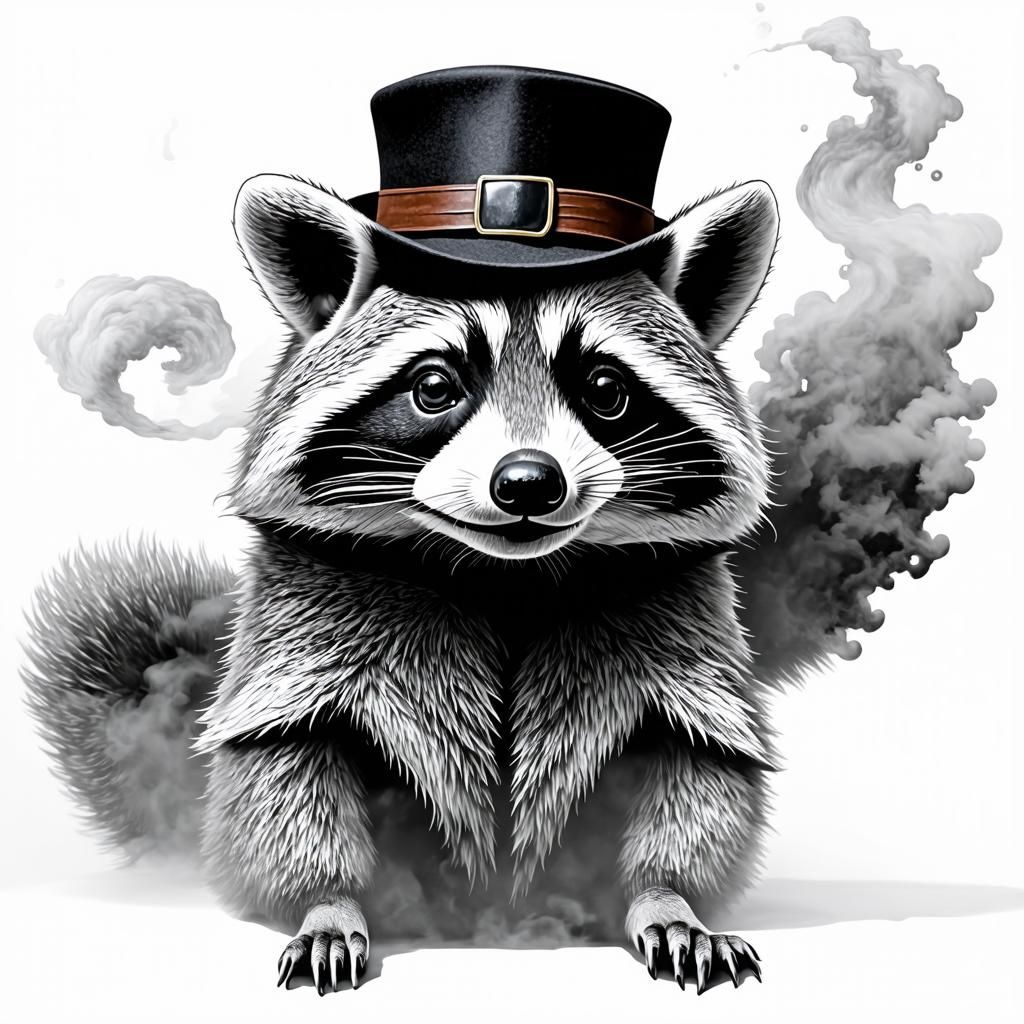}\end{minipage}%
  \begin{minipage}{0.18\textwidth}\includegraphics[width=\linewidth]{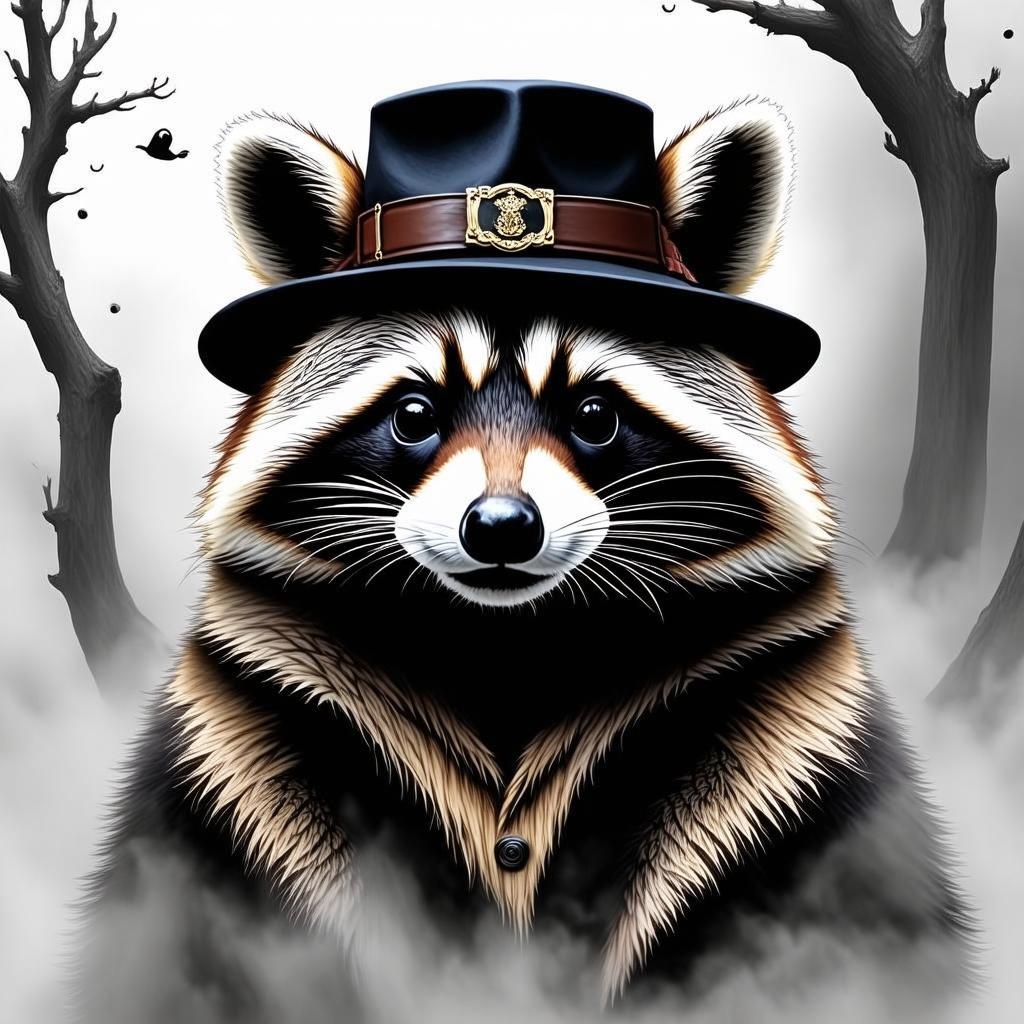}\end{minipage}%
  \begin{minipage}{0.18\textwidth}\includegraphics[width=\linewidth]{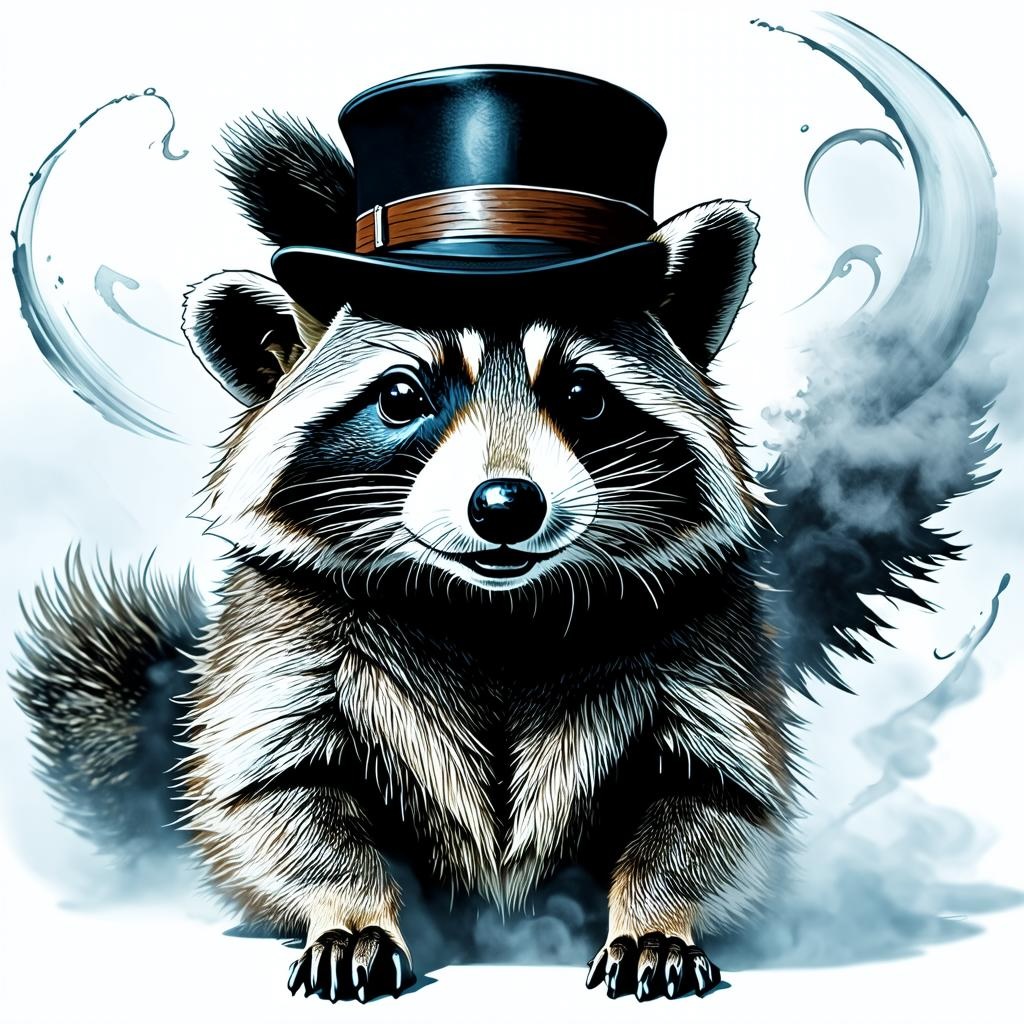}\end{minipage}%
  \begin{minipage}{0.18\textwidth}\includegraphics[width=\linewidth]{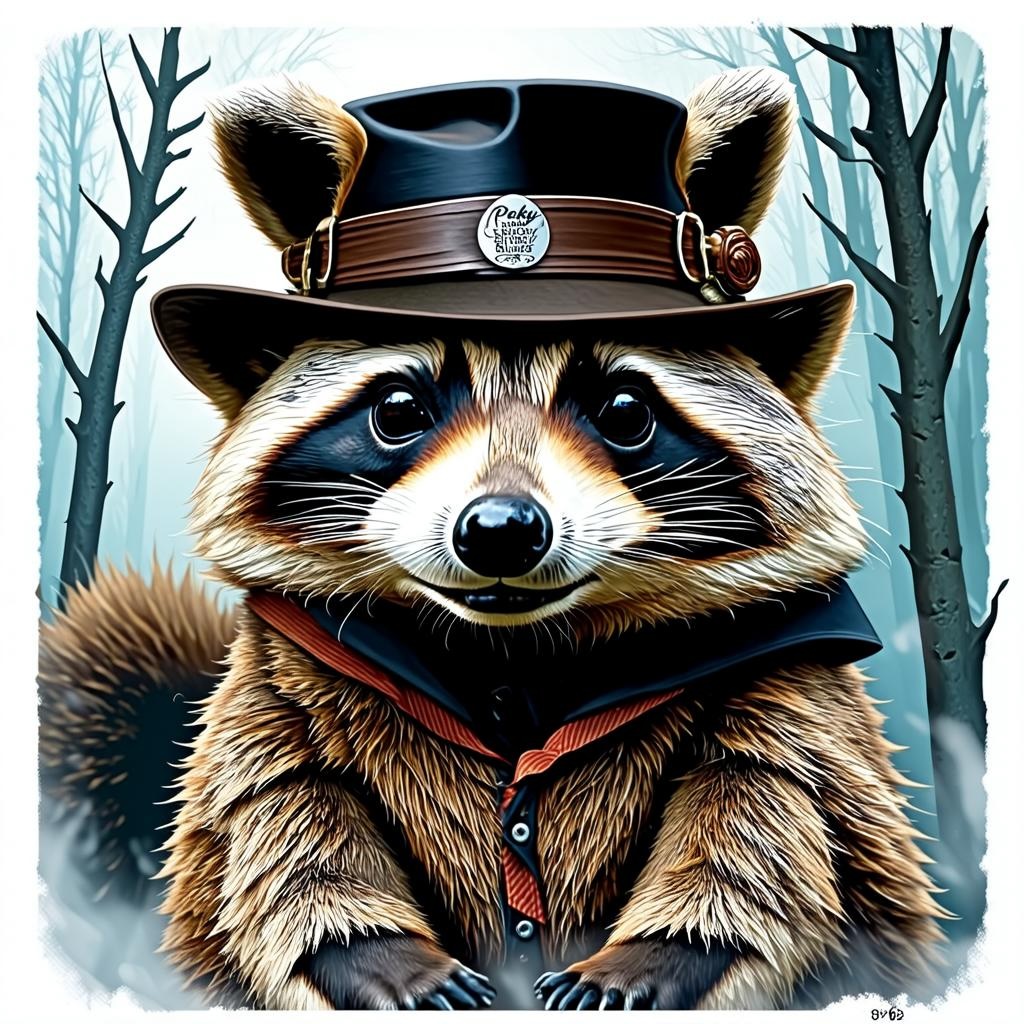}\end{minipage}%
  \end{minipage}
  \caption{Image samples generated from SD3-M fine-tuned with various methods, using validation prompts from Pick-a-Pic v2, HPDv2 and PartiPrompt.}
\end{figure*}

\end{document}